\documentclass[10pt,conference]{IEEEtran}
\IEEEoverridecommandlockouts   % allows \thanks in the title

\usepackage[utf8]{inputenc}
\usepackage[T1]{fontenc}
\usepackage{cite}              % numeric [1,2,3] citations, compresses ranges
\usepackage{amssymb,amsmath,makecell,multirow,booktabs,graphicx,threeparttable,adjustbox,xurl}
\usepackage[caption=false,font=footnotesize]{subfig}  % caption=false is required with IEEEtran

\begin{document}

\title{On the Complementarity of Quantum and Classical Features: Adaptive Hybrid Quantum-Classical Feature Fusion for Breast Cancer Classification}

\author{
  \IEEEauthorblockN{Yasmin R. Sobrinho, João R. R. Manesco, João P. Papa}
  \IEEEauthorblockA{Department of Computing, São Paulo State University (UNESP), Bauru, SP, Brazil\\
  \{yasmin.sobrinho, joao.r.manesco, joao.papa\}@unesp.br}
}

\maketitle
%% Abstract
\begin{abstract}
The integration of quantum machine learning with classical deep learning offers promising avenues for medical image analysis by mapping data into high-dimensional Hilbert spaces. However, effectively unifying these distinct paradigms remains challenging due to common optimization asymmetries. In this paper, a novel hybrid quantum-classical architecture for breast cancer diagnosis based on a dual-branch feature-extraction pipeline is proposed. Our framework extracts and unifies complementary representations from classical models and quantum circuits, exploring both trainable and deterministic (non-trainable) quantum paradigms. To integrate these embeddings, three progressive feature fusion strategies are introduced: Static Hybrid Fusion (SHF) for offline extraction, Dynamic Hybrid Fusion (DHF) for end-to-end co-adaptation, and a novel Temperature-Scaled Hybrid Fusion (TSHF). The TSHF strategy incorporates a learnable scalar, inspired by multimodal learning, that adaptively balances the gradient dynamics of the two branches and addresses optimization bottlenecks. We evaluate the framework on three publicly available breast imaging datasets---BreastMNIST, INbreast, and BUS-UCLM---spanning ultrasound and mammographic modalities, using both a ResNet-18 and a shallow CNN backbone, with five independent runs per configuration. The results support our hypothesis that unifying diverse feature representations creates a richer data context, since fusion improves over the strongest classical baseline on all three benchmarks. On BreastMNIST, the TSHF strategy pairing a ResNet backbone with a non-trainable quantum circuit reaches an accuracy of 88.33\%, an F1-score of 92.09\%, and an AUC-ROC of 92.09\%, against 84.74\% accuracy and 87.08\% AUC-ROC for the standalone ResNet-18. On INbreast, the best configuration is SHF with the same pairing, at 70.42\% accuracy, 72.96\% F1-score, and 80.28\% AUC-ROC, and on BUS-UCLM it is SHF with a trainable circuit, at 78.46\% accuracy and 84.66\% F1-score, over seven accuracy points above the best standalone branch. The benefit of adaptive fusion is therefore not universal: static fusion is the strongest strategy on the two smaller datasets, which we attribute to the additional estimation burden that co-adaptation and learned calibration impose when training data are scarce. These findings offer a methodological and exploratory contribution to hybrid quantum-classical representation learning, although clinical applicability requires further validation beyond the small, low-resolution datasets and simulated quantum circuits considered here. %We frame these findings as a methodological and exploratory contribution to hybrid quantum-classical representation learning rather than a clinically deployment-ready diagnostic system, given the reliance on small, low-resolution datasets and simulated quantum circuits.
\end{abstract}


% \centering
%     \includegraphics[width=1.0\linewidth]{Figure_0.png}

% \begin{highlights}
%     \item Hybrid quantum-classical fusion improves breast cancer classification. 
%     \item Different fusion strategies combine robust quantum and classical features. 
%     \item A quantum branch with at most 4 trainable parameters complements classical features. 
%     \item TSHF learnable scalar dynamically balances classical and quantum features.
%     \item UMAP shows that quantum integration improves latent space separability. 
% \end{highlights}

% %% Keywords
% \begin{keyword}
% quantum machine learning \sep hybrid neural networks \sep feature fusion \sep breast cancer classification \sep medical image analysis
% %% keywords here, in the form: keyword \sep keyword



% \end{keyword}

% \end{frontmatter}



\section{Introduction}
\label{s.intro}
Breast cancer remains the most commonly diagnosed cancer in women~\cite{whobreastcancer}, and the World Health Organization projects global cases to rise by nearly 40\% by 2050, with related deaths expected to surge by 68\%~\cite{who-bc-rising}. Tumors identified at earlier stages respond far better to treatment~\cite{early_detection_benefits}, yet organized screening remains inaccessible to many women, particularly in low- and middle-income countries~\cite{limited_access}. There is a pressing need for diagnostic tools that are simultaneously accurate, computationally accessible, and suitable for early detection.

Deep learning, and Convolutional Neural Networks (CNNs) in particular, has become the dominant approach for extracting discriminative features from mammograms, ultrasound, and histopathology, reducing diagnostic subjectivity and accelerating analysis~\cite{ali2025exploring, sarvamangala2022convolutional}. Nevertheless, classical models struggle to efficiently capture the highly non-linear correlations inherent to tumors: scaling them to such complexity typically demands increasingly deep, resource-heavy networks, with the associated optimization challenges and computational burden~\cite{mienye2025deep}. These limitations have motivated the exploration of Quantum Machine Learning (QML), which exploits superposition and entanglement to process information in high-dimensional Hilbert spaces, offering an inherently distinct representation of complex data~\cite{shahriyar2025advancements}.

Purely quantum approaches, however, remain impractical for large-scale medical data under current hardware constraints, particularly limited qubit counts~\cite{shahriyar2025advancements}, which has shifted research towards hybrid quantum-classical models~\cite{hafeez2024h}. Crucially, recent work indicates that quantum circuits do not merely approximate classical feature extractors, but learn features with genuinely different properties from the same input images~\cite{long2025hybrid}.

This divergence in learning paradigms forms the core of our work: we hypothesize that classical and quantum representations are complementary, and that fusing them yields a richer representation for breast cancer diagnosis. We propose a hybrid feature fusion protocol based on a dual-branch feature-extraction pipeline, and investigate whether \emph{adaptive} quantum-classical feature fusion improves classification relative to classical-only and quantum-only baselines, as well as to the non-adaptive fusion strategies we introduce as earlier stages of the same framework (SHF and DHF), and whether a learnable calibration of the quantum embedding can mitigate the optimization asymmetry that limits joint hybrid training. Our contributions, both methodological and empirical, are listed below:
\begin{itemize}
    \item A dual-branch architecture that integrates the distinct feature-extraction aspects of classical and quantum machine learning models for medical image analysis, in both trainable and deterministic (non-trainable) paradigms.

    \item A learned temperature mechanism, inspired by multimodal learning, embodied in the proposed Temperature-Scaled Hybrid Fusion (TSHF) strategy, which uses a single learnable scalar to balance classical and quantum features and directly addresses the optimization asymmetry in hybrid models.

    \item A systematic evaluation of the three fusion strategies introduced in this work---Static Hybrid Fusion (SHF), Dynamic Hybrid Fusion (DHF), and TSHF---across two classical backbones (ResNet-18 and a shallow CNN), trainable and non-trainable quantum branches, and three breast imaging datasets, framed as a controlled progression from static to adaptive fusion.

    \item A validation of our hypothesis that unifying diverse feature representations creates a richer data context, corroborated quantitatively and through UMAP analyses of the learned latent spaces.
\end{itemize}

\section{Related Works}
\label{s.relatedworks}

Breast cancer diagnosis through the lens of image analysis has followed a well-established trajectory in the literature, where hand-crafted features fed into SVMs and random forests gave way to CNNs as the dominant tool once sufficient annotated data became available, with the latter offering end-to-end feature learning directly from mammograms, ultrasounds, and histological slides~\cite{sarvamangala2022convolutional,ali2025exploring}. Within this evolution, CNNs in particular have demonstrated strong performance across multiple imaging modalities, yet their limitations are well documented: deep networks require large amounts of data to generalize and are prone to overfitting on heterogeneous clinical datasets. Moreover, they often scale poorly with the non-linear complexity of tumor morphology~\cite{mienye2025deep}. These difficulties directly constrain clinical applicability in settings where data quality and quantity cannot be assumed.

In parallel with the rise of CNNs, QML has evolved along a structurally different path. By operating in high-dimensional Hilbert spaces through superposition and entanglement, variational quantum circuits can represent data relationships that are expensive or too complex for classical networks of equivalent parameter count~\cite{shahriyar2025advancements}. Quantum SVMs, QCNNs, and QNNs have all been evaluated on classification tasks, and although they seem promising, their practical deployment is constrained by qubit counts and hardware noise, making purely quantum models unsuitable for high-resolution medical images at this stage of hardware maturity.

Thus, Hybrid Quantum-Classical Neural Networks (HQCNNs) emerged as a pragmatic response to this problem, built on the idea that variational quantum circuits can be embedded as functional layers within otherwise classical pipelines. Henderson et al.~\cite{henderson2020quanvolutional} established one of the core mechanisms for such an analysis, the so-called quanvolutions, which replace the most computationally demanding part of a CNN, the convolutional kernels, with Parameterized Quantum Circuits (PQCs) that show measurable feature extraction gains on standard benchmarks. This work was extended by Mari et al.~\cite{mari2020transfer}, who formalized quantum transfer learning and enabled pre-trained classical representations to be coupled with variational quantum classifiers. Azevedo et al.~\cite{azevedo2022quantum} applied this idea directly to breast cancer detection, pairing a frozen ResNet18 with a variational quantum classifier and reporting gains over the classical baseline.

This approach of creating a hybrid architecture with quantum and classical components continued to be extended in the literature to a broader range of architectures and datasets. Matondo-Mvula and Elleithy~\cite{matondo2024breast} applied angle-encoded 9-qubit quanvolutions to the BreastMNIST ultrasound benchmark, demonstrating that HQCNNs can match or exceed classical CNNs on medical imaging tasks at low resolution. Xie et al.~\cite{xie2025quantum}, in turn, integrated a variational quantum circuit into a Swin Transformer for breast cancer screening. At the same time, they conducted t-SNE and PCA analyses of the resulting feature spaces, which show that the quantum branch yields qualitatively distinct representations, providing empirical support for the overfitting-mitigation hypothesis. Sobrinho et al.~\cite{sobrinho2025hybrid} demonstrated that even a 4-qubit quanvolution model achieves competitive performance on mammography and ultrasound data, confirming that hardware-constrained designs remain viable.

Across this body of work, quantum and classical branches are almost universally composed sequentially, with fusion reduced to concatenation or a fixed linear projection, yet current evidence suggests that quantum and classical components may learn distinct, complementary feature representations. Matondo-Mvula and Elleithy~\cite{matondo2024breast} take a step in this direction, proposing two parallel VQCs with distinct encoding strategies whose outputs are fused with classical CNN features before classification; their analysis shows significant gains over a matched classical baseline. However, even this approach does not account for the optimization asymmetries that often happen in the joint training of branches with fundamentally different loss landscapes. Crucially, across this literature, fusion is realized either sequentially or through fixed concatenation, so the two branches are combined without ever correcting the disparity in feature scale and gradient conditioning between them. As a result, no prior approach explicitly addresses the \emph{optimization asymmetry}---the tendency of the well-conditioned classical branch to dominate joint training and suppress the quantum contribution---which we identify as the key unresolved obstacle to effective quantum--classical fusion. The complementarity between classical and quantum representations, as noted in multiple studies~\cite{xie2025quantum,long2025hybrid}, motivates a more systematic approach to such fusion mechanisms. In this work, we introduce a dual-branch pipeline with three explicit fusion strategies: SHF, DHF, and TSHF, the last of which uses a learnable scalar parameter, inspired by multimodal learning, used to dynamically reweight classical and quantum contributions and directly address the optimization asymmetry problem.

\section{Proposed Methodology}
\label{s.proposedmethods}
This section details the proposed hybrid quantum-classical architecture, designed to systematically integrate quantum-mechanical feature spaces with classical deep learning representations. We conceptualize the framework as a dual-branch feature extraction pipeline followed by specialized integration mechanisms. First, we mathematically define the core quantum operations, establishing the formulation of quanvolutional layers under both trainable and non-trainable paradigms. Subsequently, we introduce the primary contribution of our methodology: a progressive suite of feature fusion strategies. By exploring SHF, DHF, and a novel TSHF, we provide a comprehensive framework that effectively unifies heterogeneous embeddings and mitigates the inherent optimization asymmetries in hybrid models.

Throughout, we reserve $\boldsymbol{\phi}$ for the parameters of the classical
network and $\boldsymbol{\theta}_Q$ for those of the quantum circuit, so that the
two parameter spaces are never conflated. Quantum states are written in Dirac
notation: $|\psi\rangle$ denotes a normalized state vector, $\langle \psi |$ its
conjugate transpose, and $\langle \psi | O | \psi \rangle$ the expectation value
of an observable $O$ in the state $|\psi\rangle$. The symbol $\bigotimes$ denotes
the tensor (Kronecker) product, so that $\bigotimes_{i=1}^{n_q} A_i$ is the
operator acting as $A_i$ on qubit $i$ of an $n_q$-qubit register, and
$\mathcal{U}(S)$ denotes the uniform probability distribution over the set $S$.

\subsection{Quanvolutional Layer}
\begin{figure}[ht!]
\centering
\includegraphics[width=1\linewidth]{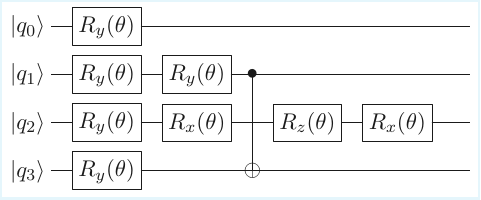}
\caption{Quantum circuit architecture utilized in the proposed quanvolutional layer.}\label{fig:circuit}
\end{figure}

Quanvolutional layers serve as the quantum analogue to classical convolutional operations~\cite{henderson2020quanvolutional}, defining a local feature extraction map $f_Q : \mathbb{R}^{c \times k \times k} \to \mathbb{R}^{c'}$ applied over sliding windows of an input tensor, where $c$ represents the number of input channels, $k$ is the spatial dimension of the square convolutional kernel, and $c'$ denotes the number of output channels. In this work, we formalize the quanvolutional operator under two distinct paradigms, trainable and non-trainable quantum circuits, using a fixed quantum circuit configuration, illustrated in Figure~\ref{fig:circuit}.

Concretely, since all datasets are single-channel ($c = 1$), given an input of
spatial dimensions $H \times W$ the layer extracts local patches $P$ of size
$k \times k$ using a sliding window of stride $s$. Each patch is flattened in
row-major order into a vector $\mathbf{x} \in \mathbb{R}^{n_q}$, and encoded onto
an $n_q$-qubit register through angle encoding, i.e., one data-dependent rotation
per qubit; the register size is set equal to the number of pixels in the patch,
$n_q = c \times k^2 = k^2$. The encoded state is then transformed by the
entangling parameterized circuit of Figure~\ref{fig:circuit}, specified in
Sections~\ref{subsec:non_trainable} and~\ref{subsec:trainable}, whose entangling
gates couple the $n_q$ qubits within the patch. Each patch is processed
independently, and a Pauli-$Z$ expectation value is read out on every qubit and
passed through a ReLU nonlinearity, producing $n_q$ output values per patch.
These values are arranged so that the qubit index becomes the channel index and
the patch location becomes the spatial position: the measurement of qubit $i$ for
the patch at grid location $(j,l)$ is written to position $(j,l)$ of output
channel $i$. The $P$ patches therefore tile each of the $n_q$ channels, yielding
an output feature map with $c' = n_q$ channels at the patch-grid resolution. The
concrete instantiation used in all experiments (patch size, stride, qubit count,
and resulting feature-map shape) is specified in Table~\ref{tab:architecture}.

\subsubsection{Non-Trainable Layer} \label{subsec:non_trainable}
\begin{figure}[ht!]
\centering
\includegraphics[width=1\linewidth]{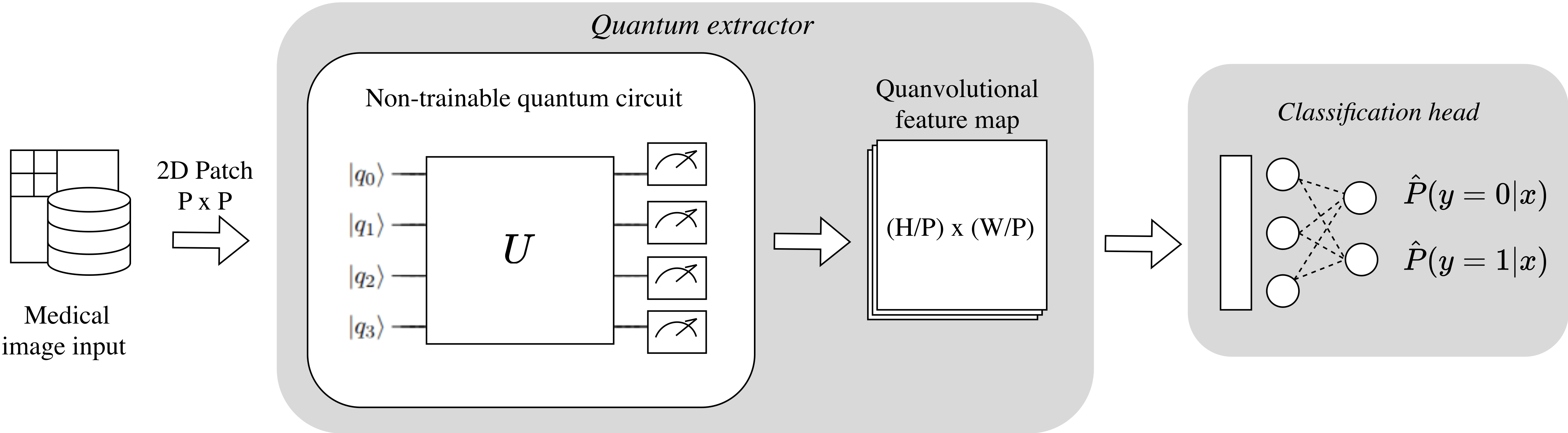}
\caption{Schematic of the hybrid model featuring a non-trainable quantum circuit for the quanvolutional layer, the resulting feature map, and the classification phase.}\label{fig:non_trainable}
\end{figure}

The non-trainable quanvolutional layer, shown in Figure~\ref{fig:non_trainable},
is designed to act as a non-trainable quantum feature map,
$\Phi : \mathbb{R}^{n_q} \to \mathbb{R}^{n_q}$. In this configuration, all
rotational gates depicted in the base architecture (Figure~\ref{fig:circuit}) are
parameterized by a constant vector
$\boldsymbol{\theta}_{\text{fix}} \sim \mathcal{U}\!\left([0, 2\pi)^{m_Q}\right)$,
where $m_Q$ is the total number of rotational parameters in the circuit and
$\mathcal{U}\!\left([0, 2\pi)^{m_Q}\right)$ denotes the uniform distribution over
the $m_Q$-dimensional hypercube $[0, 2\pi)^{m_Q}$; that is, each entry of
$\boldsymbol{\theta}_{\text{fix}}$ is drawn independently and uniformly from
$[0, 2\pi)$. This vector is sampled once prior to training and held constant
thereafter. The data are first angle-encoded through the initial column of $R_Y$
gates (as detailed in Section~\ref{subsec:trainable}), using the $\pi$-scaled
encoding specified in Table~\ref{tab:architecture}, which lets a normalized pixel
value span a wider rotation range than the bare encoding of the trainable layer.

Consequently, the quantum evolution becomes a deterministic embedding
$\mathbf{x} \mapsto \mathbf{y}$. By measuring the expectation value at the end of
each qubit wire, we obtain the output features:
\begin{equation}
y_i = \langle \psi(\mathbf{x}, \boldsymbol{\theta}_{\text{fix}}) | \sigma_z^{(i)} | \psi(\mathbf{x}, \boldsymbol{\theta}_{\text{fix}}) \rangle ,
\label{eq:non_trainable}
\end{equation}
where $|\psi(\mathbf{x}, \boldsymbol{\theta}_{\text{fix}})\rangle$ is the
$n_q$-qubit state prepared by encoding the patch $\mathbf{x}$ and applying the
circuit with the fixed parameters $\boldsymbol{\theta}_{\text{fix}}$;
$\sigma_z^{(i)}$ is the Pauli-$Z$ operator acting non-trivially on qubit $i$ and
as the identity on all remaining qubits; and
$\langle \psi | \sigma_z^{(i)} | \psi \rangle \in [-1, 1]$ is the expectation
value of that observable in the state $|\psi\rangle$, i.e.\ the mean outcome of
measuring qubit $i$ in the computational basis.

Within the broader context of the HQCNN architecture, the gradient of each output
feature with respect to the circuit parameters vanishes identically,
\begin{equation}
\nabla_{\boldsymbol{\theta}_Q} y_i = \mathbf{0} ,
\label{eq:zero_grad}
\end{equation}
where $\nabla_{\boldsymbol{\theta}_Q} y_i \in \mathbb{R}^{m_Q}$ denotes the
vector of partial derivatives of the $i$-th output feature $y_i$ with respect to
each entry of the quantum parameter vector $\boldsymbol{\theta}_Q$, and
$\mathbf{0}$ is the zero vector of the same dimension. Because the circuit
parameters are held fixed at $\boldsymbol{\theta}_{\text{fix}}$, no update signal
reaches the quantum branch, and learning is strictly confined to the parameter
space $\boldsymbol{\phi}$ of the classical layers. The non-trainable
quanvolutional layer therefore provides a stable and parameter-efficient
mechanism for quantum feature extraction. By entirely circumventing the
optimization hurdles inherent in variational circuits, this design is
particularly well-suited to hybrid models operating under NISQ-era limitations,
where circuit depth and trainability must be rigorously
controlled~\cite{preskill2018quantum}.

\subsubsection{Trainable Layer} \label{subsec:trainable}
\begin{figure}[ht!]
\centering
\includegraphics[width=1\linewidth]{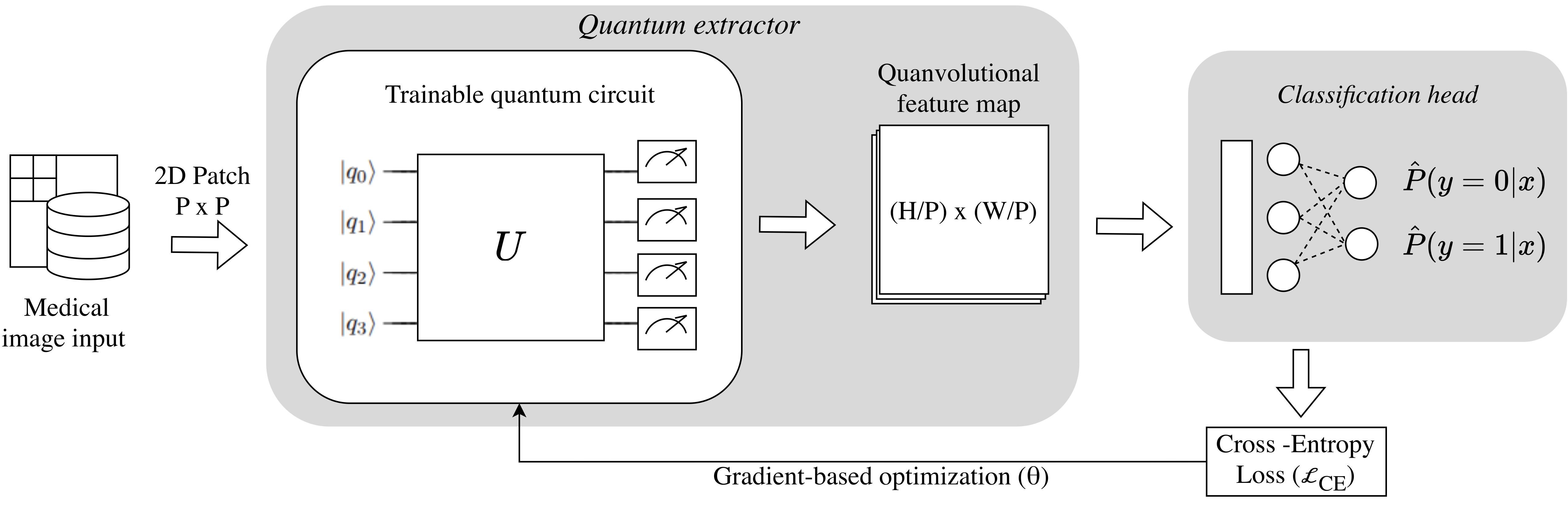}
\caption{Schematic of the hybrid model featuring a trainable quantum circuit for the quanvolutional layer, where both the quantum parameters $\boldsymbol{\theta}_Q$ and the classical parameters $\boldsymbol{\phi}$ are updated end-to-end via a gradient-based optimization loop.}\label{fig:trainable}
\end{figure}

The trainable quanvolutional layer, shown in Figure~\ref{fig:trainable}, is
modeled as a PQC, defining a transformation governed by a trainable parameter
vector $\boldsymbol{\theta}_Q \in \mathbb{R}^{m_Q}$~\cite{cerezo2021variational},
where the subscript distinguishes these circuit parameters from the classical
network weights $\boldsymbol{\phi}$. Let $\mathbf{x} \in \mathbb{R}^{n_q}$
represent a flattened local image patch, with $n_q = c \times k \times k$
corresponding to the total number of features per patch and to the required
number of qubits.

Referring to the base topology in Figure~\ref{fig:circuit}, the data is first
embedded into a quantum state via angle encoding~\cite{schuld2019quantum}. This
corresponds to the initial column of $R_Y$ gates, which act as a data-dependent
unitary
\begin{equation}
U_{\text{in}}(\mathbf{x}) = \bigotimes_{i=1}^{n_q} R_Y(x_i) ,
\label{eq:encoding}
\end{equation}
where $\bigotimes$ denotes the tensor (Kronecker) product over the $n_q$ qubits,
so that $U_{\text{in}}(\mathbf{x})$ applies the single-qubit rotation $R_Y(x_i)$
to qubit $i$ independently, for $i = 1, \dots, n_q$. Applying this unitary to the
initial ground state $|0\rangle^{\otimes n_q}$, that is, the tensor product of
$n_q$ copies of $|0\rangle$, yields the encoded state
$|\psi(\mathbf{x})\rangle = U_{\text{in}}(\mathbf{x})\,|0\rangle^{\otimes n_q}$.
Here, the generic rotation angle in the first layer of the schematic is replaced
by the normalized pixel value $x_i$, dictating a rotation around the $Y$-axis.

After data encoding, a variational \textit{ansatz} $U(\boldsymbol{\theta}_Q)$, represented
by the subsequent parameterized rotations ($R_X, R_Y, R_Z$) and the entangling
CNOT gate in Figure~\ref{fig:circuit}, is applied to produce the evolved state
$|\psi(\mathbf{x}, \boldsymbol{\theta}_Q)\rangle = U(\boldsymbol{\theta}_Q)\,|\psi(\mathbf{x})\rangle$.
The output feature vector $\mathbf{y} \in \mathbb{R}^{n_q}$ is then obtained by
evaluating the expectation values of the Pauli-$Z$ observables on each qubit $i$:
\begin{equation}
y_i = \langle \psi(\mathbf{x}, \boldsymbol{\theta}_Q) | \sigma_z^{(i)} | \psi(\mathbf{x}, \boldsymbol{\theta}_Q) \rangle ,
\label{eq:trainable}
\end{equation}
with $\sigma_z^{(i)}$ and the expectation-value notation defined as in
Eq.~\eqref{eq:non_trainable}. The two layers therefore differ only in whether the
circuit parameters are treated as constants or as free variables.

When integrated into the full HQCNN pipeline, the quantum circuit parameters
$\boldsymbol{\theta}_Q$ are optimized alongside the classical weights
$\boldsymbol{\phi}$ during training via
backpropagation~\cite{schuld2019evaluating}. This trainable configuration allows
the quantum circuit to adapt its feature extraction process to the target
classification task. However, the parameter dimensionality $m_Q$ is deliberately
kept small to maintain compatibility with NISQ-era constraints and to mitigate
optimization challenges such as the barren plateau
phenomenon~\cite{mcclean2018barren}, in which the variance of the gradient
entries, $\mathrm{Var}\!\left[\nabla_{\boldsymbol{\theta}_Q} y_i\right]$, decays
exponentially with the number of qubits and with circuit depth.

\subsection{Feature Fusion}
The core objective of our hybrid architecture is to leverage the complementary strengths of quantum and classical computing. While classical convolutional networks are optimized for hierarchical, shift-invariant spatial features~\cite{lecun2015deep}, quantum circuits, specifically quanvolutional layers, can capture complex correlations that are often inaccessible to classical kernels~\cite{henderson2020quanvolutional}.

To unify these paradigms, we propose a parallel dual-branch pipeline. In this setup, the input image is processed simultaneously by a quantum module, as described in Sections~\ref{subsec:non_trainable} and \ref{subsec:trainable}, and a classical baseline, ResNet-18~\cite{he2016deep} or a Shallow CNN (SCNN), whose full specifications and parameter counts are given in Table~\ref{tab:architecture}. This process generates two distinct embeddings: a quantum representation, $\mathbf{h}_Q$, and a classical one, $\mathbf{h}_C$.

However, integrating these heterogeneous feature spaces introduces significant challenges related to dimensionality disparity and optimization stability. Rather than proposing a single fusion mechanism in isolation, we frame our approach as a controlled progression of three strategies of increasing sophistication, in which each stage isolates one question and its answer motivates the next:
\begin{itemize}
    \item \textbf{Static Hybrid Fusion (SHF)} first asks whether quantum and classical features are complementary \emph{at all}. It fuses frozen, independently extracted embeddings and trains only the classifier, so that any observed gain cannot be attributed to joint optimization.
    \item \textbf{Dynamic Hybrid Fusion (DHF)} then allows both branches to co-adapt end-to-end. This can unlock further complementarity, but it also \emph{exposes} the central obstacle to hybrid training: an optimization asymmetry in which the well-conditioned classical branch dominates and suppresses the quantum contribution, compounded by a raw-magnitude mismatch between the two embeddings.
    \item \textbf{Temperature-Scaled Hybrid Fusion (TSHF)}, our main methodological contribution, directly addresses this diagnosed failure by introducing a learnable scalar that calibrates the quantum embedding before fusion.
\end{itemize}
All three strategies are proposed in this work as stages of a single, purpose-built fusion framework: to the best of our knowledge, no prior hybrid quantum--classical study has formulated and systematically compared this kind of progression of parallel dual-branch fusion mechanisms, spanning offline extraction (SHF), end-to-end co-adaptation (DHF), and learnable calibration (TSHF). Framed this way, SHF and DHF are not weaker alternatives competing with TSHF, but the earlier stages of the same framework that establish the problem TSHF is designed to solve: SHF demonstrates that the complementarity worth exploiting exists, DHF reveals why uncalibrated joint fusion fails to exploit it, and TSHF mitigates that specific failure through a learnable temperature scalar. Each strategy is detailed in the subsequent sections.

\subsubsection{Static Hybrid Fusion}
\begin{figure}[ht!]
\centering
\includegraphics[width=1\linewidth]{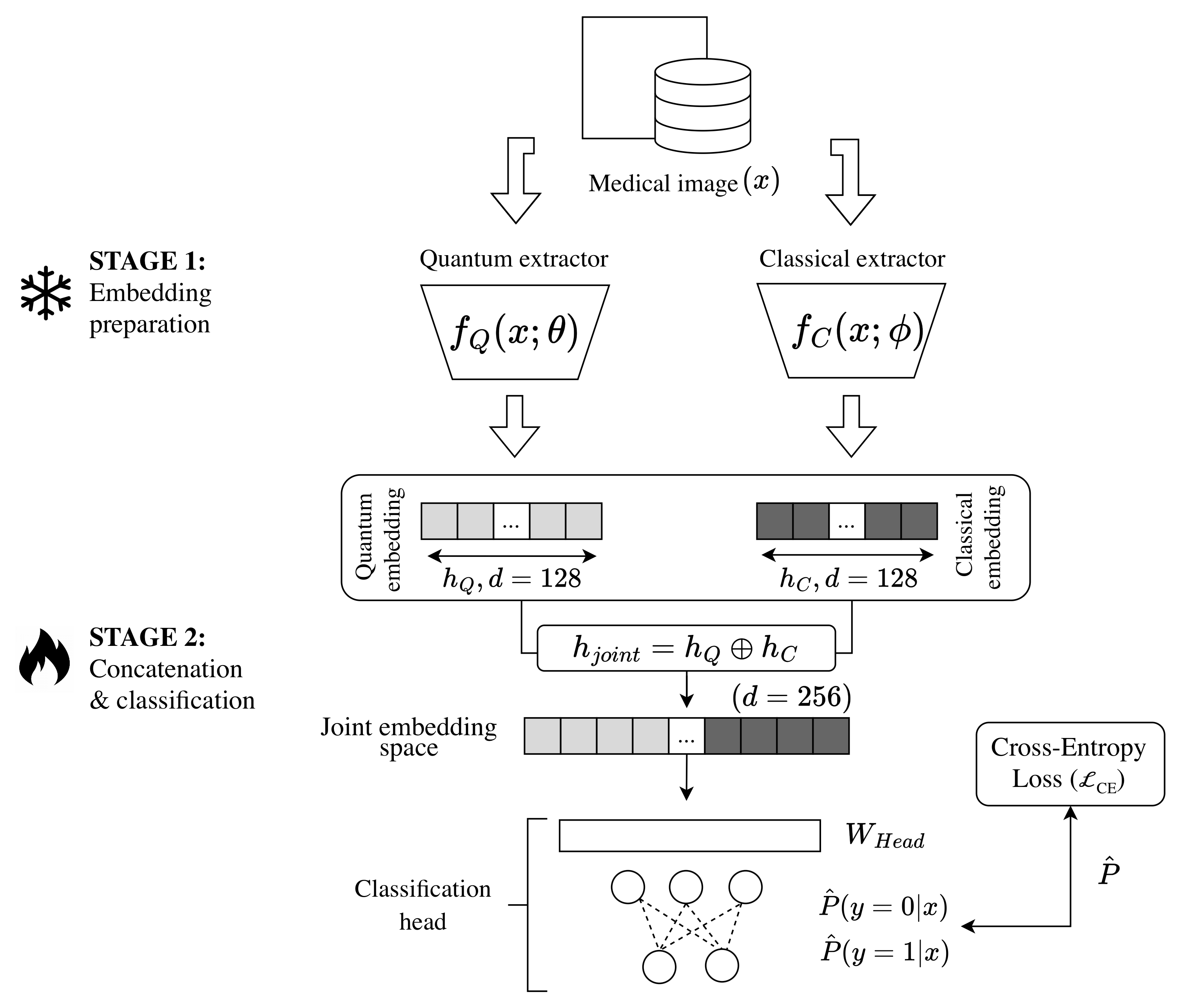}
\caption{Schematic of the SHF strategy, where features are extracted offline and fusion is restricted to the CH.}\label{fig:SHF}
\end{figure}

In this first strategy, illustrated in Figure~\ref{fig:SHF}, the quantum and classical modules operate strictly as independent feature extractors within a two-stage pipeline. The primary objective of this setup is to evaluate the raw representational capacity of the quantum embeddings, independent of the Classification Head's (CH) optimization dynamics. Consequently, no gradient flows back from the classifier to either of the feature extraction branches during the fusion stage.

To formalize the feature extraction process for this and subsequent strategies, let $f_Q$ denote the quantum feature extractor and $f_C$ represent the classical baseline extractor. The quantum extractor $f_Q$ is configured using either the trainable quanvolutional layer (Eq.~\ref{eq:trainable}) or the non-trainable deterministic feature map (Eq.~\ref{eq:non_trainable}).

For a given input image $\mathbf{x}$, these models produce their respective
embeddings, denoted as $\mathbf{h}_Q = f_Q(\mathbf{x})$ and
$\mathbf{h}_C = f_C(\mathbf{x})$. A dedicated \emph{projection head} attached to
each branch maps its raw feature vector---the flattened quantum feature map for
$\mathbf{h}_Q$ and the backbone's penultimate representation for
$\mathbf{h}_C$---to a common dimension $d$, so that the two heterogeneous
representations share the same dimensionality prior to fusion. Each projected
embedding is then $\ell_2$-normalized, so that both branches enter the fusion
with unit norm (see Table~\ref{tab:architecture}).

In the first stage of the SHF pipeline, these features are extracted offline and stored for the entire dataset (training, validation, and testing splits).
In the second stage, the pre-computed representations are concatenated to form a joint embedding space:
\begin{equation}
\mathbf{h}_{\text{joint}} = \mathbf{h}_Q \oplus \mathbf{h}_C,
\label{eq:joint}
\end{equation}
where $\oplus$ denotes the concatenation operator, which stacks two vectors of dimension $d$ into a single vector of dimension $2d$, yielding a combined feature vector $\mathbf{h}_{\text{joint}} \in \mathbb{R}^{2d}$.

This joint vector is then fed into a CH, implemented as a fully connected classification layer. During training, optimization is strictly confined to this final head, which learns to map the concatenated representations to the target class probabilities. By pairing the two quantum configurations with the two classical architectures, this offline strategy systematically yields four distinct hybrid model combinations for comparative analysis.

\subsubsection{Dynamic Hybrid Fusion}
\begin{figure}[ht!]
\centering
\includegraphics[width=1\linewidth]{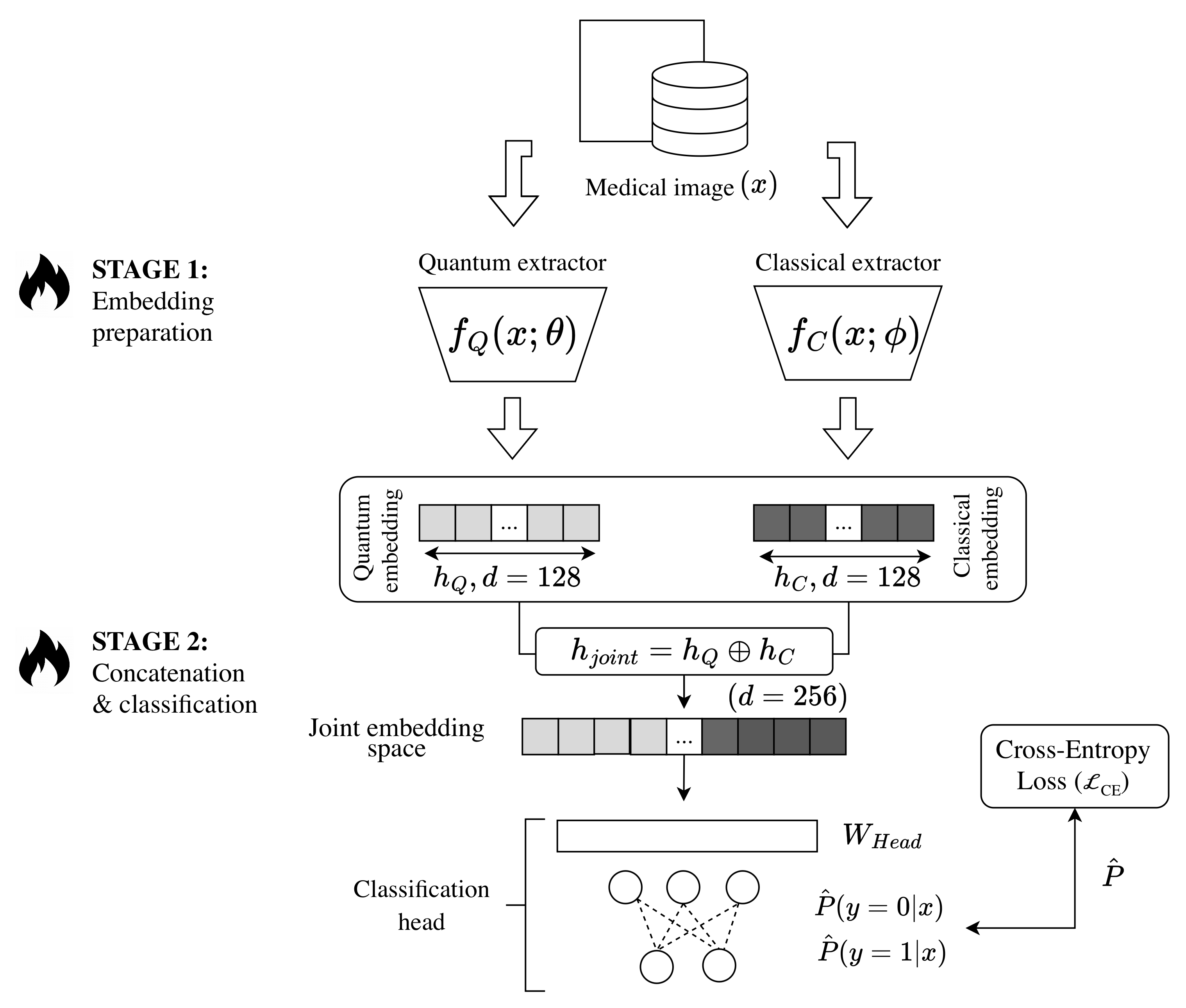}
\caption{Schematic of the DHF strategy, illustrating the end-to-end gradient flow through both branches.}\label{fig:DHF}
\end{figure}

Unlike the offline pipeline, where representations are isolated from the classification objective during feature extraction, this strategy, as depicted in Figure~\ref{fig:DHF}, forces the quantum and classical branches to co-adapt. The primary goal of this fully differentiable approach is to allow the network to discover synergistic feature interactions, optimizing both extractors simultaneously for the specific classification task.

In this paradigm, the input image $\mathbf{x}$ is propagated through $f_Q$ and $f_C$ in a single forward pass. The resulting embeddings $\mathbf{h}_Q$ and $\mathbf{h}_C$ are dynamically fused at runtime via the concatenation operation defined in Eq.~\ref{eq:joint}.

This joint representation $\mathbf{h}_{\text{joint}}$ is immediately passed to the CH, which outputs the predicted class probabilities. During the backward pass, the classification error is quantified using the Cross-Entropy Loss function. Crucially, the gradient of this loss is propagated backward through the CH and then bifurcates, flowing directly into both the classical parameters $\boldsymbol{\phi}$ of $f_C$ and the quantum parameters $\boldsymbol{\theta}_Q$ of $f_Q$ (when utilizing the trainable quanvolutional layer).

The entire hybrid architecture is optimized jointly end-to-end (see Table~\ref{tab:hyperparameters} for the per-branch optimization settings). While this synchronous training scheme theoretically enables the discovery of highly complementary hybrid features, it introduces significant optimization asymmetries. Because CNNs generally exhibit more stable and well-scaled gradient landscapes compared to PQCs, the classical branch can dominate the learning process~\cite{mcclean2018barren}. This classical gradient dominance can inadvertently suppress the learning capacity of the quantum circuit, rendering the $\mathbf{h}_Q$ representation under-optimized, a challenge that directly motivates the introduction of TSHF.

\subsubsection{Temperature-Scaled Hybrid Fusion}
\begin{figure}[ht!]
\centering
\includegraphics[width=1\linewidth]{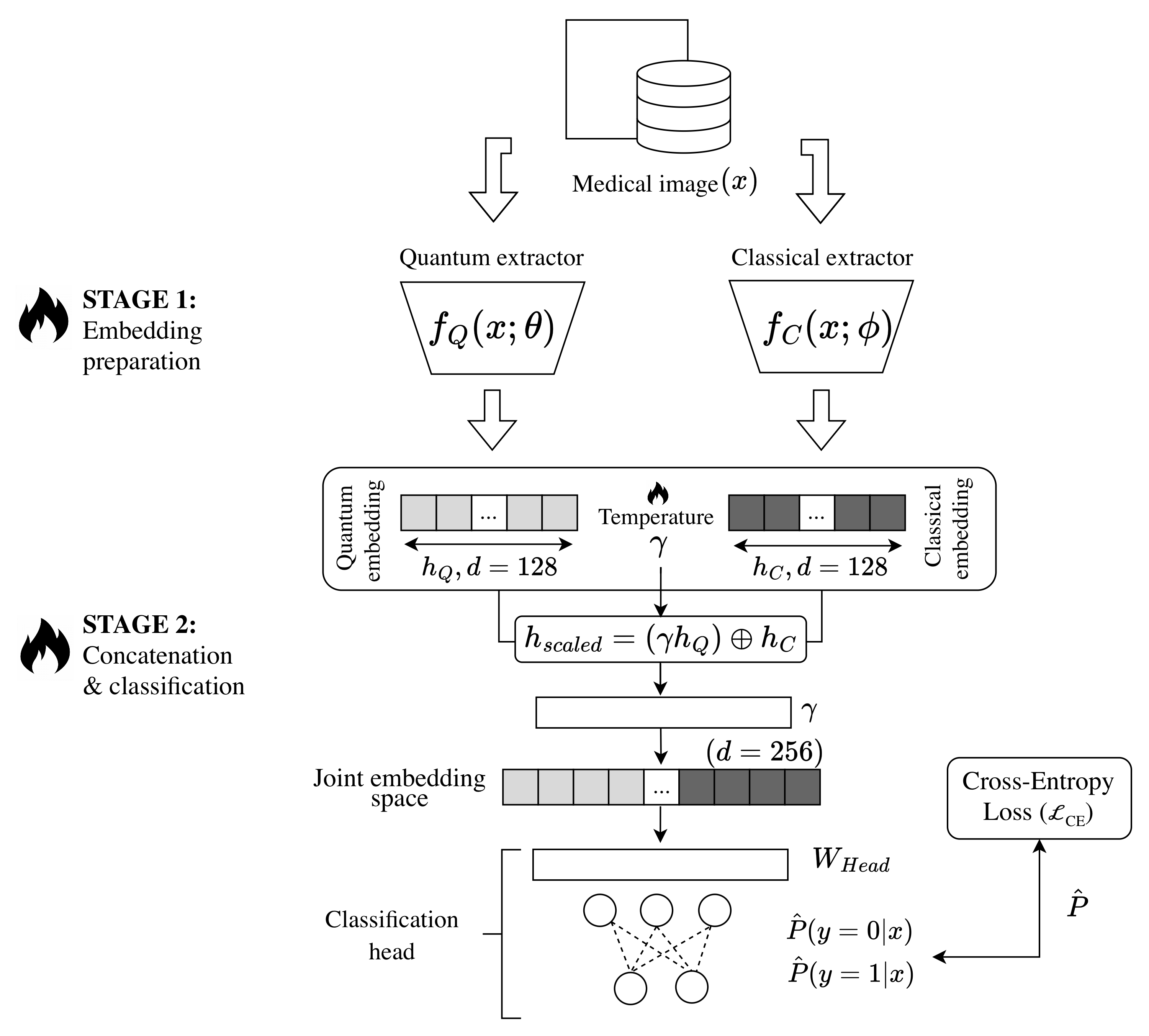}
\caption{Schematic of the TSHF strategy, featuring the learnable scalar $\gamma$ for quantum feature modulation.}\label{fig:TSHF}
\end{figure}

As established in the previous section, a critical bottleneck in the joint optimization of hybrid architectures is the inherent disparity in scale, variance, and representational capacity between quantum and classical embeddings~\cite{liang2022mind}. The classical network, benefiting from a highly parameterized space and unconstrained gradient flows, often dominates the learning objective. To mitigate this asymmetric co-adaptation and prevent the suppression of quantum features, we introduce the TSHF strategy, shown in Figure~\ref{fig:TSHF}.

To move beyond the constraints of fixed concatenation, we formulate a dynamic integration strategy by introducing a trainable, scalar temperature parameter $\gamma \in \mathbb{R}$, inspired by multimodal architectures~\cite{radford2021learning, guo2017calibration}. This parameter modulates the magnitude of the quantum feature vector prior to its fusion with the classical baseline. The temperature-scaled joint representation is formally defined as:
\begin{equation}
\mathbf{h}_{\text{scaled}} = (\gamma \, \mathbf{h}_Q) \oplus \mathbf{h}_C,
\label{eq:scaled_fusion}
\end{equation}
where the scalar multiplication is applied element-wise across $\mathbf{h}_Q$, and $\oplus$ denotes the concatenation operator of Eq.~\eqref{eq:joint}, yielding $\mathbf{h}_{\text{scaled}} \in \mathbb{R}^{2d}$.

The parameter $\gamma$ is initialized to $\gamma = 1.0$, which mathematically reduces to standard concatenation at the onset of training, and is iteratively updated via backpropagation alongside the network weights. The theoretical advantage of this formulation becomes particularly evident during the backward pass. By the chain rule, the gradient of the loss function $\mathcal{L}$ with respect to the quantum representations is directly scaled by this parameter:
\begin{equation}
\frac{\partial \mathcal{L}}{\partial \mathbf{h}_Q} = \gamma \, \frac{\partial \mathcal{L}}{\partial (\gamma \mathbf{h}_Q)} ,
\label{eq:chain_rule}
\end{equation}
where $\partial \mathcal{L} / \partial \mathbf{h}_Q$ denotes the gradient of the
loss with respect to the unscaled quantum embedding, and
$\partial \mathcal{L} / \partial (\gamma \mathbf{h}_Q)$ the gradient with respect
to its scaled counterpart, as received from the classification head.

This adaptive scaling may therefore act as an implicit, data-driven
regularization mechanism~\cite{wang2021understanding}, giving the optimizer the
flexibility to amplify or attenuate the contribution of the quantum
representation relative to the classical features during training. By
rescaling both the quantum embedding and the gradient that reaches the quantum
circuit, the scalar allows the relative contribution of the quantum branch to be
tuned after per-branch normalization, instead of being fixed in advance, which
mitigates the risk of the quantum features being overshadowed by classical
gradient dominance.

\section{Experimental Setup}
\label{s.setup}
This section describes the datasets and preprocessing pipeline, followed by the
implementation environment and the complete set of settings required to reproduce our
experiments, consolidated in Tables~\ref{tab:reproducibility},
\ref{tab:hyperparameters}, and~\ref{tab:architecture}.

\subsection{Datasets and Preprocessing}
\begin{figure}[ht!]
    \centering
    \subfloat[BreastMNIST]{\includegraphics[width=0.3\linewidth]{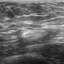}}
    \hfill
    \subfloat[INbreast]{\includegraphics[width=0.3\linewidth]{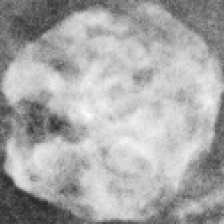}}
    \hfill
    \subfloat[BUS-UCLM]{
    \includegraphics[width=0.3\linewidth, height=0.3\linewidth]{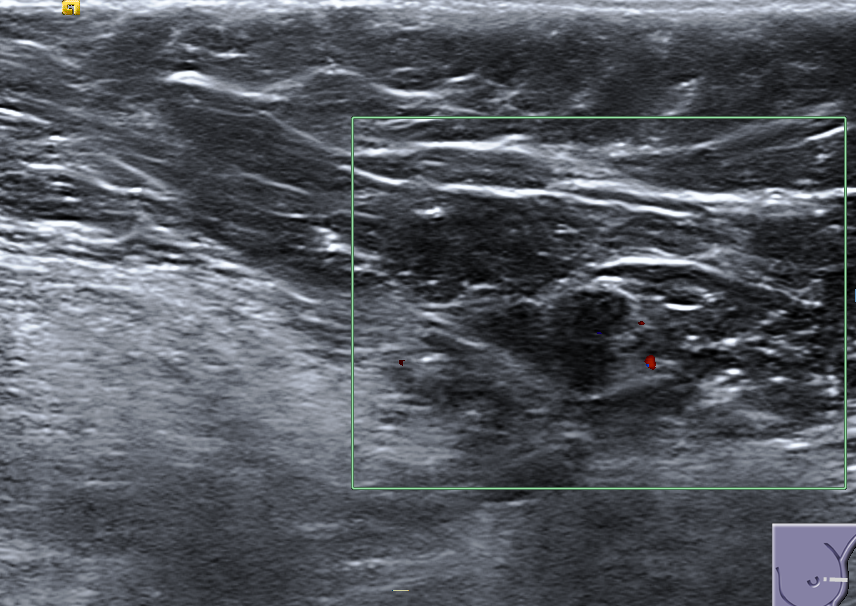}}
    \hfill

    \caption{Sample images from the BreastMNIST, INbreast, and BUS-UCLM datasets used for model training and evaluation.}
    \label{fig:datasets}
\end{figure}

This study employs three publicly available medical imaging datasets for breast cancer
classification: BreastMNIST~\cite{yang2023medmnist}, BUS-UCLM~\cite{vallez2025bus}, and
INbreast~\cite{moreira2012inbreast}, comprising breast ultrasound and mammographic
images for the binary discrimination of benign from malignant cases. BreastMNIST
consists of grayscale ultrasound images; BUS-UCLM provides annotated ultrasound images;
and INbreast includes digital mammograms with expert annotations. Figure~\ref{fig:datasets}
shows representative samples, and Table~\ref{tab:datasets} summarizes the class
distribution of each split.

\begin{table}[ht!]
\centering
\caption{Class distribution of the BreastMNIST, INbreast, and BUS-UCLM splits.}
\label{tab:datasets}
\footnotesize
\setlength{\tabcolsep}{8pt}
\renewcommand{\arraystretch}{1.1}
\resizebox{\linewidth}{!}{%
\begin{tabular}{llrrrr}
\toprule
\textbf{Dataset} & \textbf{Split} & \textbf{Benign} & \textbf{Malignant} & \textbf{Total} & \textbf{Malig.\ \%} \\
\midrule
\multirow{4}{*}{BreastMNIST}
 & Train & 399 & 147 & 546 & 26.9 \\
 & Val   & 57  & 21  & 78  & 26.9 \\
 & Test  & 114 & 42  & 156 & 26.9 \\
 & \textbf{Overall} & \textbf{570} & \textbf{210} & \textbf{780} & \textbf{26.9} \\
\midrule
\multirow{4}{*}{INbreast}
 & Train & 160 & 150 & 310 & 48.4 \\
 & Val   & 38  & 37  & 75  & 49.3 \\
 & Test  & 50  & 46  & 96  & 47.9 \\
 & \textbf{Overall} & \textbf{248} & \textbf{233} & \textbf{481} & \textbf{48.4} \\
\midrule
\multirow{4}{*}{BUS-UCLM}
 & Train & 135 & 64 & 199 & 32.2 \\
 & Val   & 32  & 7  & 39  & 17.9 \\
 & Test  & 7   & 19 & 26  & 73.1 \\
 & \textbf{Overall} & \textbf{174} & \textbf{90} & \textbf{264} & \textbf{34.1} \\
\bottomrule
\end{tabular}
}%
\end{table}

All datasets are processed into single-channel grayscale inputs at $28 \times 28$
resolution. BreastMNIST is used exactly as distributed by MedMNIST, at its native
resolution and with its official train/validation/test partition, which enables
direct comparison with prior work on the same benchmark. For INbreast, the expert
annotations are first used to crop the suspicious tumor regions (some images
contain more than one region of interest, each of which is treated as a separate
sample); the cropped regions are then enhanced with Contrast Limited Adaptive
Histogram Equalization (CLAHE), normalized with per-image min--max scaling to
$[0,1]$, and resized to $28 \times 28$. BUS-UCLM images are only normalized and
resized, with no cropping or contrast enhancement applied. For both INbreast and
BUS-UCLM, data partitions are constructed at the patient level: all samples
derived from a given patient are assigned to a single subset (training,
validation, or test), so that no patient appears in more than one subset,
preventing data leakage between training and evaluation.

\subsection{Implementation and Reproducibility Details}
All experiments were implemented in PyTorch, with quantum circuits built in PennyLane
and integrated through the \texttt{TorchLayer} interface. Quantum circuits were
simulated on CPU with the \texttt{lightning.qubit} state-vector backend in analytic
mode, while all classical components were executed on a CUDA-enabled NVIDIA L40S GPU.
Table~\ref{tab:reproducibility} consolidates the environment and protocol settings.

\begin{table}[htpb]
\centering
\caption{Experimental environment and evaluation protocol. Architectural details are
given in Table~\ref{tab:architecture} and training hyperparameters in
Table~\ref{tab:hyperparameters}.}
\label{tab:reproducibility}
\footnotesize
\setlength{\tabcolsep}{5pt}
\renewcommand{\arraystretch}{1.15}
\begin{adjustbox}{max width=\linewidth}
\begin{tabular}{@{}ll@{}}
\toprule
\textbf{Setting} & \textbf{Value} \\
\midrule
Framework & PyTorch (\texttt{TorchLayer} interface for trainable circuits). \\
Quantum simulation & PennyLane \texttt{lightning.qubit}, analytic mode (no shots), CPU. \\
Hardware & NVIDIA L40S GPU (classical components); CPU (quantum simulation). \\
Fixed circuit params & $\boldsymbol{\theta}_{\text{fix}} \sim \mathcal{U}([0,2\pi)^m)$, sampled once before training. \\
Optimizer / loss & Adam (per-group learning rates) / cross-entropy. \\
Runs & 5 independent seeds per configuration. \\
Data splits & official partition (BreastMNIST); patient-level partitions (INbreast, BUS-UCLM); see Table~\ref{tab:datasets}. \\
\bottomrule
\end{tabular}
\end{adjustbox}
\end{table}
All models were optimized with Adam~\cite{kingma2014adam} under the cross-entropy loss;
the fusion models employ branch-specific learning rates for the classical backbone, the
quantum circuit, and the classification head. The temperature scalar $\gamma$ of TSHF
was initialized to $1.0$, reducing to plain concatenation at the onset of training, and
updated jointly with the network weights. The deterministic and trainable quantum
paradigms share the same circuit architecture, illustrated in Figure~\ref{fig:circuit},
and differ only in whether the rotation parameters are held fixed or optimized. The
complete set of training hyperparameters is reported in
Table~\ref{tab:hyperparameters}, and the full layer-by-layer specification of the
pipeline, including exact parameter counts for every component, is given in
Table~\ref{tab:architecture}. Every configuration is trained with 5 independent seeds,
and all results are reported as mean $\pm$ standard deviation over these runs.

\begin{table*}[ht!]
\centering
\caption{Training hyperparameters for the standalone baselines and the hybrid fusion strategies. Fusion hyperparameters are shared across all classical backbone and quantum circuit combinations; the quantum circuit hyperparameters (qubits, layers, patch size) follow the corresponding standalone configuration. The learnable temperature scalar $\gamma$ applies only to TSHF. All experiments use $28 \times 28$ input images and 5 independent runs.}
\label{tab:hyperparameters}
\footnotesize
\setlength{\tabcolsep}{5pt}
\renewcommand{\arraystretch}{1.2}
\begin{adjustbox}{max width=\linewidth}
\begin{tabular}{l cccc ccc}
\toprule
 & \multicolumn{4}{c}{\textbf{Standalone baselines}} & \multicolumn{3}{c}{\textbf{Fusion strategies}} \\
\cmidrule(lr){2-5} \cmidrule(lr){6-8}
\textbf{Hyperparameter} & \textbf{SCNN} & \textbf{ResNet-18} & \textbf{Non-trainable} & \textbf{Trainable} & \textbf{SHF} & \textbf{DHF} & \textbf{TSHF} \\
\midrule
Epochs per run          & 20     & 20     & 10      & 10      & 10  & 10  & 10 \\
Batch size              & 16      & 16      & 16      & 16      & 16  & 16  & 16 \\
Learning rate           & $10^{-3}$ & $10^{-3}$ & $3 \times 10^{-3}$ & $10^{-2}$ & -- & -- & -- \\
LR classical branch     & --      & --      & --      & --      & $10^{-4}$ & $10^{-4}$ & $10^{-4}$ \\
LR quantum branch       & --      & --      & --      & --      & $10^{-2}$ & $10^{-2}$ & $10^{-2}$ \\
LR fusion head          & --      & --      & --      & --      & $10^{-3}$ & $10^{-3}$ & $10^{-1}$ \\
LR $\gamma$             & --      & --      & --      & --      & N/A & N/A & $10^{-2}$ \\
Fused feature dim.      & --      & --      & --      & --      & 128 & 128 & 128 \\
Number of qubits        & --      & --      & 4       & 4       & 4   & 4   & 4 \\
Circuit layers          & --      & --      & 1       & 1       & 1 & 1 & 1 \\
Patch size              & 2      & 2      & 2       & 2       & 2   & 2   & 2 \\
Stride              & 2      & 2      & 2       & 2       & 2   & 2   & 2 \\
\bottomrule
\end{tabular}
\end{adjustbox}
\end{table*}

\begin{table*}[t]
\centering
\caption{Architecture of the hybrid feature-fusion pipeline. Shapes are given for a
$1\times28\times28$ grayscale input; parameter counts are exact. Training
hyperparameters are listed in Table~\ref{tab:hyperparameters}.}
\label{tab:architecture}
\scriptsize
\setlength{\tabcolsep}{3pt}
\renewcommand{\arraystretch}{1.1}
\begin{adjustbox}{max width=\linewidth}
\begin{tabular}{@{}llllr@{}}
\toprule
\textbf{Block} & \textbf{Layer / setting} & \textbf{Configuration} & \textbf{Output} & \textbf{Params} \\
\midrule
Input & Preprocessing & grayscale, resize $28\times28$, $\mathrm{Norm.}(\mu{=}0.5,\sigma{=}0.5)$ & $1\times28\times28$ & -- \\
\midrule
\multirow{6}{*}{\shortstack[l]{SCNN\\(from scratch)}}
 & Conv block 1 & $\mathrm{Conv}(1{\to}32,3{\times}3,s{=}2,p{=}1)$+BN+ReLU+MaxPool$(2,2)$ & $32\times7\times7$ & 384 \\
 & Conv block 2 & $\mathrm{Conv}(32{\to}64,3{\times}3,s{=}2,p{=}1)$+BN+ReLU+MaxPool$(2,2)$ & $64\times2\times2$ & 18{,}624 \\
 & Conv block 3 & $\mathrm{Conv}(64{\to}128,3{\times}3,s{=}2,p{=}1)$+BN+ReLU+GAP & $128\times1\times1$ & 74{,}112 \\
 & Feature tap & flatten & 128 & -- \\
 & Baseline head & $\mathrm{Linear}(128{\to}2)$ & 2 & 258 \\
 & \textit{Total} & & & \textbf{93{,}378} \\
\midrule
\multirow{5}{*}{\shortstack[l]{ResNet-18\\(ImageNet init.)}}
 & Stem adaptation & \texttt{conv1} $\to$ $\mathrm{Conv}(1{\to}64,7{\times}7,s{=}2,p{=}3)$, no bias & $64\times14\times14$ & 3{,}136 \\
 & layer1--layer4 & unchanged BasicBlocks $[2,2,2,2]$, widths $64{\to}512$ & $512\times1\times1$ & 11{,}166{,}912 \\
 & Head adaptation & \texttt{fc} $\to$ $\mathrm{Linear}(512{\to}2)$ (was $512{\to}1000$) & 2 & 1{,}026 \\
 & Feature tap & backbone up to GAP, flattened & 512 & -- \\
 & \textit{Total} & & & \textbf{11{,}171{,}266} \\
\midrule
\multirow{4}{*}{\shortstack[l]{Quantum branch\\(shared)}}
 & Patching & non-overlapping $2\times2$ patches, stride 2 & 196 patches & -- \\
 & Qubits & 4 qubits (one per patch pixel) & -- & -- \\
 & Measurement & $\langle Z_j\rangle$, $j=1,\dots,4$ & $4\times14\times14$ & -- \\
 & Post-processing & bilinear resize + ReLU + flatten & 784 & -- \\
\cmidrule(l){2-5}
\multirow{3}{*}{\quad QC-fixed}
 & Encoding & $R_Y(\pi x_j)$ on wire $j$ & -- & -- \\
 & \textit{Ansatz} & \texttt{QuantumCircuit}$(\Theta)$, $\Theta\!\sim\!\mathcal{U}(0,2\pi)$, $(1,4)$, frozen & -- & 0 \\
 & Head & $\mathrm{Linear}(784{\to}2)$ & 2 & 1{,}570 \\
\cmidrule(l){2-5}
\multirow{3}{*}{\quad QC-trainable}
 & Encoding & \texttt{AngleEmbedding}$(x,\text{rot}{=}Y)$ on 4 wires & -- & -- \\
 & \textit{Ansatz} & \texttt{QuantumCircuit}$(W)$, $W\in\mathbb{R}^{1\times4}$, \texttt{TorchLayer} & -- & 4 \\
 & Head & $\mathrm{Linear}(784{\to}2)$ & 2 & 1{,}570 \\
\midrule
\multirow{3}{*}{\shortstack[l]{Projection\\heads}}
 & Classical & $\mathrm{Linear}(128{\to}128)$ / $\mathrm{Linear}(512{\to}128)$ & 128 & 16{,}512 / 65{,}664 \\
 & Quantum & $\mathrm{Linear}(784{\to}128)$ & 128 & 100{,}480 \\
 & normalization & per-branch $\ell_2$ projection onto the unit sphere & 128 & 0 \\
\midrule
\multirow{4}{*}{Fusion layer}
 & Temperature & $\gamma=\exp(\log\gamma)$; none ($\gamma{=}1$) / fixed / trainable & scalar & 1 \\
 & Concatenation & $[\,\mathbf{h}_C\,;\,\gamma\,\mathbf{h}_Q\,]$ & 256 & -- \\
 & Activation & ReLU & 256 & -- \\
 & Classifier & $\mathrm{Linear}(256{\to}2)$ & 2 & 514 \\
\midrule
\shortstack[l]{Classification\\head}
 & MLP & $\mathrm{Lin.}(d{\to}128)$--ReLU--Drop$(0.5)$--$\mathrm{Lin.}(128{\to}2)$ & 2 & $128d+514$ \\
\bottomrule
\end{tabular}
\end{adjustbox}
\end{table*}

The complete implementation, including the per-branch normalization and the
patient-level split files, is publicly available (see Code and Data
Availability).

\section{Results}
\label{s.results}

This section reports the outcome of a systematic evaluation comprising three
fusion strategies (SHF, DHF, TSHF), two classical backbones (ResNet-18 and
SCNN), two quantum branches (trainable and deterministic), and three breast
imaging datasets, for a total of 36 fusion configurations plus 12 standalone
baselines, each trained with five independent seeds. We first state the
evaluation protocol, then report the standalone baselines, the fusion results
per dataset, an ablation on the temperature scalar, a comparison against prior
hybrid models, and the computational cost of the quantum branch.
Metric-specific analysis is deferred to Section~\ref{s.discussion}, where it is
combined with the representation-level evidence. 

\subsection{Evaluation Protocol}
\label{subsec:protocol}

All configurations are trained with five independent seeds and evaluated on the
held-out test split of each dataset; every number reported below is the mean
$\pm$ standard deviation over those five runs, and model selection within each
run is performed on the validation split (Table~\ref{tab:reproducibility}). We
report accuracy, precision, recall, $F_{1}$-score and the area under the ROC curve.
Accuracy and AUC are invariant to the choice of positive class and are therefore
used whenever an observation is stated across datasets; precision, recall and F1
are class-dependent and are consequently interpreted only within each dataset.

The positive class follows the convention of each source. For BreastMNIST we
retain the official MedMNIST labeling, in which the positive class is the
benign/normal category, so that our figures remain directly comparable with
prior work on the same benchmark. For INbreast and BUS-UCLM, the positive class is the malignant category. This is restated at the
start of each dataset subsection. Per-model F1-scores are summarized in
~\ref{app:f1_curves}. Confidence intervals are shown in~\ref{app:ci}.

\subsection{Standalone Baselines}
\label{subsec:baselines}

Table~\ref{tab:standalone_baselines} reports each paradigm operating
independently. Among the classical backbones, ResNet-18 leads on BreastMNIST
(84.74\% vs.\ 82.82\% accuracy) and marginally on INbreast, whereas on BUS-UCLM
the shallow SCNN is both more accurate (66.92\% vs.\ 65.38\%) and markedly more
separable (91.88\% vs.\ 86.62\% AUC), indicating that the deeper backbone does
not pay off on the smallest and most imbalanced training set. Among the quantum
branches, the deterministic circuit is the more stable of the two on
BreastMNIST and INbreast, consistent with the known difficulty of optimizing
variational circuits, while the trainable circuit attains the best accuracy on
BUS-UCLM at the price of a very large spread across seeds ($\pm 15.53$).

\begin{table*}[htbp]
\centering
\caption{Test-set performance of the standalone classical and quantum
architectures, reported as mean $\pm$ standard deviation over five independent
runs. All values in \%. The positive class is the benign category for
BreastMNIST and the malignant category for INbreast and BUS-UCLM. The best
result per dataset and paradigm is highlighted in bold.}
\label{tab:standalone_baselines}
\footnotesize
\setlength{\tabcolsep}{4pt}
\renewcommand{\arraystretch}{1.2}
\begin{adjustbox}{max width=\linewidth}
\begin{tabular}{lll ccccc}
\toprule
\textbf{Dataset} & \textbf{Paradigm} & \textbf{Architecture} & \textbf{Acc.} & \textbf{Prec.} & \textbf{Rec.} & \textbf{F1} & \textbf{AUC} \\
\midrule
\multirow{4}{*}{BreastMNIST}
 & \multirow{2}{*}{Classical}
 & SCNN      & 82.82 $\pm$ 2.70 & 86.27 $\pm$ 2.26 & \textbf{91.05 $\pm$ 3.79} & 88.55 $\pm$ 1.93 & 84.49 $\pm$ 2.23 \\
 & & ResNet-18 & \textbf{84.74 $\pm$ 2.99} & \textbf{88.67 $\pm$ 1.69} & 90.70 $\pm$ 2.82 & \textbf{89.67 $\pm$ 2.10} & \textbf{87.08 $\pm$ 3.07} \\
\cmidrule{2-8}
 & \multirow{2}{*}{Quantum}
 & Non-trainable & \textbf{78.21 $\pm$ 1.98} & \textbf{82.69 $\pm$ 4.80} & 89.47 $\pm$ 4.92 & \textbf{85.73 $\pm$ 0.54} & \textbf{79.42 $\pm$ 3.13} \\
 & & Trainable     & 77.82 $\pm$ 2.06 & 81.18 $\pm$ 3.33 & \textbf{91.05 $\pm$ 4.45} & 85.71 $\pm$ 1.26 & 78.32 $\pm$ 1.67 \\
\midrule
\multirow{4}{*}{INbreast}
 & \multirow{2}{*}{Classical}
 & SCNN      & 67.92 $\pm$ 2.38 & 65.05 $\pm$ 4.16 & \textbf{73.04 $\pm$ 5.87} & \textbf{68.56 $\pm$ 1.13} & \textbf{78.78 $\pm$ 1.61} \\
 & & ResNet-18 & \textbf{68.33 $\pm$ 2.72} & \textbf{72.91 $\pm$ 13.87} & 66.09 $\pm$ 24.04 & 65.15 $\pm$ 7.93 & 77.30 $\pm$ 4.54 \\
\cmidrule{2-8}
 & \multirow{2}{*}{Quantum}
 & Non-trainable & 69.58 $\pm$ 2.49 & 65.78 $\pm$ 2.71 & \textbf{76.52 $\pm$ 5.83} & \textbf{70.63 $\pm$ 2.85} & \textbf{78.63 $\pm$ 1.32} \\
 & & Trainable     & \textbf{70.00 $\pm$ 2.49} & \textbf{70.00 $\pm$ 6.74} & 68.26 $\pm$ 9.55 & 68.44 $\pm$ 1.97 & 77.15 $\pm$ 2.86 \\
\midrule
\multirow{4}{*}{BUS-UCLM}
 & \multirow{2}{*}{Classical}
 & SCNN      & \textbf{66.92 $\pm$ 9.26} & 93.20 $\pm$ 4.19 & \textbf{58.95 $\pm$ 12.57} & \textbf{71.66 $\pm$ 10.05} & \textbf{91.88 $\pm$ 3.58} \\
 & & ResNet-18 & 65.38 $\pm$ 9.02 & \textbf{95.14 $\pm$ 4.58} & 55.79 $\pm$ 13.72 & 69.36 $\pm$ 11.04 & 86.62 $\pm$ 5.27 \\
\cmidrule{2-8}
 & \multirow{2}{*}{Quantum}
 & Non-trainable & 69.23 $\pm$ 9.02 & \textbf{97.24 $\pm$ 3.79} & 60.00 $\pm$ 14.22 & 73.15 $\pm$ 11.26 & \textbf{86.47 $\pm$ 3.94} \\
 & & Trainable     & \textbf{70.77 $\pm$ 15.53} & 86.67 $\pm$ 8.81 & \textbf{72.63 $\pm$ 25.35} & \textbf{76.64 $\pm$ 15.40} & 81.80 $\pm$ 12.01 \\
\bottomrule
\end{tabular}
\end{adjustbox}
\end{table*}

Two observations matter for what follows. First, the standalone quantum branches
are not competitive with the classical backbones on BreastMNIST---roughly seven
accuracy points below ResNet-18---but are on par with them on INbreast and
BUS-UCLM, where the classical models themselves are weak. Second, the two
paradigms fail differently: on each dataset the two paradigms settle on different precision–recall trade-offs, rather than one being uniformly worse. Different failure modes are a necessary precondition for complementarity, and are what the fusion strategies are designed to exploit.

\subsection{Fusion Strategies}
\label{subsec:fusion_results}

\subsubsection{BreastMNIST}
\label{subsubsec:res_breastmnist}

On BreastMNIST the positive class is the benign category, so precision and
recall below refer to that class. Table~\ref{tab:metrics_breastmnist} shows a
clear ordering with the ResNet-18 backbone: the best configuration of each
strategy improves from the standalone baseline (84.74\%) to DHF (86.28\%) and
then to TSHF (88.33\% $\pm$ 2.00), which also attains the best F1 of the table
(92.09\% $\pm$ 1.25); static fusion matches the baseline exactly (84.74\%)
without exceeding it. The gain in separability is larger than the gain in
accuracy: with the single exception of SHF paired with the deterministic
circuit, every ResNet-based fusion configuration reaches an AUC between 92.0\%
and 92.5\%, against 87.08\% $\pm$ 3.07 for the standalone ResNet-18, an
improvement of roughly five points that is also considerably more stable across
seeds.

\begin{table*}[htbp]
\centering
\caption{Test-set performance of the fusion strategies on \textbf{BreastMNIST},
reported as mean $\pm$ standard deviation over five independent runs. The
positive class is the benign category. Metric values are given in \%, whereas
$\gamma$ is the learned fusion scaling factor of TSHF (dimensionless). The best
result per metric is highlighted in bold.}
\label{tab:metrics_breastmnist}
\footnotesize
\setlength{\tabcolsep}{4pt}
\renewcommand{\arraystretch}{1.1}
\begin{adjustbox}{max width=\linewidth}
\begin{tabular}{lll ccccc c}
\toprule
\textbf{Fusion} & \textbf{Classical} & \textbf{Quantum} & \textbf{Acc.} & \textbf{Prec.} & \textbf{Rec.} & \textbf{F1} & \textbf{AUC} & \boldmath$\gamma$ \\
\midrule
\multirow{4}{*}{Baseline}
 & ResNet-18             & \multirow{2}{*}{N/A} & 84.74 $\pm$ 2.99 & 88.67 $\pm$ 1.69 & 90.70 $\pm$ 2.82 & 89.67 $\pm$ 2.10 & 87.08 $\pm$ 3.07 & N/A \\
 & SCNN                  &                      & 82.82 $\pm$ 2.70 & 86.27 $\pm$ 2.26 & 91.05 $\pm$ 3.79 & 88.55 $\pm$ 1.93 & 84.49 $\pm$ 2.23 & N/A \\
\cmidrule{2-9}
 & \multirow{2}{*}{N/A}  & Trainable            & 77.82 $\pm$ 2.06 & 81.18 $\pm$ 3.33 & 91.05 $\pm$ 4.45 & 85.71 $\pm$ 1.26 & 78.32 $\pm$ 1.67 & N/A \\
 &                       & Non-trainable        & 78.21 $\pm$ 1.98 & 82.69 $\pm$ 4.80 & 89.47 $\pm$ 4.92 & 85.73 $\pm$ 0.54 & 79.42 $\pm$ 3.13 & N/A \\
\midrule
\multirow{4}{*}{SHF}
 & \multirow{2}{*}{ResNet-18} & Trainable       & 84.74 $\pm$ 2.10 & \textbf{94.85 $\pm$ 0.86} & 83.68 $\pm$ 3.49 & 88.88 $\pm$ 1.73 & 92.12 $\pm$ 0.18 & N/A \\
 &                            & Non-trainable   & 78.46 $\pm$ 2.89 & 83.55 $\pm$ 4.32 & 88.60 $\pm$ 8.95 & 85.61 $\pm$ 2.70 & 81.14 $\pm$ 1.24 & N/A \\
\cmidrule{2-9}
 & \multirow{2}{*}{SCNN}      & Trainable       & 80.77 $\pm$ 1.28 & 85.53 $\pm$ 1.63 & 88.77 $\pm$ 2.94 & 87.08 $\pm$ 1.01 & 83.56 $\pm$ 1.19 & N/A \\
 &                            & Non-trainable   & 80.77 $\pm$ 1.50 & 86.37 $\pm$ 1.56 & 87.54 $\pm$ 2.73 & 86.92 $\pm$ 1.14 & 83.07 $\pm$ 0.66 & N/A \\
\midrule
\multirow{4}{*}{DHF}
 & \multirow{2}{*}{ResNet-18} & Trainable       & 84.87 $\pm$ 1.79 & 92.96 $\pm$ 3.56 & 86.14 $\pm$ 6.31 & 89.21 $\pm$ 1.73 & \textbf{92.48 $\pm$ 0.52} & N/A \\
 &                            & Non-trainable   & 86.28 $\pm$ 2.20 & 93.83 $\pm$ 1.40 & 87.02 $\pm$ 4.36 & 90.22 $\pm$ 1.84 & 92.39 $\pm$ 0.24 & N/A \\
\cmidrule{2-9}
 & \multirow{2}{*}{SCNN}      & Trainable       & 80.00 $\pm$ 3.52 & 84.55 $\pm$ 2.72 & 89.30 $\pm$ 9.27 & 86.52 $\pm$ 3.54 & 82.45 $\pm$ 0.55 & N/A \\
 &                            & Non-trainable   & 81.54 $\pm$ 1.39 & 85.88 $\pm$ 1.24 & 89.47 $\pm$ 2.32 & 87.62 $\pm$ 1.03 & 83.46 $\pm$ 0.86 & N/A \\
\midrule
\multirow{4}{*}{TSHF}
 & \multirow{2}{*}{ResNet-18} & Trainable       & 86.15 $\pm$ 1.48 & 94.12 $\pm$ 0.99 & 86.49 $\pm$ 3.01 & 90.11 $\pm$ 1.26 & 92.18 $\pm$ 0.22 & 1.280 $\pm$ 0.331 \\
 &                            & Non-trainable   & \textbf{88.33 $\pm$ 2.00} & 91.56 $\pm$ 3.69 & 92.81 $\pm$ 3.19 & \textbf{92.09 $\pm$ 1.25} & 92.09 $\pm$ 0.44 & 1.025 $\pm$ 0.039 \\
\cmidrule{2-9}
 & \multirow{2}{*}{SCNN}      & Trainable       & 81.54 $\pm$ 0.84 & 83.30 $\pm$ 2.29 & \textbf{93.68 $\pm$ 4.04} & 88.10 $\pm$ 0.80 & 83.71 $\pm$ 0.64 & 0.930 $\pm$ 0.053 \\
 &                            & Non-trainable   & 81.92 $\pm$ 0.95 & 84.15 $\pm$ 1.31 & 92.81 $\pm$ 2.94 & 88.23 $\pm$ 0.81 & 83.29 $\pm$ 0.84 & 1.019 $\pm$ 0.057 \\
\bottomrule
\end{tabular}
\end{adjustbox}
\end{table*}

The advantage does not transfer to the shallow backbone. All SCNN-based
configurations fall between 80.0\% and 81.9\% accuracy, and none improves on the
standalone SCNN in AUC. Static fusion is also the most sensitive to the choice of
quantum branch: SHF with ResNet-18 reaches 84.74\% with the trainable circuit but
only 78.46\% with the deterministic one, the widest gap between paired
configurations in the table. Once the branches are allowed to co-adapt (DHF) or are explicitly calibrated
(TSHF), that gap narrows to 1.4 and 2.2 points, respectively, with the same
backbone.

\begin{figure}[htbp]
    \centering
    \subfloat[SCNN]{%
        \includegraphics[width=0.43\linewidth]{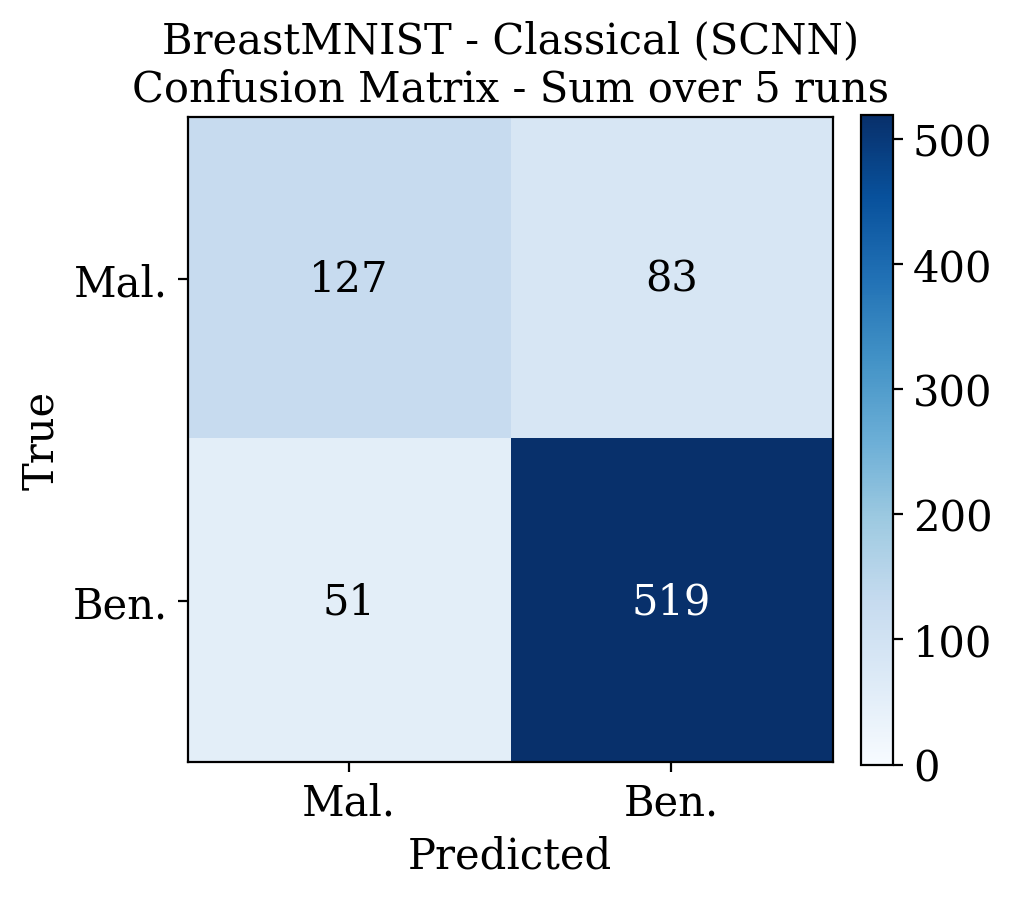}}
    \subfloat[ResNet-18]{%
        \includegraphics[width=0.43\linewidth]{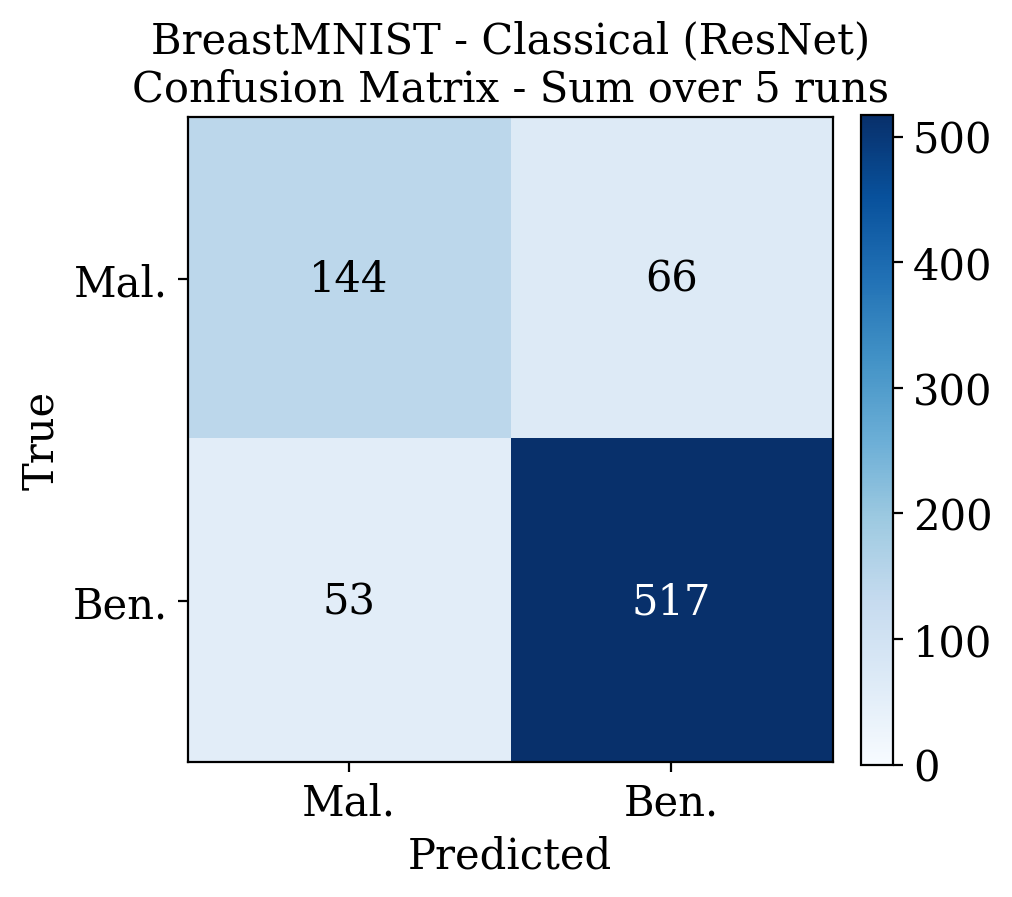}}

    \vspace{0.1em}
    
    \subfloat[Quantum, non-trainable]{%
        \includegraphics[width=0.43\linewidth]{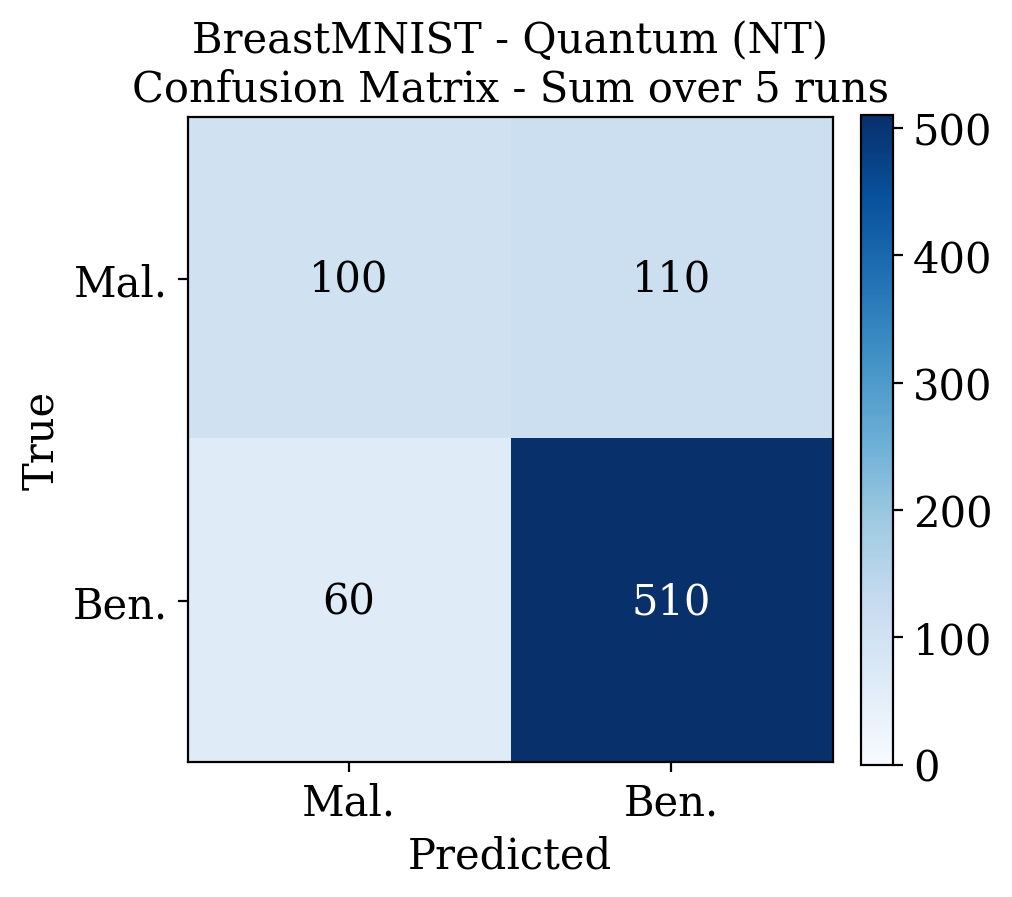}}
    \subfloat[Quantum, trainable]{%
        \includegraphics[width=0.43\linewidth]{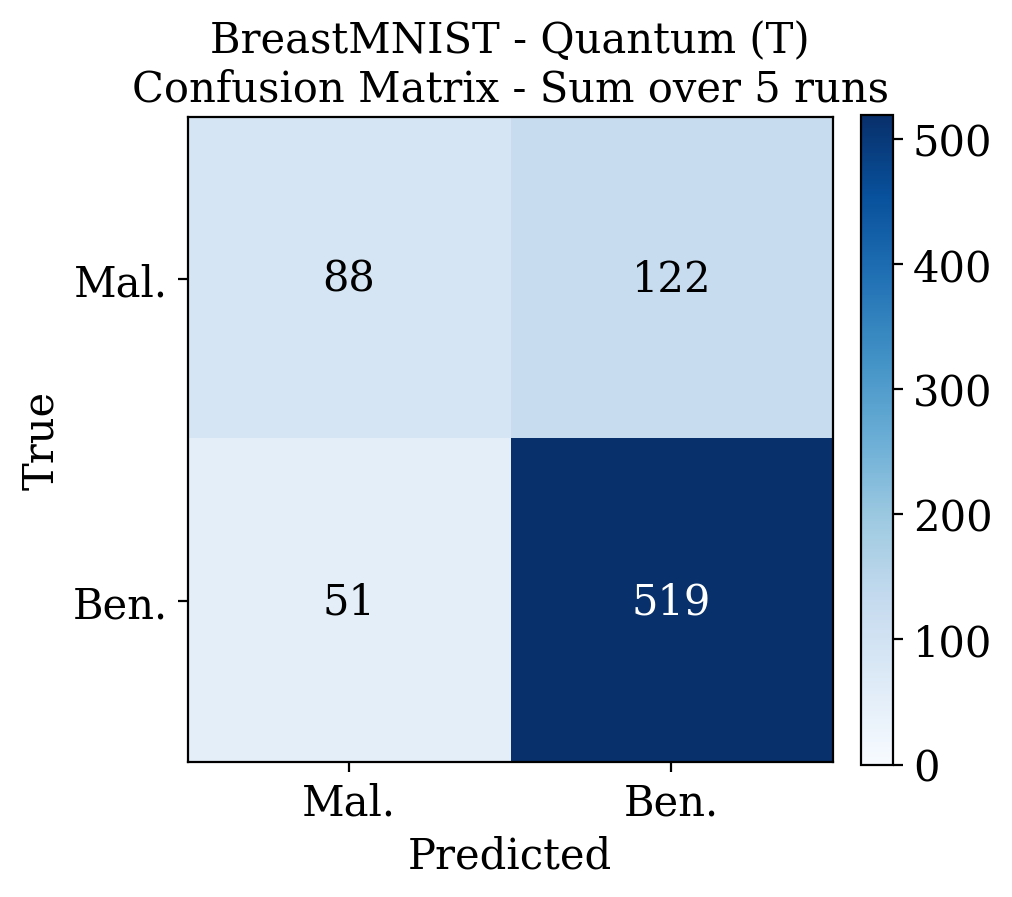}}
        
    \vspace{0.1em}

    \subfloat[TSHF (ResNet-18 + non-trainable)]{%
        \includegraphics[width=0.43\linewidth]{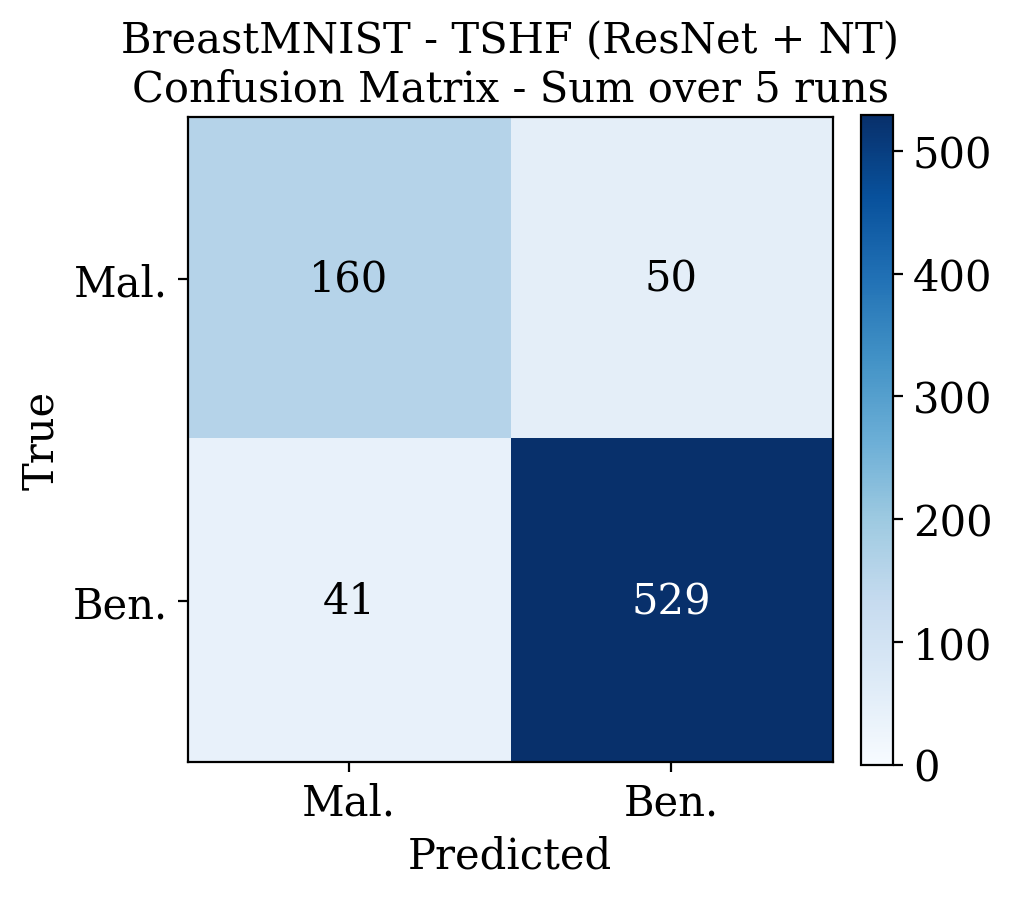}}

    \caption{Confusion matrices on the BreastMNIST test set, summed over five
    independent runs.}
    \label{fig:cm_breastmnist}
\end{figure}

Figure~\ref{fig:cm_breastmnist} shows where the improvement comes from. The best
fusion configuration reduces the number of malignant cases classified as benign
from 66 (ResNet-18) and 110 (deterministic quantum) to 47, and also reduces the
number of benign cases classified as malignant, from 53 and 60 to 41. The gain is
therefore not obtained by shifting the operating point toward one class: both
error types decrease, with the larger reduction on the harder, minority class.

\subsubsection{INbreast}
\label{subsubsec:res_inbreast}

On INbreast the positive class is the malignant category. The dataset is
approximately balanced (47.9\% malignant in the test split) and small, and
Table~\ref{tab:metrics_inbreast} shows a much flatter landscape than
BreastMNIST: the sixteen configurations span roughly five accuracy points, and
several of them overlap within one standard deviation. The best result is
obtained by SHF with ResNet-18 and the deterministic circuit
(70.42\% $\pm$ 3.73 accuracy, 72.96\% $\pm$ 1.29 F1, 80.28\% $\pm$ 0.46 AUC),
the only configuration to improve on all four baselines simultaneously, although
by a modest margin: $+0.42$ accuracy points over the best standalone quantum
branch and $+2.09$ over ResNet-18.

\begin{table*}[htbp]
\centering
\caption{Test-set performance of the fusion strategies on \textbf{INbreast},
reported as mean $\pm$ standard deviation over five independent runs. The
positive class is the malignant category. Metric values are given in \%, whereas
$\gamma$ is the learned fusion scaling factor of TSHF (dimensionless). The best
result per metric is highlighted in bold.}
\label{tab:metrics_inbreast}
\footnotesize
\setlength{\tabcolsep}{4pt}
\renewcommand{\arraystretch}{1.1}
\begin{adjustbox}{max width=\linewidth}
\begin{tabular}{lll ccccc c}
\toprule
\textbf{Fusion} & \textbf{Classical} & \textbf{Quantum} & \textbf{Acc.} & \textbf{Prec.} & \textbf{Rec.} & \textbf{F1} & \textbf{AUC} & \boldmath$\gamma$ \\
\midrule
\multirow{4}{*}{Baseline}
 & ResNet-18             & \multirow{2}{*}{N/A} & 68.33 $\pm$ 2.72 & \textbf{72.91 $\pm$ 13.87} & 66.09 $\pm$ 24.04 & 65.15 $\pm$ 7.93 & 77.30 $\pm$ 4.54 & N/A \\
 & SCNN                  &                      & 67.92 $\pm$ 2.38 & 65.05 $\pm$ 4.16 & 73.04 $\pm$ 5.87 & 68.56 $\pm$ 1.13 & 78.78 $\pm$ 1.61 & N/A \\
\cmidrule{2-9}
 & \multirow{2}{*}{N/A}  & Trainable            & 70.00 $\pm$ 2.49 & 70.00 $\pm$ 6.74 & 68.26 $\pm$ 9.55 & 68.44 $\pm$ 1.97 & 77.15 $\pm$ 2.86 & N/A \\
 &                       & Non-trainable        & 69.58 $\pm$ 2.49 & 65.78 $\pm$ 2.71 & 76.52 $\pm$ 5.83 & 70.63 $\pm$ 2.85 & 78.63 $\pm$ 1.32 & N/A \\
\midrule
\multirow{4}{*}{SHF}
 & \multirow{2}{*}{ResNet-18} & Trainable       & 66.87 $\pm$ 4.32 & 64.83 $\pm$ 8.87 & 77.39 $\pm$ 17.36 & 68.73 $\pm$ 3.57 & 79.62 $\pm$ 0.09 & N/A \\
 &                            & Non-trainable   & \textbf{70.42 $\pm$ 3.73} & 65.74 $\pm$ 5.92 & 83.04 $\pm$ 7.11 & \textbf{72.96 $\pm$ 1.29} & \textbf{80.28 $\pm$ 0.46} & N/A \\
\cmidrule{2-9}
 & \multirow{2}{*}{SCNN}      & Trainable       & 68.33 $\pm$ 0.93 & 68.16 $\pm$ 5.62 & 66.96 $\pm$ 13.08 & 66.55 $\pm$ 3.76 & 76.83 $\pm$ 2.20 & N/A \\
 &                            & Non-trainable   & 68.75 $\pm$ 1.47 & 69.57 $\pm$ 4.82 & 63.48 $\pm$ 9.40 & 65.81 $\pm$ 3.63 & 77.70 $\pm$ 0.98 & N/A \\
\midrule
\multirow{4}{*}{DHF}
 & \multirow{2}{*}{ResNet-18} & Trainable       & 67.08 $\pm$ 2.16 & 62.18 $\pm$ 1.65 & 80.00 $\pm$ 5.41 & 69.91 $\pm$ 2.56 & 74.67 $\pm$ 1.26 & N/A \\
 &                            & Non-trainable   & 69.79 $\pm$ 1.95 & 66.51 $\pm$ 2.95 & 75.65 $\pm$ 11.75 & 70.30 $\pm$ 4.09 & 76.25 $\pm$ 2.04 & N/A \\
\cmidrule{2-9}
 & \multirow{2}{*}{SCNN}      & Trainable       & 68.33 $\pm$ 2.51 & 68.47 $\pm$ 4.86 & 65.22 $\pm$ 14.90 & 65.76 $\pm$ 5.99 & 76.55 $\pm$ 2.48 & N/A \\
 &                            & Non-trainable   & 68.54 $\pm$ 2.80 & 69.98 $\pm$ 5.71 & 61.74 $\pm$ 9.91 & 64.99 $\pm$ 4.79 & 77.12 $\pm$ 1.50 & N/A \\
\midrule
\multirow{4}{*}{TSHF}
 & \multirow{2}{*}{ResNet-18} & Trainable       & 67.92 $\pm$ 1.55 & 64.35 $\pm$ 3.62 & 76.52 $\pm$ 11.44 & 69.32 $\pm$ 3.12 & 76.53 $\pm$ 1.25 & 1.080 $\pm$ 0.169 \\
 &                            & Non-trainable   & 65.42 $\pm$ 3.71 & 59.79 $\pm$ 3.35 & \textbf{86.96 $\pm$ 4.61} & 70.72 $\pm$ 1.71 & 78.12 $\pm$ 1.22 & 1.095 $\pm$ 0.150 \\
\cmidrule{2-9}
 & \multirow{2}{*}{SCNN}      & Trainable       & 67.71 $\pm$ 6.46 & 66.53 $\pm$ 8.10 & 70.43 $\pm$ 6.98 & 67.80 $\pm$ 3.07 & 75.26 $\pm$ 2.54 & 0.977 $\pm$ 0.053 \\
 &                            & Non-trainable   & 68.13 $\pm$ 3.50 & 64.65 $\pm$ 4.18 & 75.22 $\pm$ 5.24 & 69.36 $\pm$ 2.46 & 75.70 $\pm$ 2.76 & 1.045 $\pm$ 0.084 \\
\bottomrule
\end{tabular}
\end{adjustbox}
\end{table*}

\begin{figure}[htbp]
    \centering
    \subfloat[SCNN]{%
        \includegraphics[width=0.43\linewidth]{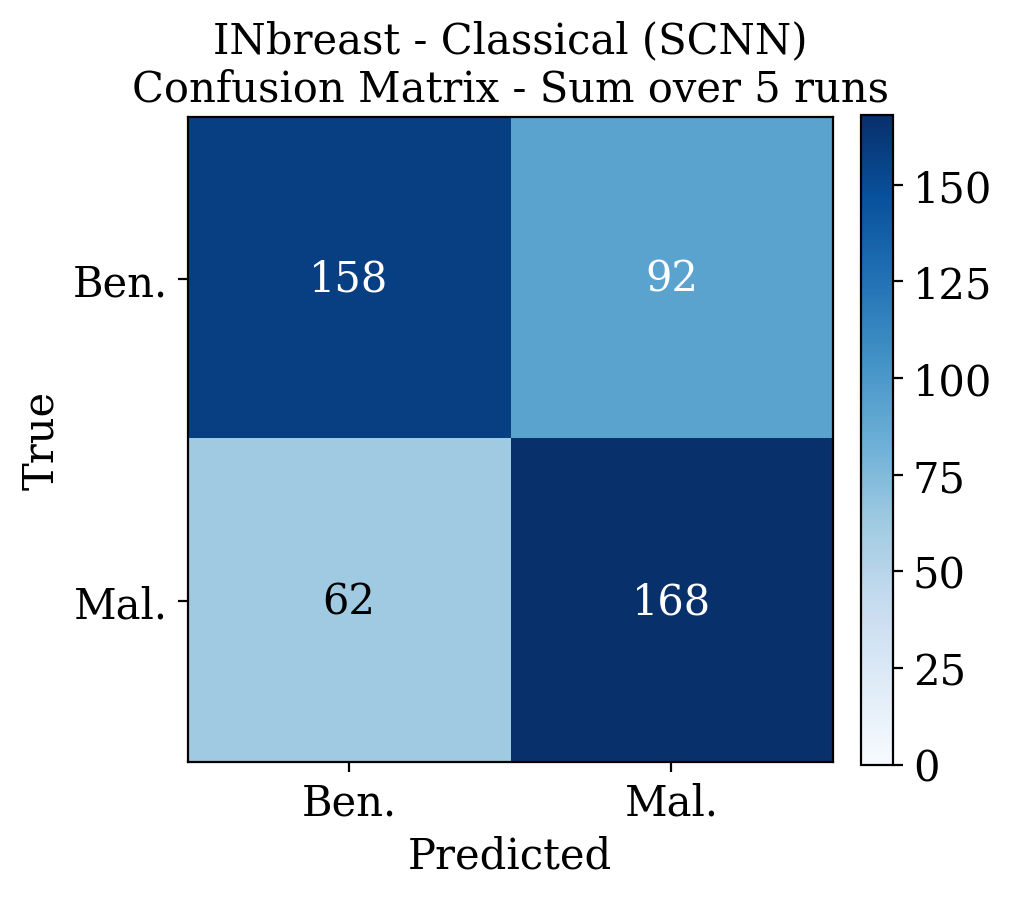}}
    \subfloat[ResNet-18]{%
        \includegraphics[width=0.43\linewidth]{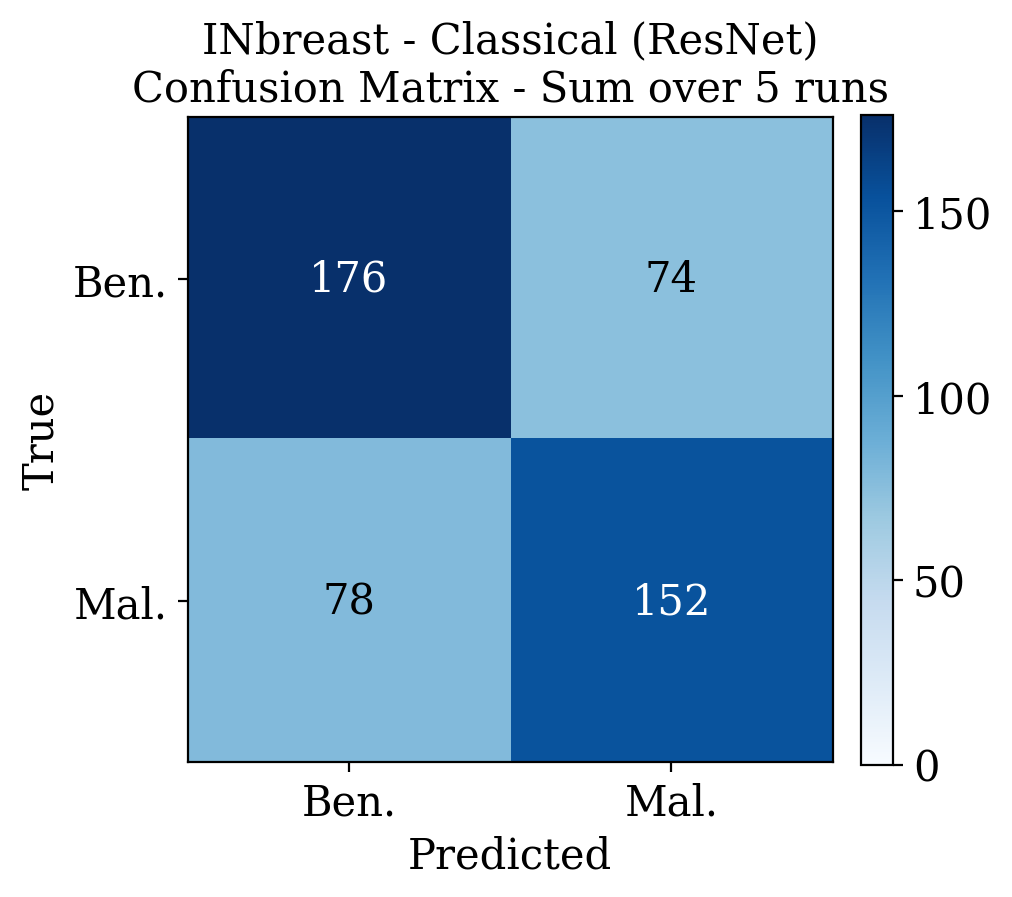}}
        
    \vspace{0.1em}
    
    \subfloat[Quantum, non-trainable]{%
        \includegraphics[width=0.43\linewidth]{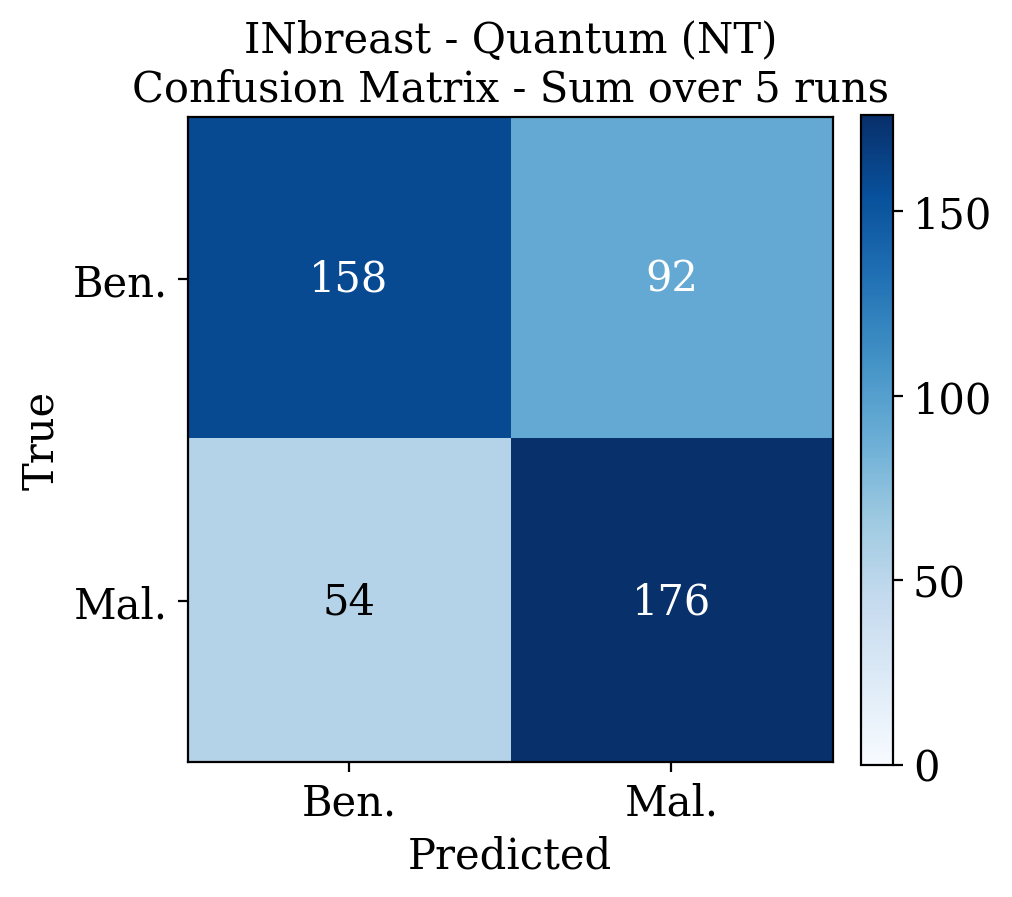}}
    \subfloat[Quantum, trainable]{%
        \includegraphics[width=0.43\linewidth]{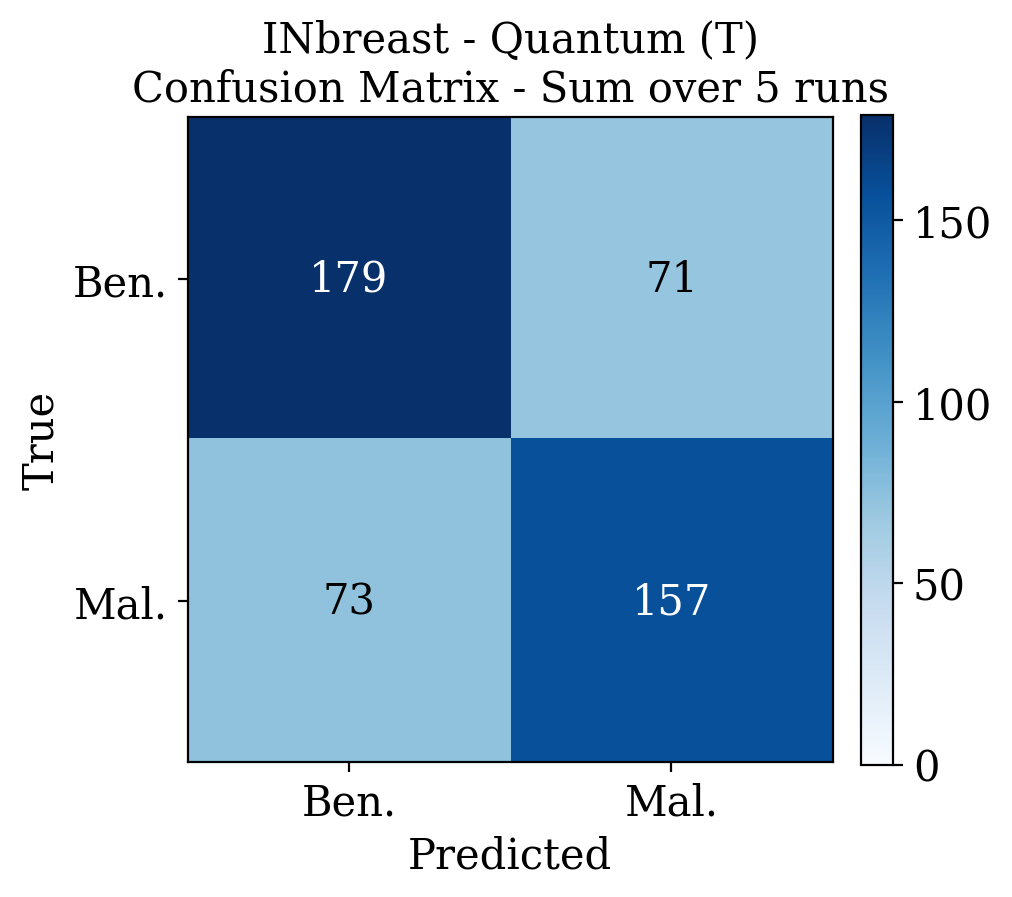}}
        
    \vspace{0.1em}
    
    \subfloat[SHF (ResNet-18 + non-trainable)]{%
        \includegraphics[width=0.43\linewidth]{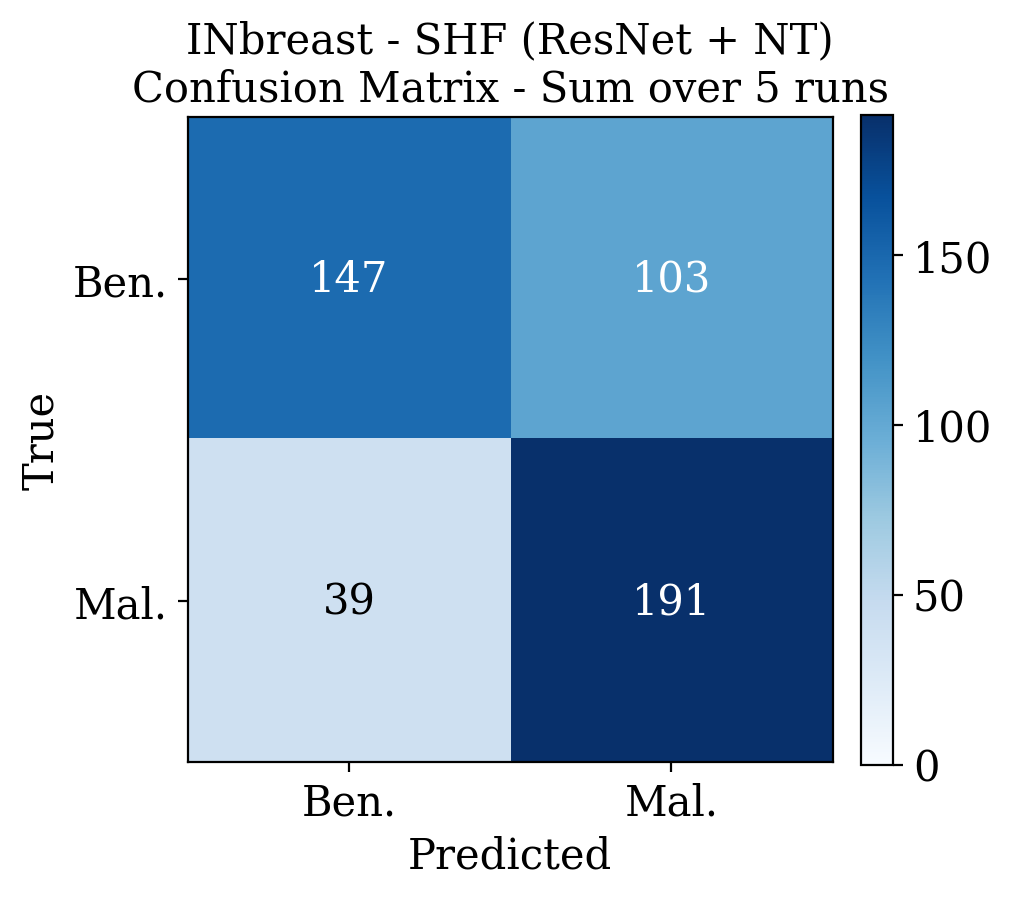}}

    \caption{Confusion matrices on the INbreast test set, summed over five
    independent runs.}
    \label{fig:cm_inbreast}
\end{figure}

Unlike BreastMNIST, the progression does not continue beyond static fusion. DHF
is at best equal to SHF and, with ResNet-18 and the trainable circuit, degrades
to 74.67\% AUC, below every baseline. TSHF likewise does not recover the SHF
result: its ResNet-18 configurations reach 76.5--78.1\% AUC against 79.6--80.3\%
for SHF. The confusion matrices in Figure~\ref{fig:cm_inbreast} indicate a
different mechanism from BreastMNIST: the best fusion model reduces missed
malignant cases from 54--78 in the baselines to 39, but raises false alarms on
the benign class from 71--92 to 103. Here fusion trades specificity for
sensitivity rather than improving both.

\subsubsection{BUS-UCLM}
\label{subsubsec:res_busuclm}

On BUS-UCLM the positive class is the malignant category, which is the majority
of the test split (73.1\%). This is the smallest dataset in the study, with only
26 test images, and every standard deviation in
Table~\ref{tab:metrics_busuclm} should be read with that in mind: a single
misclassified image shifts accuracy by 3.8 points.

\begin{table*}[htbp]
\centering
\caption{Test-set performance of the fusion strategies on \textbf{BUS-UCLM},
reported as mean $\pm$ standard deviation over five independent runs. The
positive class is the malignant category. Metric values are given in \%, whereas
$\gamma$ is the learned fusion scaling factor of TSHF (dimensionless). The best
result per metric is highlighted in bold.}
\label{tab:metrics_busuclm}
\footnotesize
\setlength{\tabcolsep}{4pt}
\renewcommand{\arraystretch}{1.1}
\begin{adjustbox}{max width=\linewidth}
\begin{tabular}{lll ccccc c}
\toprule
\textbf{Fusion} & \textbf{Classical} & \textbf{Quantum} & \textbf{Acc.} & \textbf{Prec.} & \textbf{Rec.} & \textbf{F1} & \textbf{AUC} & \boldmath$\gamma$ \\
\midrule
\multirow{4}{*}{Baseline}
 & ResNet-18             & \multirow{2}{*}{N/A} & 65.38 $\pm$ 9.02 & 95.14 $\pm$ 4.58 & 55.79 $\pm$ 13.72 & 69.36 $\pm$ 11.04 & 86.62 $\pm$ 5.27 & N/A \\
 & SCNN                  &                      & 66.92 $\pm$ 9.26 & 93.20 $\pm$ 4.19 & 58.95 $\pm$ 12.57 & 71.66 $\pm$ 10.05 & \textbf{91.88 $\pm$ 3.58} & N/A \\
\cmidrule{2-9}
 & \multirow{2}{*}{N/A}  & Trainable            & 70.77 $\pm$ 15.53 & 86.67 $\pm$ 8.81 & 72.63 $\pm$ 25.35 & 76.64 $\pm$ 15.40 & 81.80 $\pm$ 12.01 & N/A \\
 &                       & Non-trainable        & 69.23 $\pm$ 9.02 & 97.24 $\pm$ 3.79 & 60.00 $\pm$ 14.22 & 73.15 $\pm$ 11.26 & 86.47 $\pm$ 3.94 & N/A \\
\midrule
\multirow{4}{*}{SHF}
 & \multirow{2}{*}{ResNet-18} & Trainable       & \textbf{78.46 $\pm$ 8.85} & 89.32 $\pm$ 9.72 & \textbf{82.11 $\pm$ 12.12} & \textbf{84.66 $\pm$ 6.57} & 89.47 $\pm$ 3.28 & N/A \\
 &                            & Non-trainable   & 73.08 $\pm$ 7.69 & 94.17 $\pm$ 3.42 & 67.37 $\pm$ 10.79 & 78.16 $\pm$ 7.38 & 90.38 $\pm$ 2.09 & N/A \\
\cmidrule{2-9}
 & \multirow{2}{*}{SCNN}      & Trainable       & 62.31 $\pm$ 6.88 & 98.00 $\pm$ 4.47 & 49.47 $\pm$ 8.81 & 65.39 $\pm$ 7.81 & 87.52 $\pm$ 2.64 & N/A \\
 &                            & Non-trainable   & 56.15 $\pm$ 9.65 & \textbf{98.33 $\pm$ 3.73} & 41.05 $\pm$ 14.60 & 56.46 $\pm$ 14.79 & 90.23 $\pm$ 4.15 & N/A \\
\midrule
\multirow{4}{*}{DHF}
 & \multirow{2}{*}{ResNet-18} & Trainable       & 58.46 $\pm$ 7.40 & 91.92 $\pm$ 4.88 & 47.37 $\pm$ 10.53 & 61.98 $\pm$ 9.01 & 82.56 $\pm$ 6.43 & N/A \\
 &                            & Non-trainable   & 69.23 $\pm$ 9.02 & 92.05 $\pm$ 1.80 & 63.16 $\pm$ 12.34 & 74.41 $\pm$ 10.00 & 89.17 $\pm$ 3.39 & N/A \\
\cmidrule{2-9}
 & \multirow{2}{*}{SCNN}      & Trainable       & 59.23 $\pm$ 4.39 & 98.00 $\pm$ 4.47 & 45.26 $\pm$ 6.00 & 61.69 $\pm$ 5.64 & 89.77 $\pm$ 2.41 & N/A \\
 &                            & Non-trainable   & 55.38 $\pm$ 5.83 & \textbf{98.33 $\pm$ 3.73} & 40.00 $\pm$ 10.26 & 56.10 $\pm$ 8.69 & 88.57 $\pm$ 3.66 & N/A \\
\midrule
\multirow{4}{*}{TSHF}
 & \multirow{2}{*}{ResNet-18} & Trainable       & 65.38 $\pm$ 13.60 & 92.93 $\pm$ 5.08 & 56.84 $\pm$ 19.48 & 68.98 $\pm$ 16.03 & 83.16 $\pm$ 13.44 & 0.989 $\pm$ 0.045 \\
 &                            & Non-trainable   & 53.08 $\pm$ 14.24 & 95.83 $\pm$ 5.89 & 37.89 $\pm$ 20.52 & 51.13 $\pm$ 24.55 & 81.95 $\pm$ 11.09 & 1.005 $\pm$ 0.024 \\
\cmidrule{2-9}
 & \multirow{2}{*}{SCNN}      & Trainable       & 62.31 $\pm$ 15.24 & 94.53 $\pm$ 5.19 & 51.58 $\pm$ 21.18 & 64.45 $\pm$ 19.75 & 88.57 $\pm$ 4.13 & 0.959 $\pm$ 0.062 \\
 &                            & Non-trainable   & 54.62 $\pm$ 20.24 & 76.67 $\pm$ 43.08 & 40.00 $\pm$ 29.45 & 50.51 $\pm$ 33.42 & 87.67 $\pm$ 4.68 & 1.025 $\pm$ 0.045 \\
\bottomrule
\end{tabular}
\end{adjustbox}
\end{table*}

\begin{figure}[htbp]
    \centering
    \subfloat[SCNN]{%
        \includegraphics[width=0.43\linewidth]{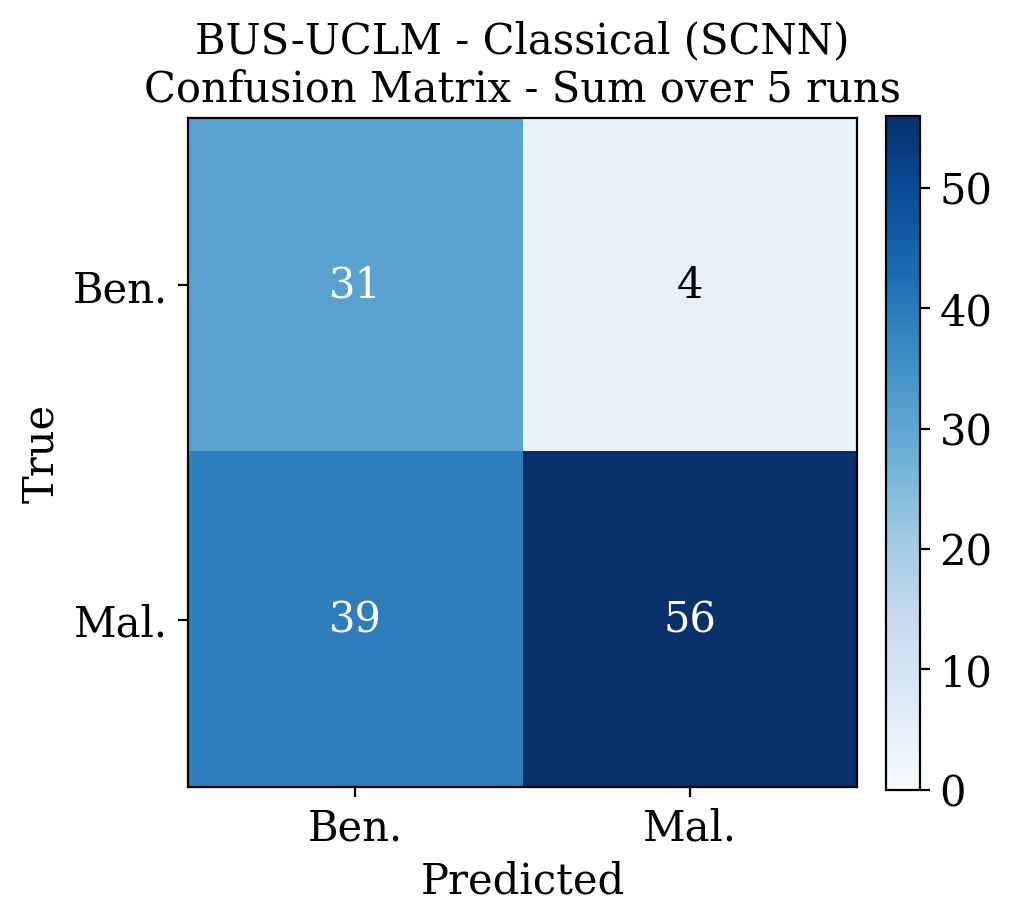}}
    \subfloat[ResNet-18]{%
        \includegraphics[width=0.43\linewidth]{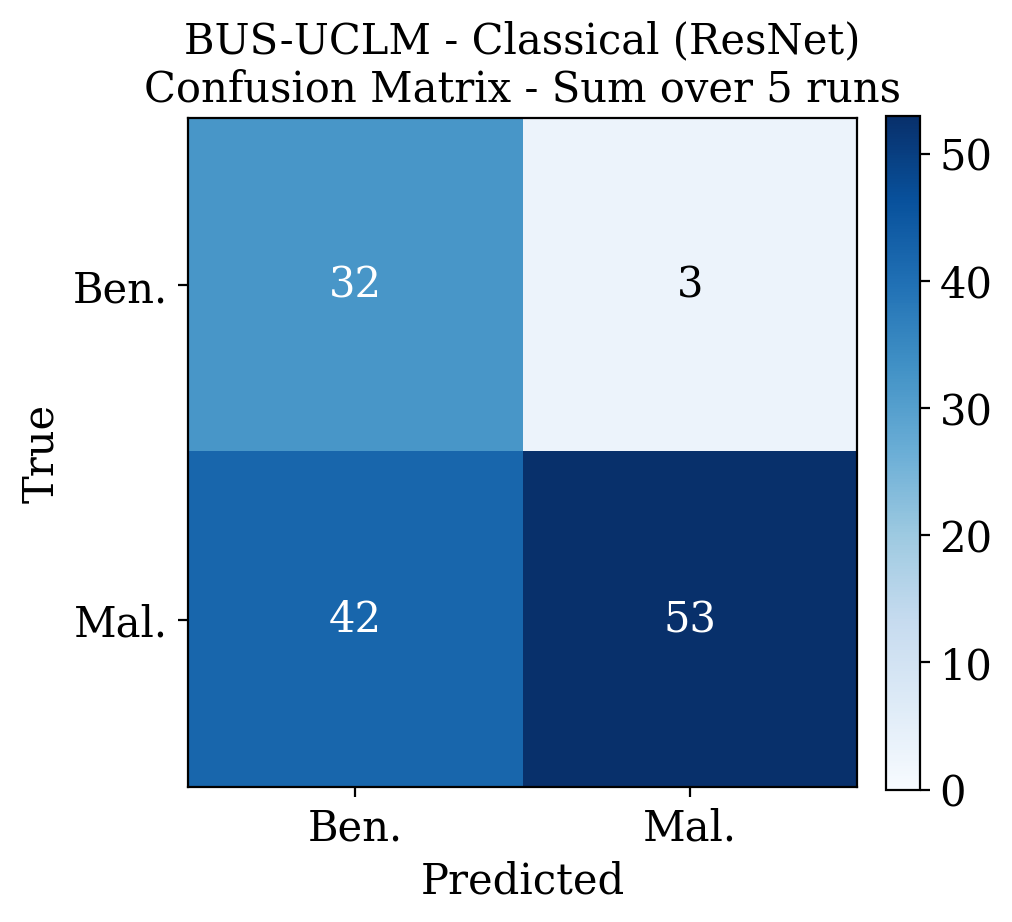}}

    \vspace{0.1em}
    
    \subfloat[Quantum, non-trainable]{%
        \includegraphics[width=0.43\linewidth]{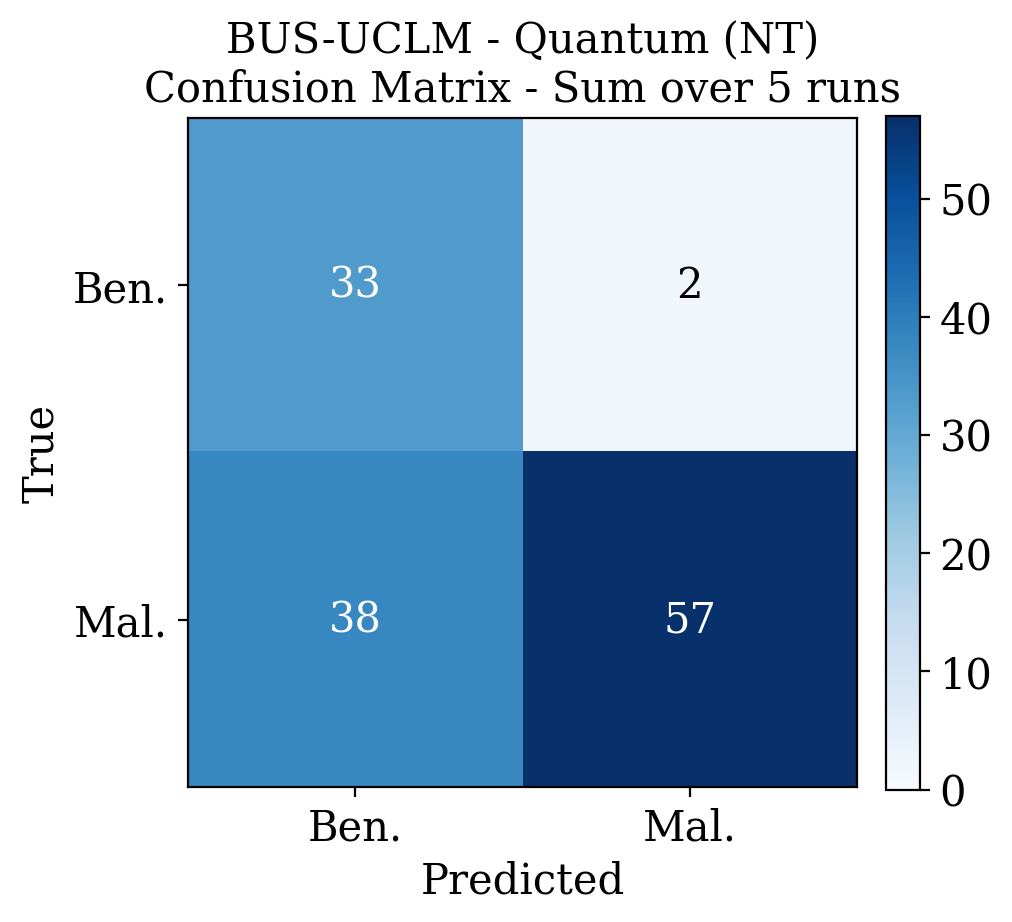}}
    \subfloat[Quantum, trainable]{%
        \includegraphics[width=0.43\linewidth]{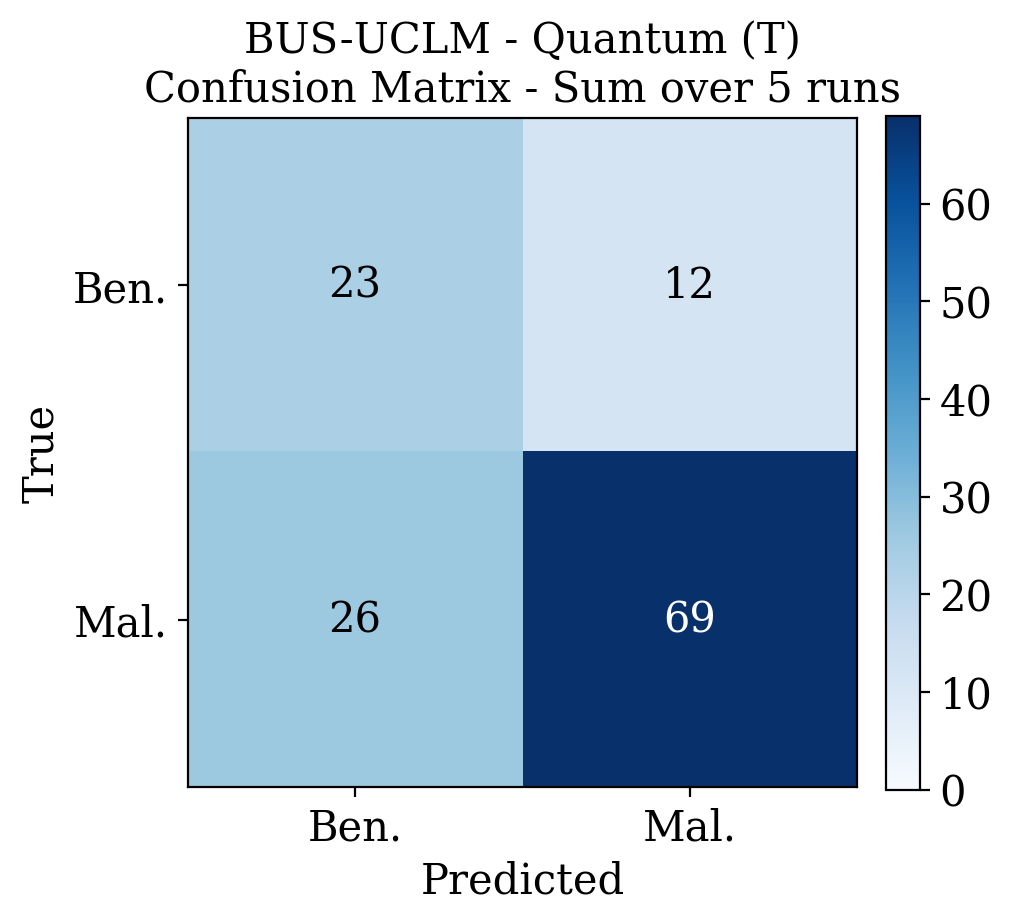}}
    
    \vspace{0.1em}
    
    \subfloat[SHF (ResNet-18 + trainable)]{%
        \includegraphics[width=0.43\linewidth]{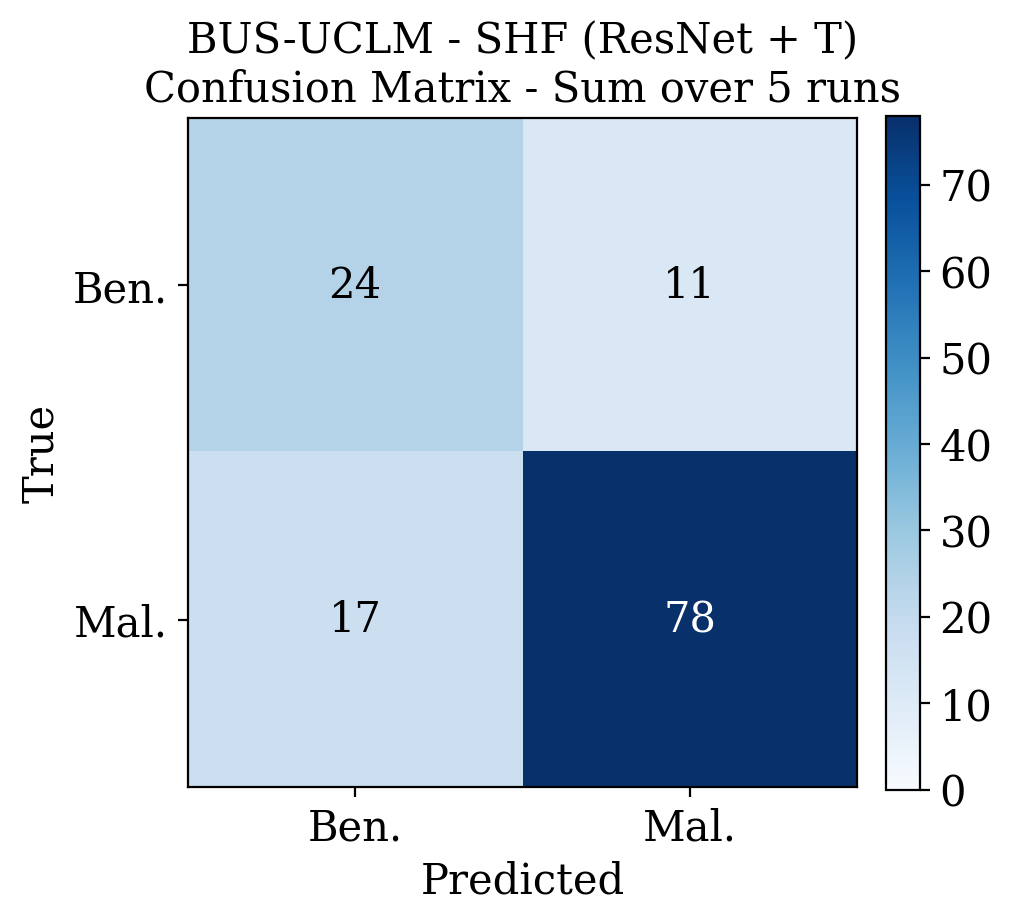}}

    \caption{Confusion matrices on the BUS-UCLM test set, summed over five
    independent runs.}
    \label{fig:cm_busuclm}
\end{figure}

Within those limits, SHF with ResNet-18 and the trainable circuit produces the
largest relative improvement observed anywhere in this work
(78.46\% $\pm$ 8.85 accuracy and 84.66\% $\pm$ 6.57 F1, against 70.77\% and
76.64\% for the best standalone branch). The baselines on this dataset are
markedly conservative---precision between 86.7\% and 97.2\% with recall between
55.8\% and 72.6\%---and the fused model recovers a substantial part of the
missed malignant cases, as Figure~\ref{fig:cm_busuclm} shows.

Again, the more elaborate strategies do not help. DHF and TSHF fall below the
baselines in accuracy for most configurations, and all four TSHF configurations
on this dataset show wide seed-to-seed dispersion, from $\pm 13.60$ to
$\pm 20.24$ accuracy points; the latter, obtained with the SCNN backbone and the
deterministic circuit, is the largest dispersion recorded in the entire study. Separability tells a slightly
different story from accuracy: the highest AUC on this dataset belongs to the
standalone SCNN (91.88\% $\pm$ 3.58), which no fusion configuration matches. The confusion matrices of all sixteen configurations of each dataset are
provided in~\ref{app:confusion}.

\subsection{Effect of the Temperature Scalar}
\label{subsec:ablation}

Table~\ref{tab:ablation_gamma} isolates the contribution of $\gamma$ on
BreastMNIST with the ResNet-18 backbone, comparing no scaling (DHF), five fixed
values spanning three orders of magnitude, per-branch $\ell_1$ normalization, and
the scalar learned end-to-end (TSHF). Three observations follow.

First, the model is tolerant to a fixed $\gamma$ over a wide range but not
indifferent to it. Across the ten fixed-$\gamma$ runs there is no monotonic
trend: $\gamma = 0.1$ is the worst setting in both blocks and $\gamma = 2$ is
among the best in both, but the ordering of the remaining values differs between
them. The spread across fixed values is $1.8$ accuracy points with the
deterministic circuit and $3.7$ with the trainable one. Since the best fixed
value is neither consistent across blocks nor predictable from the others,
obtaining it would require a search.

Second, the learned scalar is not uniformly superior to a simpler normalization.
With the deterministic circuit it is the best setting in accuracy, recall and F1,
and no fixed value or normalization matches it. With the trainable circuit,
however, $\ell_1$ normalization attains the best accuracy (87.69\% $\pm$ 1.46),
precision and F1 of that block, ahead of both the learned scalar and every fixed
value.

\begin{table*}[htbp]
\centering
\caption{Ablation on the temperature scalar $\gamma$, evaluated on the
\textbf{BreastMNIST} test set with the ResNet-18 backbone. All configurations
share the same training protocol and differ only in how the quantum feature
vector is scaled prior to fusion: no scaling (DHF), a fixed scalar, $\ell_1$
normalization, or a scalar learned end-to-end (TSHF). Results are mean $\pm$
standard deviation over five independent runs; metric values in \%, while
$\gamma_{\text{eff}}$ is dimensionless. The best result per metric within each
quantum configuration is highlighted in bold.}
\label{tab:ablation_gamma}
\footnotesize
\setlength{\tabcolsep}{4pt}
\renewcommand{\arraystretch}{1.1}
\begin{adjustbox}{max width=\linewidth}
\begin{tabular}{ll ccccc c}
\toprule
\textbf{Quantum} & \textbf{Scaling} & \textbf{Acc.} & \textbf{Prec.} & \textbf{Rec.} & \textbf{F1} & \textbf{AUC} & \boldmath$\gamma_{\text{eff}}$ \\
\midrule
\multirow{8}{*}{Non-trainable}
 & none (DHF)       & 86.28 $\pm$ 2.20 & \textbf{93.83 $\pm$ 1.40} & 87.02 $\pm$ 4.36 & 90.22 $\pm$ 1.84 & \textbf{92.39 $\pm$ 0.24} & N/A \\
 & $\gamma = 0.001$ & 85.00 $\pm$ 1.79 & 92.83 $\pm$ 3.60 & 86.49 $\pm$ 6.46 & 89.32 $\pm$ 1.75 & 92.14 $\pm$ 0.10 & 0.001 \\
 & $\gamma = 0.01$  & 85.64 $\pm$ 2.42 & 91.25 $\pm$ 5.81 & 89.65 $\pm$ 6.61 & 90.11 $\pm$ 1.62 & 91.93 $\pm$ 0.25 & 0.010 \\
 & $\gamma = 0.1$   & 84.74 $\pm$ 2.05 & 92.65 $\pm$ 2.63 & 86.14 $\pm$ 5.49 & 89.13 $\pm$ 1.85 & 91.92 $\pm$ 0.28 & 0.100 \\
 & $\gamma = 0.5$   & 85.26 $\pm$ 2.60 & 92.77 $\pm$ 2.99 & 86.84 $\pm$ 6.56 & 89.51 $\pm$ 2.29 & 91.83 $\pm$ 1.28 & 0.500 \\
 & $\gamma = 2$     & 86.54 $\pm$ 1.28 & 92.30 $\pm$ 2.36 & 89.12 $\pm$ 3.37 & 90.62 $\pm$ 1.05 & 91.53 $\pm$ 0.93 & 2.000 \\
 & $\ell_1$ norm.   & 87.82 $\pm$ 1.87 & 92.99 $\pm$ 1.93 & 90.18 $\pm$ 2.35 & 91.54 $\pm$ 1.34 & 90.99 $\pm$ 1.12 & N/A \\
 & learnable (TSHF) & \textbf{88.33 $\pm$ 2.00} & 91.56 $\pm$ 3.69 & \textbf{92.81 $\pm$ 3.19} & \textbf{92.09 $\pm$ 1.25} & 92.09 $\pm$ 0.44 & 1.025 $\pm$ 0.039 \\
\midrule
\multirow{8}{*}{Trainable}
 & none (DHF)       & 84.87 $\pm$ 1.79 & 92.96 $\pm$ 3.56 & 86.14 $\pm$ 6.31 & 89.21 $\pm$ 1.73 & \textbf{92.48 $\pm$ 0.52} & N/A \\
 & $\gamma = 0.001$ & 86.54 $\pm$ 1.01 & 89.42 $\pm$ 4.29 & \textbf{92.98 $\pm$ 4.80} & 90.99 $\pm$ 0.51 & 92.13 $\pm$ 0.39 & 0.001 \\
 & $\gamma = 0.01$  & 84.23 $\pm$ 1.48 & 94.32 $\pm$ 1.52 & 83.51 $\pm$ 3.42 & 88.53 $\pm$ 1.31 & 92.29 $\pm$ 0.11 & 0.010 \\
 & $\gamma = 0.1$   & 82.82 $\pm$ 3.05 & 91.36 $\pm$ 7.51 & 85.79 $\pm$ 8.07 & 87.95 $\pm$ 1.79 & 92.08 $\pm$ 0.15 & 0.100 \\
 & $\gamma = 0.5$   & 86.15 $\pm$ 2.67 & 90.01 $\pm$ 4.11 & 91.58 $\pm$ 6.22 & 90.58 $\pm$ 2.10 & 91.92 $\pm$ 0.78 & 0.500 \\
 & $\gamma = 2$     & 86.28 $\pm$ 2.06 & 94.30 $\pm$ 0.85 & 86.49 $\pm$ 3.80 & 90.18 $\pm$ 1.69 & 92.46 $\pm$ 0.27 & 2.000 \\
 & $\ell_1$ norm.   & \textbf{87.69 $\pm$ 1.46} & \textbf{94.93 $\pm$ 1.17} & 87.89 $\pm$ 3.06 & \textbf{91.24 $\pm$ 1.22} & 90.96 $\pm$ 1.32 & N/A \\
 & learnable (TSHF) & 86.15 $\pm$ 1.48 & 94.12 $\pm$ 0.99 & 86.49 $\pm$ 3.01 & 90.11 $\pm$ 1.26 & 92.18 $\pm$ 0.22 & 1.280 $\pm$ 0.331 \\
\bottomrule
\end{tabular}
\end{adjustbox}
\end{table*}

Third, the two scaling schemes differ in what they optimize. The $\ell_1$
variant records the lowest AUC of both blocks (90.99\% and 90.96\%, against
91.5--92.5\% for every other setting), so its threshold-level gains come with a
loss in ranking quality. The learned scalar, by contrast, achieves the best
threshold-level metrics of the deterministic block while remaining within
$0.3$ AUC points of the best value in that block.

The practical reading is therefore not that a particular $\gamma$ is optimal, but
that learning it removes a free choice at negligible cost: TSHF adds exactly one
trainable parameter to the pipeline (Table~\ref{tab:architecture}) and requires
no additional tuning stage, while recovering a result that a fixed value only
reaches when correctly guessed. The competitiveness of $\ell_1$ normalization
also indicates that part of the benefit attributed to the temperature mechanism
is available from simpler per-branch normalization alone, and that the learnable
scalar is most useful when the ranking behavior of the scores matters.

The learned values themselves are reported alongside the metrics in
Tables~\ref{tab:metrics_breastmnist}, \ref{tab:metrics_inbreast}
and~\ref{tab:metrics_busuclm}. Across all twelve TSHF configurations $\gamma$
remains within $[0.930, 1.280]$, that is, within roughly $30\%$ of its
initialization, and in eleven of them within $10\%$. The single configuration that
departs appreciably from unity is ResNet-18 with the trainable circuit on
BreastMNIST ($1.280 \pm 0.331$), which is also the configuration with the largest
seed-to-seed variation in $\gamma$. The learned scalar therefore does not by
itself separate the datasets on which TSHF succeeds from those on which it does
not, and we refrain from reading dataset difficulty off its value.

\subsection{Comparison with Prior Hybrid Models}
\label{subsec:literature}

BreastMNIST is the only dataset in this study with an official fixed partition
used by prior work, which makes it the natural point of comparison.
Table~\ref{tab:across_literature} reports our best configuration per fusion
strategy against three previously published hybrid quantum--classical models
evaluated on the same held-out test set, so that the comparison is free of
split-induced variability.

\begin{table*}[htbp]
\centering
\caption{Comparison of the proposed fusion strategies against prior hybrid
quantum--classical models on \textbf{BreastMNIST}. All works use the official
MedMNIST test split, in which the positive class is the benign category. Our
results are reported as mean $\pm$ standard deviation over five independent runs
and correspond to the best configuration of each fusion strategy; prior works
report single runs. Values in \%.}
\label{tab:across_literature}
\footnotesize
\setlength{\tabcolsep}{5pt}
\renewcommand{\arraystretch}{1.15}
\begin{adjustbox}{max width=\linewidth}
\begin{tabular}{lccccc}
\toprule
\textbf{Reference} & \textbf{Acc.} & \textbf{Prec.} & \textbf{Recall} & \textbf{F1-score} & \textbf{AUC} \\
\midrule
Matondo \textit{et al.}~\cite{matondo2024breast}   & 76.66  & 71.00  & 67.00 & 68.00 & -- \\
Yurtseven \textit{et al.}~\cite{yurtseven2025parallel} & 86.54 & 86.70 & \textbf{96.32} & 91.31 & -- \\
Sobrinho \textit{et al.}~\cite{sobrinho2025hybrid} & 82.10  & 81.70  & 82.00 & 80.30 & 82.60 \\
\midrule
ResNet-18 + Train.\ $|$ SHF (Ours)      & 84.74 $\pm$ 2.10 & \textbf{94.85 $\pm$ 0.86} & 83.68 $\pm$ 3.49 & 88.88 $\pm$ 1.73 & 92.12 $\pm$ 0.18 \\
ResNet-18 + Train.\ $|$ DHF (Ours)      & 84.87 $\pm$ 1.79 & 92.96 $\pm$ 3.56 & 86.14 $\pm$ 6.31 & 89.21 $\pm$ 1.73 & \textbf{92.48 $\pm$ 0.52} \\
ResNet-18 + Non-Train.\ $|$ DHF (Ours)  & 86.28 $\pm$ 2.20 & 93.83 $\pm$ 1.40 & 87.02 $\pm$ 4.36 & 90.22 $\pm$ 1.84 & 92.39 $\pm$ 0.24 \\
ResNet-18 + Non-Train.\ $|$ TSHF (Ours) & \textbf{88.33 $\pm$ 2.00} & 91.56 $\pm$ 3.69 & 92.81 $\pm$ 3.19 & \textbf{92.09 $\pm$ 1.25} & 92.09 $\pm$ 0.44 \\
\bottomrule
\end{tabular}
\end{adjustbox}
\end{table*}

TSHF with ResNet-18 and the deterministic circuit attains the highest accuracy
of the group (88.33\% $\pm$ 2.00), ahead of the parallel multi-circuit fusion of
Yurtseven \textit{et al.}~\cite{yurtseven2025parallel} (86.54\%) and of the
lightweight quanvolution model of Sobrinho \textit{et
al.}~\cite{sobrinho2025hybrid} (82.10\%); its F1-score (92.09\% $\pm$ 1.25) also
exceeds the best previously reported value (91.31\%). DHF with ResNet-18 and the
deterministic circuit reaches 86.28\%, within $0.3$ points of the best prior
result and ahead of the two remaining ones. Two caveats belong with these
numbers: our figures are five-seed averages rather than single best runs,
which makes the comparison conservative on our side, and only one of the prior works reports AUC (82.60\%~\cite{sobrinho2025hybrid}), which all our configurations exceed.

\subsection{Computational Cost}
\label{subsec:cost}

Table~\ref{tab:computational_cost} separates hardware-independent cost from
measured wall-clock time. The quantum branch is extremely parameter-efficient:
the deterministic circuit contributes no trainable parameters at all, and the
variational circuit contributes four, against $11{,}171{,}266$ for ResNet-18. The
fused models are therefore dominated by the classical backbone and the projection
heads. TSHF adds a single learnable scalar on top of DHF, so both configurations
share the same analytical cost; the differences in their measured timings fall
within run-to-run dispersion and should not be attributed to the scaling
mechanism.

\begin{table*}[htbp]
\centering
\caption{Computational cost of the standalone baselines and of all hybrid fusion
configurations (SHF, DHF, and TSHF). Analytical quantities (left) are
hardware-independent; wall-clock measurements (right) are medians over timed
repeats on a 32-thread CPU with batch size 16 and $28 \times 28$ inputs. The
\textbf{Params} column reports the total number of parameters, all of which are
trainable except for SHF, where the number of trainable parameters is given in
parentheses since the feature extractors are kept frozen. The fixed circuit
contributes no trainable parameters and the variational circuit contributes four
(Table~\ref{tab:architecture}). Slowdown is the training step time relative to
the SCNN baseline. Since \texttt{lightning.qubit} simulation runs on CPU, the
quantum branch does not benefit from GPU execution.}
\label{tab:computational_cost}
\footnotesize
\setlength{\tabcolsep}{5pt}
\renewcommand{\arraystretch}{1.15}
\begin{adjustbox}{max width=\linewidth}
\begin{tabular}{@{}l rrr rrr@{}}
\toprule
& \multicolumn{3}{c}{\textbf{Analytical}} & \multicolumn{3}{c}{\textbf{Measured (CPU)}} \\
\cmidrule(lr){2-4} \cmidrule(lr){5-7}
\textbf{Model} & \textbf{Params} & \textbf{MACs/img} & \textbf{Circuits/img} &
\textbf{Infer.\ (ms/img)} & \textbf{Train (ms/step)} & \textbf{Slowdown} \\
\midrule
SCNN             & 93{,}378        & 425{,}344      & 0   & 0.03 & 2.4      & $1\times$ \\
ResNet-18        & 11{,}171{,}266  & 33{,}006{,}848 & 0   & 0.22 & 55       & $23\times$ \\
\midrule
QC-fixed         & 1{,}570         & 1{,}568        & 196 & 118  & 2{,}357  & ${\sim}9.6 \times 10^{2}$ \\
QC-trainable     & 1{,}574         & 1{,}568        & 196 & 143  & 8{,}304  & ${\sim}3.4 \times 10^{3}$ \\
\midrule
SHF + SCNN       & 210{,}888 (117{,}506)       & 542{,}336      & 196 & 176 & 2{,}864  & ${\sim}1.2 \times 10^{3}$ \\
SHF + ResNet-18  & 11{,}337{,}928 (166{,}658)  & 33{,}172{,}224 & 196 & 134 & 2{,}140  & ${\sim}8.7 \times 10^{2}$ \\
\midrule
DHF + SCNN       & 210{,}888       & 542{,}336      & 196 & 149  & 8{,}839  & ${\sim}3.6 \times 10^{3}$ \\
DHF + ResNet-18  & 11{,}337{,}928  & 33{,}172{,}224 & 196 & 169  & 12{,}785 & ${\sim}5.2 \times 10^{3}$ \\
\midrule
TSHF + SCNN      & 210{,}889       & 542{,}336      & 196 & 190  & 12{,}283 & ${\sim}5.0 \times 10^{3}$ \\
TSHF + ResNet-18 & 11{,}337{,}929  & 33{,}172{,}224 & 196 & 180  & 9{,}442  & ${\sim}3.9 \times 10^{3}$ \\
\bottomrule
\end{tabular}
\end{adjustbox}
\end{table*}

That efficiency does not carry over to runtime under classical simulation. Each
image requires 196 circuit evaluations, and the hybrid models are roughly three
orders of magnitude slower per training step than the SCNN baseline (between
${\sim}8.7 \times 10^{2}$ and ${\sim}5.2 \times 10^{3}$ times). SHF is the least expensive hybrid configuration to train, since its feature extractors are kept frozen and only the fusion and classification layers are updated, whereas DHF and TSHF backpropagate through both branches. Because \texttt{lightning.qubit}
simulates the circuit on CPU, this overhead does not shrink with a better GPU:
repeating the benchmark on an NVIDIA L40S did not systematically reduce the
hybrid timings, which varied in both directions within run-to-run dispersion,
while the purely classical baselines sped up by approximately $1.6\times$ (SCNN)
and $10\times$ (ResNet-18). These figures reflect the cost of simulating the
quantum branch and are not a property of the algorithm itself; on quantum
hardware the 196 circuits per image would be evaluated natively and the cost
profile would differ substantially. One practical consequence follows directly
from the deterministic design: because the fixed circuit does not change during
training, its features can be pre-computed once for the whole dataset and
cached, converting a per-epoch cost into a single preprocessing pass.

% =====================================================================
%  DISCUSSION
% =====================================================================
\section{Discussion}
\label{s.discussion}

\subsection{When Does Fusion Help?}
\label{subsec:when_fusion_helps}

Taken together, the three datasets do not support a single ordering of the
fusion strategies. Using accuracy and AUC, the two metrics that are comparable
across datasets, the best configuration per dataset is TSHF on BreastMNIST
(88.33\%), SHF on INbreast (70.42\%) and SHF on BUS-UCLM (78.46\%). The
progression from static to adaptive fusion that holds on BreastMNIST is
therefore not reproduced on the two smaller datasets, where the additional
flexibility of joint co-adaptation and learned calibration does not pay for
itself.

The amount of training data is the factor that most clearly tracks this pattern.
BreastMNIST provides 546 training images and a fixed benchmark split; INbreast
and BUS-UCLM provide 310 and 199. Deciding how much to weight a branch, and
allowing both branches to co-adapt, are themselves estimation problems, and the
two smaller datasets do not appear to supply enough signal to solve them
reliably: on those datasets the strategy that freezes both extractors and trains
only the classifier is the one that generalizes best. The dispersion figures
support this reading, with the TSHF configurations on BUS-UCLM showing the
widest seed-to-seed spread of the entire study ($\pm 13.60$ and $\pm 20.24$
accuracy points).

We note that the learned $\gamma$ values do not by themselves distinguish the
successful from the unsuccessful fusion cases: they remain within $30\%$ of unity in all
twelve TSHF configurations, including the best one on BreastMNIST
($1.025 \pm 0.039$). What the scalar contributes is therefore not a large
re-weighting but the removal of a trivial choice, as the ablation in
Table~\ref{tab:ablation_gamma} shows: performance varies by $1.8$ to $3.7$
accuracy points across fixed values of $\gamma$, depending on the quantum
branch, and learning it recovers the upper end of that range without a search. That ablation also shows that a fixed $\ell_1$
normalization of both branches is a competitive alternative at the default
operating point, at the cost of the lowest AUC of the comparison, which suggests
that the two mechanisms address partly overlapping problems.

The confusion matrices in Figures~\ref{fig:cm_breastmnist}--\ref{fig:cm_busuclm}
show that even when fusion helps, it does not always help in the same way. On
BreastMNIST the fused model improves mainly by reducing errors on both classes, most markedly on the minority malignant class. On INbreast and
BUS-UCLM the improvement comes from recovering malignant cases at the cost of
additional false alarms. The two behaviors have different operational
implications, and we regard the fact that they differ as a limitation on how far
a single summary statement about fusion can be pushed.

\begin{figure*}[htbp]
    \centering
    \subfloat[ResNet-18 backbone\label{fig:roc_resnet}]{%
        \includegraphics[width=0.48\linewidth]{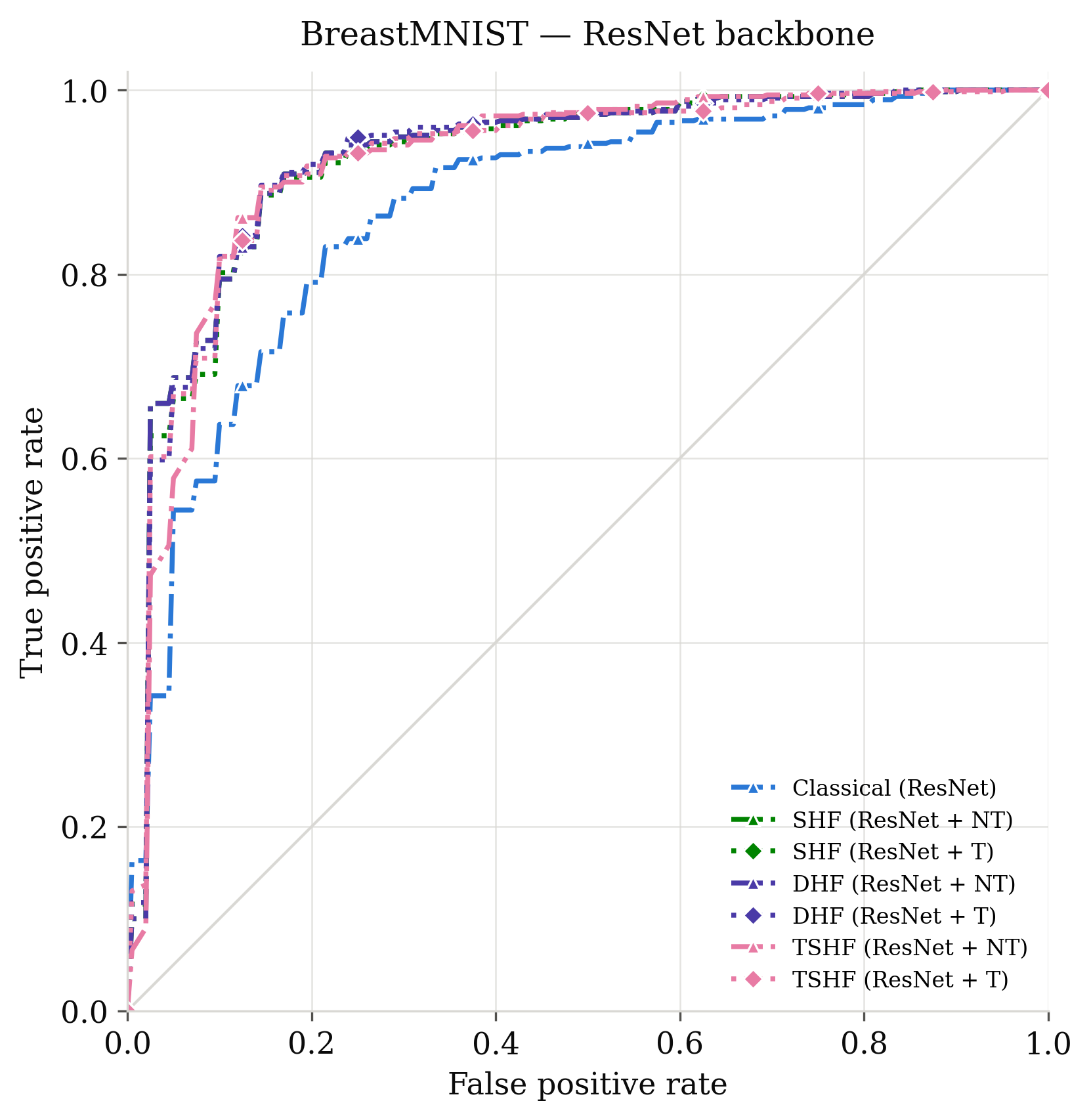}}
    \hfill
    \subfloat[SCNN backbone\label{fig:roc_scnn}]{%
        \includegraphics[width=0.48\linewidth]{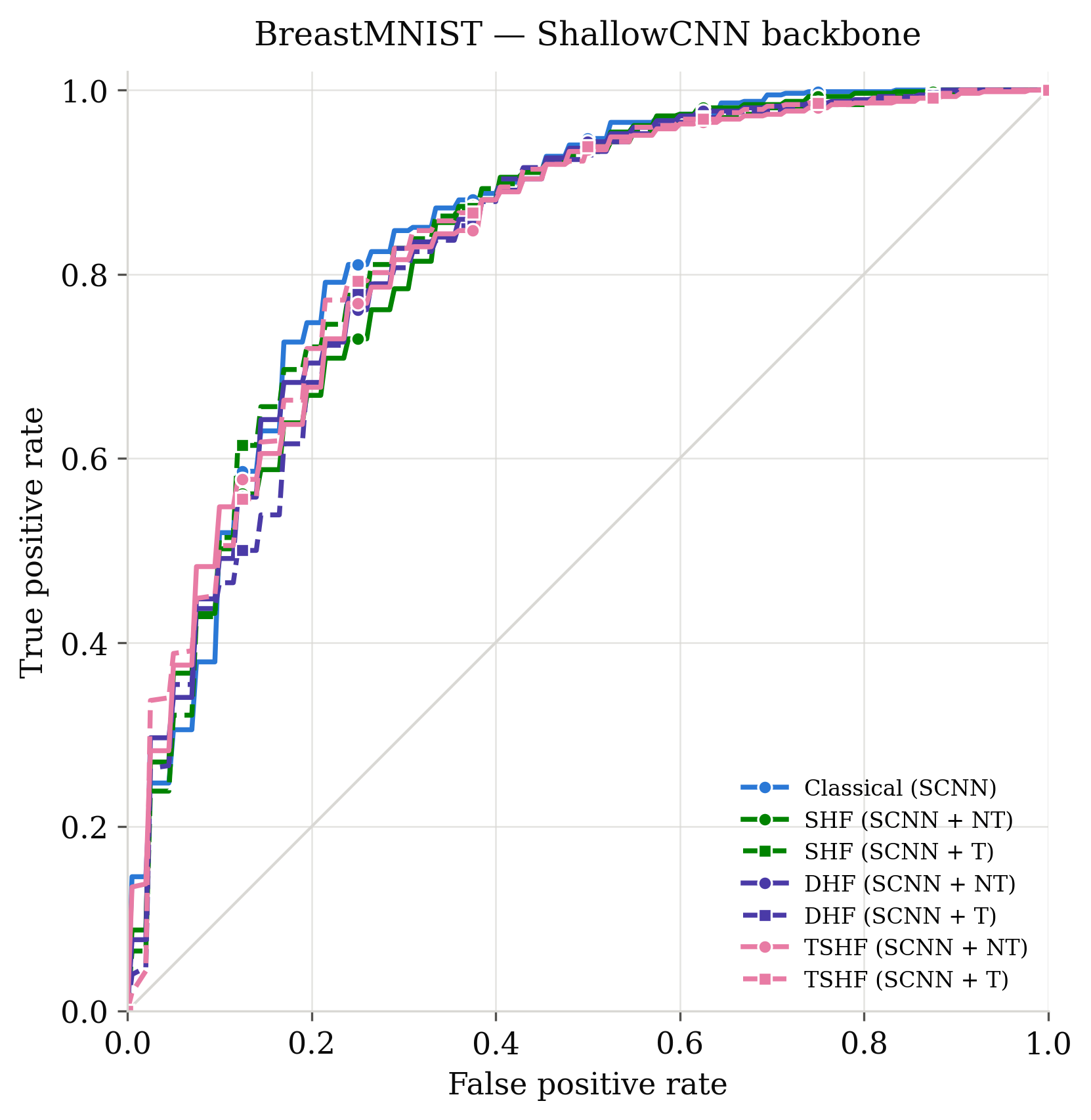}}

    \vspace{0.5em}

    \subfloat[Non-trainable quantum circuit\label{fig:roc_nt}]{%
        \includegraphics[width=0.48\linewidth]{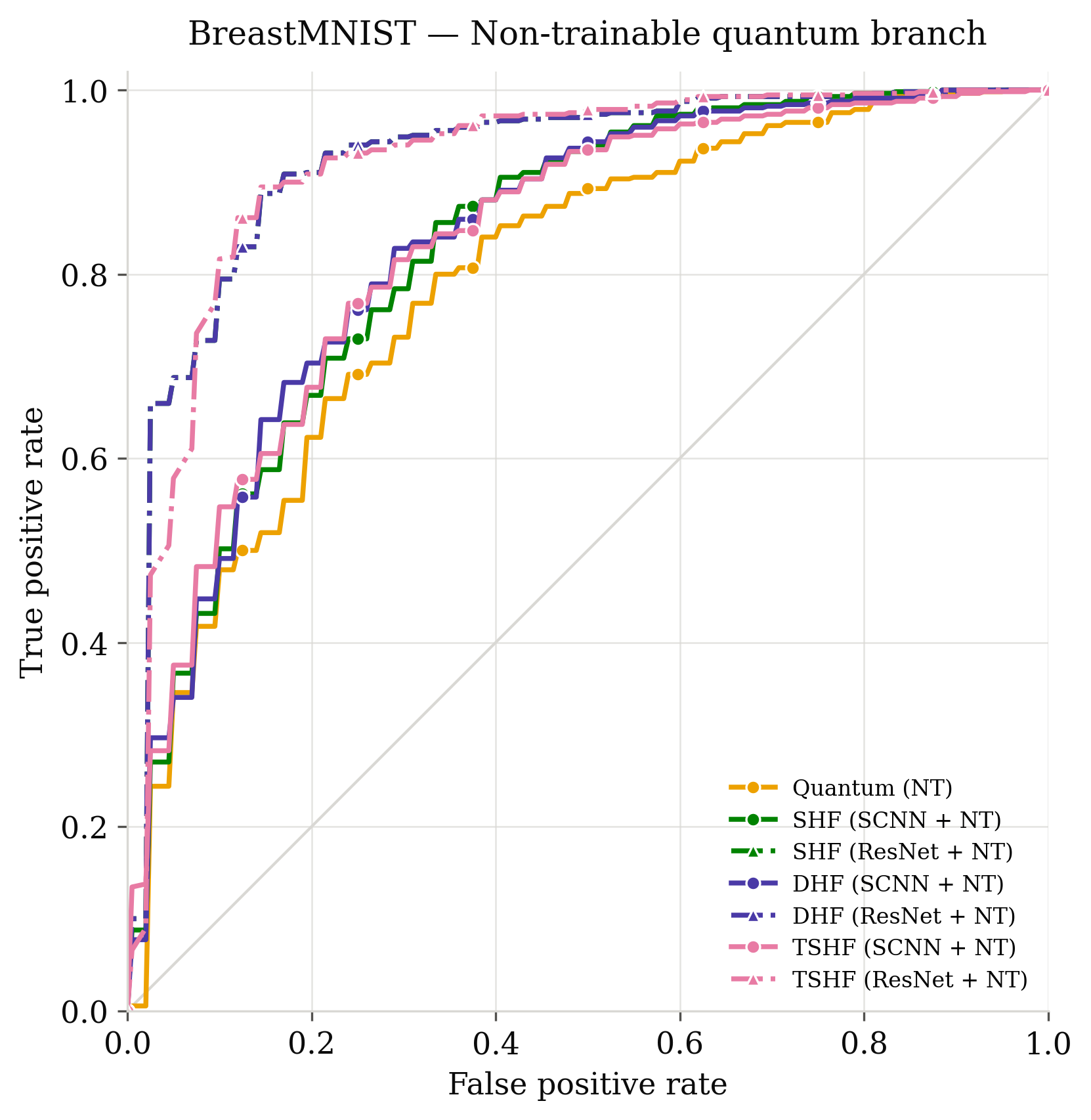}}
    \hfill
    \subfloat[Trainable quantum circuit\label{fig:roc_t}]{%
        \includegraphics[width=0.48\linewidth]{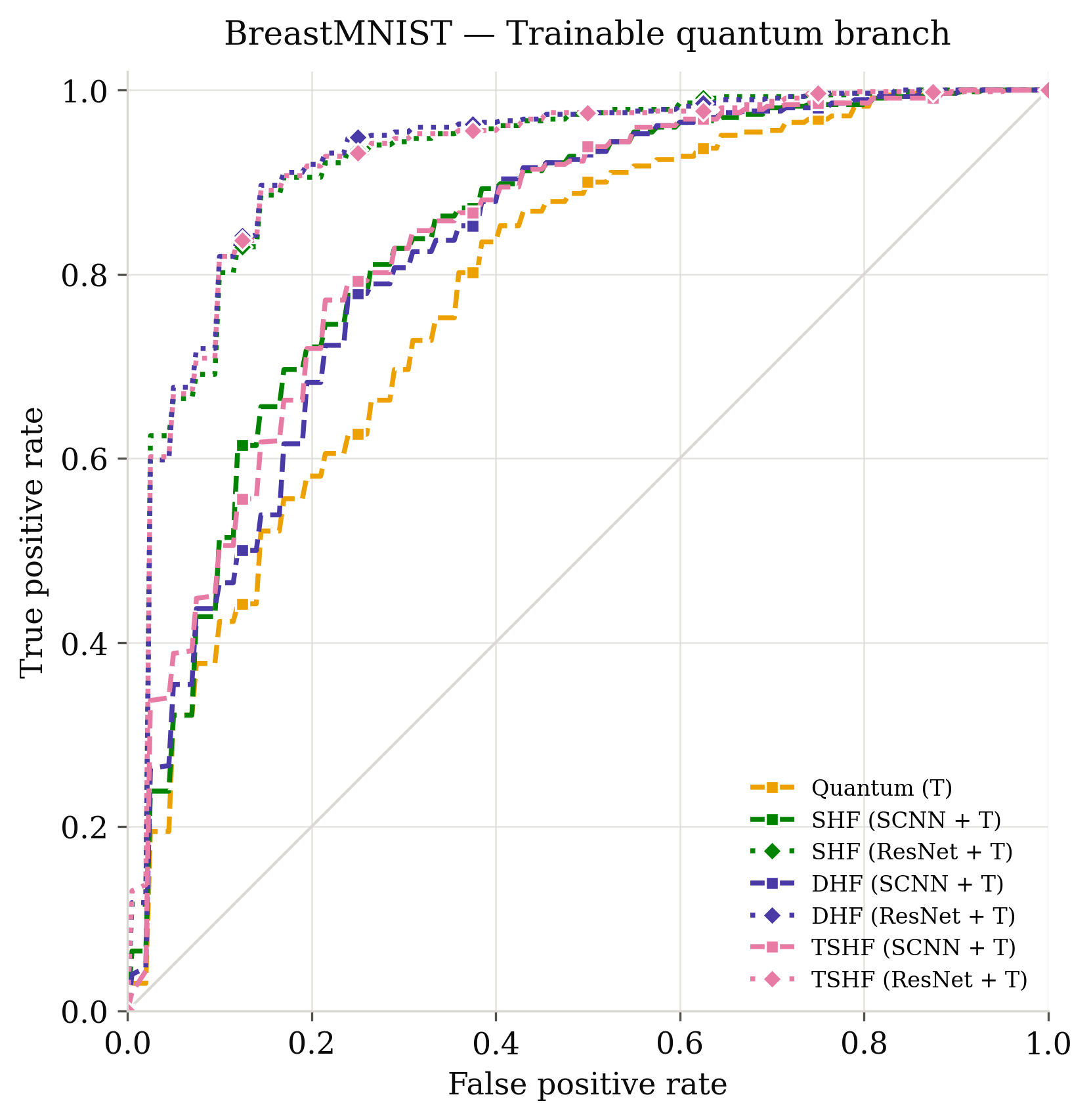}}

    \caption{ROC curves on the BreastMNIST test set.}
    \label{fig:roc_groups}
\end{figure*}

Figure~\ref{fig:roc_groups} presents the same comparison in threshold-free form
for BreastMNIST, grouping the strategies by classical backbone and by quantum
circuit so that each panel isolates one factor. The ResNet-based fusion curves
dominate the standalone classical baseline over essentially the whole
false-positive range, which is the behavior behind the five-point AUC gain
reported in Section~\ref{subsubsec:res_breastmnist}. The corresponding curves for
INbreast, BUS-UCLM and BreastMNIST are provided in~\ref{app:curves}, where the
separation between strategies is visibly smaller.

\subsection{The Role of the Learned Scale}
\label{subsec:branch_magnitudes}

\begin{figure*}[htbp]
    \centering
    \includegraphics[width=\linewidth]{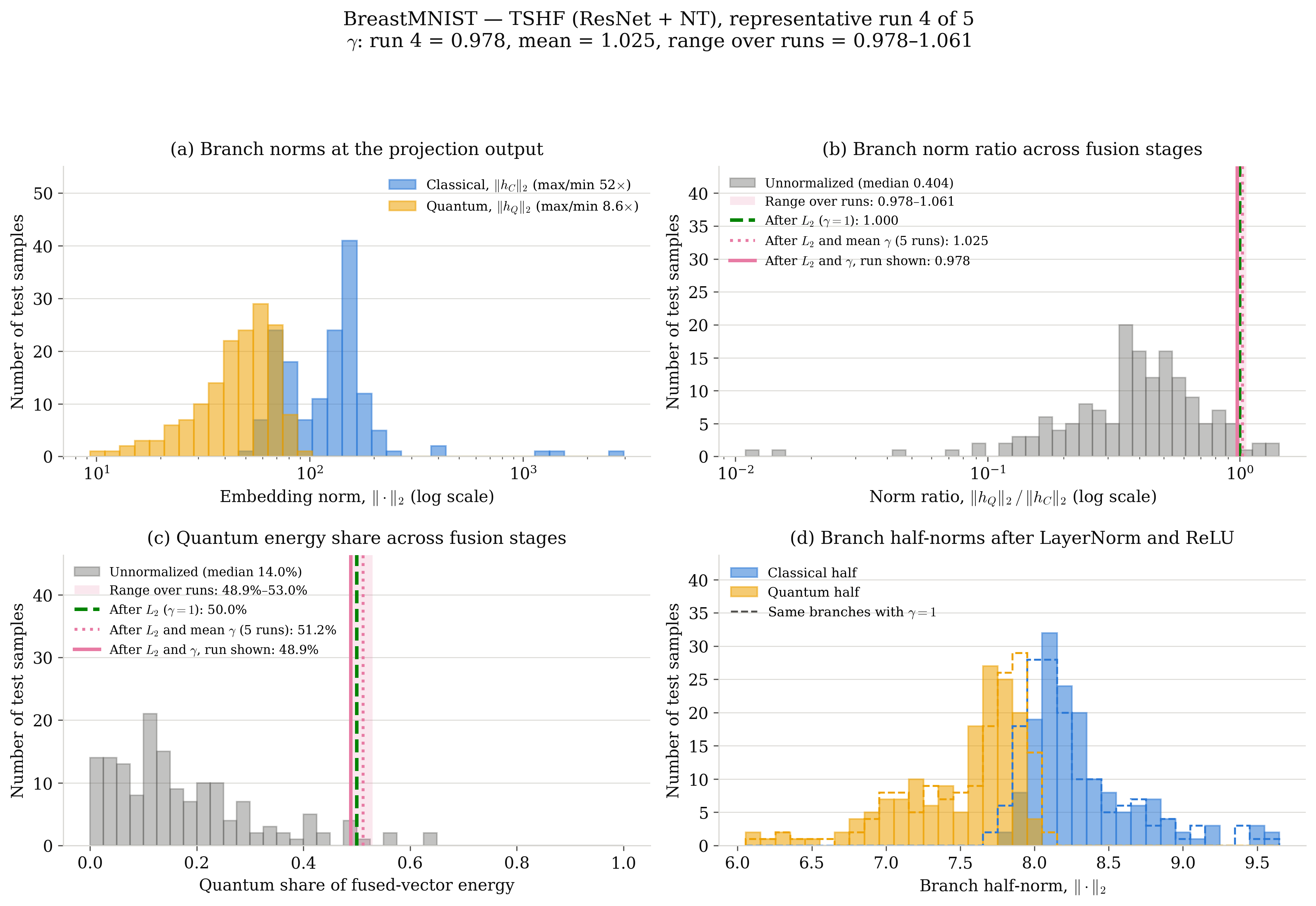}
    \caption{Histograms of branch magnitudes across the TSHF fusion stages
(BreastMNIST, ResNet-18 + non-trainable circuit; the best-performing configuration; representative run 4 of 5).
\textbf{(a)} At the projection output, the quantum norm is systematically smaller
than the classical norm, and the classical norm is more dispersed ($52\times$
across samples against $8.6\times$ for the quantum norm). \textbf{(b, c)} Without
normalization, the branch ratio is input-dependent (median $0.404$) and the
quantum branch receives a median of $14.0\%$ of the fused vector's energy. The
per-branch $\ell_2$ normalization maps the ratio to unity for every sample
($\gamma = 1$, i.e., plain concatenation of the normalized branches at the start
of TSHF training), and $\gamma$ then sets it to a single constant for the whole
test set. Vertical lines show the value learned in the representative run
($\gamma = 0.978$, quantum share $48.9\%$, solid) and the mean over the five runs
($\gamma = 1.025 \pm 0.039$, quantum share $\gamma^2/(1+\gamma^2) = 51.2\%$,
dotted); the shaded band marks the range across runs ($0.978$--$1.061$, i.e.,
$48.9\%$--$53.0\%$). \textbf{(d)} The ratio reaches the classifier: the fusion
\textsc{LayerNorm} carries no affine gain and therefore cannot rescale the
branches independently; dashed outlines show the same branches with $\gamma = 1$.}
    \label{fig:branch_magnitudes}
\end{figure*}

Figure~\ref{fig:branch_magnitudes} makes explicit what $\gamma$ acts upon. At the
projection output the two branches differ both in typical magnitude and in
dispersion: the quantum norm is systematically smaller (median ratio
$\|h_Q\|_2/\|h_C\|_2 = 0.404$), and the classical norm varies by a factor of
$52$ across test samples, against $8.6$ for the quantum norm, whose bounded
circuit expectations concentrate it. As a consequence, the ratio between the
branches is both biased and input-dependent, ranging over approximately
$0.01$--$1.4$ and giving the quantum branch a median of only $14.0\%$ of the
fused vector's energy (between ${\approx}0\%$ and ${\approx}65\%$ depending on
the sample).

The per-branch $\ell_2$ normalization removes both the bias and the per-sample
dependence, mapping the ratio to unity for every sample, and $\gamma$ then sets
it to a single constant: $0.978$ in the run shown, corresponding to a $48.9\%$
quantum share. Because the fusion \textsc{LayerNorm} carries no affine gain, it
cannot rescale the branches independently, so this ratio is what actually reaches
the classifier. The qualitative behavior (biased and input-dependent before
normalization, constant afterwards) holds in all five runs, and the learned
constant remains close to unity, between $0.978$ and $1.061$.

This clarifies what the mechanism does and does not do. Most of the correction
is performed by the per-branch normalization, which removes both the systematic
under-weighting of the quantum branch and the per-sample variability in how much
each branch contributes. The learnable scalar then exposes the remaining balance
as a single degree of freedom; in this configuration, training keeps it close to
an even split, which suggests that, once magnitudes are equalized, neither branch
needs to be up- or down-weighted. That is also why the mechanism is general:
nothing in its formulation is specific to quantum features, and the same scalar
could calibrate any pair of branches whose relative scale is not known a priori.

\subsection{Trainable versus Deterministic Quantum Feature Maps}
\label{subsec:trainable_vs_fixed}

Comparing the two quantum branches under identical conditions is one of the
practically useful outcomes of this study. The deterministic circuit provides the
best configuration on two of the three datasets---TSHF on BreastMNIST and SHF on
INbreast---and on BreastMNIST it is also the more stable of the two under static
fusion. It achieves this with no trainable quantum parameters, which means it is
immune to barren plateaus~\cite{mcclean2018barren} by construction, requires no
gradient estimation through the circuit, and, as noted in
Section~\ref{subsec:cost}, can be pre-computed and cached.

The trainable circuit is not uniformly worse: it provides the best configuration
on BUS-UCLM and the highest precision under SHF on BreastMNIST. But with four
variational parameters in a single layer over four qubits, its expressivity is
severely limited, and the fact that it does not consistently outperform a
randomly initialized fixed circuit should be read as a statement about this
particular \textit{ansatz} rather than about variational circuits in general. Wider or
deeper circuits, with more qubits per patch or several variational layers, would
give the trainable branch a genuinely larger hypothesis space; whether the
resulting gain survives the associated optimization difficulties is an open
question that we leave to future work.

Under present NISQ constraints, however, the deterministic option is the more
deployable of the two, and the fact that it remains competitive is arguably a
more useful finding than a marginal advantage for the variational one would have
been.

\subsection{Latent Space Topology}
\label{subsec:umap_analysis}

\begin{figure}[htbp]
    \centering
    \subfloat[Classical: ResNet-18]{%
        \includegraphics[width=0.48\linewidth]{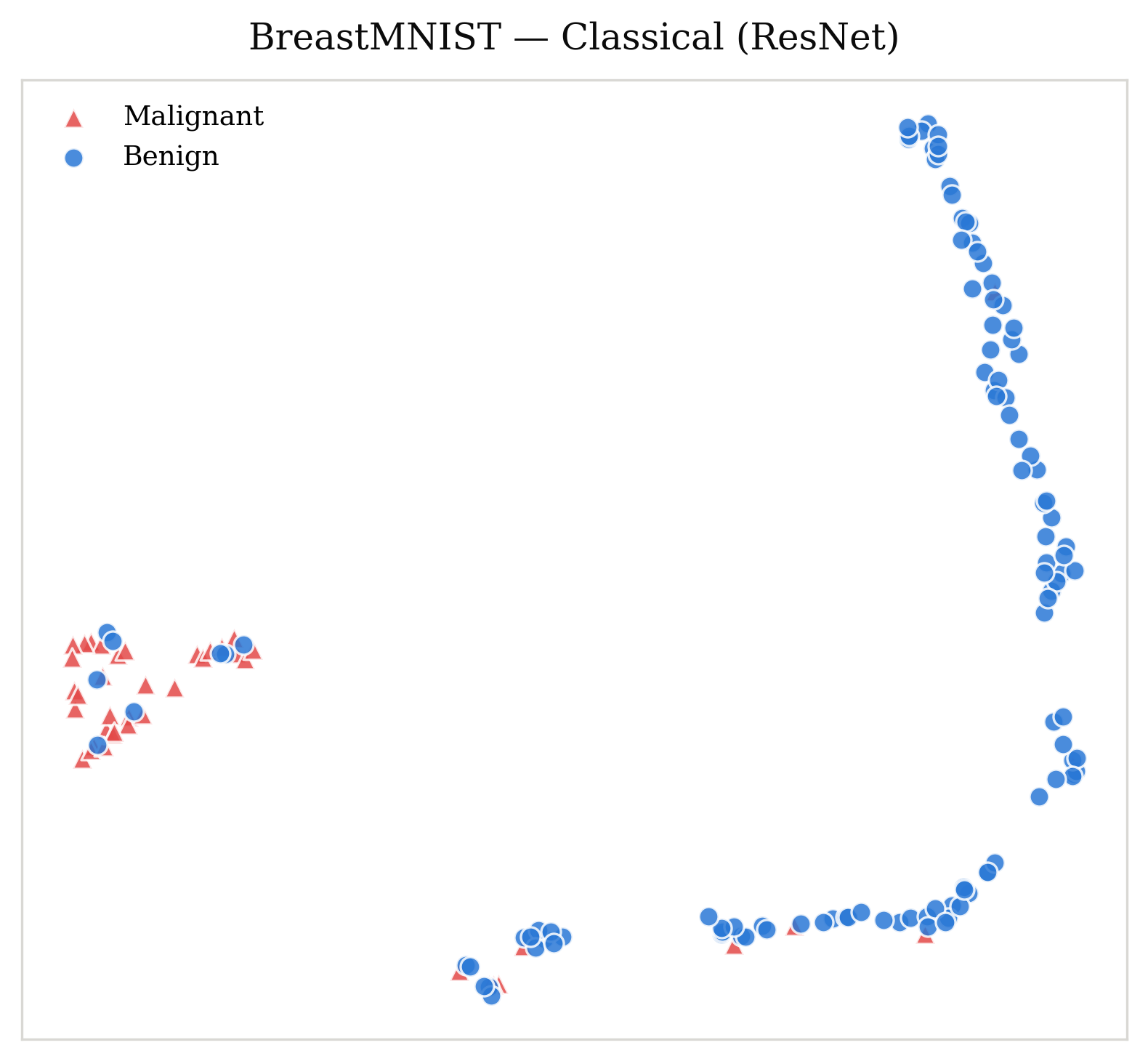}}
    \hfill
    \subfloat[Quantum: non-trainable]{%
        \includegraphics[width=0.48\linewidth]{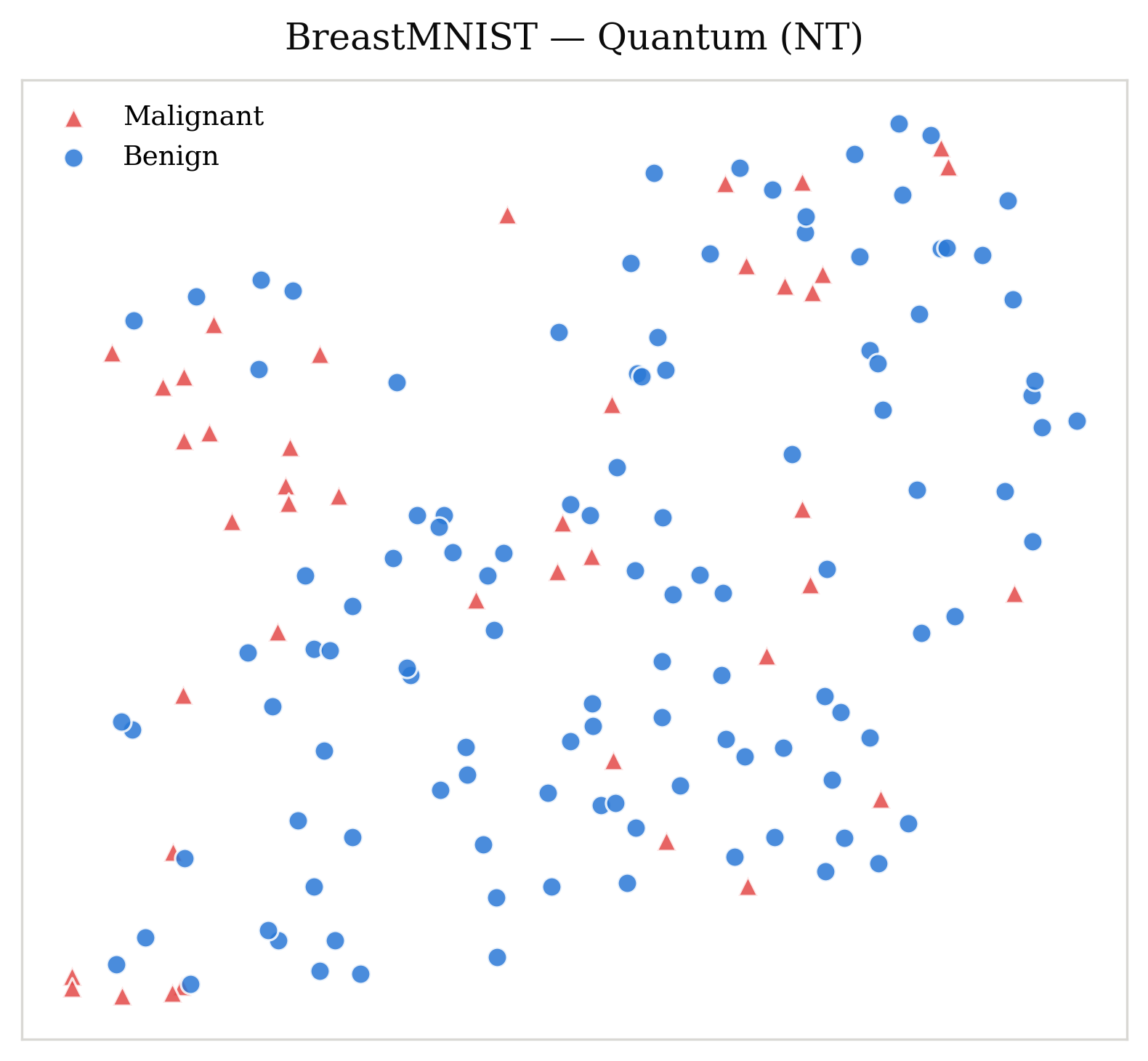}}

    \vspace{0.4em}

    \subfloat[Classical: SCNN]{%
        \includegraphics[width=0.48\linewidth]{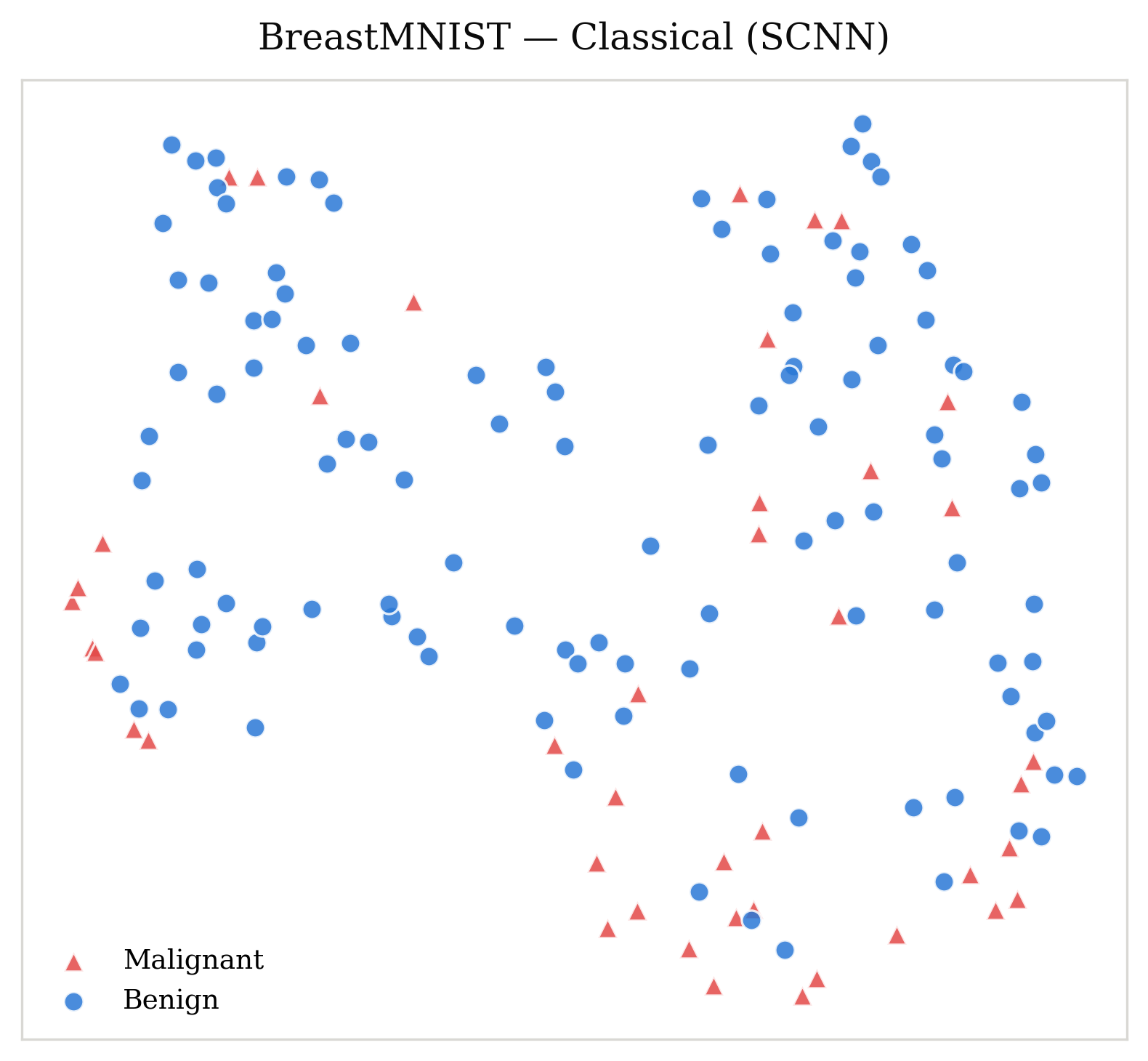}}
    \hfill
    \subfloat[Quantum: trainable]{%
        \includegraphics[width=0.48\linewidth]{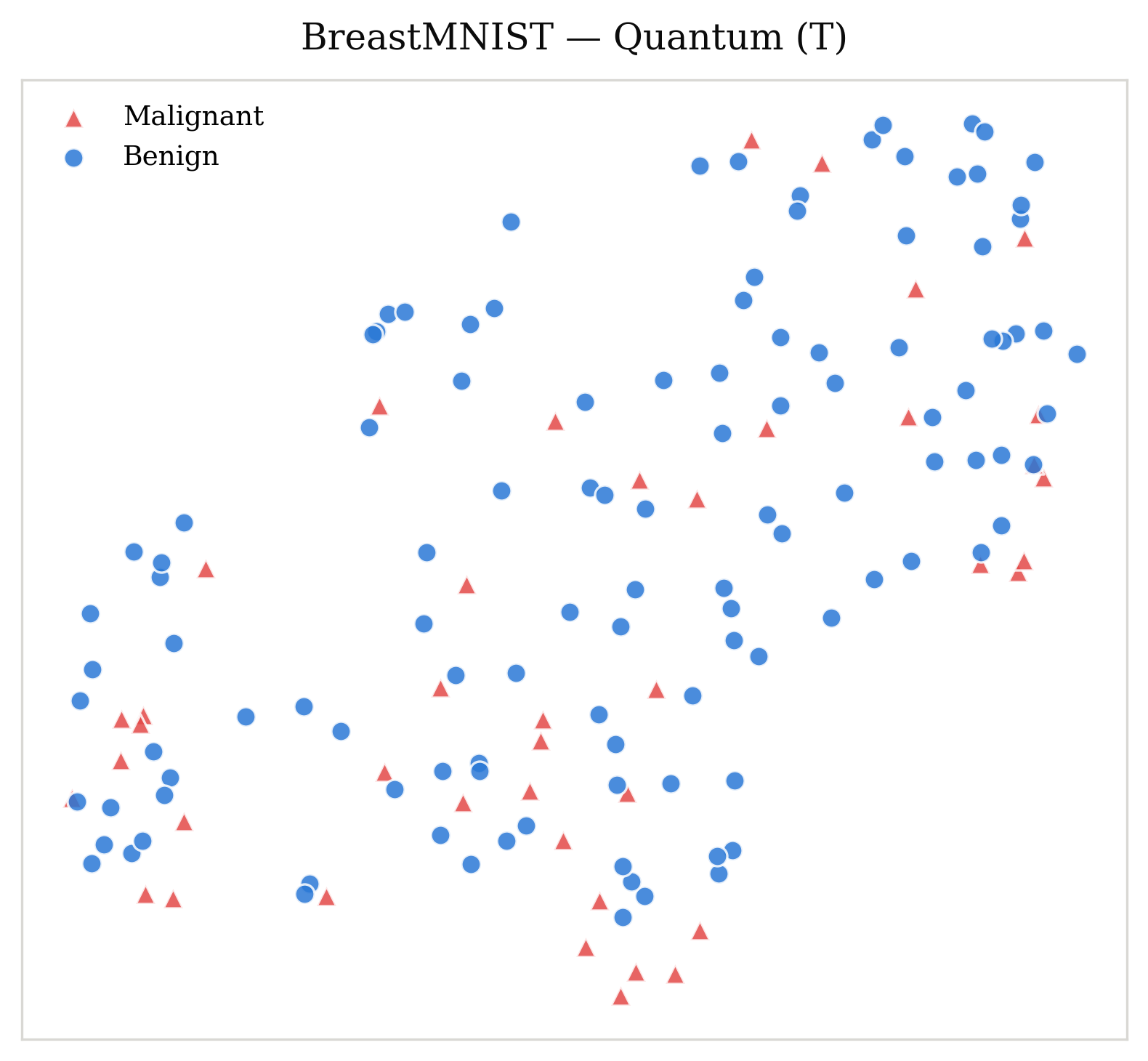}}

    \caption{UMAP projections of the embeddings learned by the standalone
    baselines on the BreastMNIST test set. The left column shows the classical
    backbones and the right column the quantum branches, illustrating how the two
    paradigms organize the same samples differently before any fusion is
    applied.}
    \label{fig:umap_baselines}
\end{figure}

\begin{figure}[htbp]
    \centering
    \subfloat[ResNet-18 + non-trainable]{%
        \includegraphics[width=0.48\linewidth]{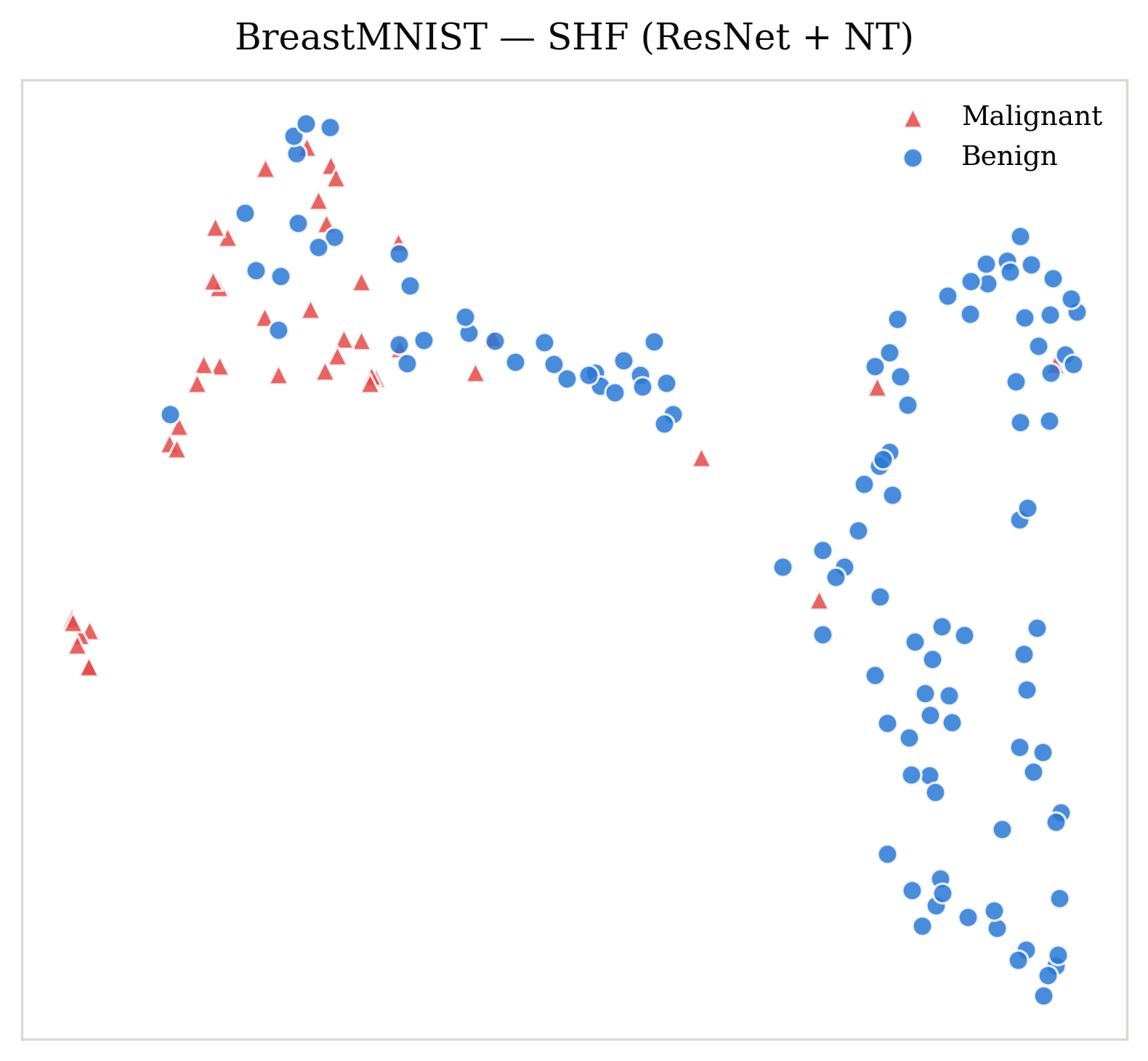}}
    \hfill
    \subfloat[ResNet-18 + trainable]{%
        \includegraphics[width=0.48\linewidth]{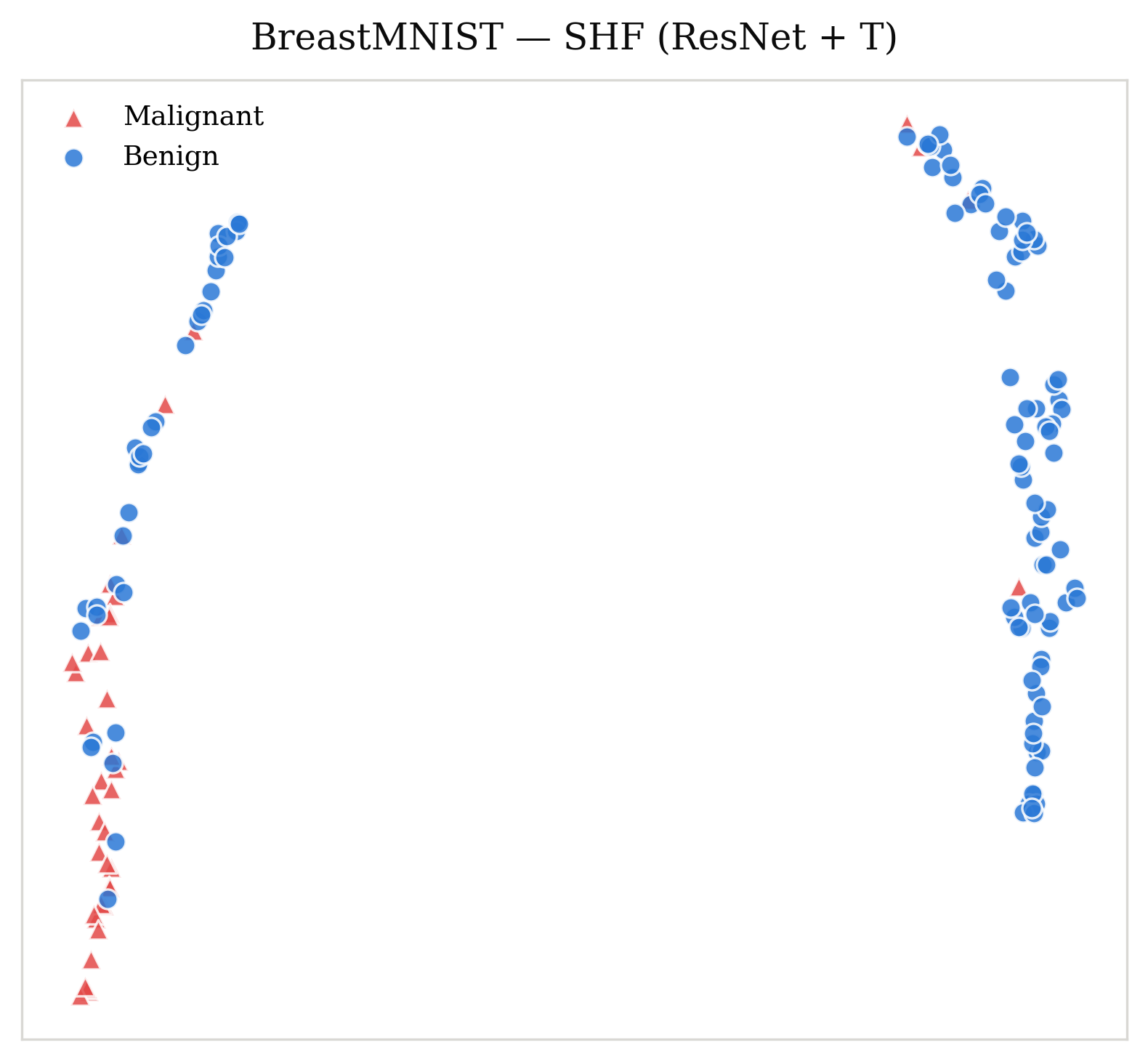}}

    \vspace{0.4em}

    \subfloat[SCNN + non-trainable]{%
        \includegraphics[width=0.48\linewidth]{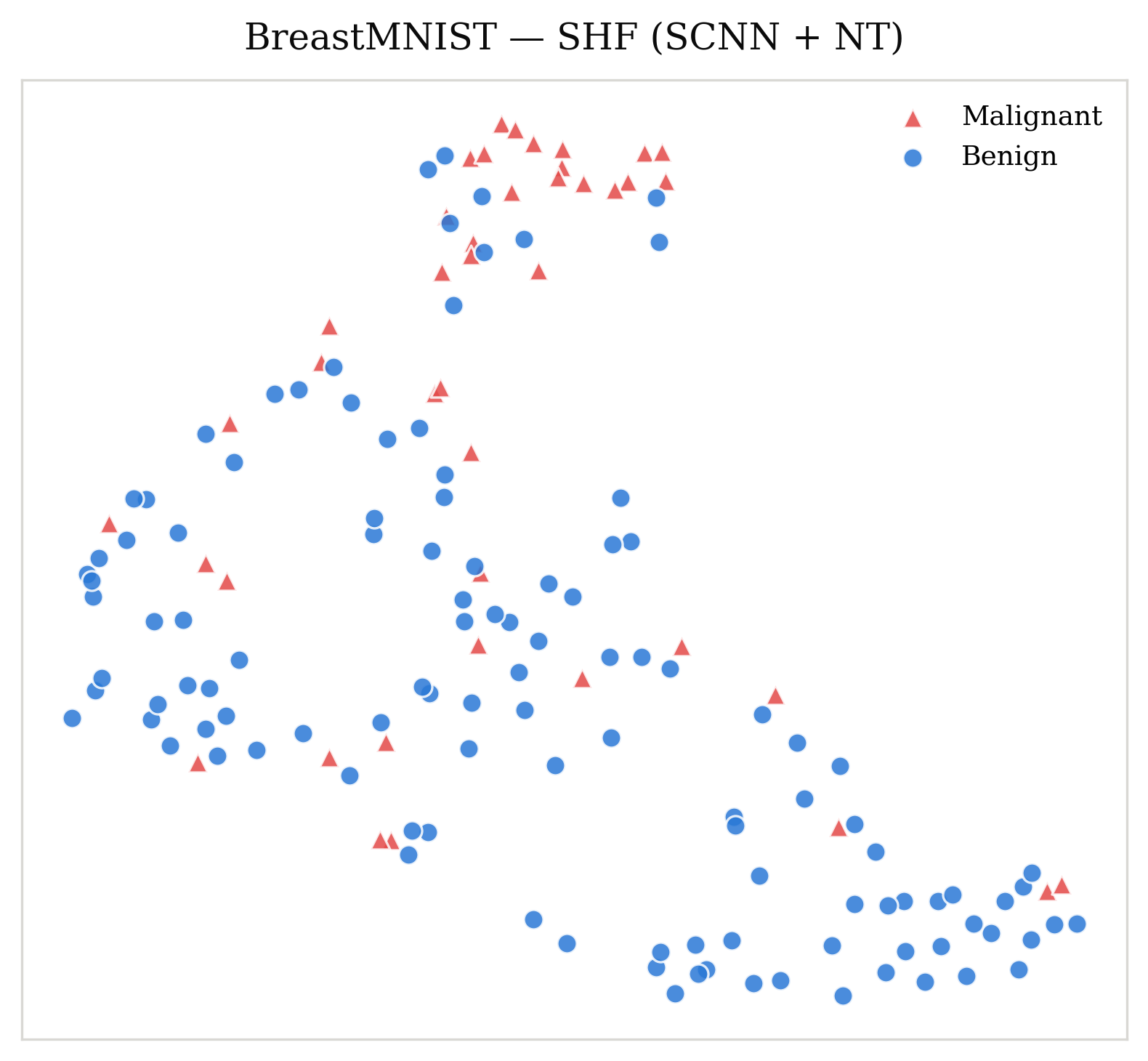}}
    \hfill
    \subfloat[SCNN + trainable]{%
        \includegraphics[width=0.48\linewidth]{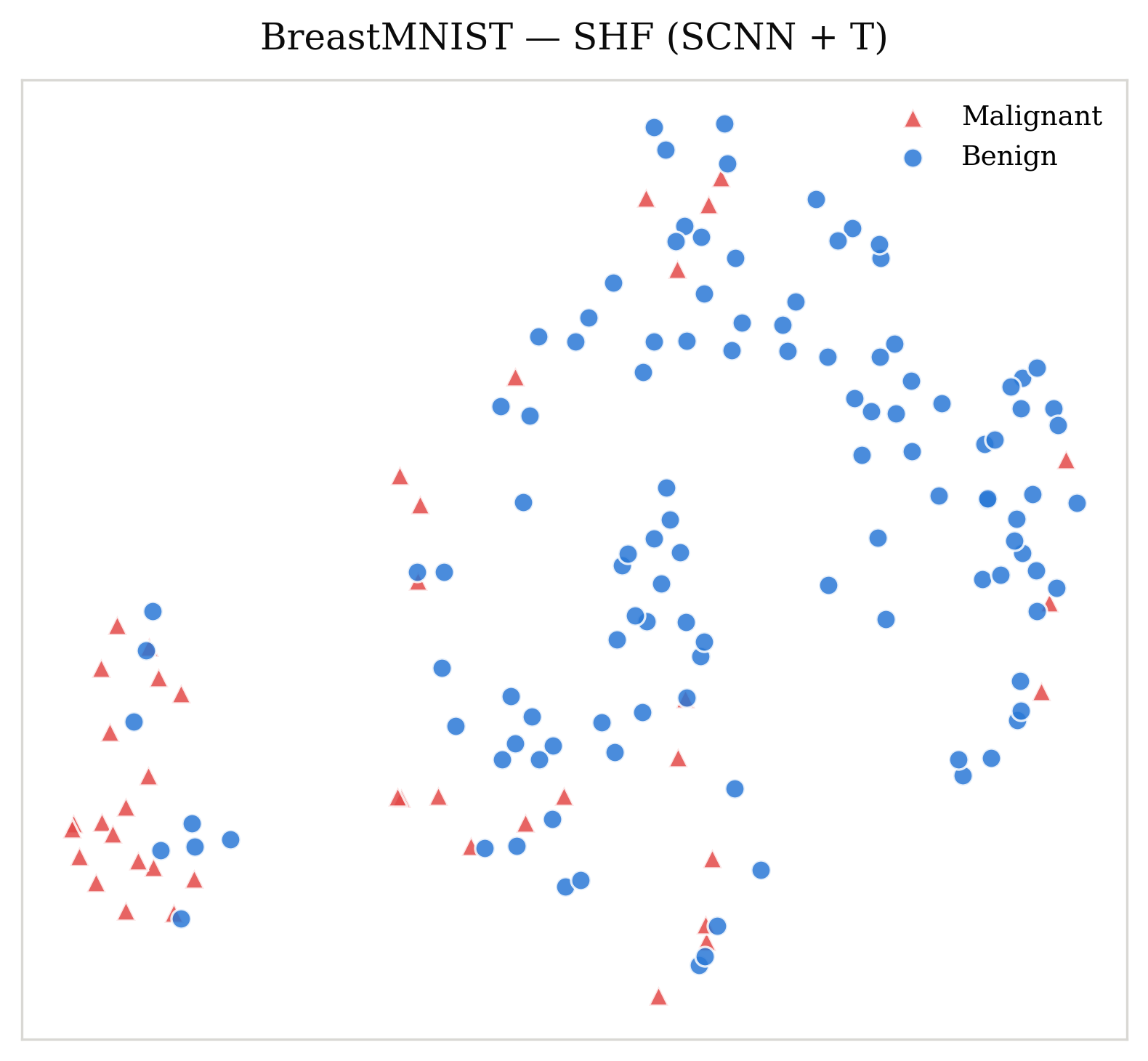}}

    \caption{UMAP projections of the fused embeddings on BreastMNIST under the
    \textbf{SHF} strategy. Rows correspond to the classical backbone and columns
    to the quantum circuit.}
    \label{fig:umap_shf}
\end{figure}

\begin{figure}[htbp]
    \centering
    \subfloat[ResNet-18 + non-trainable]{%
        \includegraphics[width=0.48\linewidth]{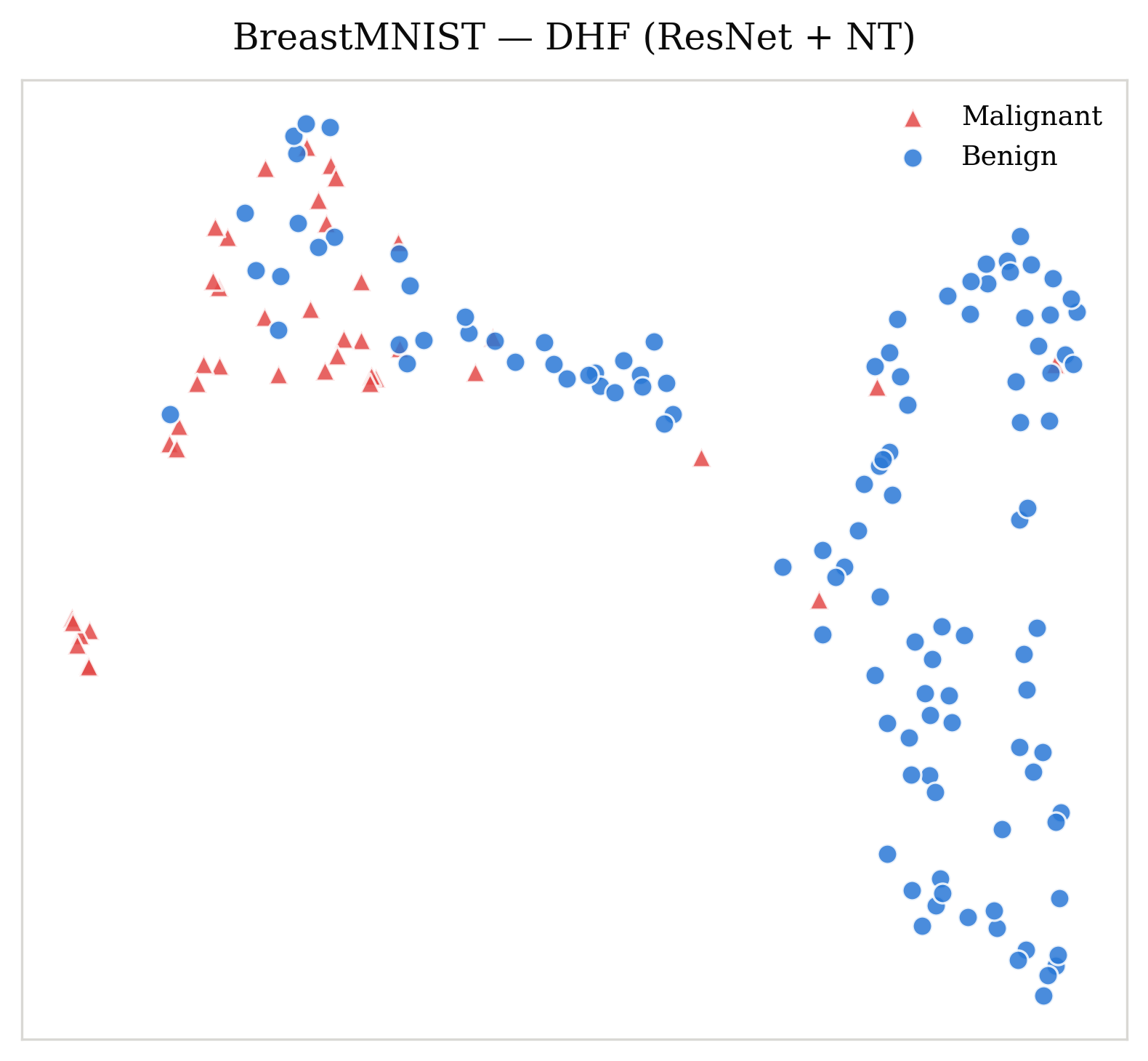}}
    \hfill
    \subfloat[ResNet-18 + trainable]{%
        \includegraphics[width=0.48\linewidth]{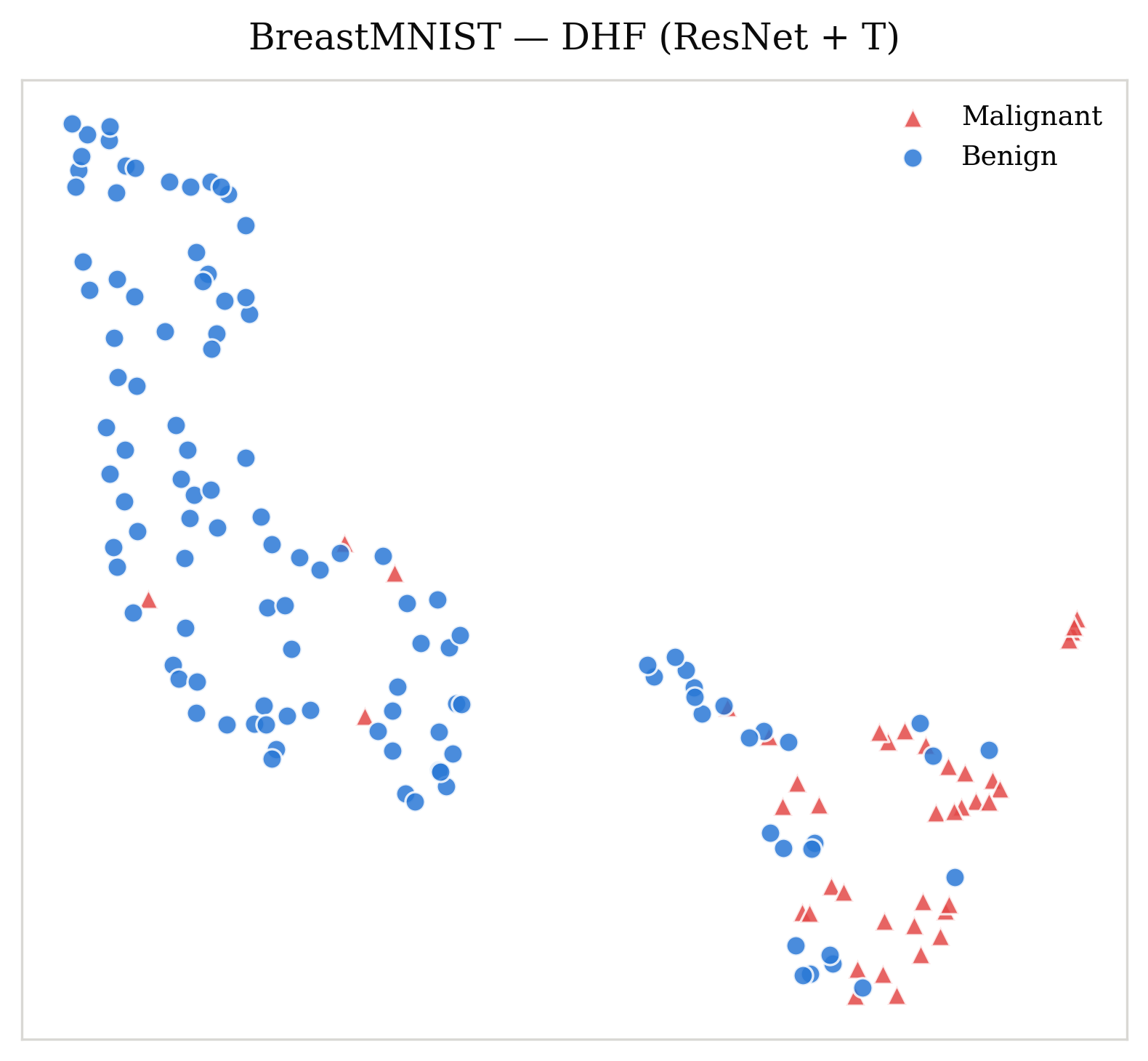}}

    \vspace{0.4em}

    \subfloat[SCNN + non-trainable]{%
        \includegraphics[width=0.48\linewidth]{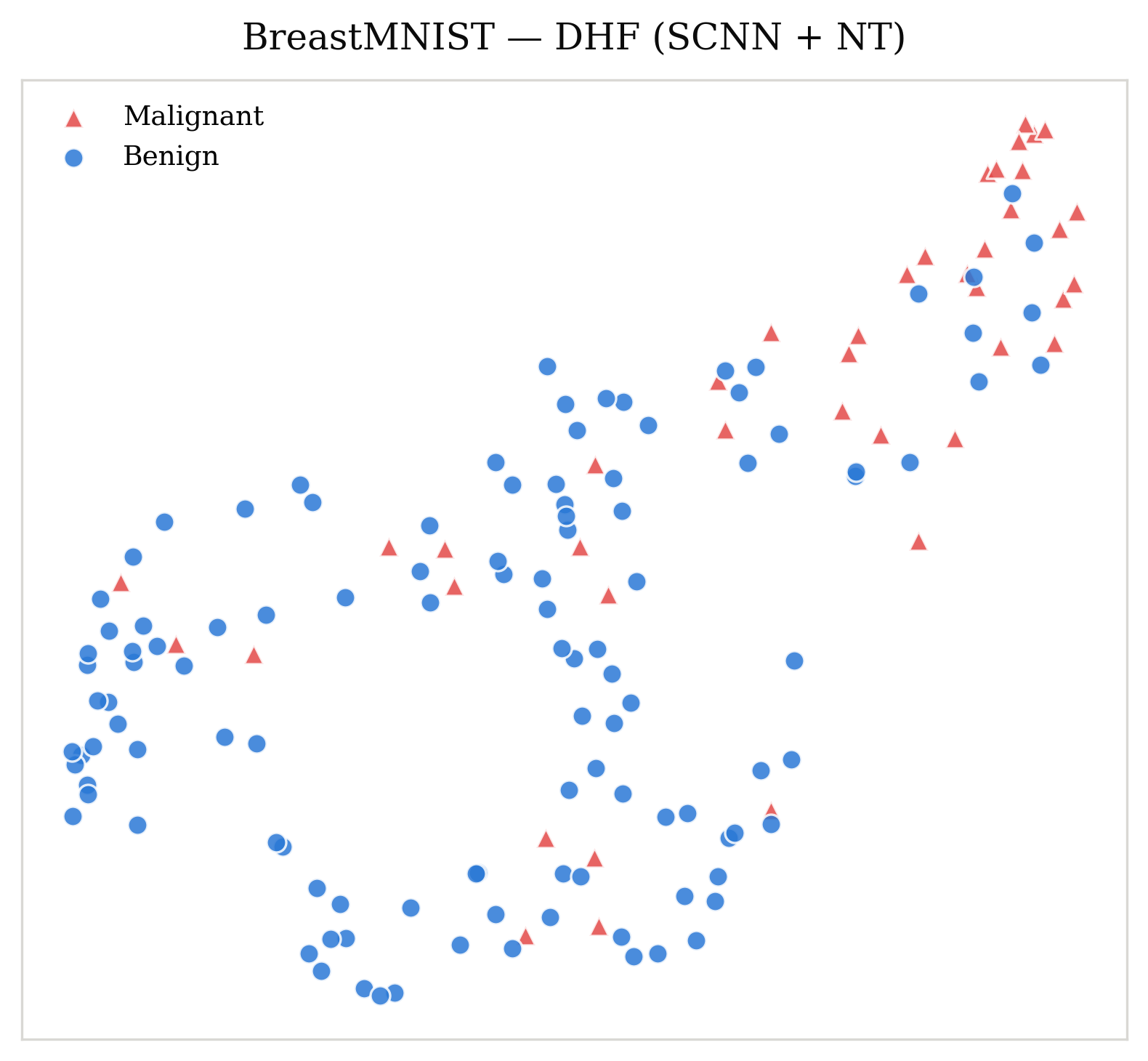}}
    \hfill
    \subfloat[SCNN + trainable]{%
        \includegraphics[width=0.48\linewidth]{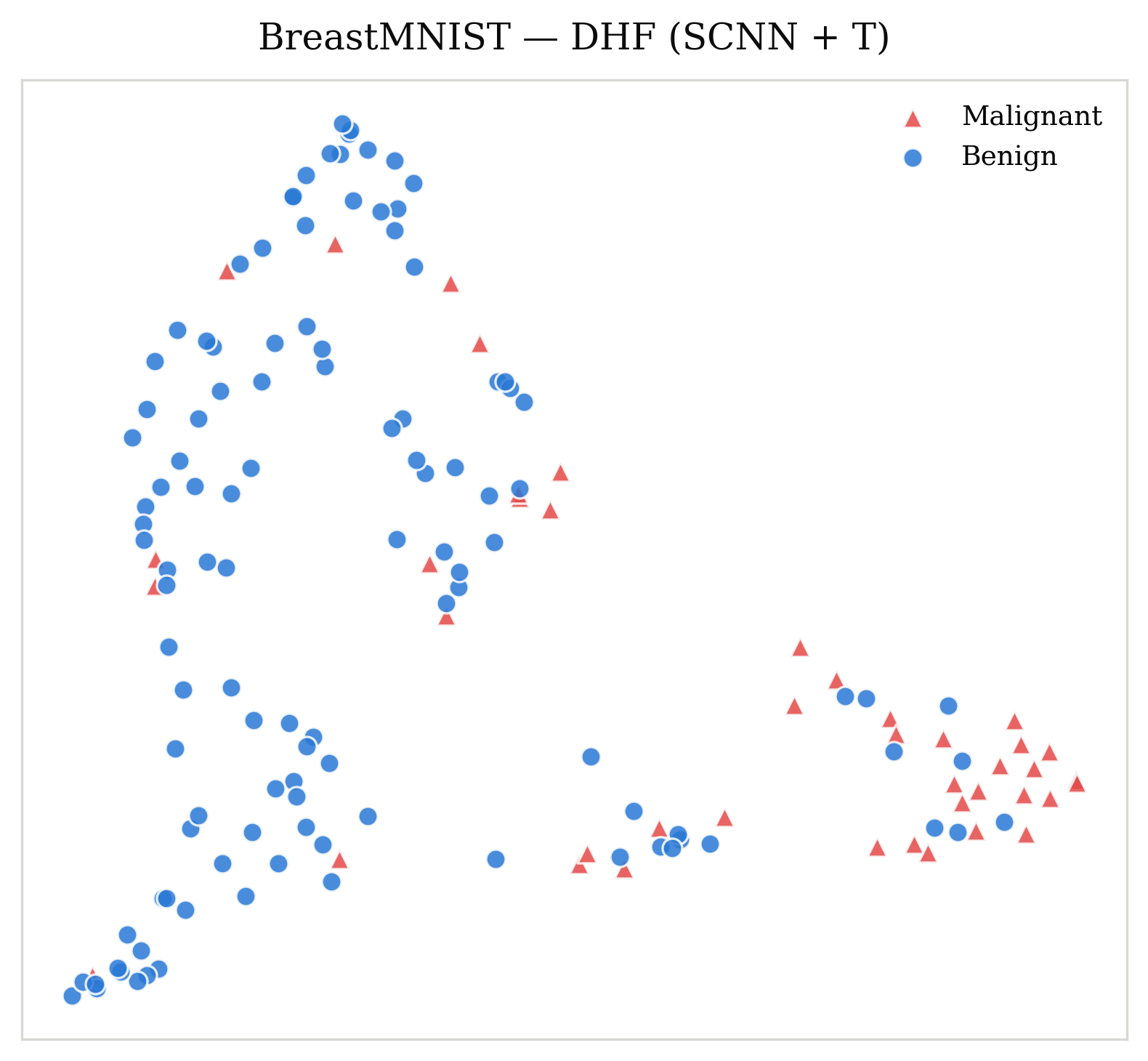}}

    \caption{UMAP projections of the fused embeddings on BreastMNIST under the
    \textbf{DHF} strategy. Rows correspond to the classical backbone and columns
    to the quantum circuit.}
    \label{fig:umap_dhf}
\end{figure}

\begin{figure}[htbp]
    \centering
    \subfloat[ResNet-18 + non-trainable]{%
        \includegraphics[width=0.48\linewidth]{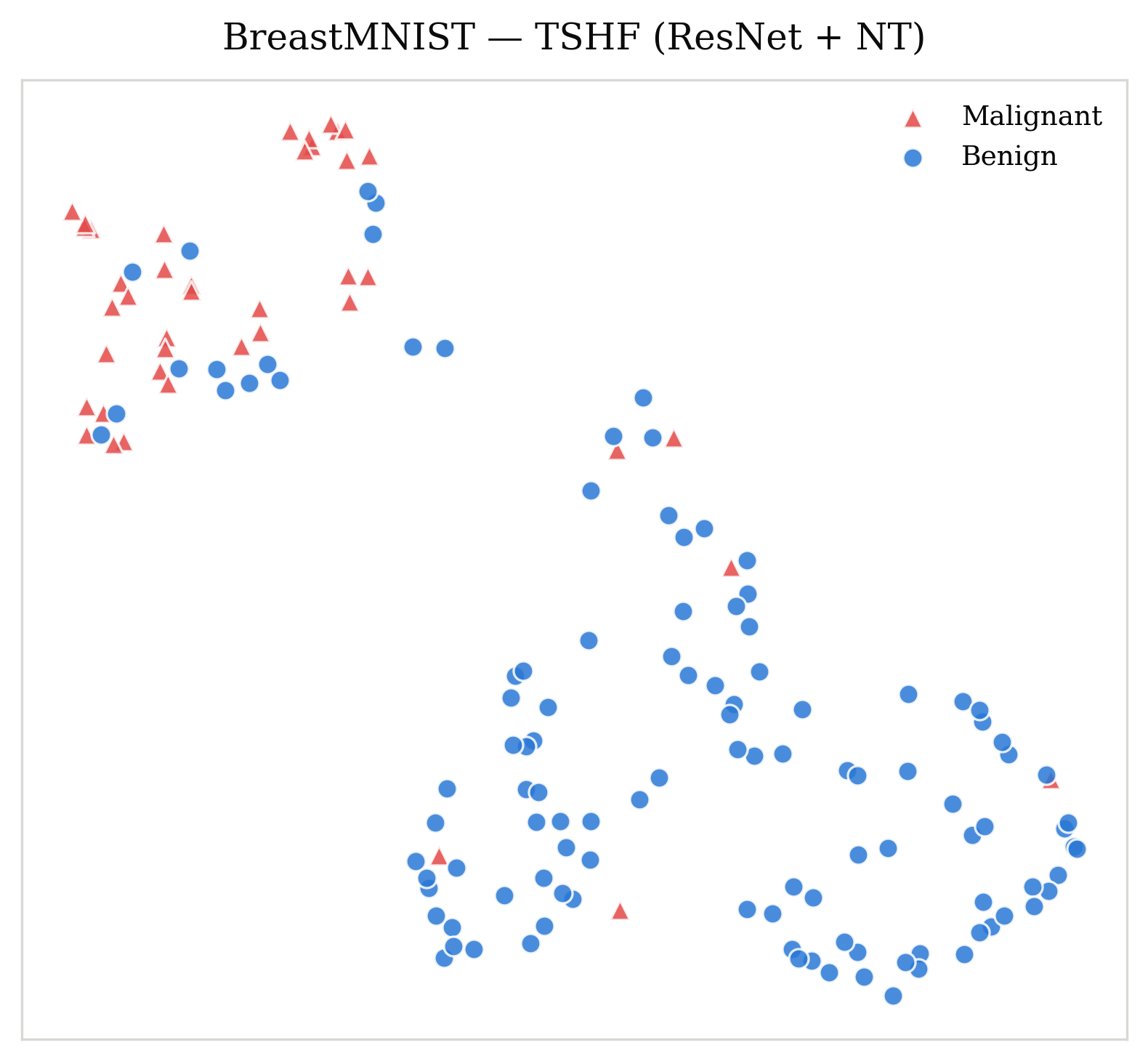}}
    \hfill
    \subfloat[ResNet-18 + trainable]{%
        \includegraphics[width=0.48\linewidth]{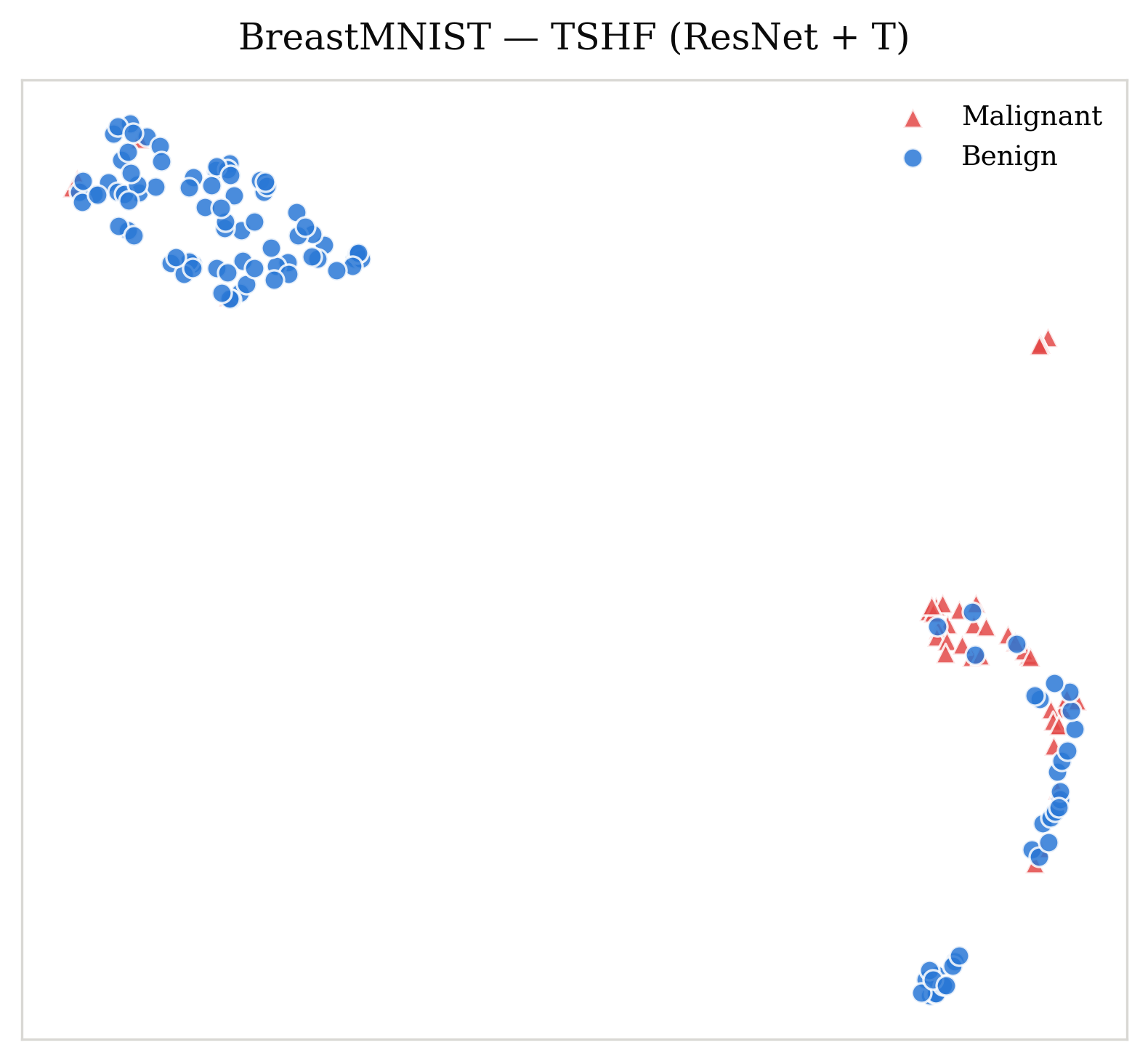}}

    \vspace{0.4em}

    \subfloat[SCNN + non-trainable]{%
        \includegraphics[width=0.48\linewidth]{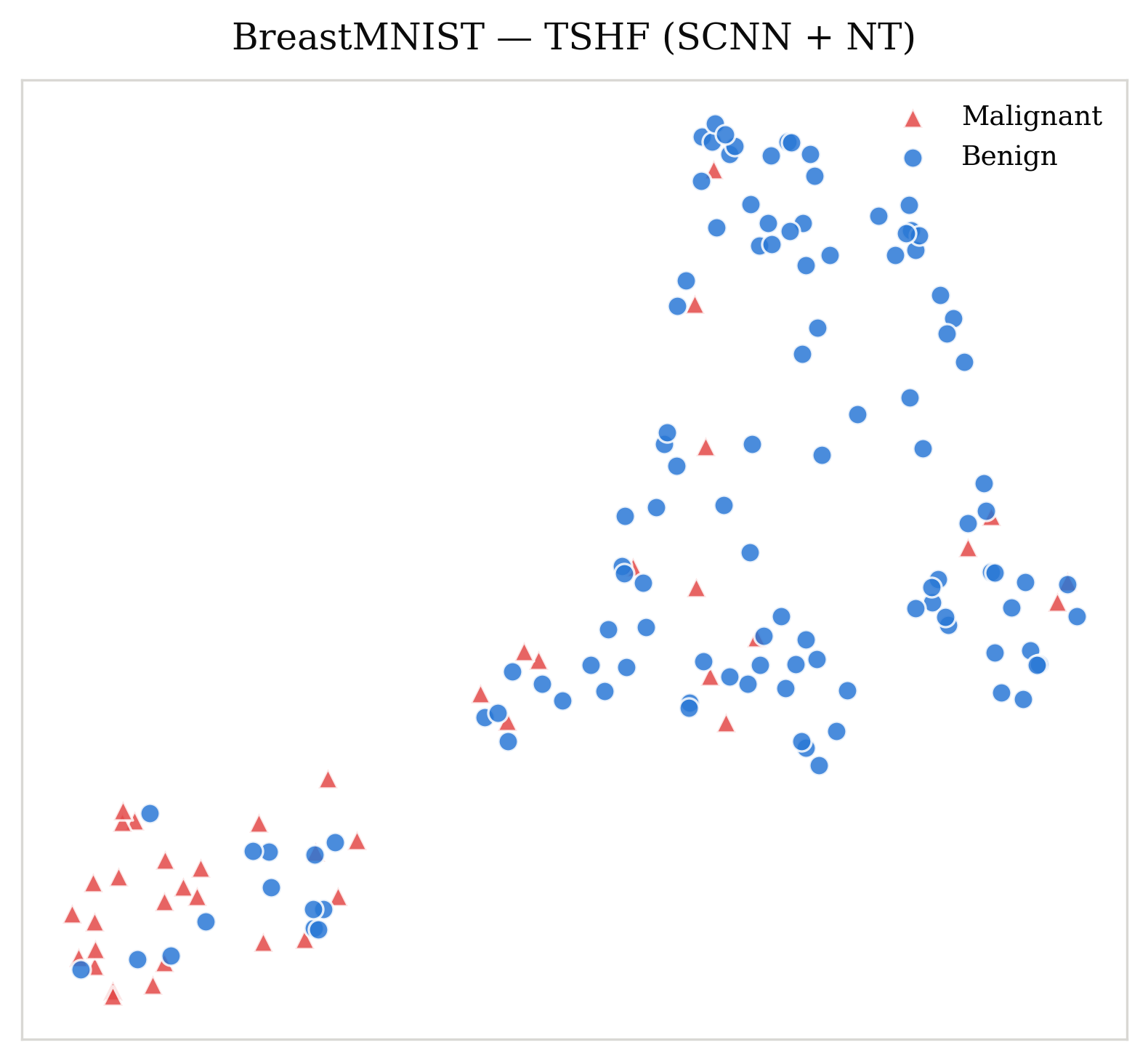}}
    \hfill
    \subfloat[SCNN + trainable]{%
        \includegraphics[width=0.48\linewidth]{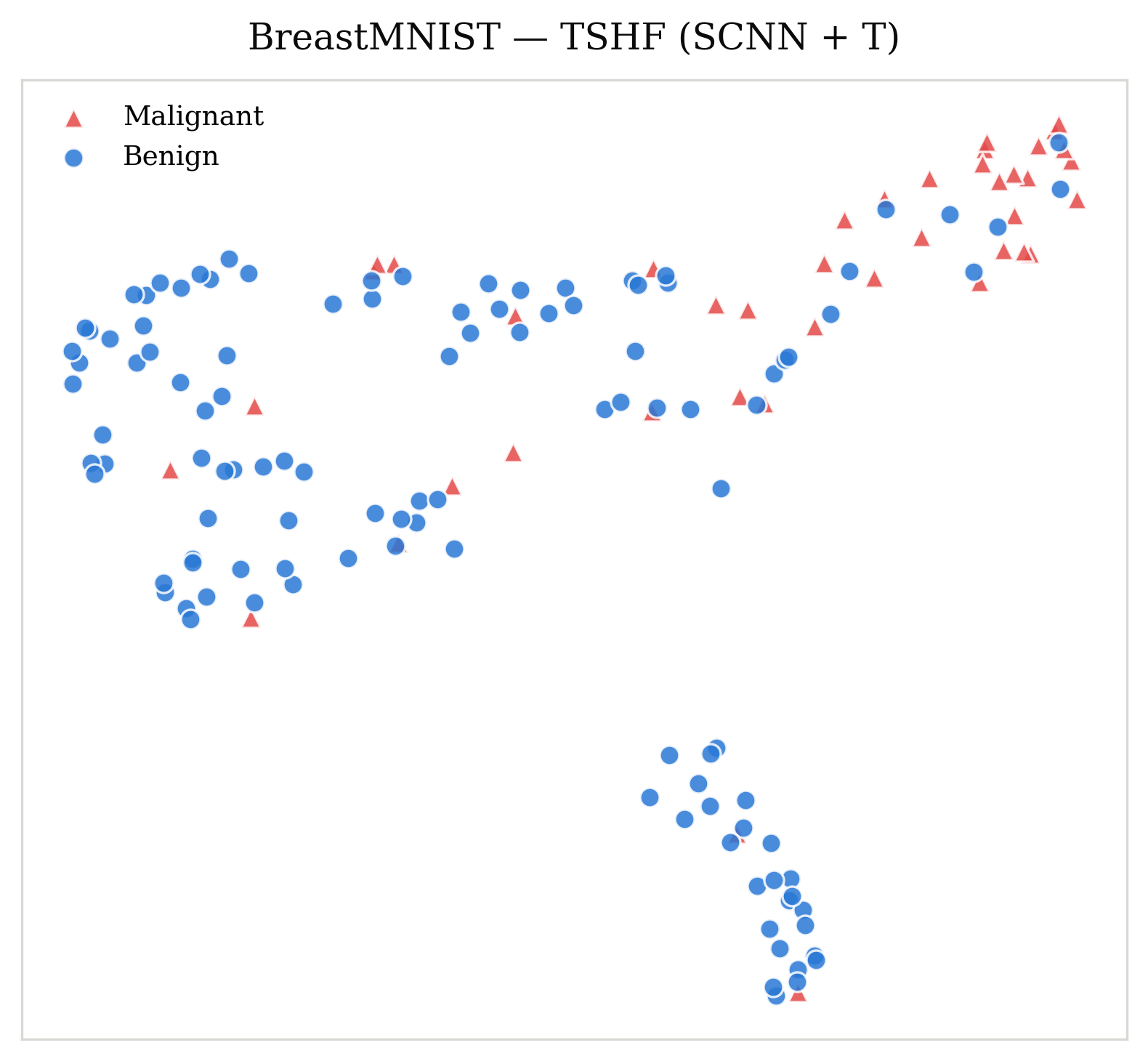}}

    \caption{UMAP projections of the fused embeddings on BreastMNIST under the
    \textbf{TSHF} strategy. Rows correspond to the classical backbone and columns
    to the quantum circuit.}
    \label{fig:umap_tshf}
\end{figure}

Figures~\ref{fig:umap_baselines}--\ref{fig:umap_tshf} project the learned
embeddings of the BreastMNIST test set with UMAP~\cite{McInnes2018}; the
corresponding projections for INbreast and BUS-UCLM are provided in
~\ref{app:umap}. The standalone baselines already show that the two
paradigms organize the same samples differently, which is the geometric
counterpart of the complementarity hypothesis.

Among the fused representations, the configurations that perform best
quantitatively also produce the most structured projections, with the two classes
occupying largely distinct regions and a residual overlap at the boundary. Two
caveats apply to this analysis. UMAP is a non-linear, stochastic embedding whose
apparent cluster separation depends on its hyperparameters and does not
constitute a measure of separability, and the projections shown come from a
single run. We therefore treat them as illustrative of the decision geometry
rather than as independent evidence, with the quantitative separability claim
resting on the AUC values instead.

\subsection{Limitations}
\label{subsec:limitations}

The scope of this work is methodological and exploratory, and several limitations
bound the conclusions that can be drawn from it.

All experiments use small, low-resolution public datasets: $28 \times 28$
grayscale inputs, with test splits of 156, 96 and 26 images. At these sizes,
differences of one or two accuracy points fall within seed-to-seed variation, and
on BUS-UCLM a single image is worth 3.8 accuracy points. We report five seeds and
standard deviations throughout for this reason, and avoid drawing conclusions
from differences smaller than the observed dispersion.

No formal statistical significance testing was performed between competing
methods. Given the small number of independent runs, the reported differences
should be read as descriptive trends rather than statistically established
improvements; sample-level paired analyses are left to future work.

On BUS-UCLM, enforcing patient-level partitions on a dataset with few patients
produces a marked shift in malignant prevalence across subsets (32.2\%, 17.9\%,
and 73.1\% in training, validation, and test), so results on this dataset
should be regarded as indicative.

The quantum branch is classically simulated in analytic mode, without shot noise,
decoherence or gate errors. The reported behavior is therefore that of the
idealized circuit, and results on physical hardware may differ. The circuit
itself is deliberately small---four qubits, one variational layer, $2 \times 2$
patches---which bounds what the quantum branch can represent.

Finally, no claim of clinical readiness is intended or supported. We have not
performed calibration analysis, external validation on independent cohorts, or
reader studies, and the datasets used are not representative of a screening
population. Statements about accuracy and separability in this paper refer to
held-out test splits of public research datasets; establishing operating
characteristics suitable for diagnostic use would require evidence of a kind that
lies outside the scope of this study.

% =====================================================================
%  CONCLUSION
% =====================================================================
\section{Conclusion and Future Work}
\label{s.conclusion}

This work presented a systematic empirical study of quantum--classical feature
fusion for breast cancer image classification, organized around the hypothesis
that classical and quantum branches extract complementary representations. We
evaluated three fusion strategies of increasing adaptivity---SHF, DHF and
TSHF---across two classical backbones, trainable and deterministic quanvolutional
branches, and three datasets, with five seeds per configuration.

The evidence supports the complementarity hypothesis in a qualified form. On
BreastMNIST, fusion improves over the strongest classical baseline in both
accuracy, from 84.74\% to 88.33\% with the learned calibration of TSHF, and
separability, from 87.08\% to between 92.0\% and 92.5\% AUC across the
ResNet-based fusion configurations. That TSHF result is the best of the study on
this benchmark and exceeds previously published hybrid models on the same
official split. On
INbreast and BUS-UCLM, static fusion yields the best configurations while the
adaptive strategies do not improve further, and the learned scalar remains close
to its initialization---an outcome we attribute to the limited training data
available on those datasets rather than to a failure of the mechanism.

Two aspects of the study are of practical value independently of which strategy
prevails. The quantum branch is extremely parameter-efficient, contributing four
trainable parameters in its variational form and none in its deterministic form,
against more than eleven million for the classical backbone. And the
deterministic circuit remains competitive with---and on two of three datasets
preferable to---the variational one, while avoiding trainability issues entirely
and permitting its features to be pre-computed and cached, an appealing property
under NISQ constraints. TSHF itself is deliberately minimal: a single learnable
scalar, requiring no tuning stage, that adapts the contribution of one branch
during training and whose learned value provides a diagnostic read-out of that
branch's relative weight.

Future work should extend the quantum branch beyond the four-qubit, single-layer
circuit used here, since the limited expressivity of this \textit{ansatz} is the most
likely reason the trainable variant does not consistently outperform the fixed
one; wider registers and deeper variational layers would test whether a larger
hypothesis space repays its optimization cost. Execution on physical quantum
hardware would additionally expose the effects of noise and decoherence that
analytic simulation hides. Beyond the quantum side, evaluation on larger,
higher-resolution and multi-modal datasets, together with calibration analysis
and external validation, would be required before any statement about clinical
utility could be made.

\section*{Code and Data Availability}
The implementation, configuration files, and scripts used to reproduce all
experiments, together with the patient-level split files for INbreast and
BUS-UCLM, are publicly available at \url{https://github.com/yrsobrinho/BreastCancerHybridFeatureFusion}. BreastMNIST is available through the MedMNIST benchmark, and INbreast and BUS-UCLM can be
obtained from their original providers under their respective access terms.

\section*{Acknowledgement}

The authors are grateful to the São Paulo Research Foundation (FAPESP) grants 2013/07375-0, 2023/14427-0, 2024/00117-0, 2024/22853-0, 2026/11390-4, 2025/07171-2, and by the Brazilian National Council for Scientific and Technological Development (CNPq) process number 308529/2021-9.

\bibliographystyle{elsarticle-num} 
\bibliography{refs}

% =====================================================================
%  APPENDIX
% =====================================================================
\appendices   

\section{Additional ROC and Precision--Recall Curves}
\label{app:curves}

\begin{figure*}[htbp]
    \centering
    \subfloat[ResNet-18 backbone\label{fig:roc_inb_resnet}]{%
        \includegraphics[width=0.48\linewidth]{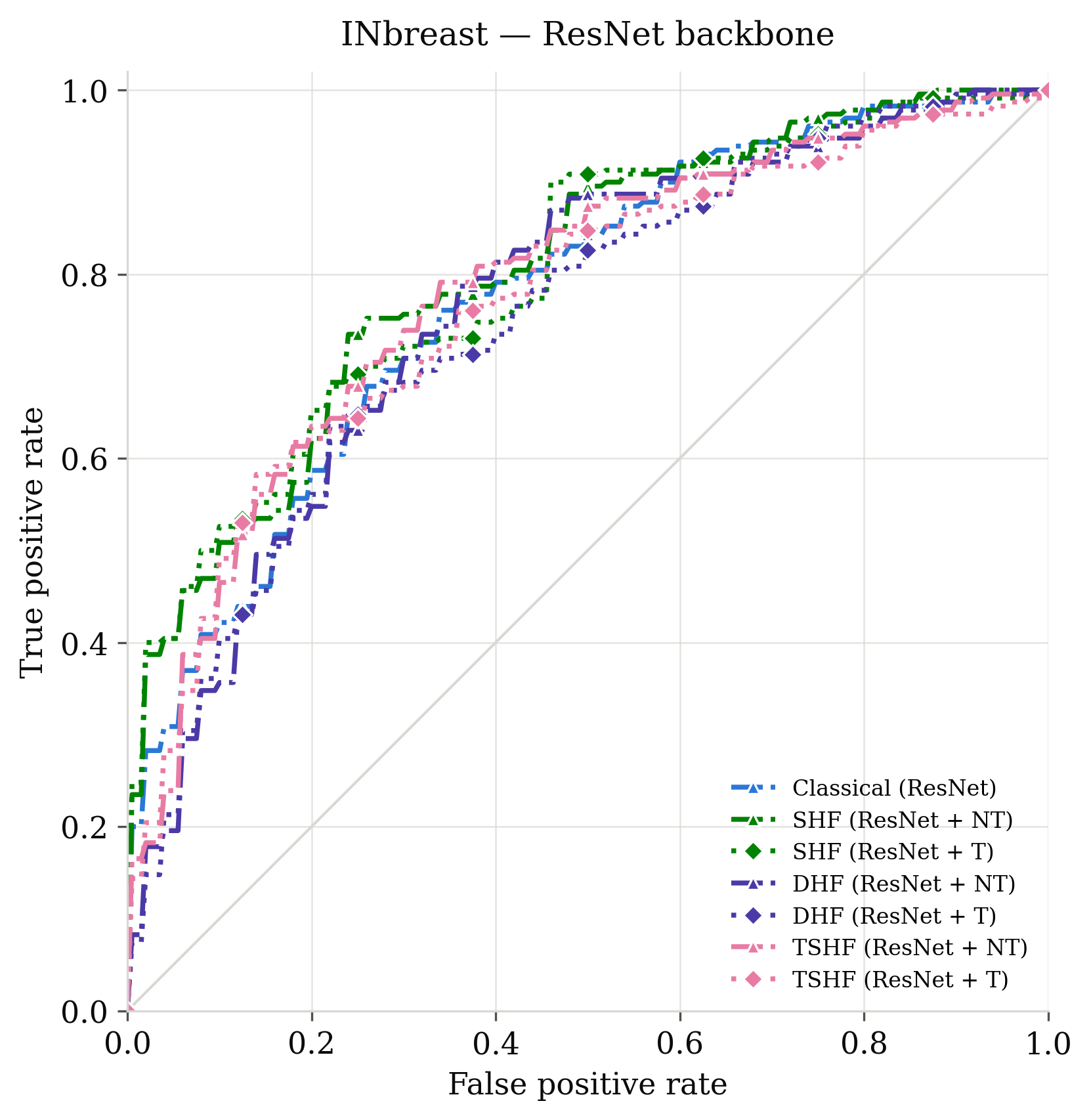}}
    \hfill
    \subfloat[SCNN backbone\label{fig:roc_inb_scnn}]{%
        \includegraphics[width=0.48\linewidth]{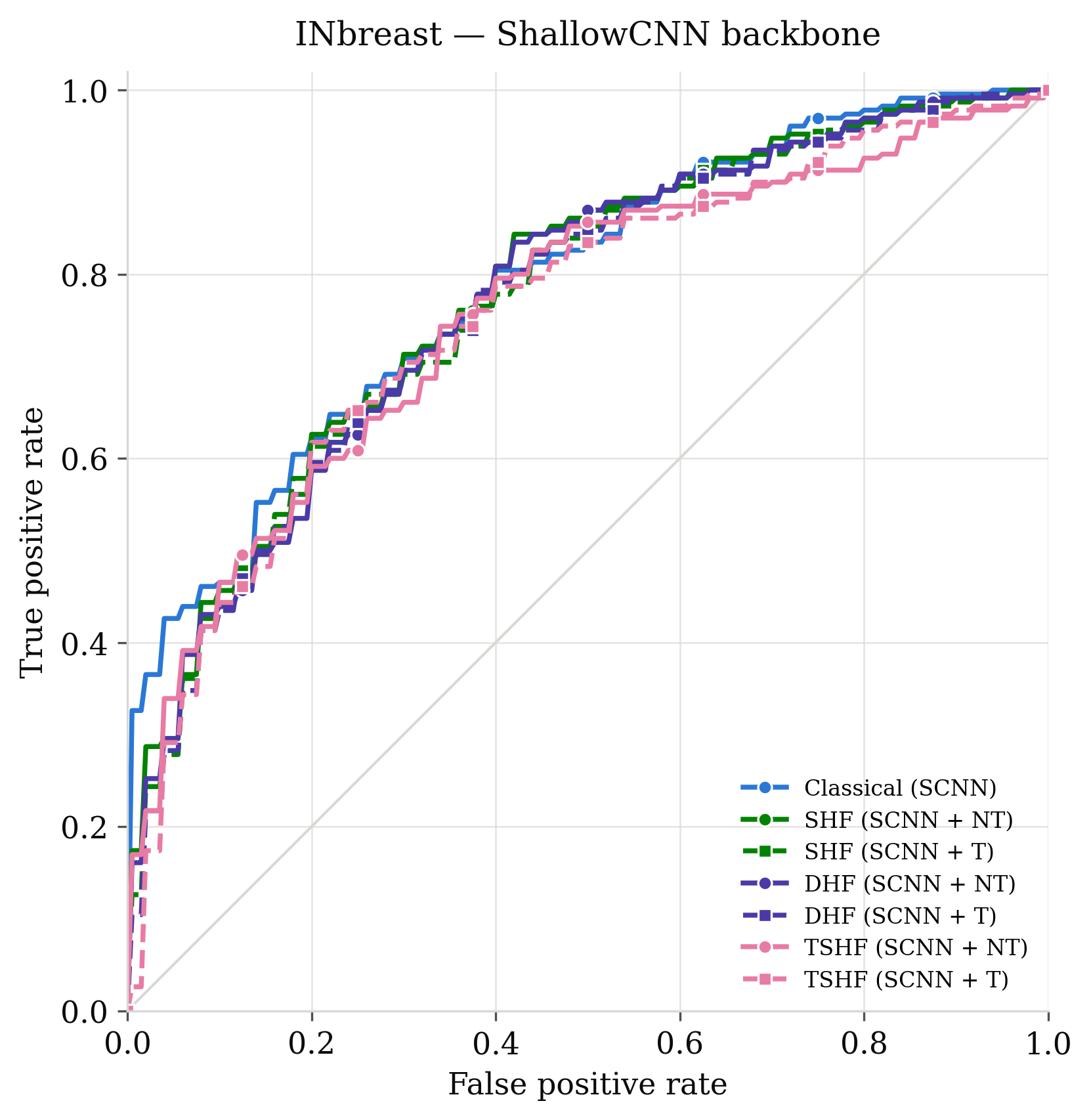}}

    \vspace{0.5em}

    \subfloat[Non-trainable quantum circuit\label{fig:roc_inb_nt}]{%
        \includegraphics[width=0.48\linewidth]{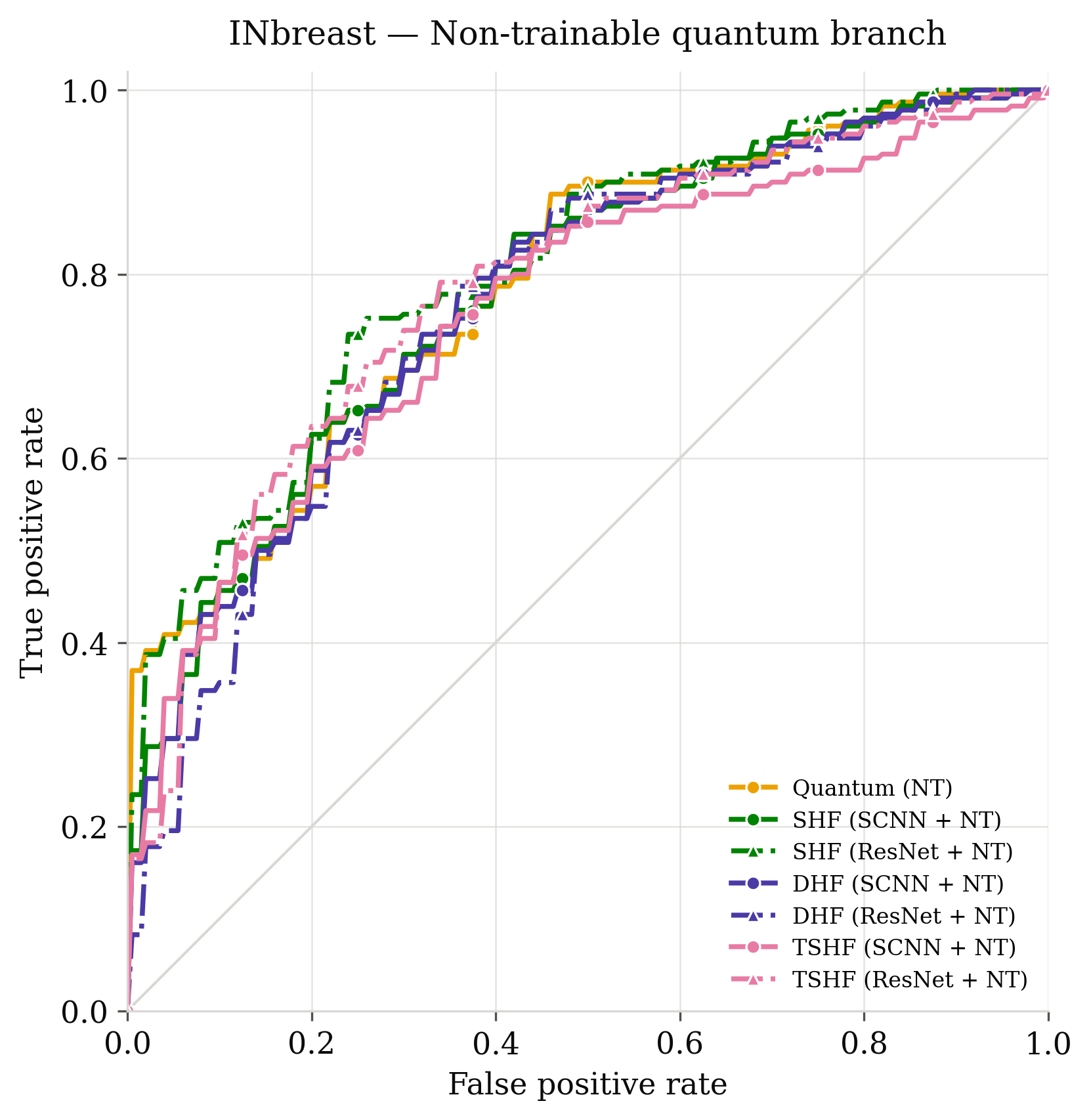}}
    \hfill
    \subfloat[Trainable quantum circuit\label{fig:roc_inb_t}]{%
        \includegraphics[width=0.48\linewidth]{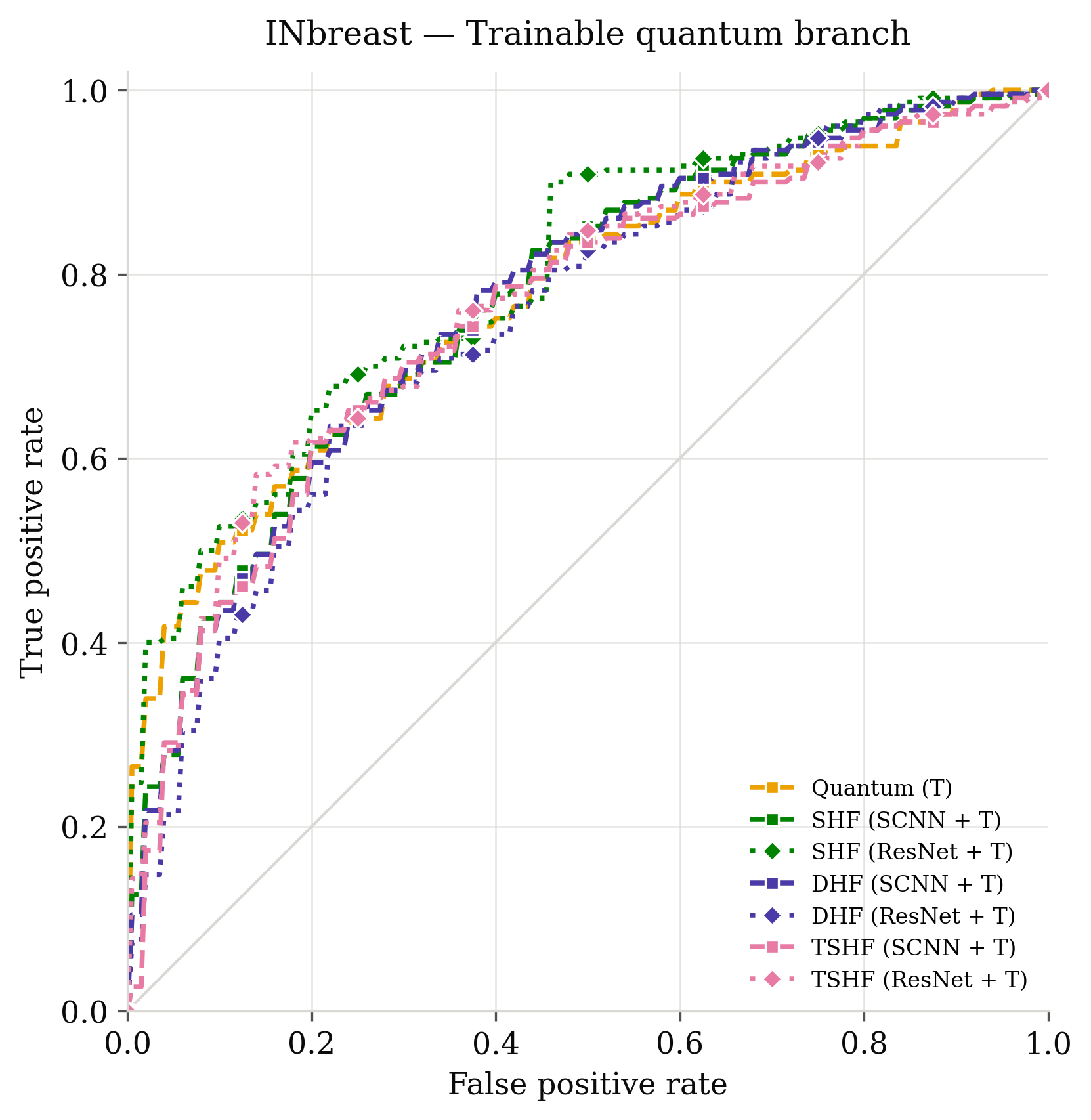}}

    \caption{ROC curves on the INbreast test set, averaged over five independent
    runs, grouped by classical backbone (top row) and by quantum circuit (bottom
    row).}
    \label{fig:roc_inbreast_groups}
\end{figure*}

\begin{figure*}[htbp]
    \centering
    \subfloat[ResNet-18 backbone\label{fig:roc_bus_resnet}]{%
        \includegraphics[width=0.48\linewidth]{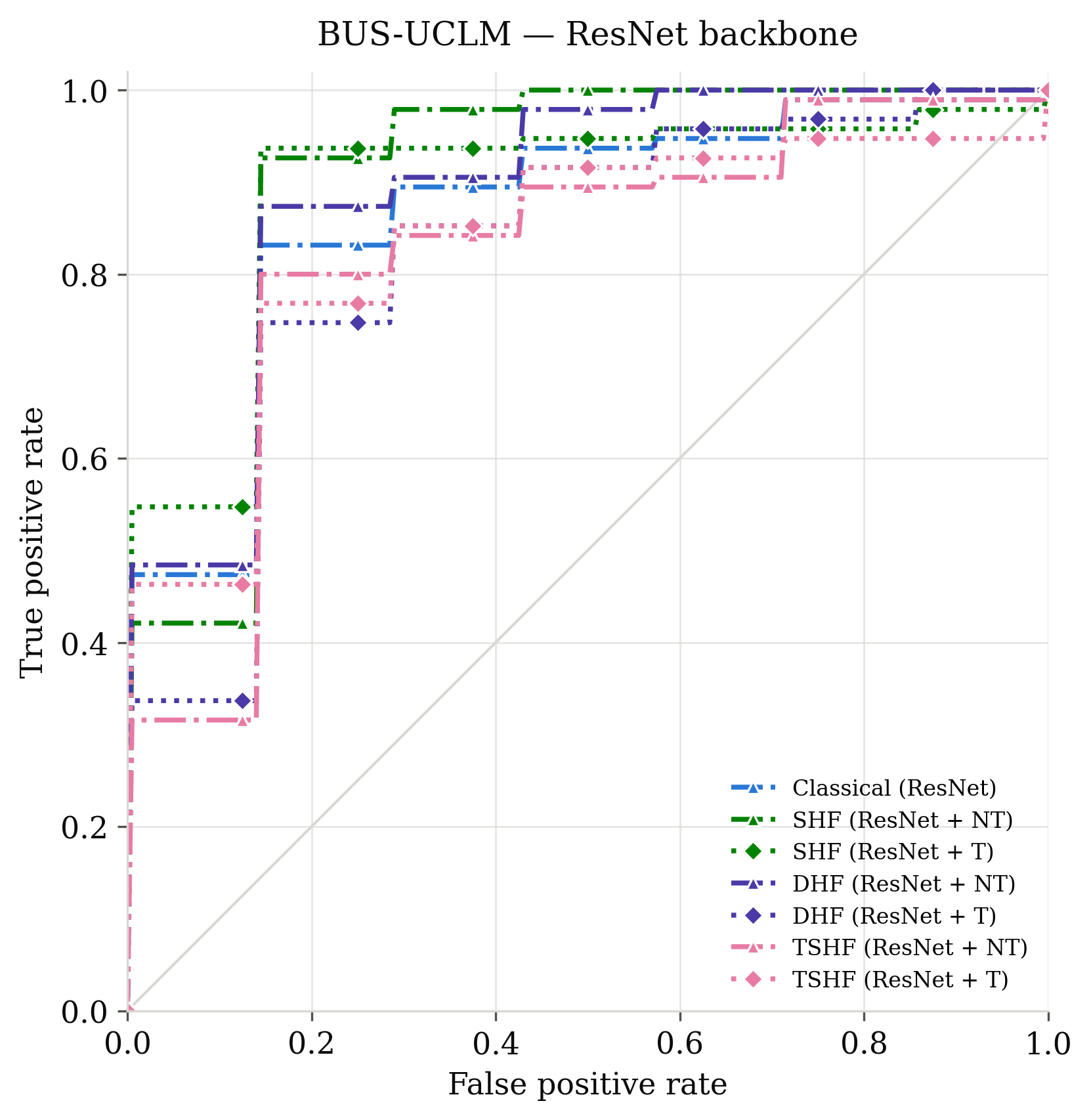}}
    \hfill
    \subfloat[SCNN backbone\label{fig:roc_bus_scnn}]{%
        \includegraphics[width=0.48\linewidth]{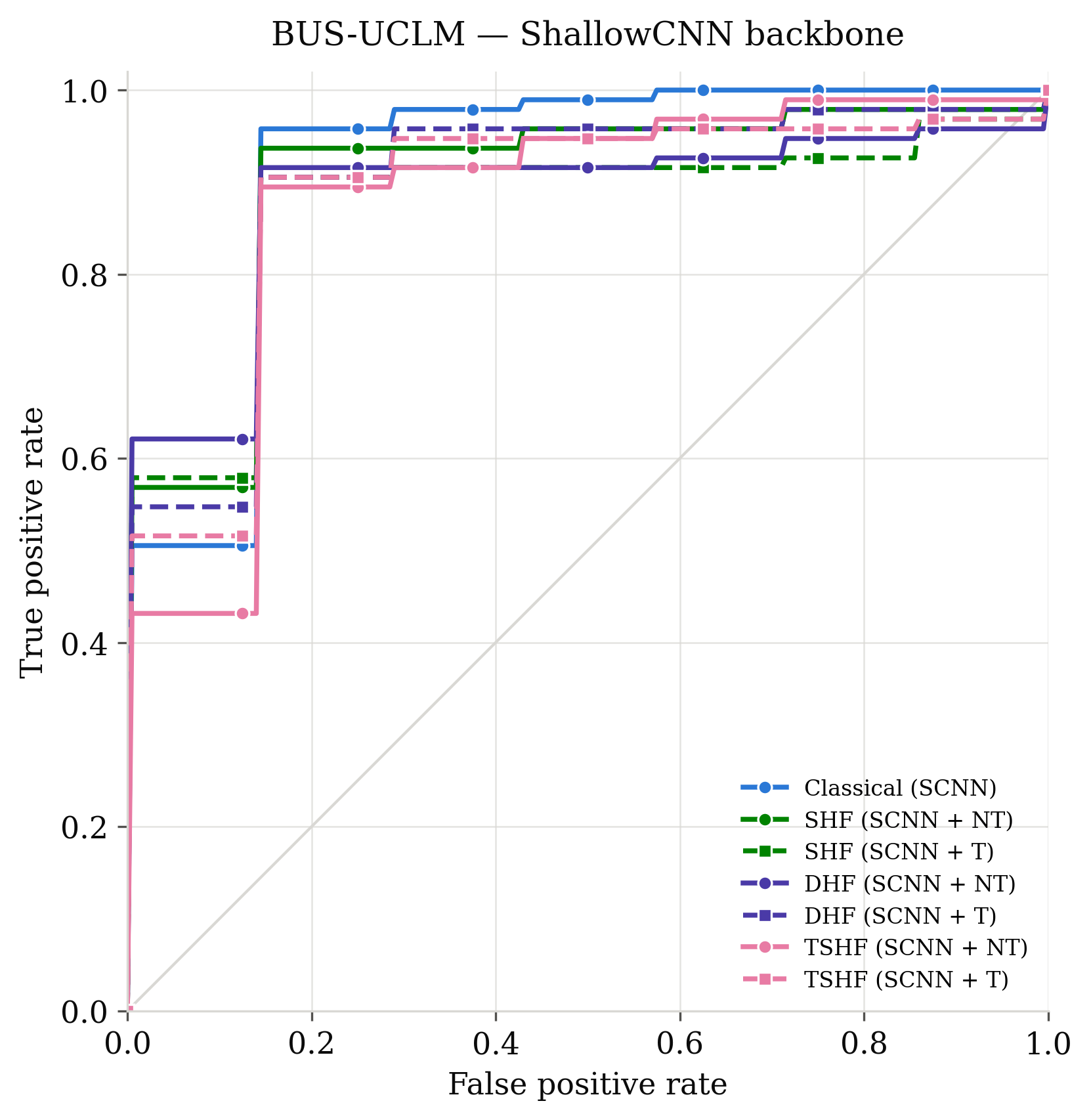}}

    \vspace{0.5em}

    \subfloat[Non-trainable quantum circuit\label{fig:roc_bus_nt}]{%
        \includegraphics[width=0.48\linewidth]{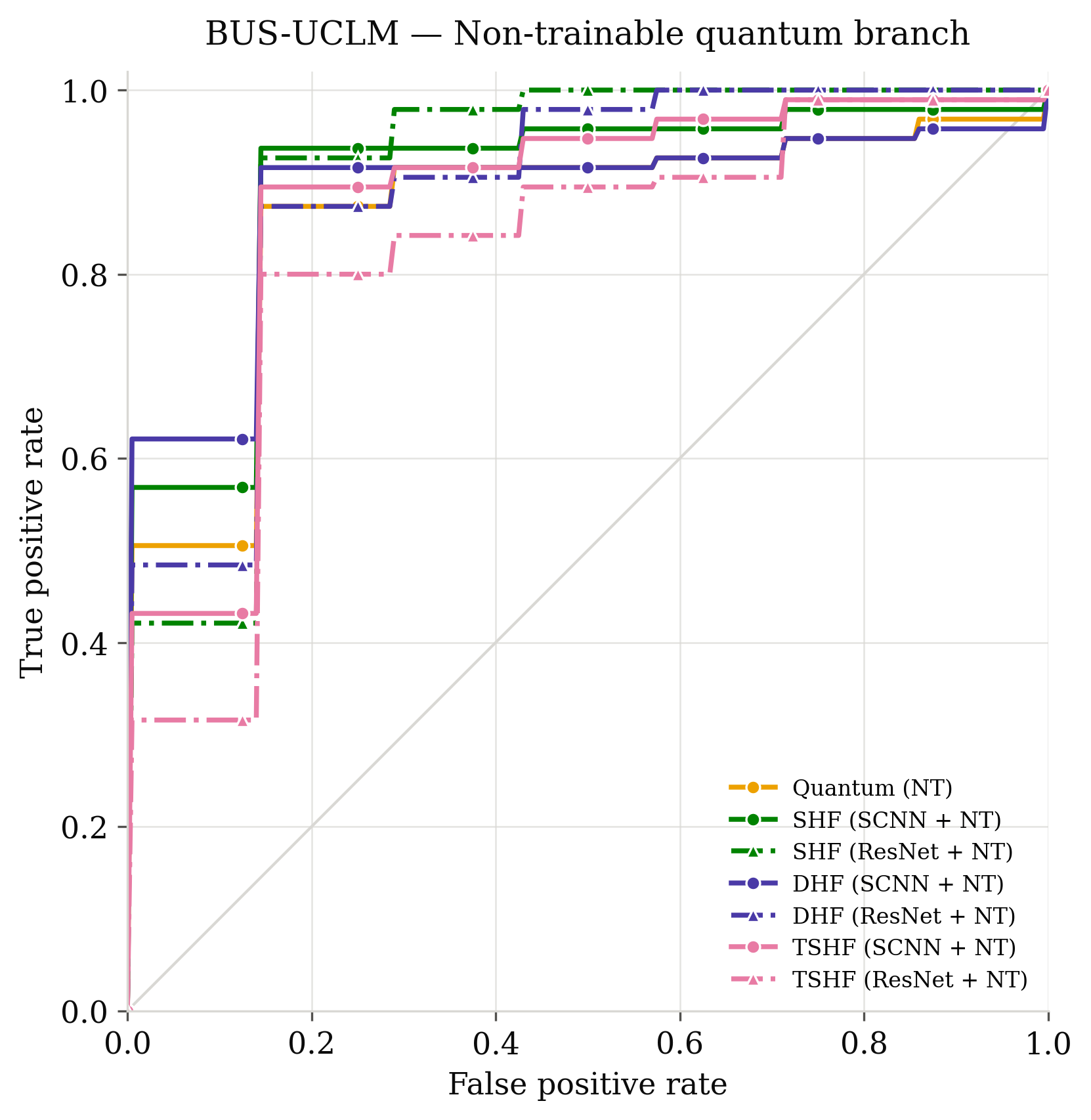}}
    \hfill
    \subfloat[Trainable quantum circuit\label{fig:roc_bus_t}]{%
        \includegraphics[width=0.48\linewidth]{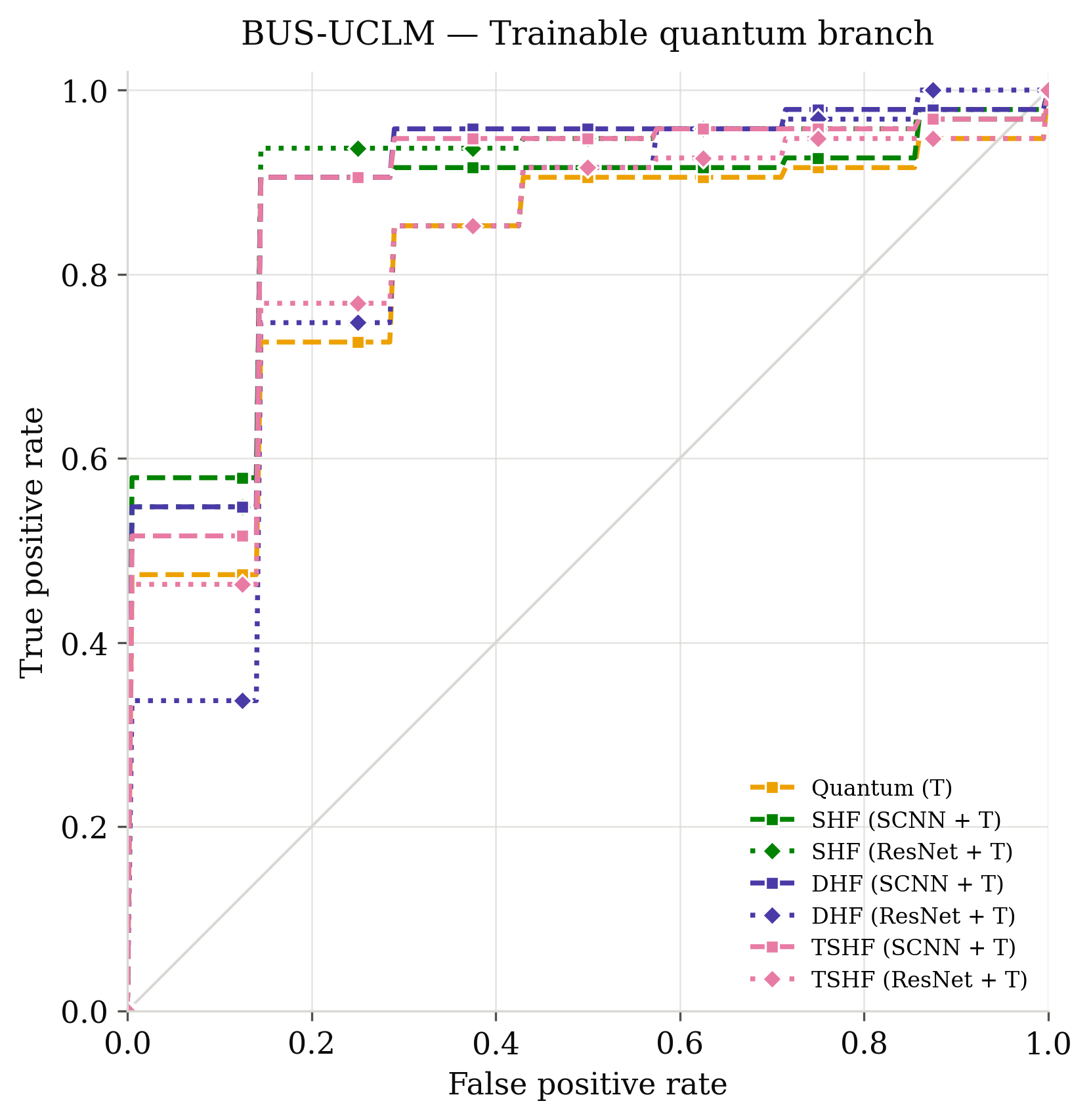}}

    \caption{ROC curves on the BUS-UCLM test set, averaged over five independent
    runs, grouped by classical backbone (top row) and by quantum circuit (bottom
    row).}
    \label{fig:roc_busuclm_groups}
\end{figure*}

Figures~\ref{fig:roc_inbreast_groups} and~\ref{fig:roc_busuclm_groups} show the
ROC curves for INbreast and BUS-UCLM, following the same grouping used for
BreastMNIST in the main text: the top row holds the classical backbone fixed and
the bottom row the quantum circuit, so that each panel isolates the effect of the
fusion strategy.

\begin{figure*}[htbp]
    \centering
    \subfloat[ResNet-18 backbone\label{fig:pr_inb_resnet}]{%
        \includegraphics[width=0.48\linewidth]{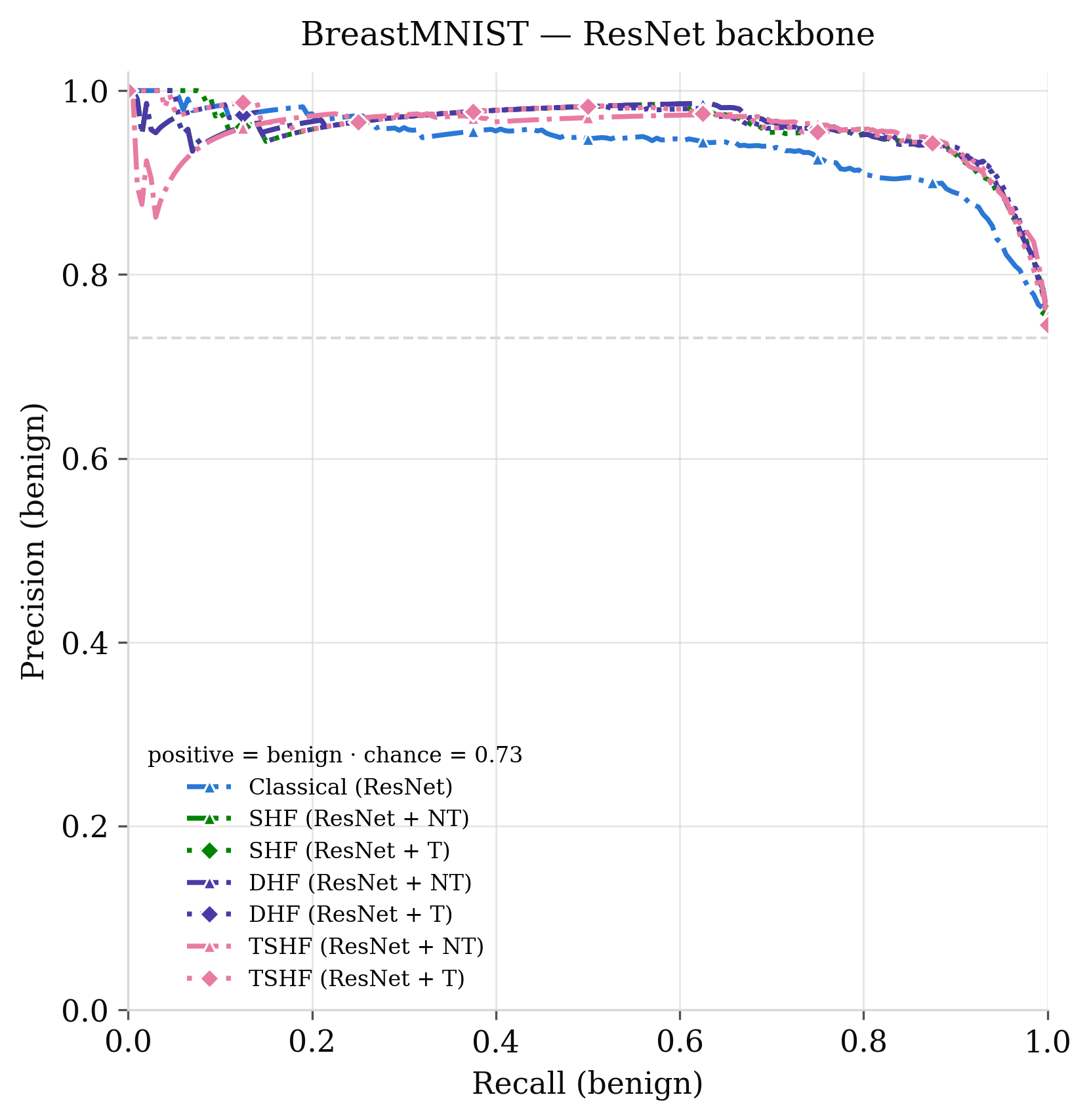}}
    \hfill
    \subfloat[SCNN backbone\label{fig:pr_inb_scnn}]{%
        \includegraphics[width=0.48\linewidth]{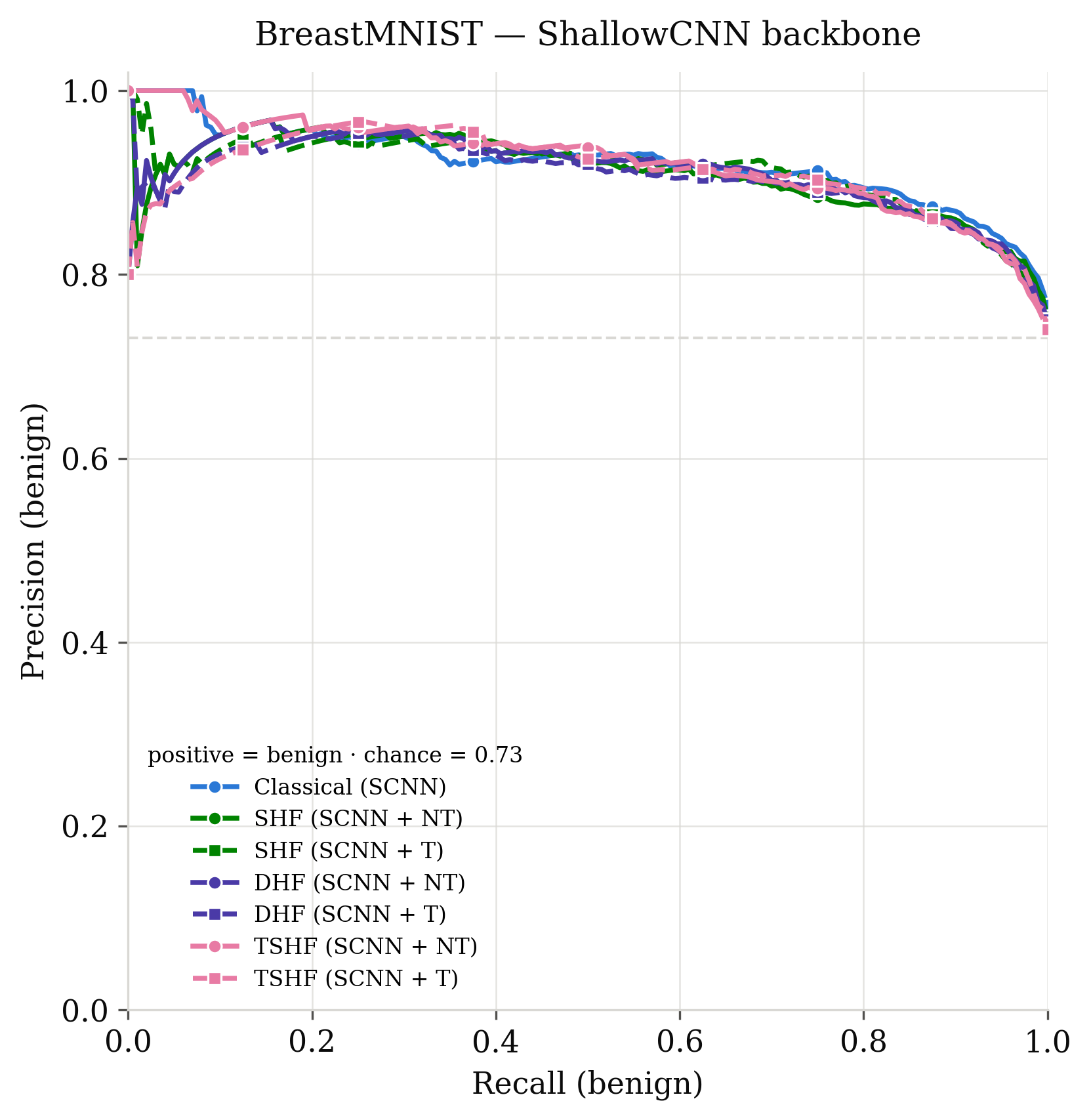}}

    \vspace{0.5em}

    \subfloat[Non-trainable quantum circuit\label{fig:pr_inb_nt}]{%
        \includegraphics[width=0.48\linewidth]{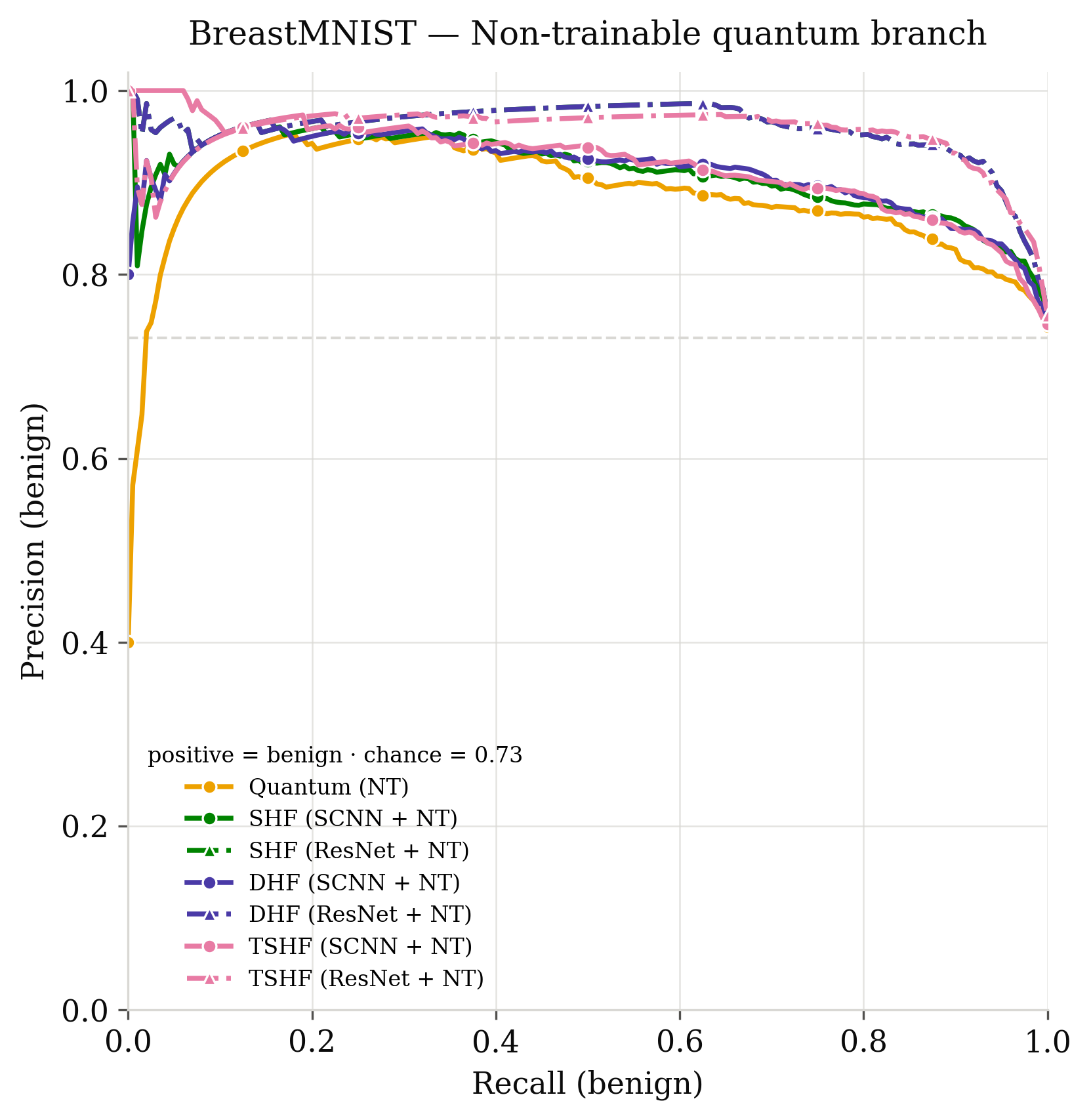}}
    \hfill
    \subfloat[Trainable quantum circuit\label{fig:pr_inb_t}]{%
        \includegraphics[width=0.48\linewidth]{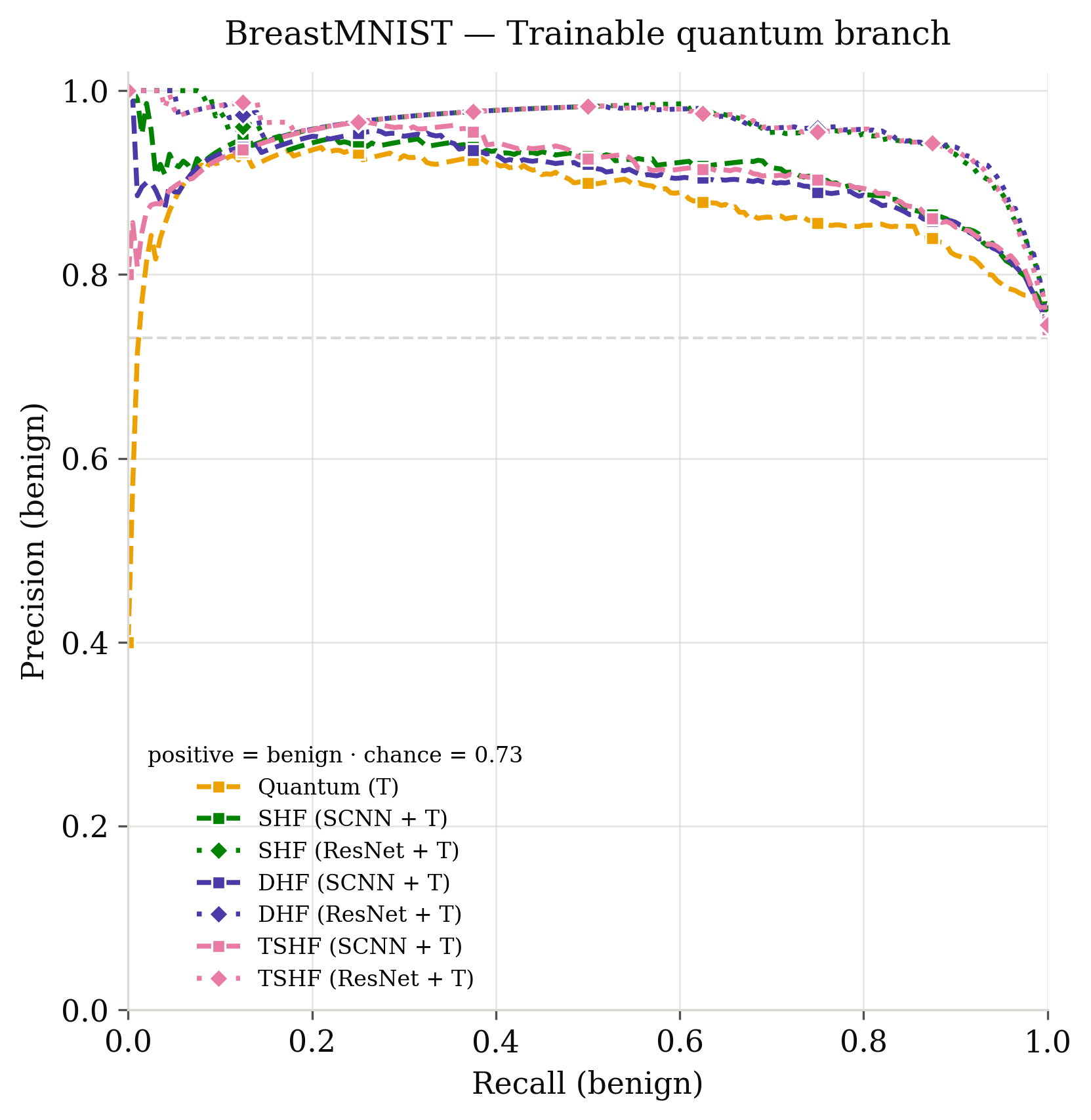}}

    \caption{Precision--recall curves on the INbreast test set, averaged over
    five independent runs. The dashed line marks the no-skill baseline, given by
    the positive-class prevalence of the split (chance $= 0.48$).}
    \label{fig:pr_inbreast_groups}
\end{figure*}

\begin{figure*}[htbp]
    \centering
    \subfloat[ResNet-18 backbone\label{fig:pr_inb_resnetb}]{%
        \includegraphics[width=0.48\linewidth]{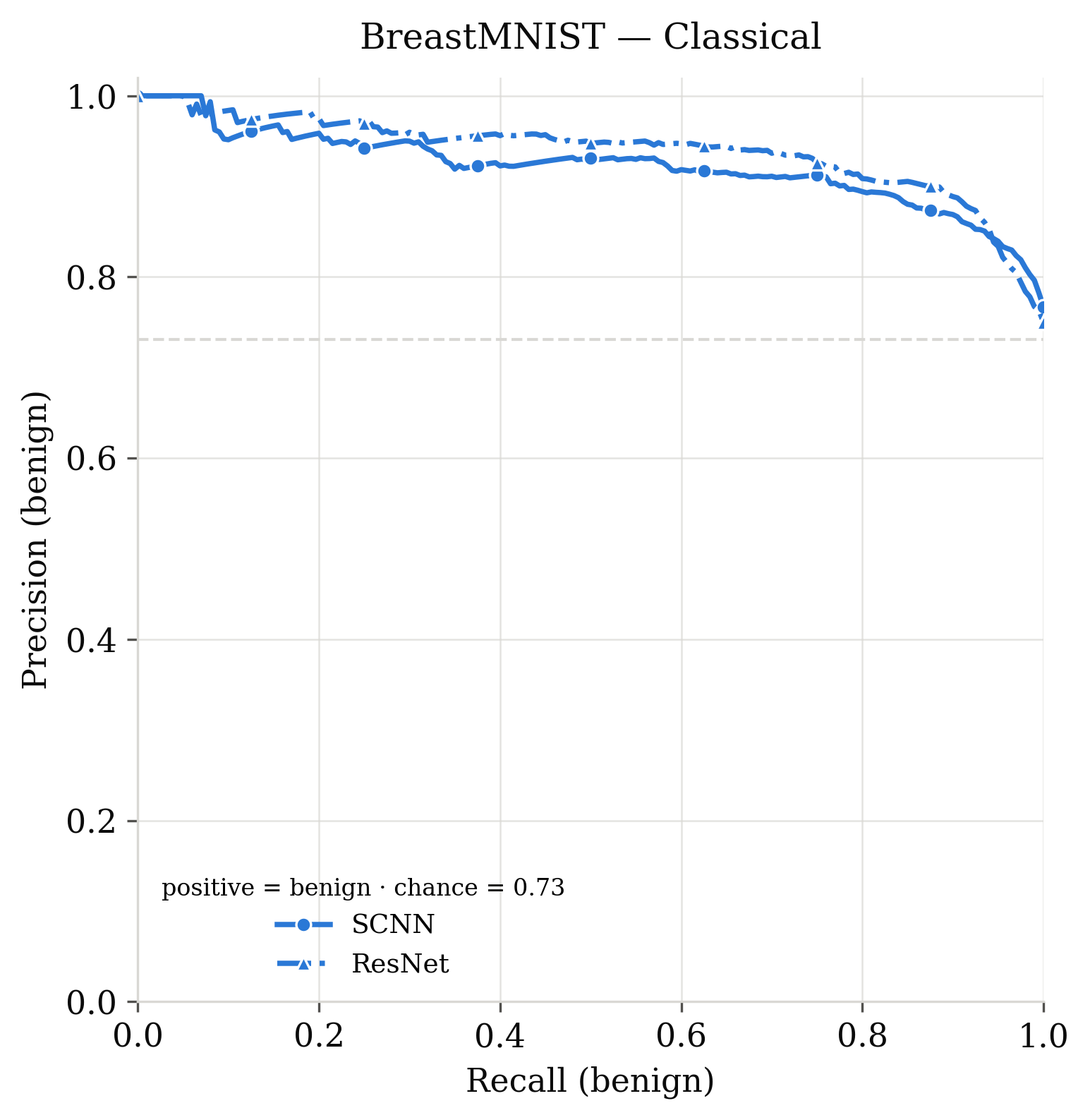}}
    \hfill
    \subfloat[SCNN backbone\label{fig:pr_inb_scnnb}]{%
        \includegraphics[width=0.48\linewidth]{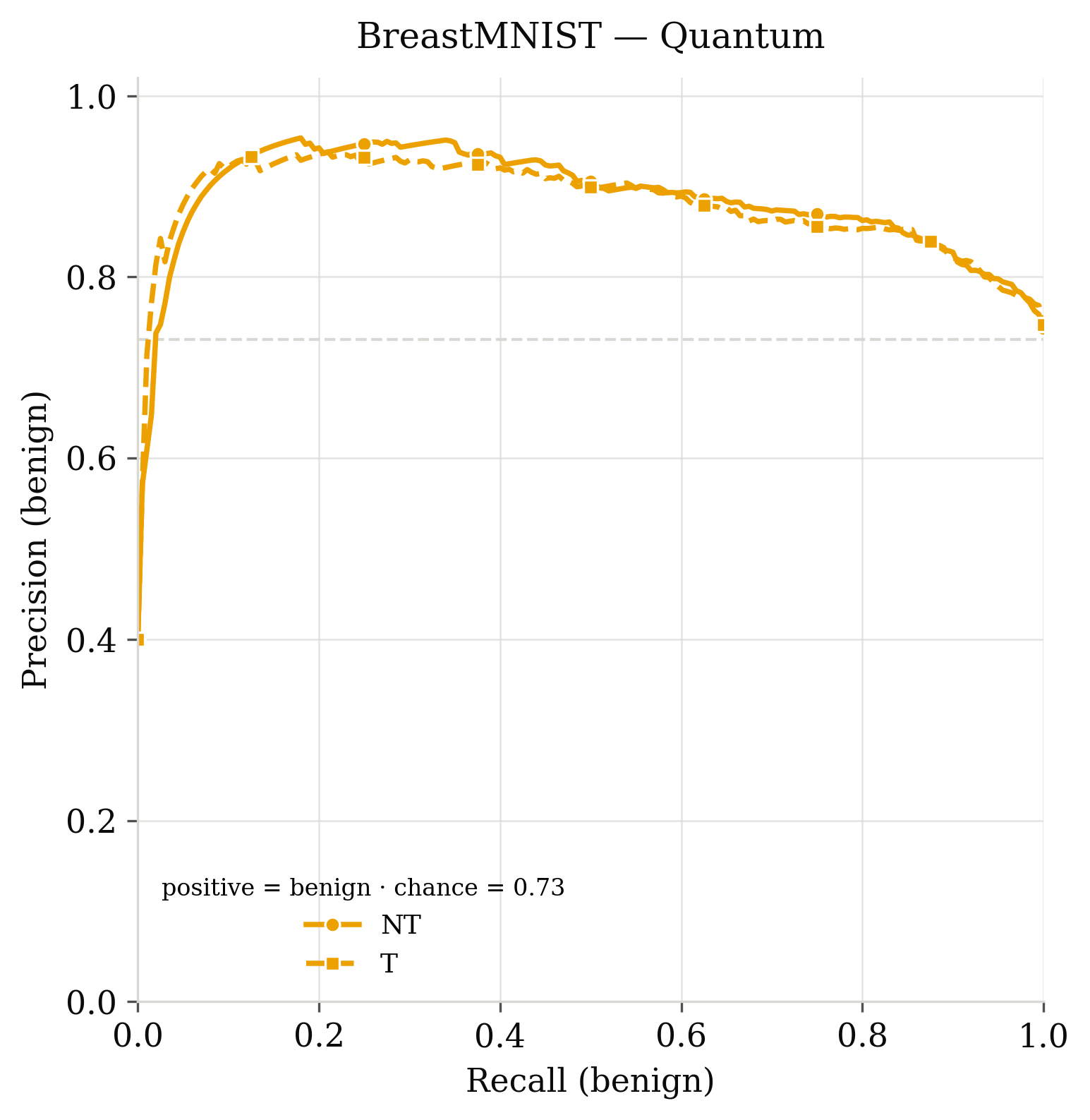}}

    \caption{Precision--recall curves for baseline models on the BreastMNIST test set, averaged over
    five independent runs. }
    \label{fig:pr_inbreast_groupsb}
\end{figure*}

\clearpage

\section{Additional UMAP Projections}
\label{app:umap}

\begin{figure}[htbp]
    \centering
    \subfloat[Classical: ResNet-18]{%
        \includegraphics[width=0.48\linewidth]{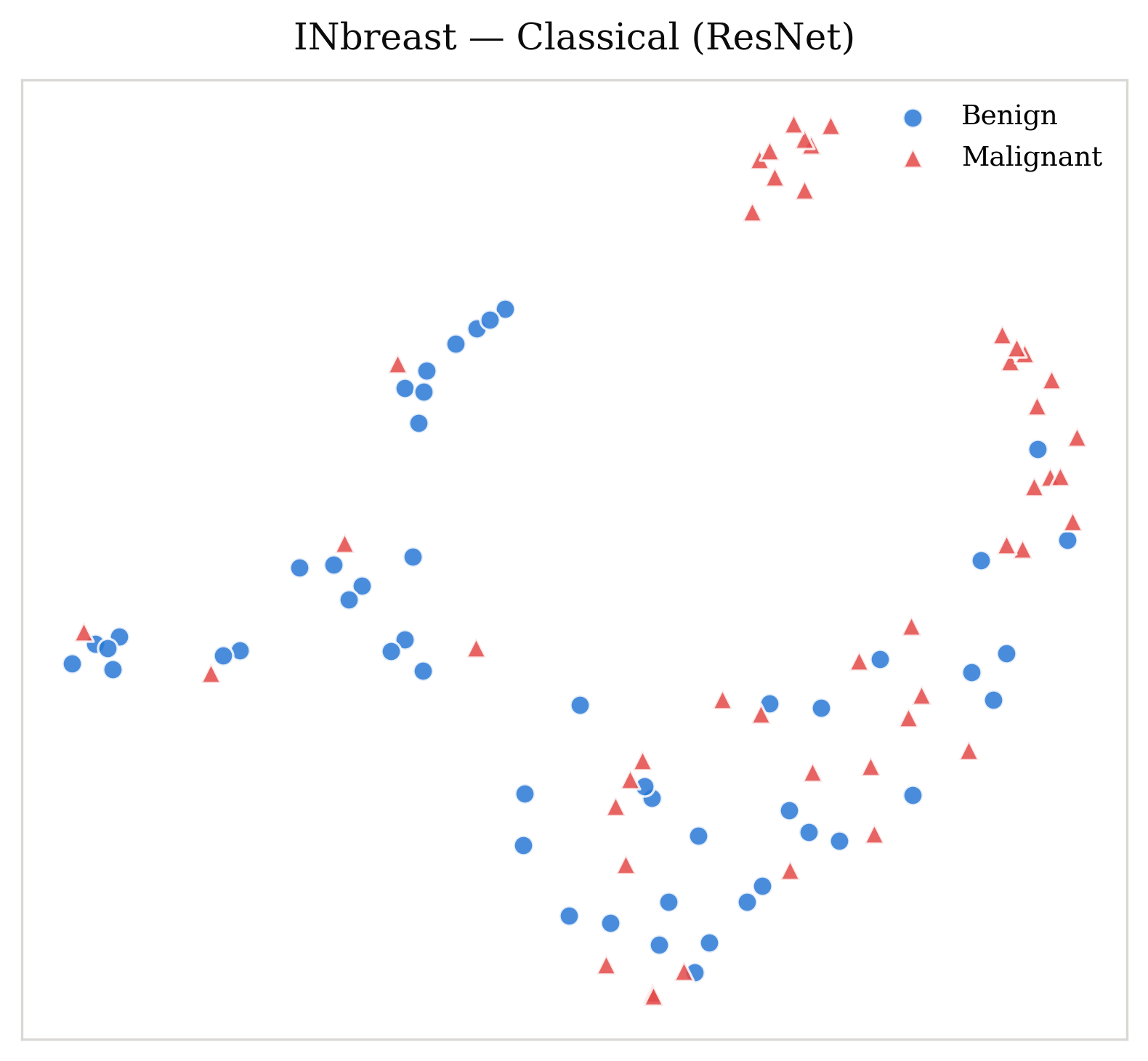}}
    \hfill
    \subfloat[Quantum: non-trainable]{%
        \includegraphics[width=0.48\linewidth]{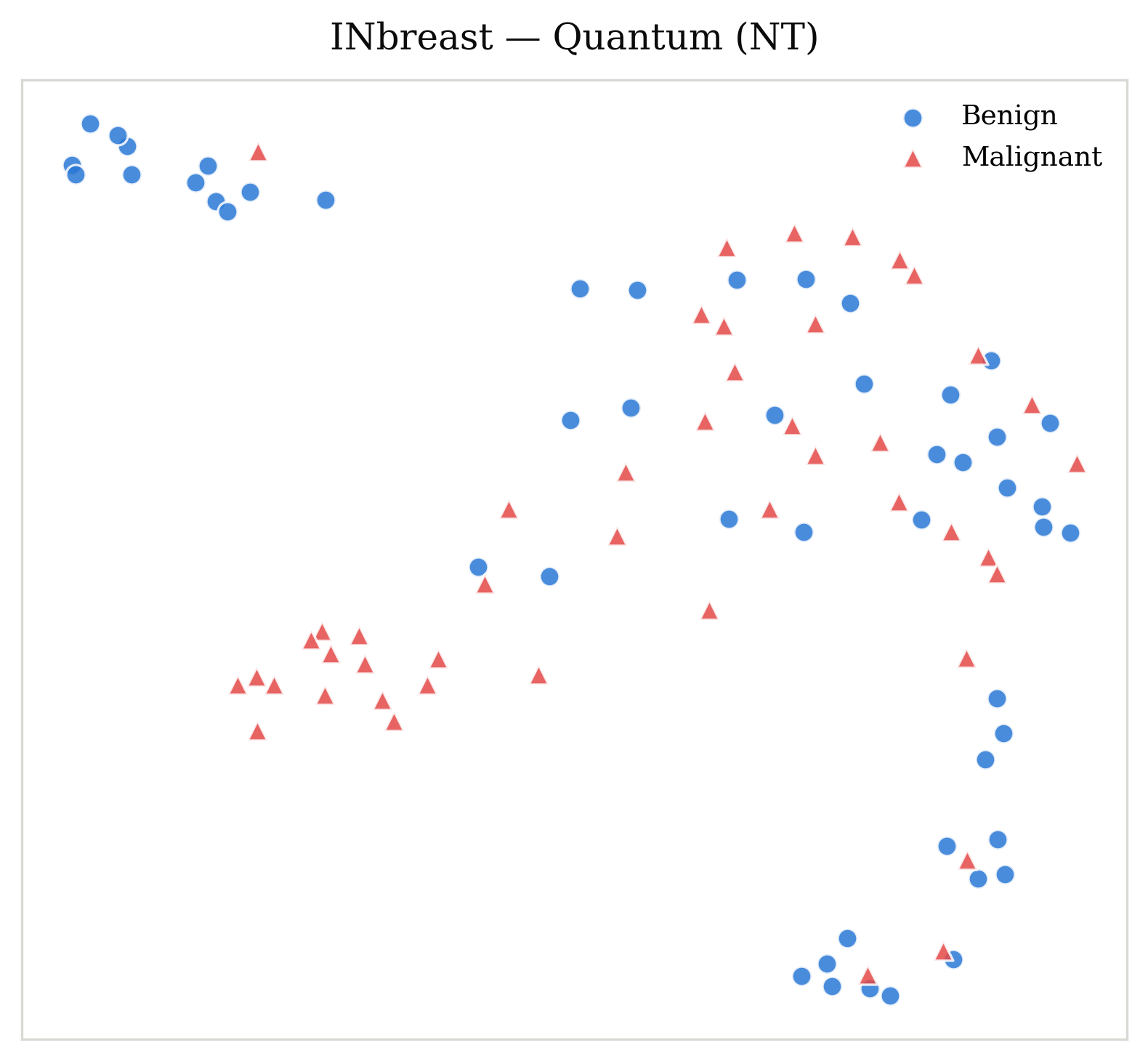}}

    \vspace{0.4em}

    \subfloat[Classical: SCNN]{%
        \includegraphics[width=0.48\linewidth]{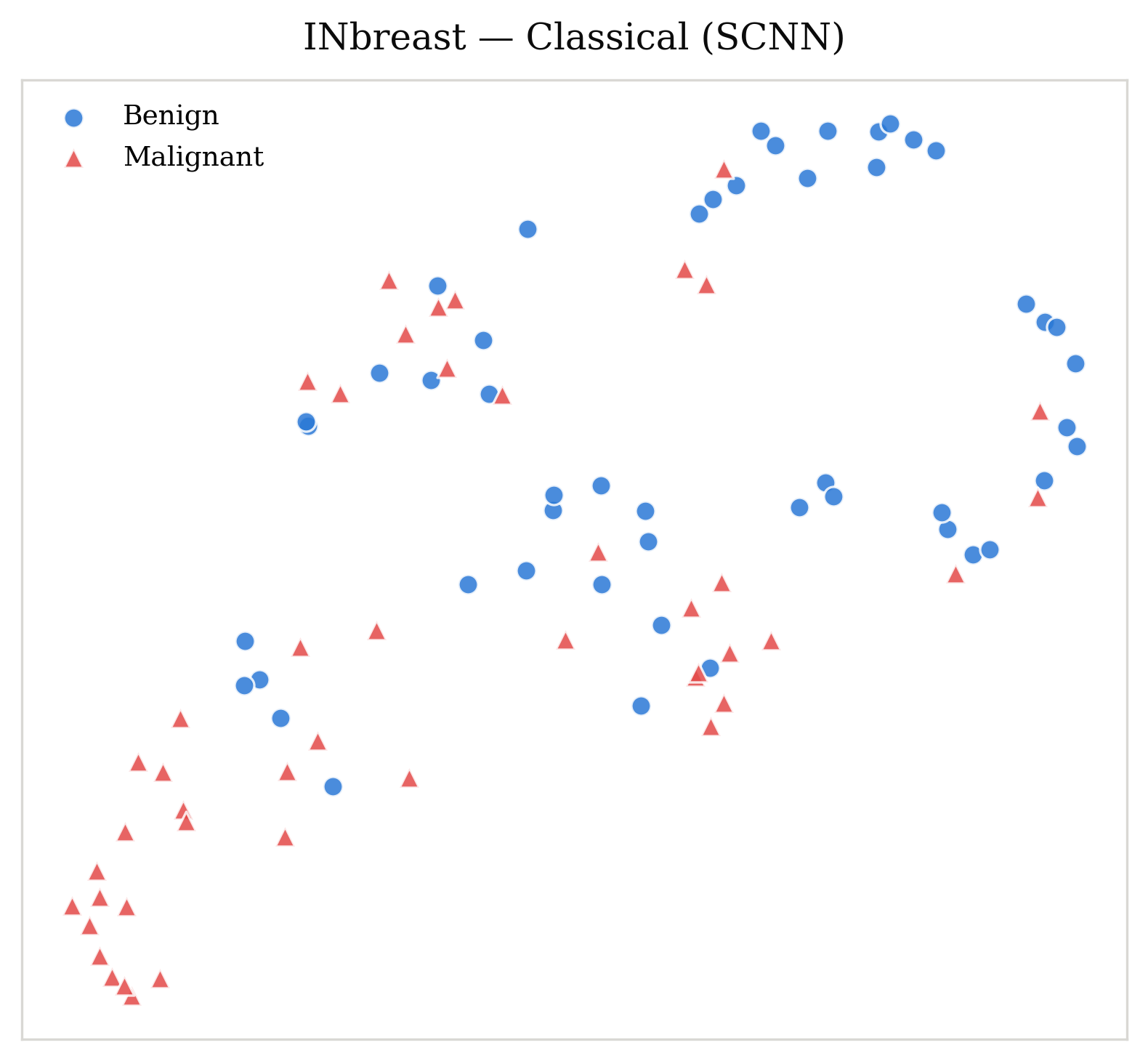}}
    \hfill
    \subfloat[Quantum: trainable]{%
        \includegraphics[width=0.48\linewidth]{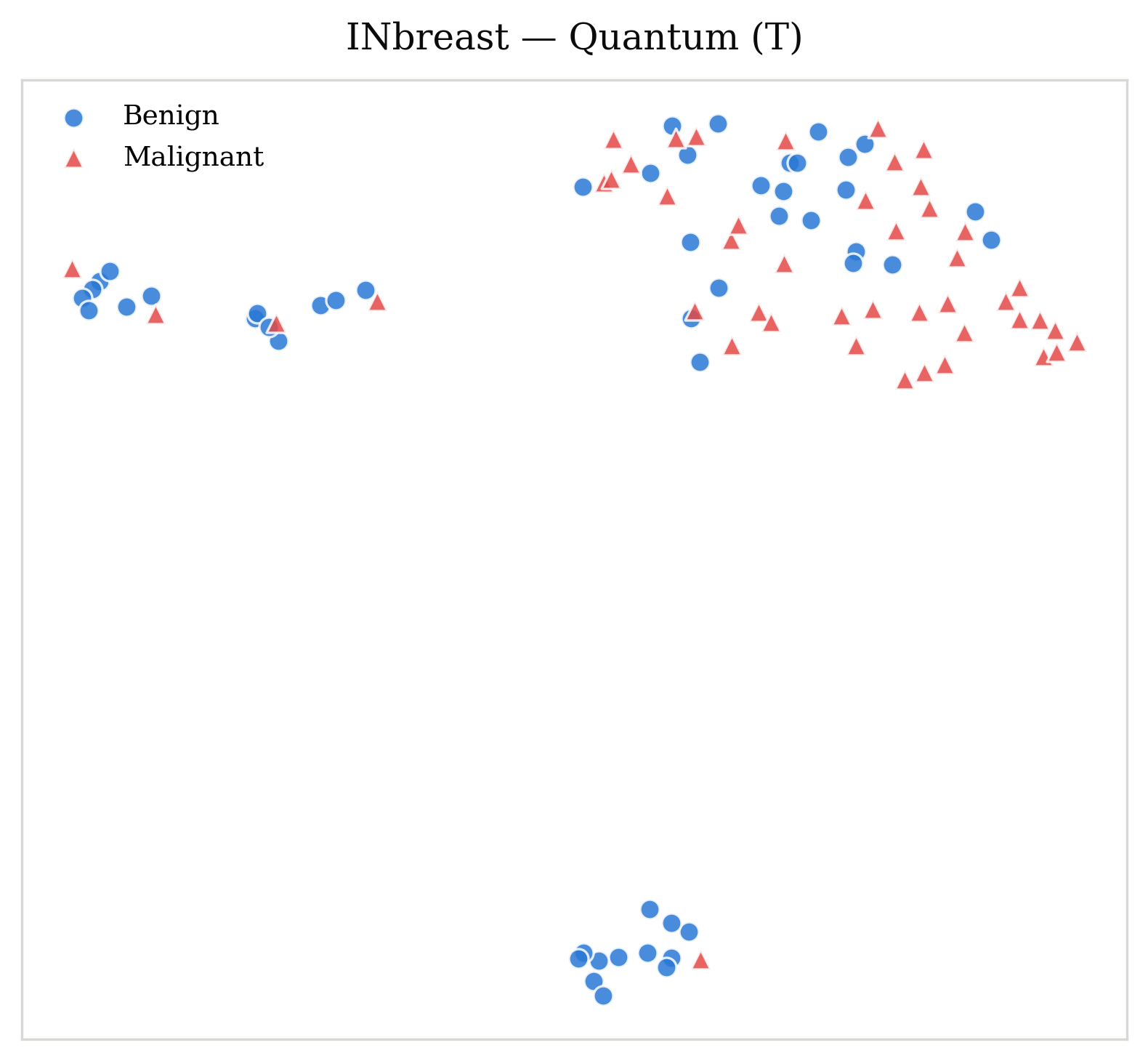}}

    \caption{UMAP projections of the standalone baseline embeddings on the
    INbreast test set.}
    \label{fig:umap_inbreast_baselines}
\end{figure}

\begin{figure}[htbp]
    \centering
    \subfloat[ResNet-18 + non-trainable]{%
        \includegraphics[width=0.48\linewidth]{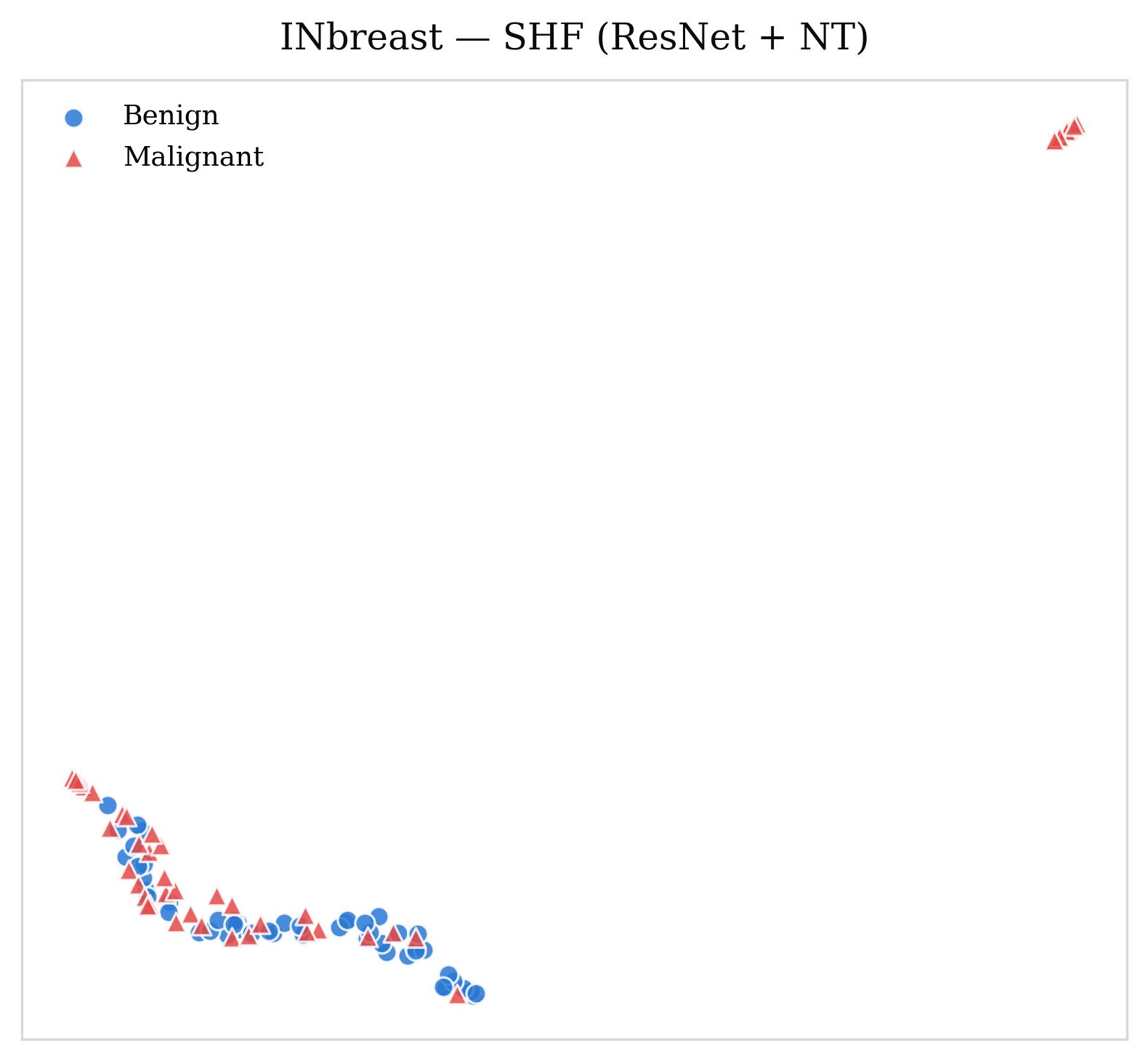}}
    \hfill
    \subfloat[ResNet-18 + trainable]{%
        \includegraphics[width=0.48\linewidth]{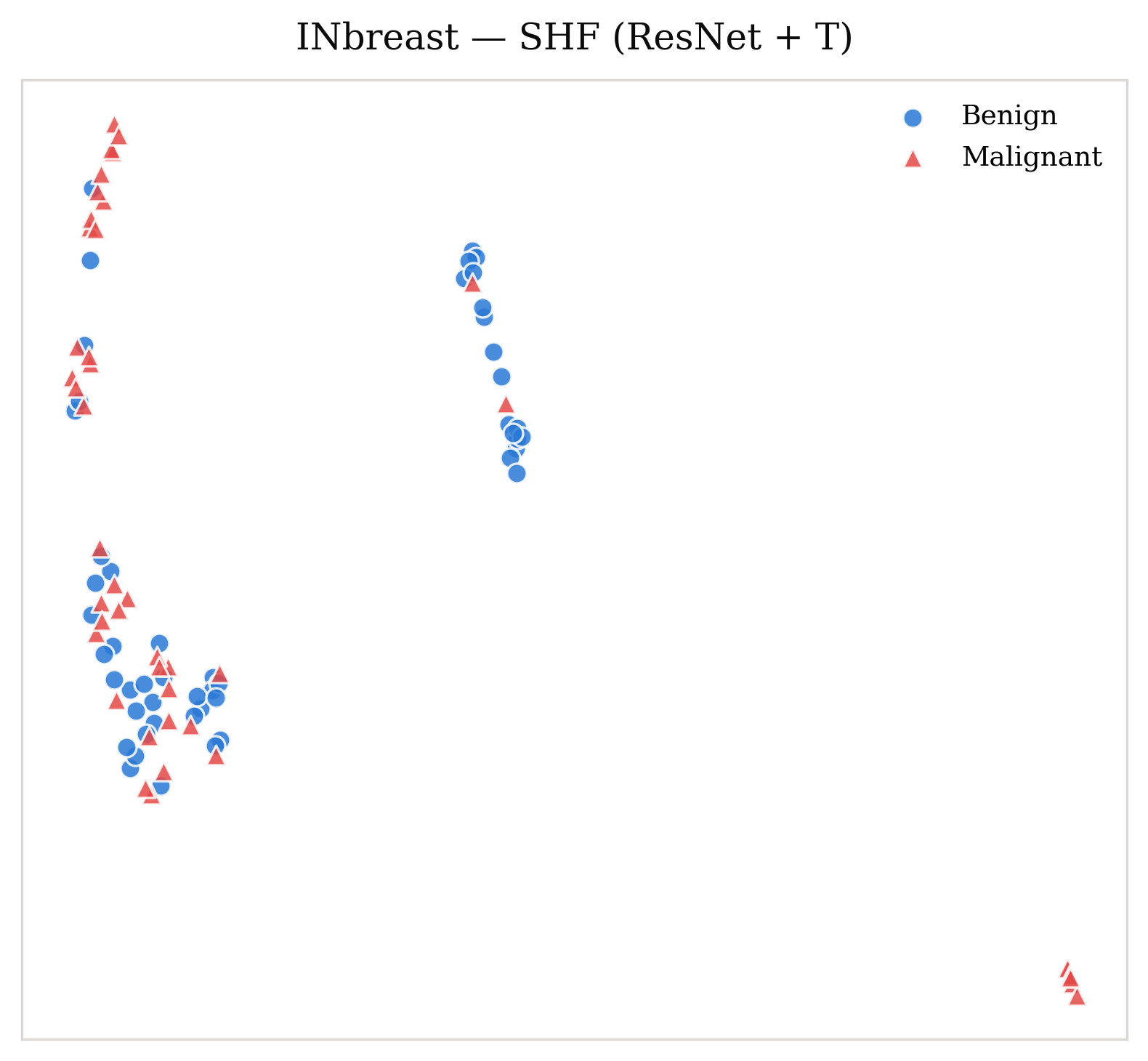}}

    \vspace{0.4em}

    \subfloat[SCNN + non-trainable]{%
        \includegraphics[width=0.48\linewidth]{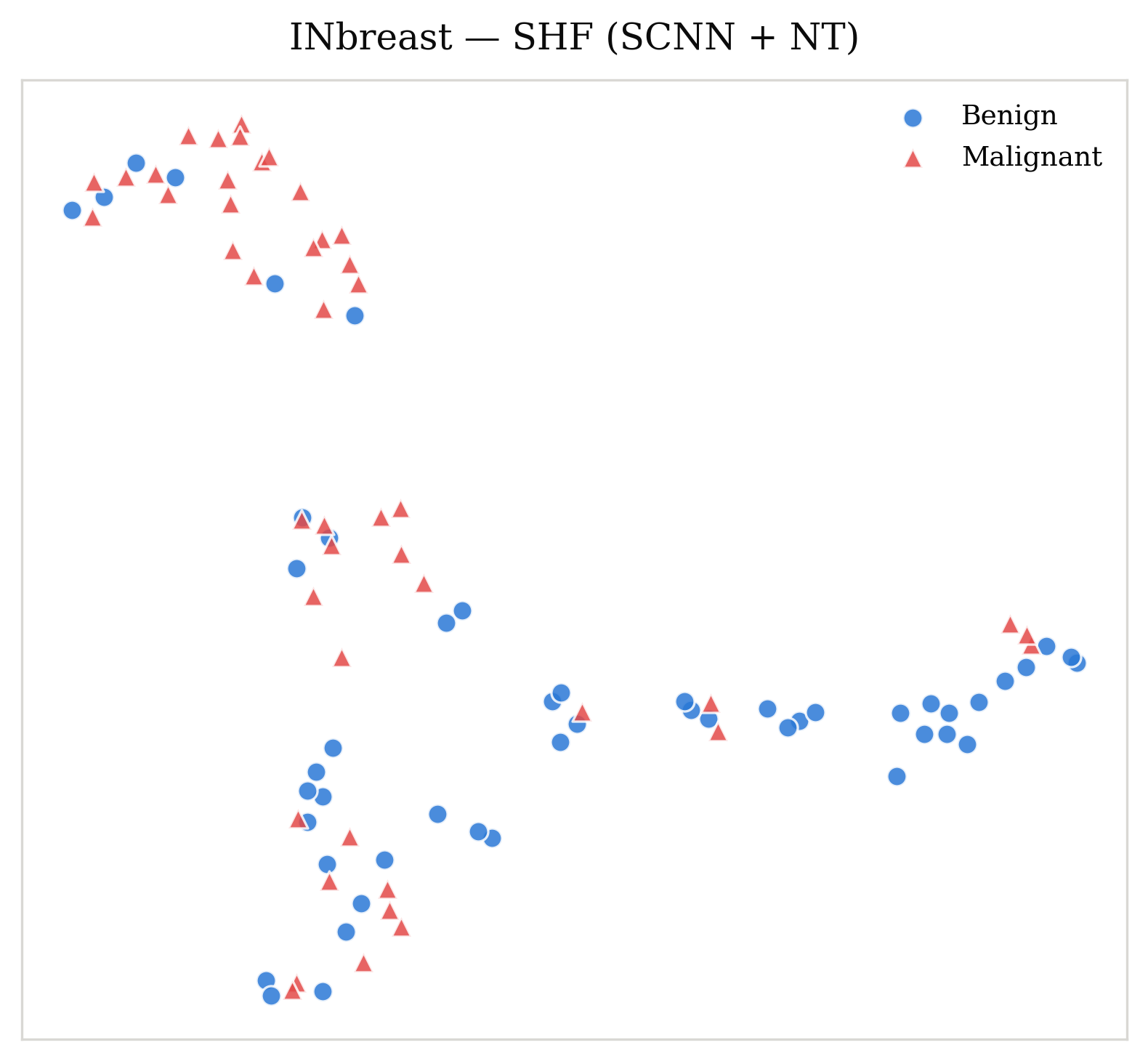}}
    \hfill
    \subfloat[SCNN + trainable]{%
        \includegraphics[width=0.48\linewidth]{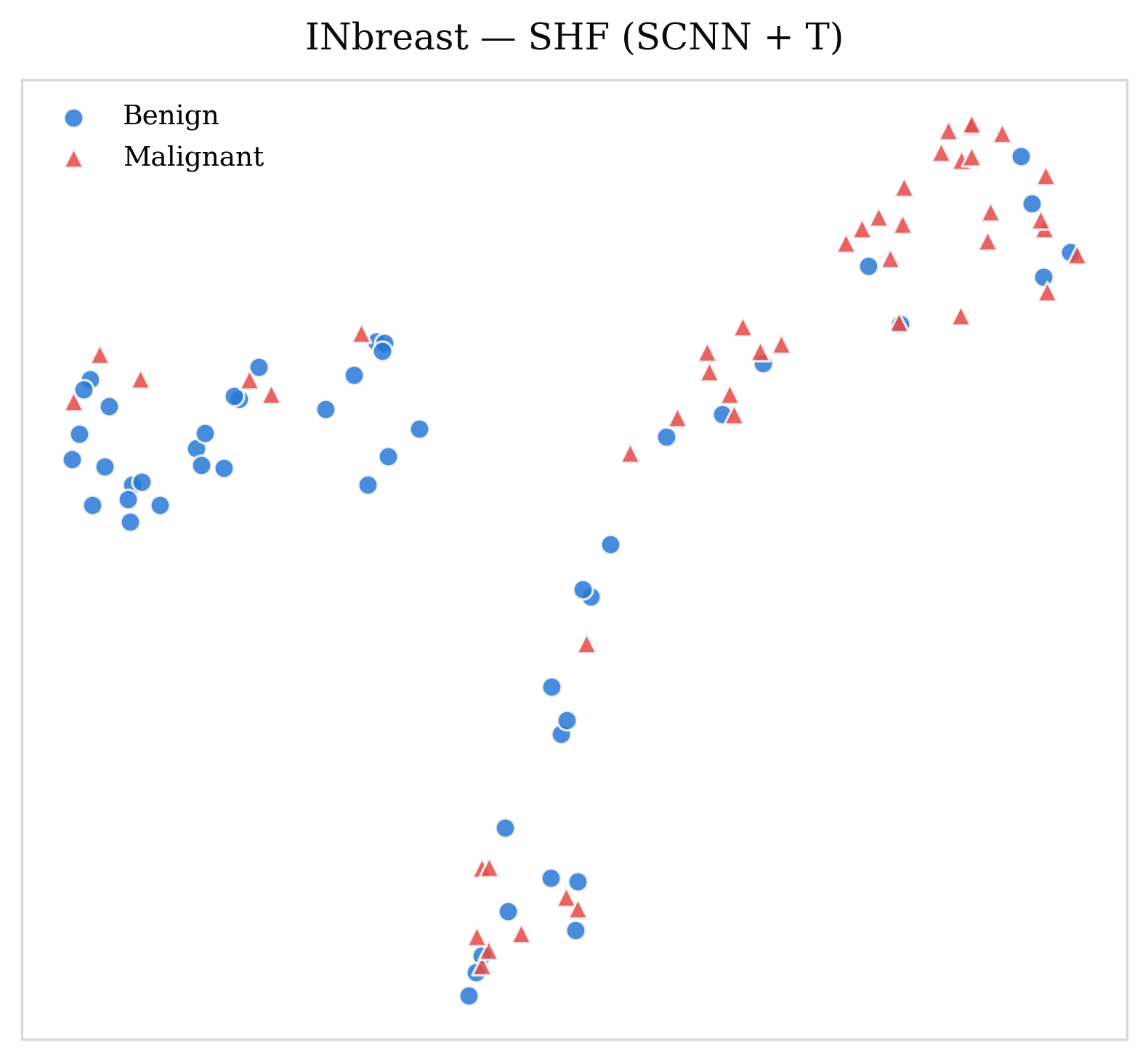}}

    \caption{UMAP projections of the fused embeddings on INbreast under the
    \textbf{SHF} strategy.}
    \label{fig:umap_inbreast_shf}
\end{figure}

\begin{figure}[htbp]
    \centering
    \subfloat[ResNet-18 + non-trainable]{%
        \includegraphics[width=0.48\linewidth]{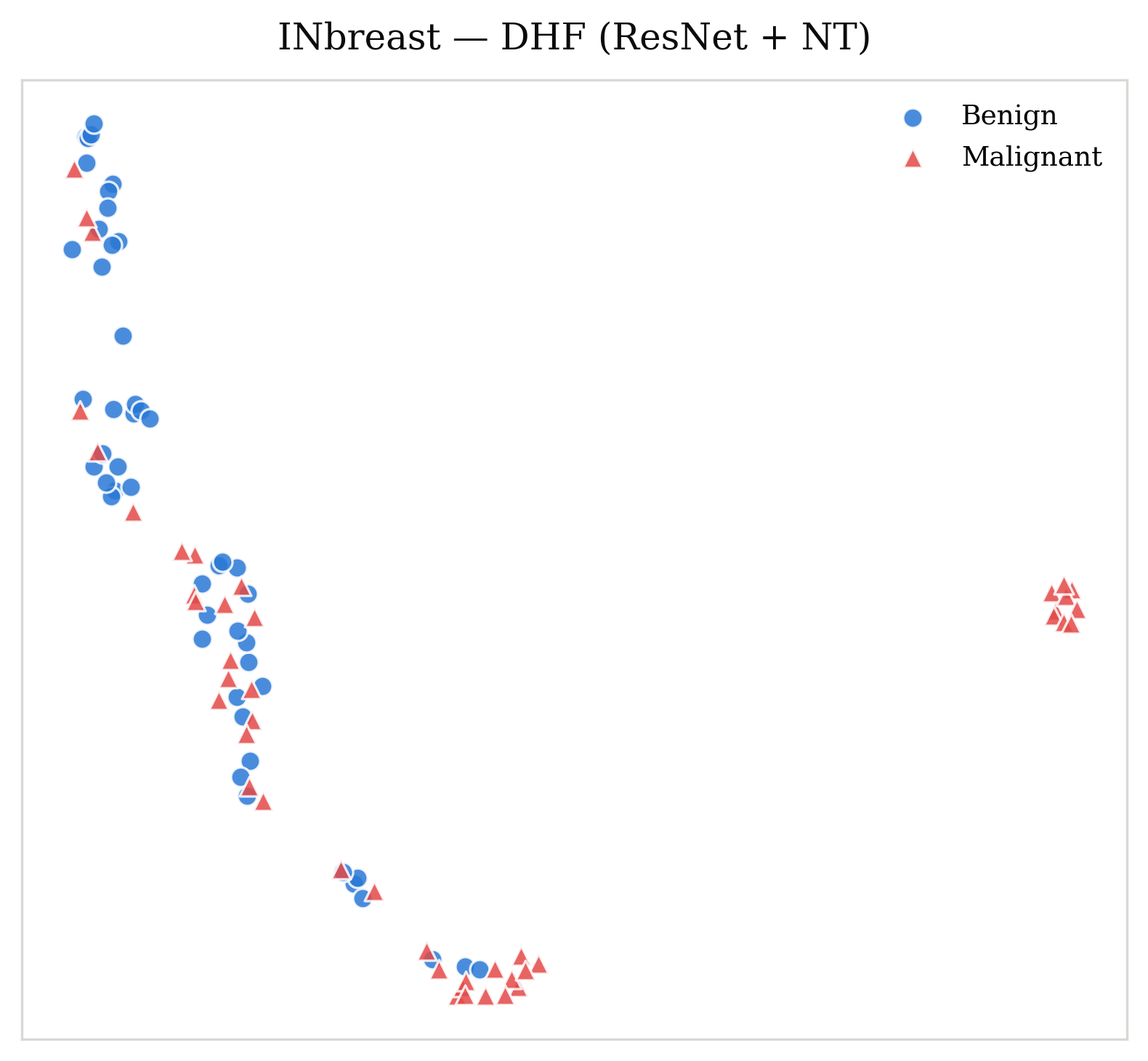}}
    \hfill
    \subfloat[ResNet-18 + trainable]{%
        \includegraphics[width=0.48\linewidth]{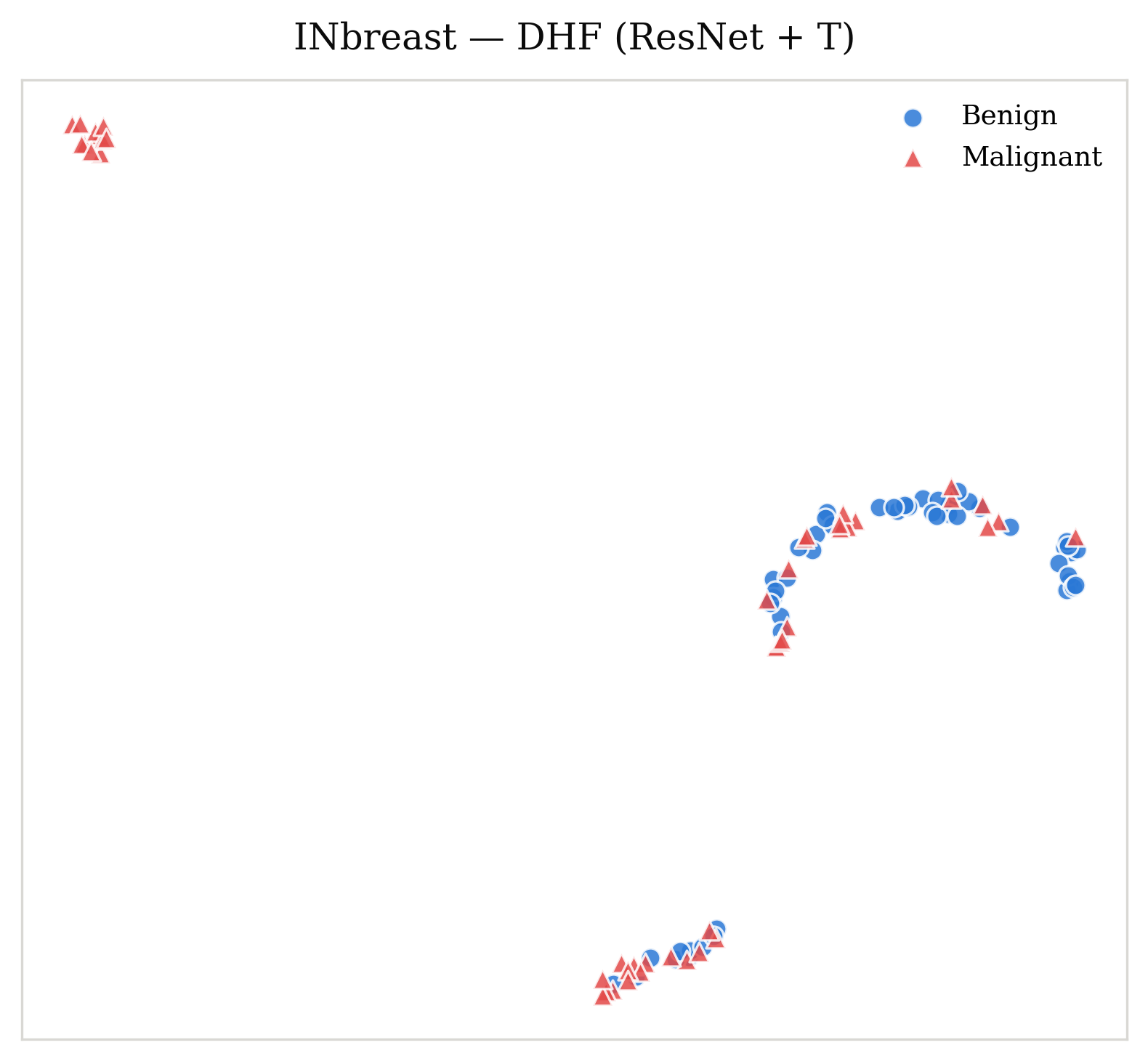}}

    \vspace{0.4em}

    \subfloat[SCNN + non-trainable]{%
        \includegraphics[width=0.48\linewidth]{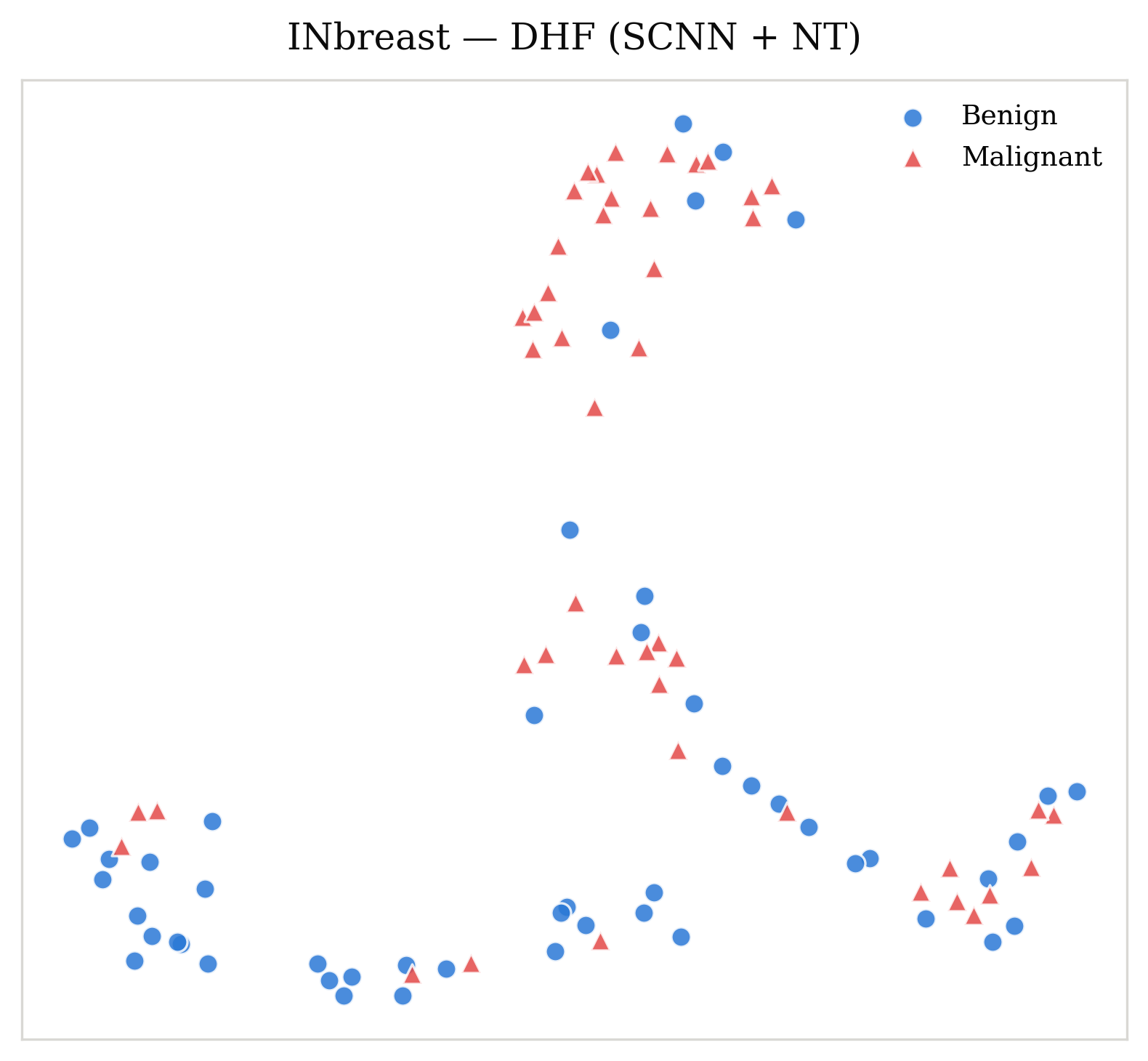}}
    \hfill
    \subfloat[SCNN + trainable]{%
        \includegraphics[width=0.48\linewidth]{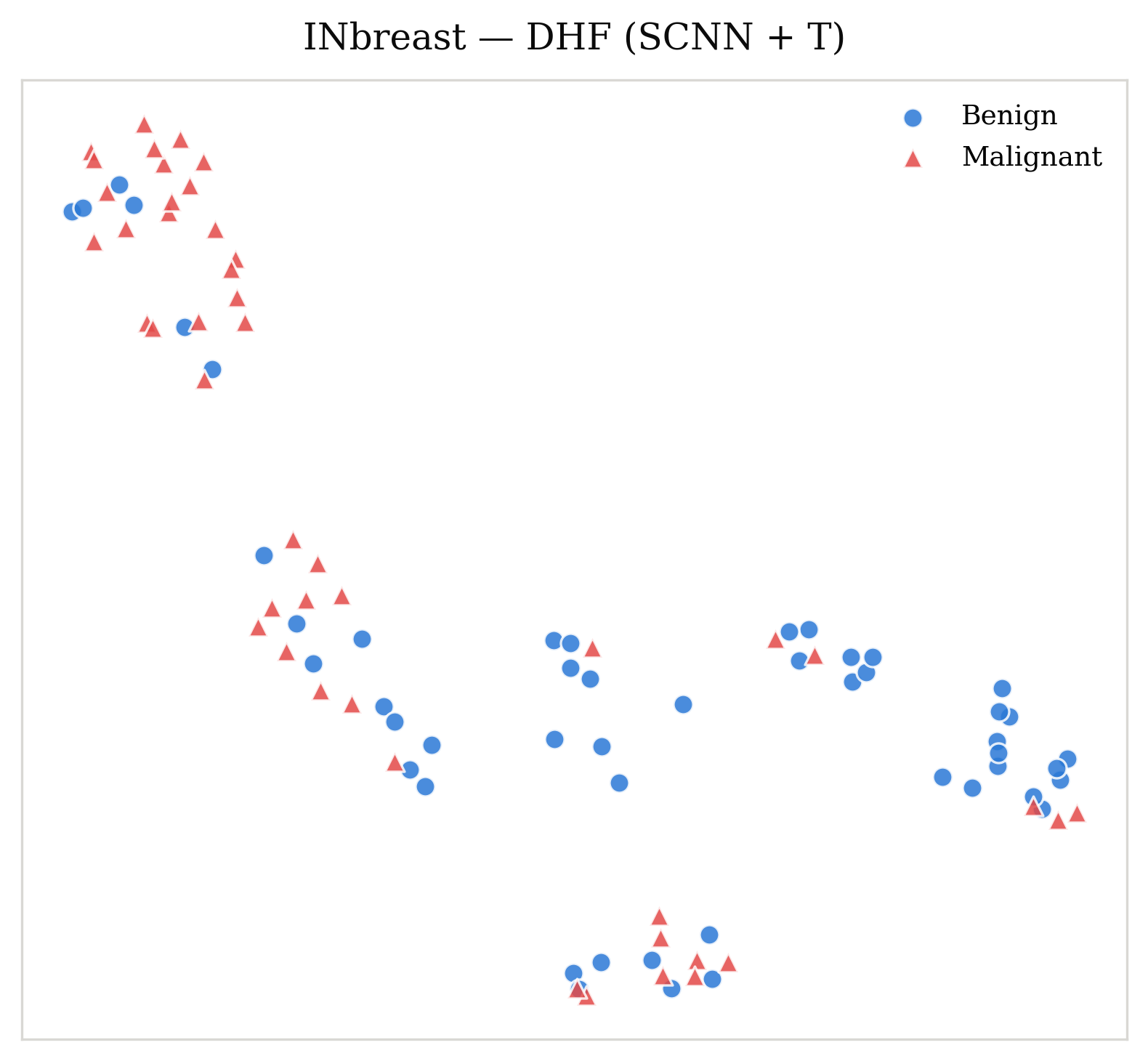}}

    \caption{UMAP projections of the fused embeddings on INbreast under the
    \textbf{DHF} strategy.}
    \label{fig:umap_inbreast_dhf}
\end{figure}

\begin{figure}[htbp]
    \centering
    \subfloat[ResNet-18 + non-trainable]{%
        \includegraphics[width=0.48\linewidth]{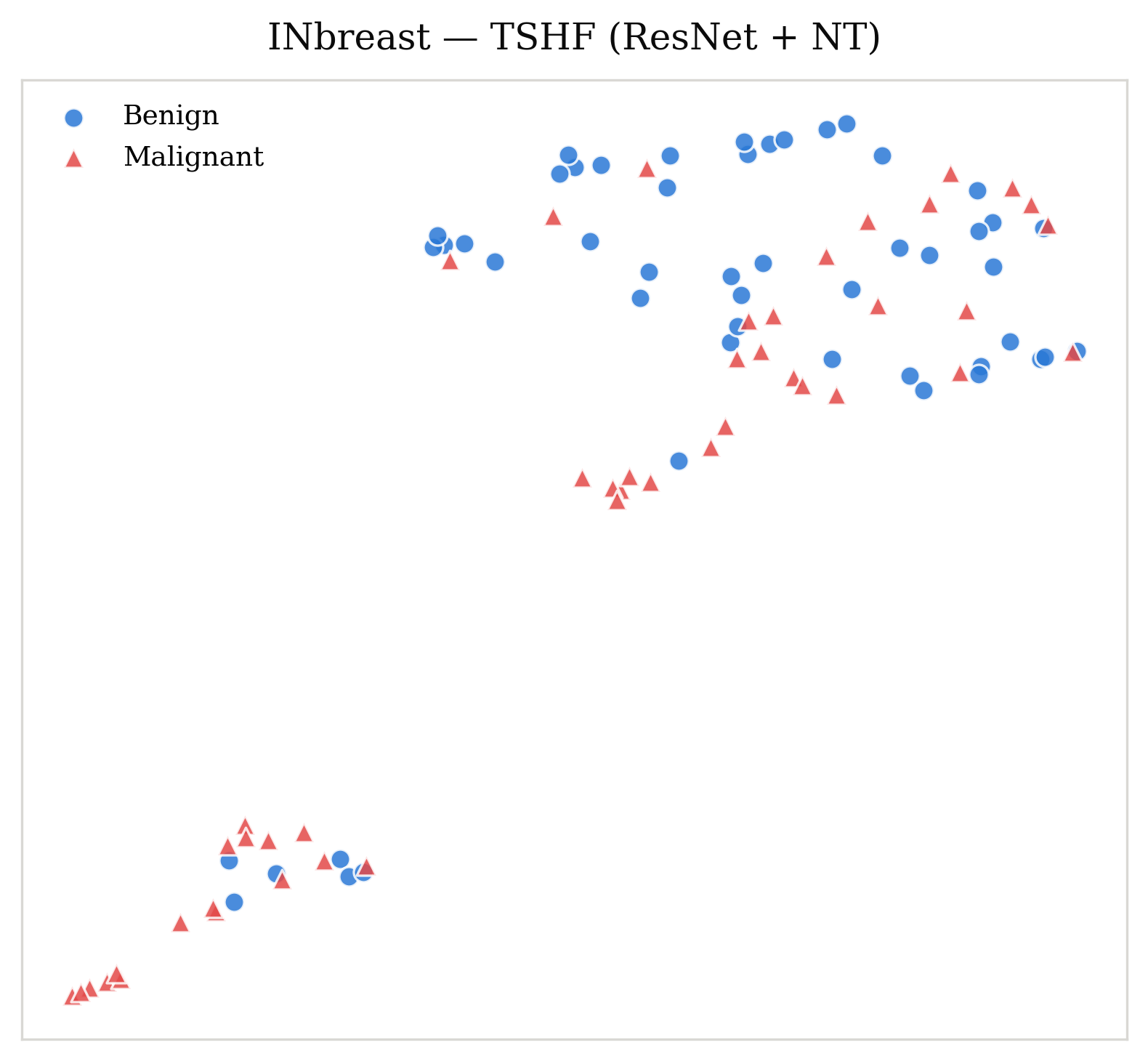}}
    \hfill
    \subfloat[ResNet-18 + trainable]{%
        \includegraphics[width=0.48\linewidth]{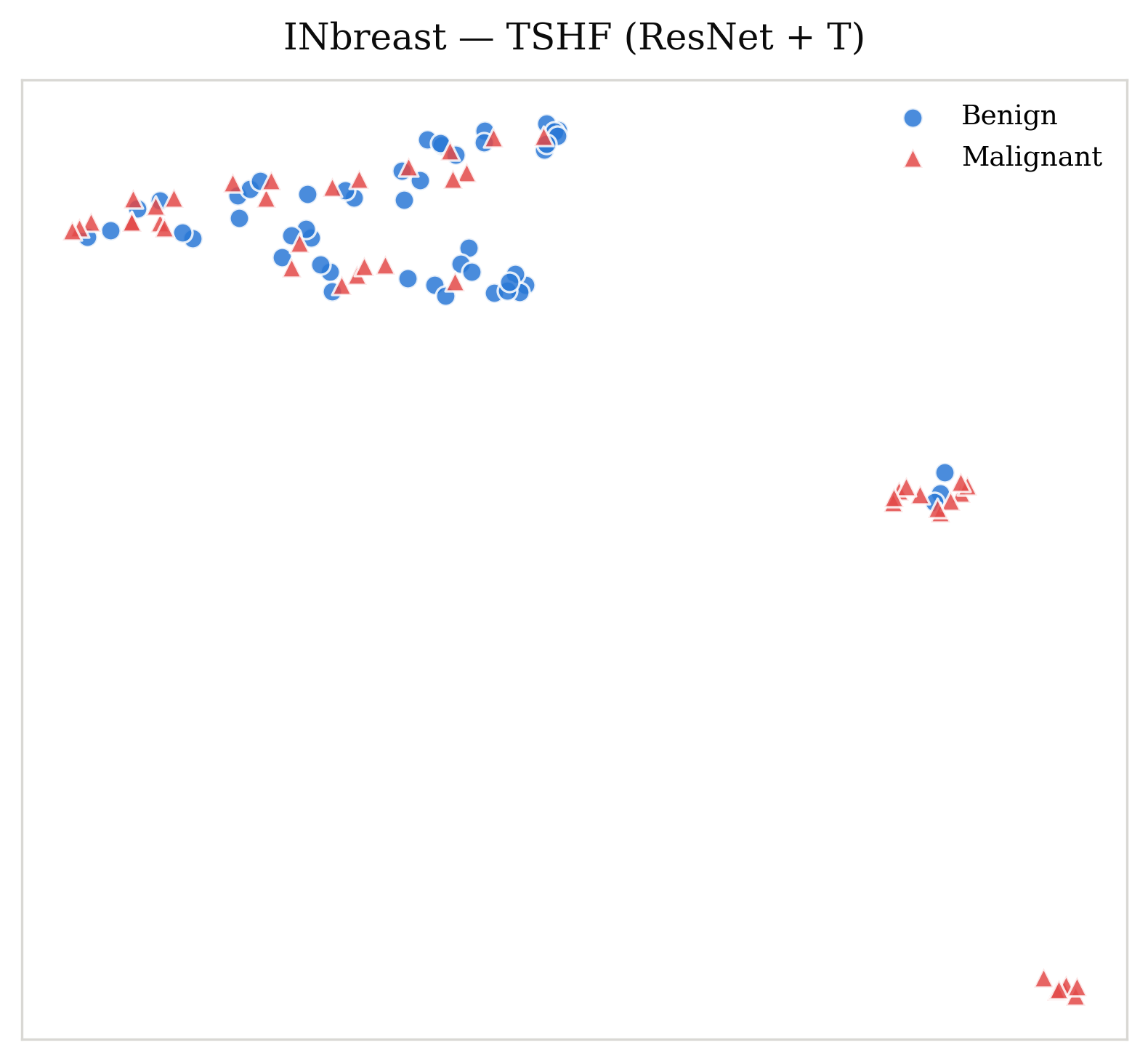}}

    \vspace{0.4em}

    \subfloat[SCNN + non-trainable]{%
        \includegraphics[width=0.48\linewidth]{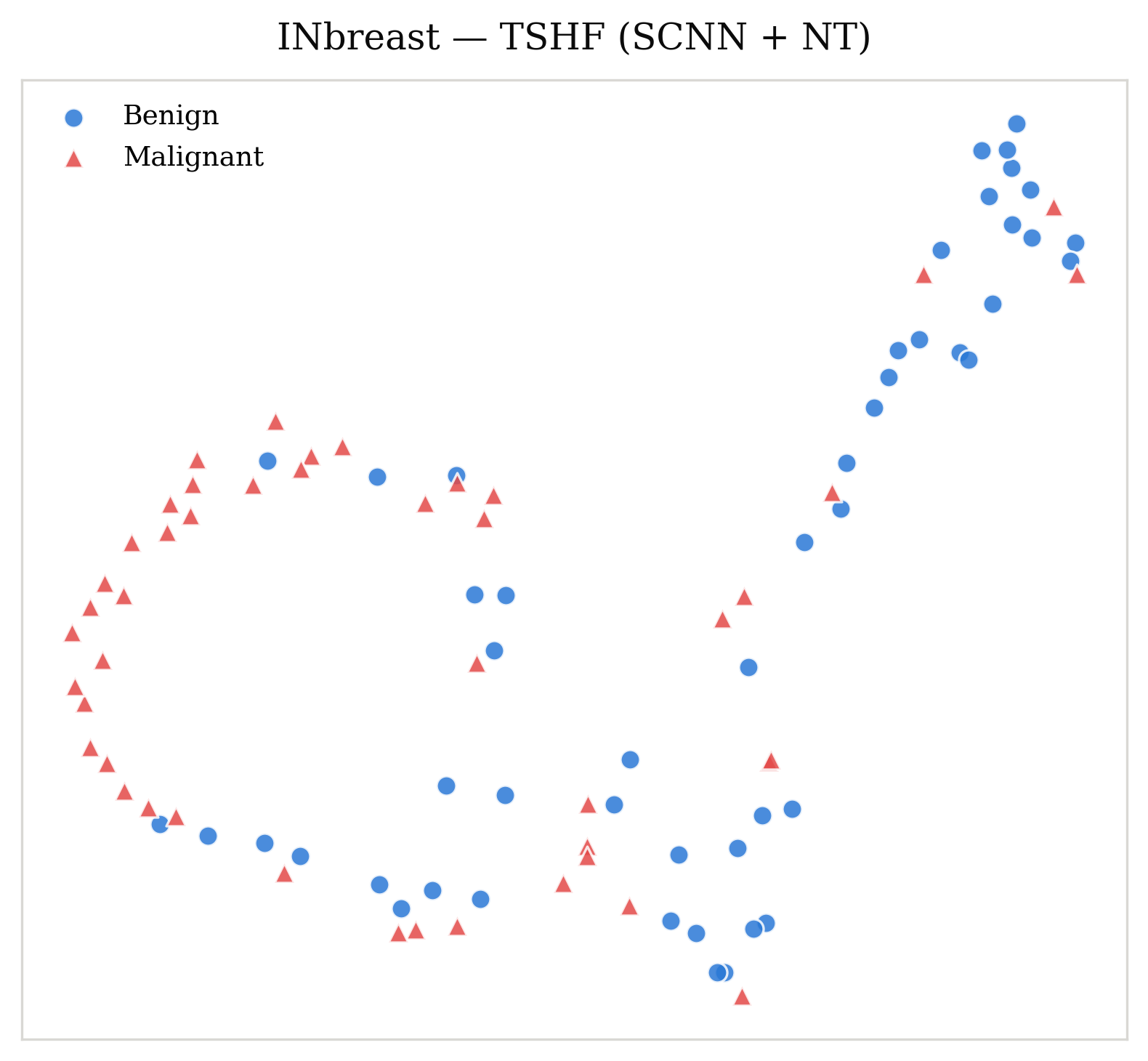}}
    \hfill
    \subfloat[SCNN + trainable]{%
        \includegraphics[width=0.48\linewidth]{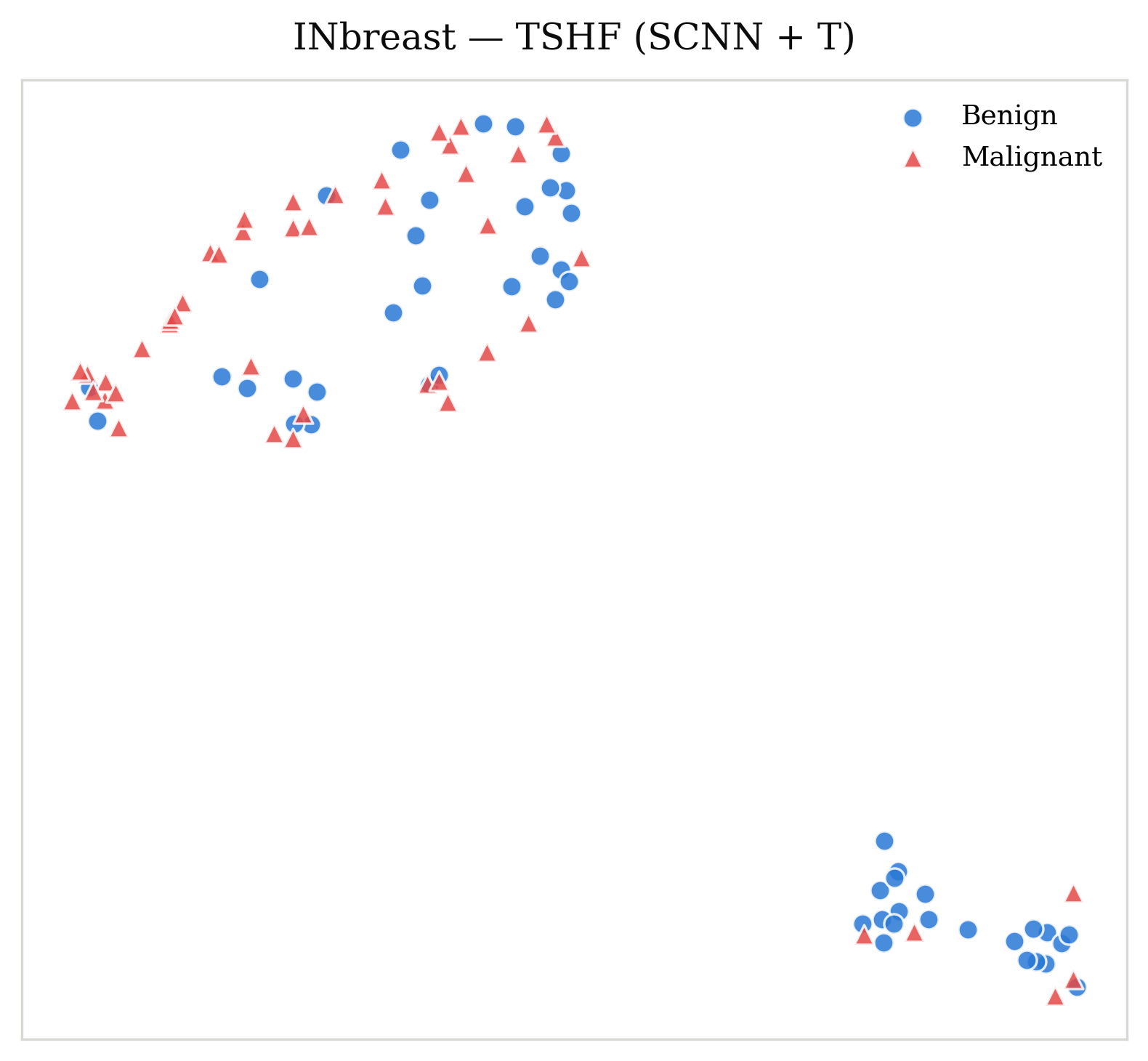}}

    \caption{UMAP projections of the fused embeddings on INbreast under the
    \textbf{TSHF} strategy.}
    \label{fig:umap_inbreast_tshf}
\end{figure}

\begin{figure}[htbp]
    \centering
    \subfloat[Classical: ResNet-18]{%
        \includegraphics[width=0.48\linewidth]{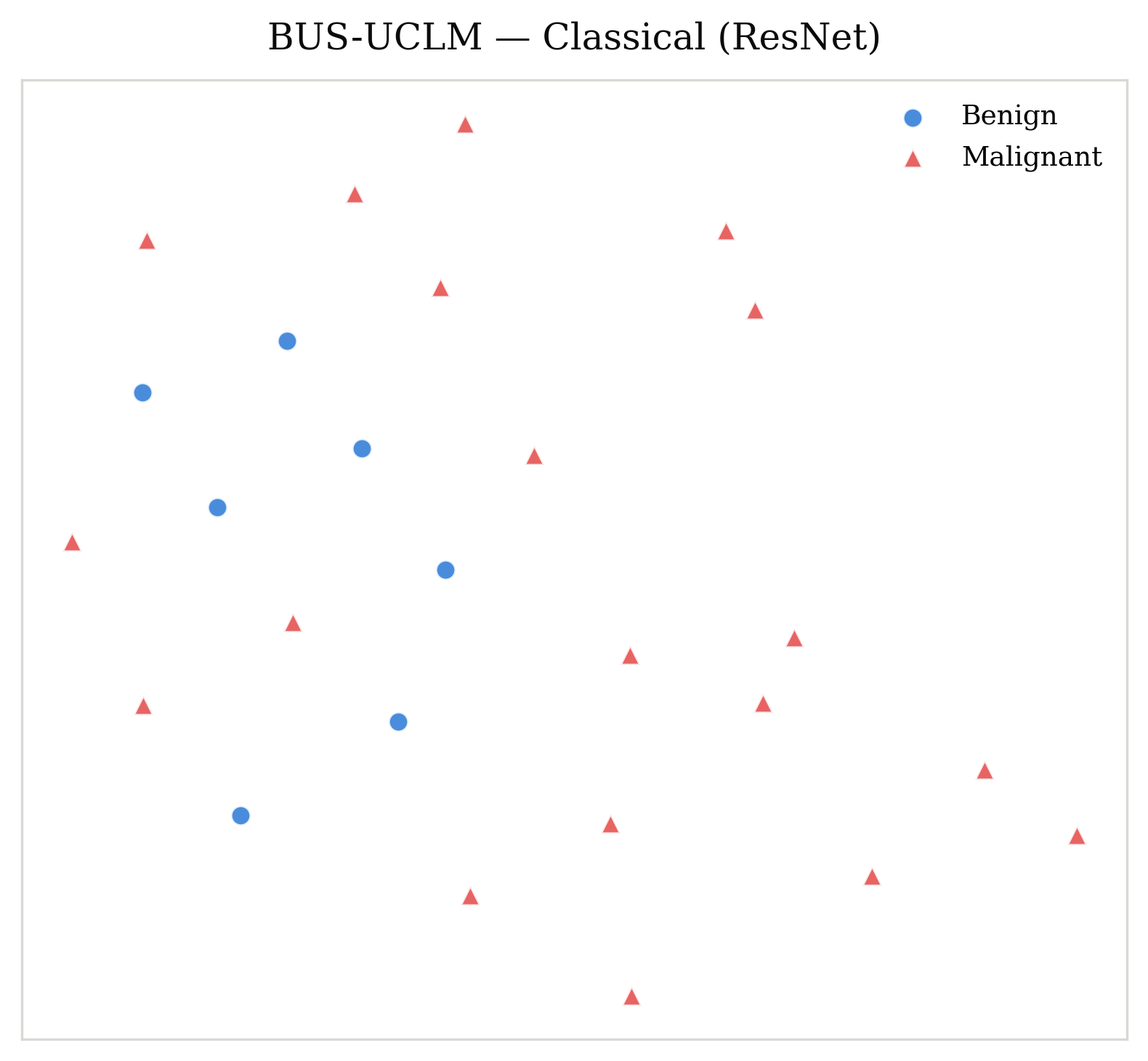}}
    \hfill
    \subfloat[Quantum: non-trainable]{%
        \includegraphics[width=0.48\linewidth]{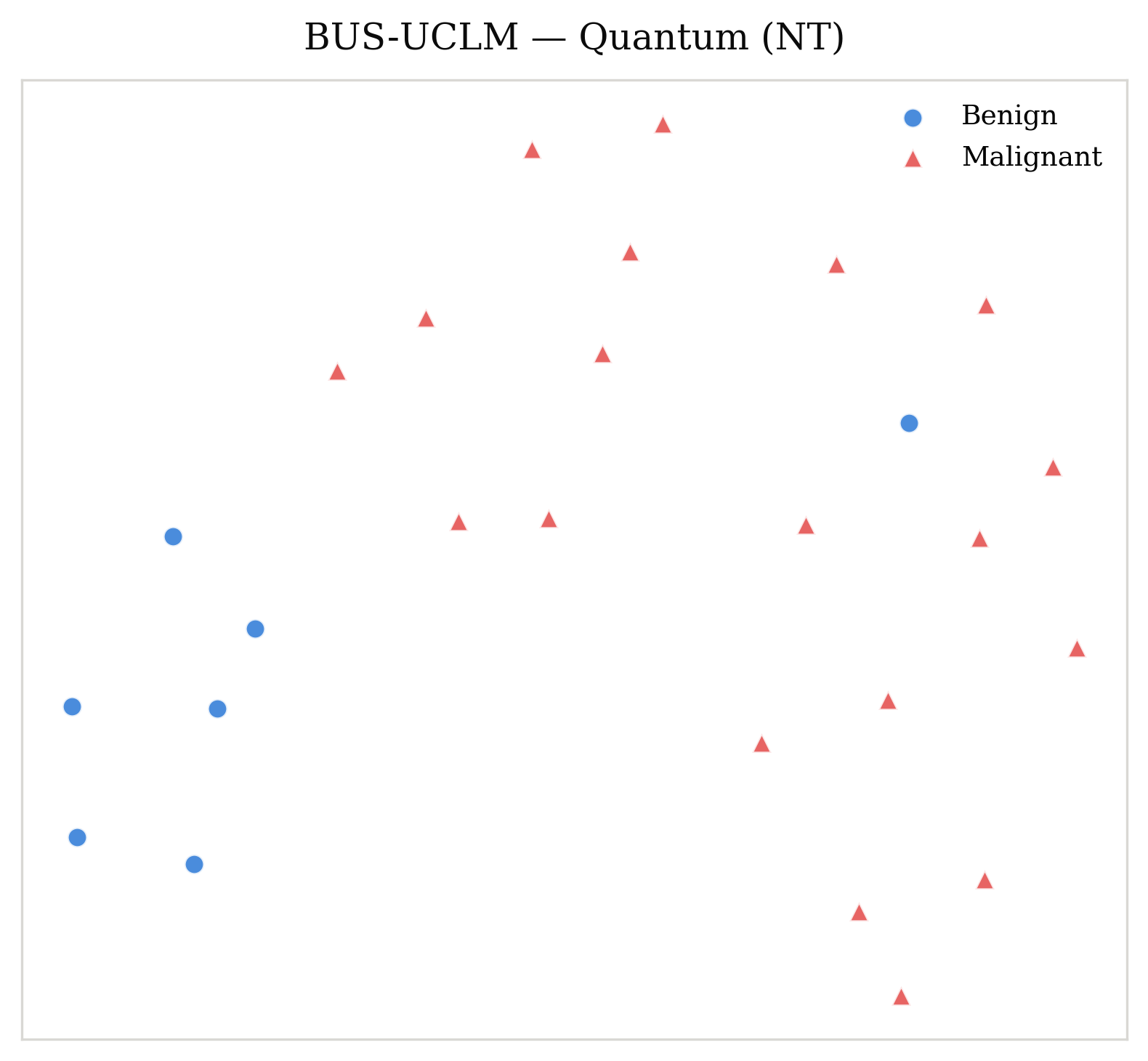}}

    \vspace{0.4em}

    \subfloat[Classical: SCNN]{%
        \includegraphics[width=0.48\linewidth]{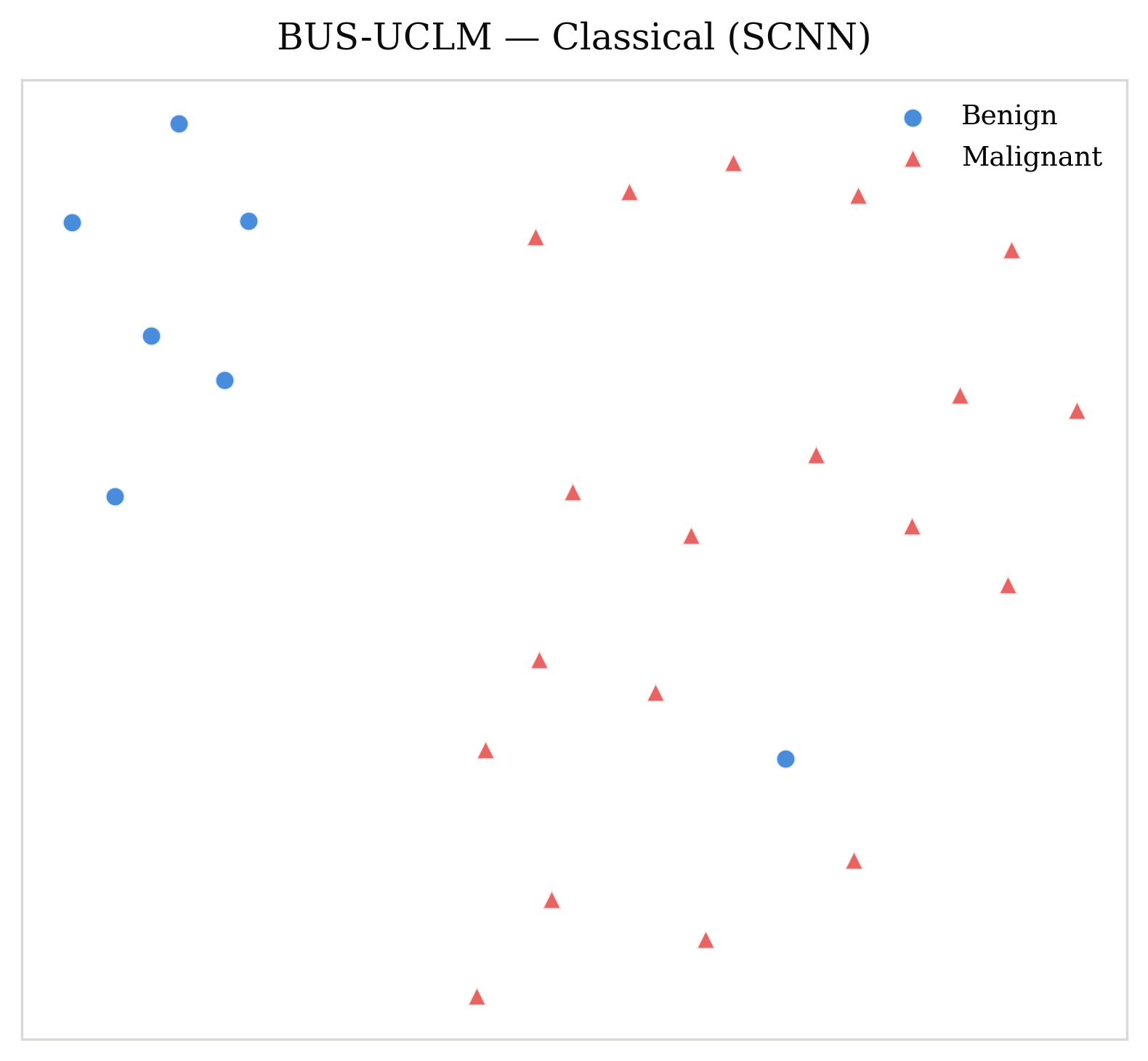}}
    \hfill
    \subfloat[Quantum: trainable]{%
        \includegraphics[width=0.48\linewidth]{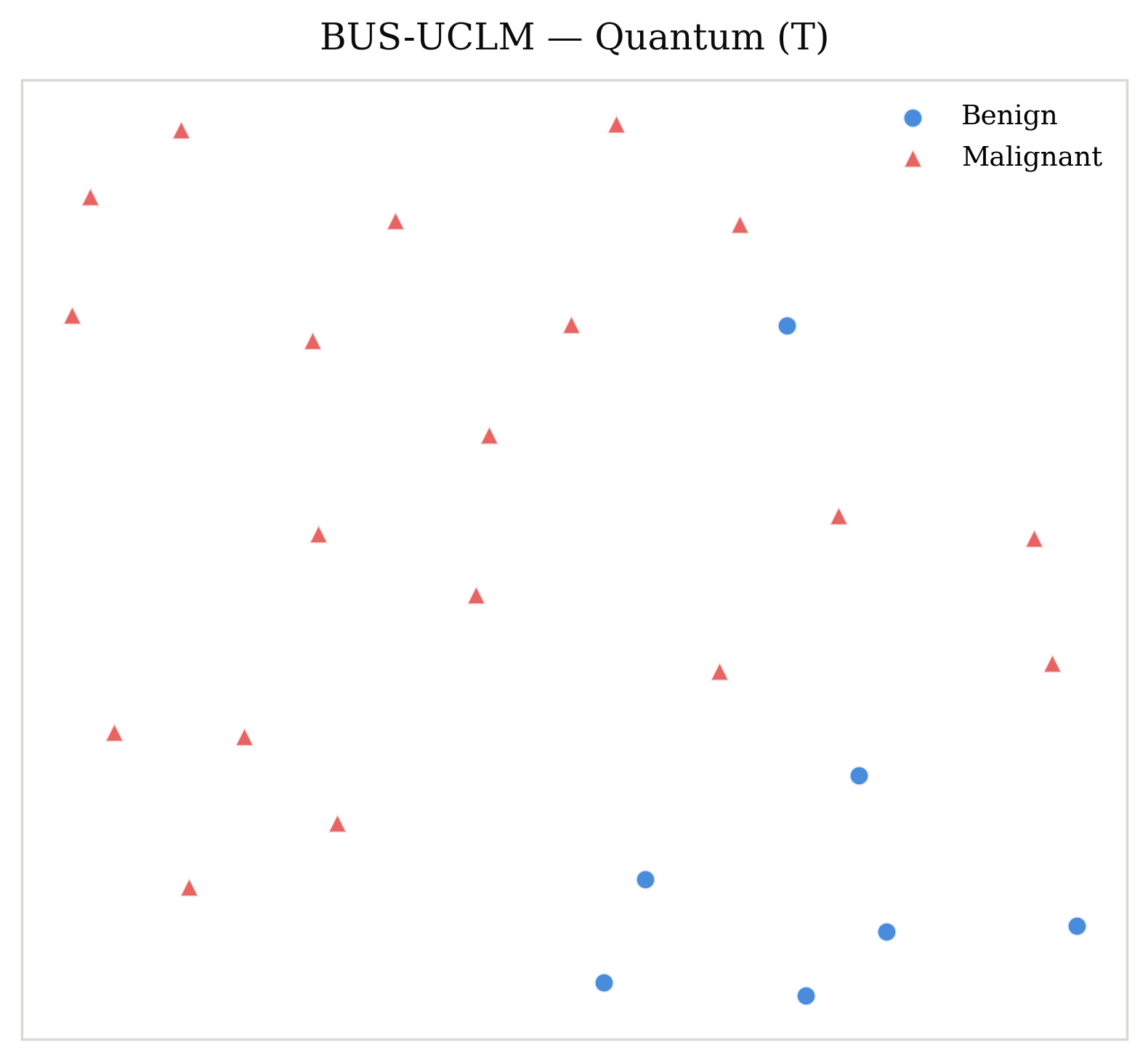}}

    \caption{UMAP projections of the standalone baseline embeddings on the
    BUS-UCLM test set.}
    \label{fig:umap_busuclm_baselines}
\end{figure}

\begin{figure}[htbp]
    \centering
    \subfloat[ResNet-18 + non-trainable]{%
        \includegraphics[width=0.48\linewidth]{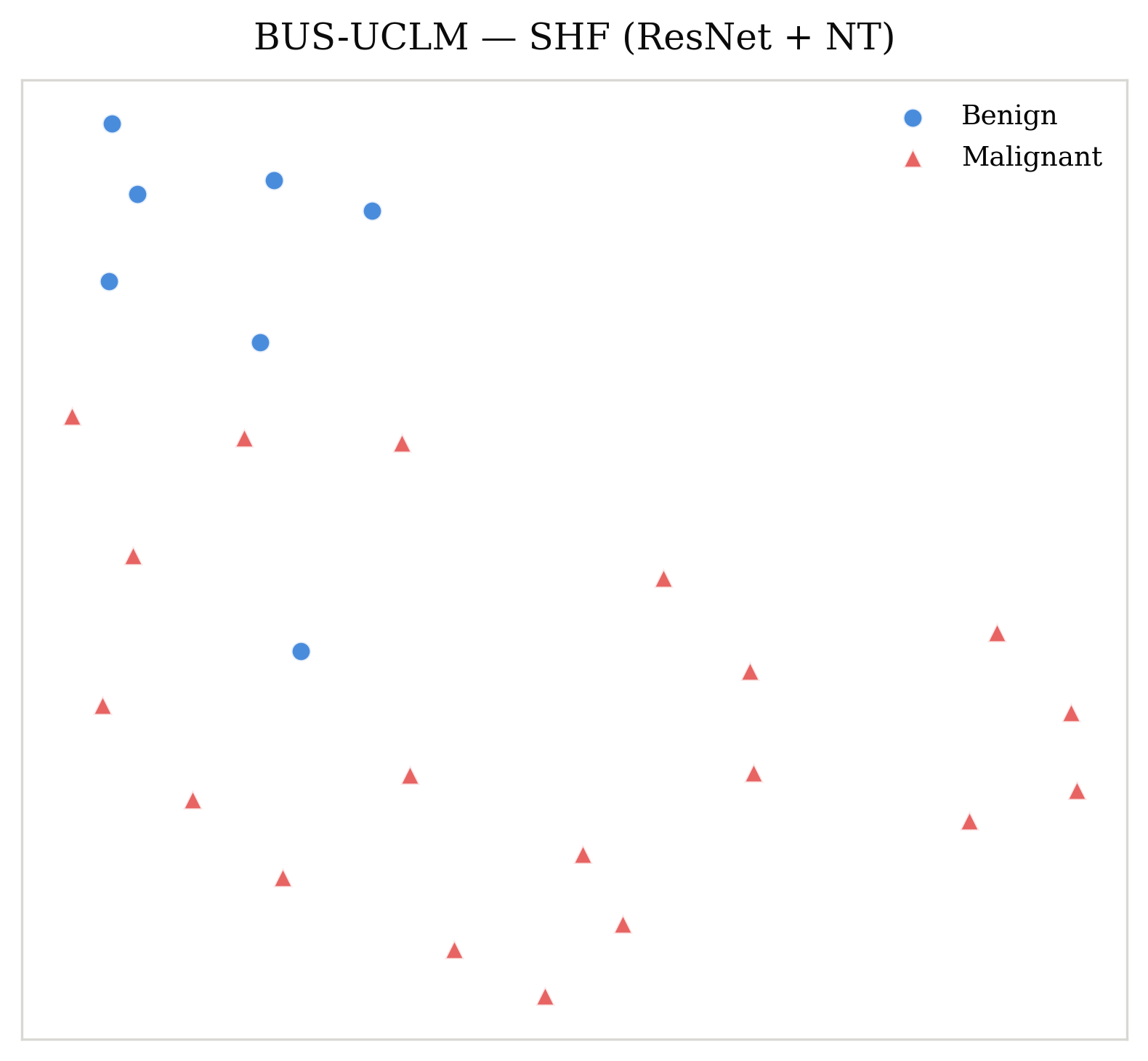}}
    \hfill
    \subfloat[ResNet-18 + trainable]{%
        \includegraphics[width=0.48\linewidth]{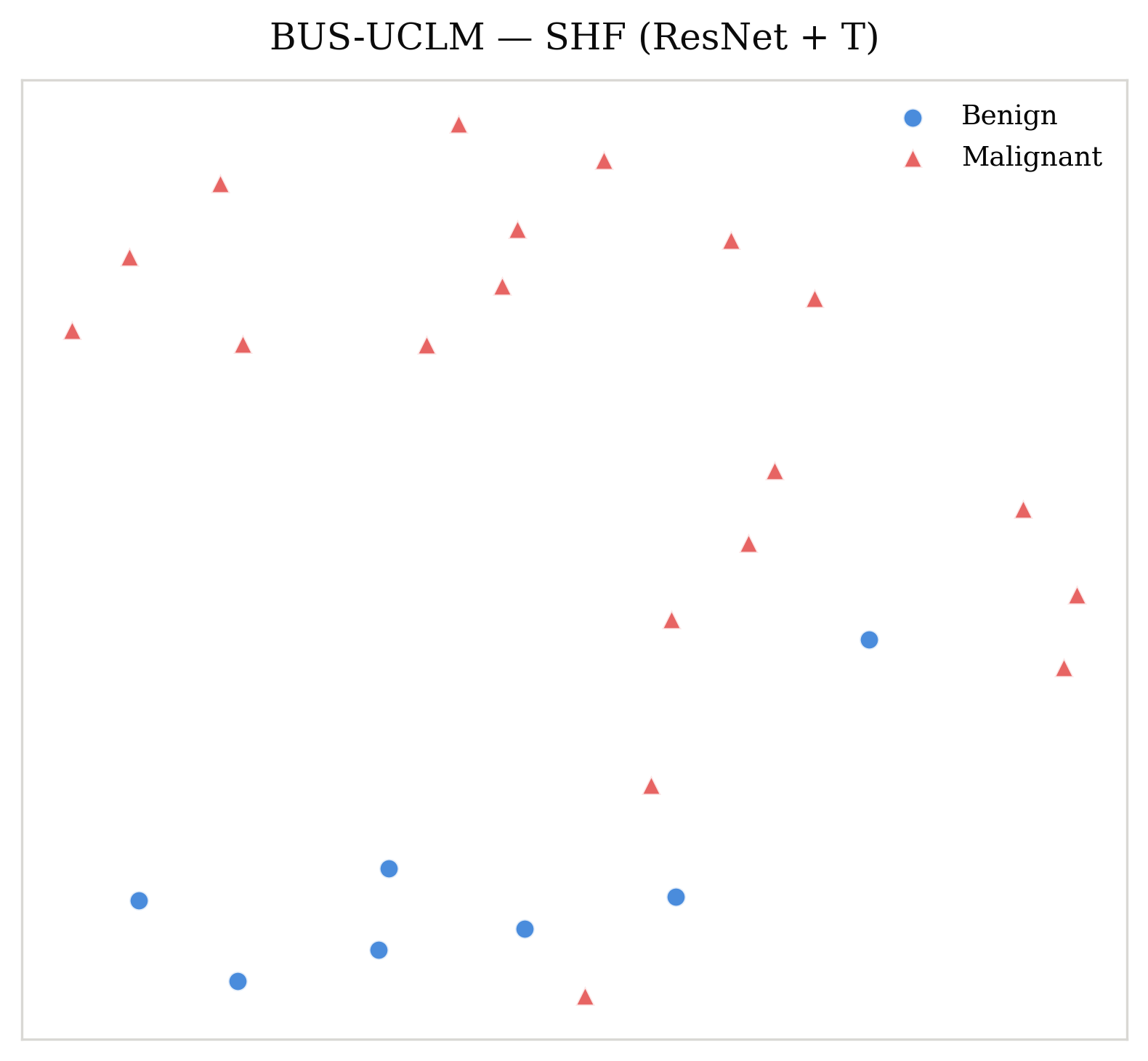}}

    \vspace{0.4em}

    \subfloat[SCNN + non-trainable]{%
        \includegraphics[width=0.48\linewidth]{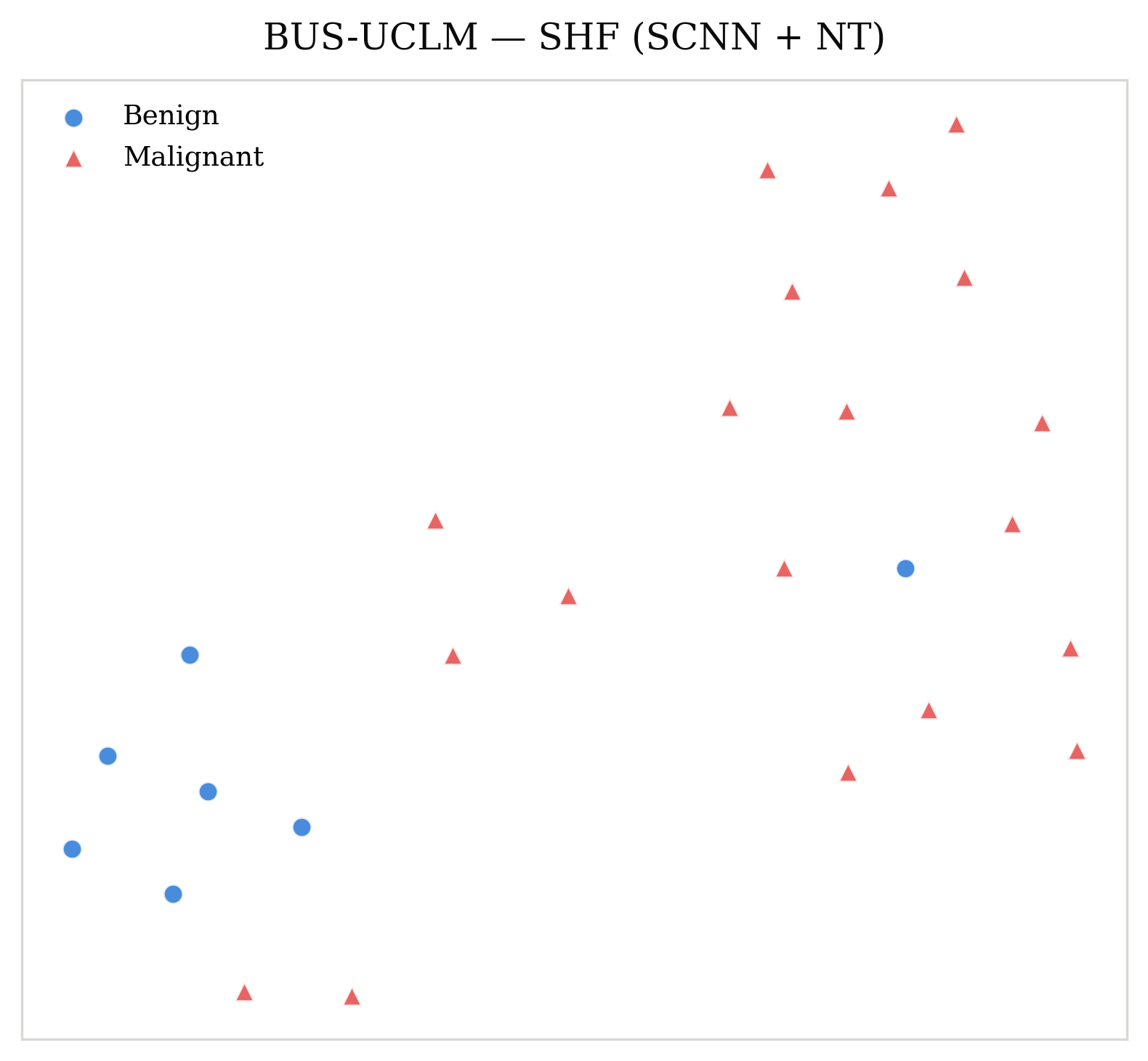}}
    \hfill
    \subfloat[SCNN + trainable]{%
        \includegraphics[width=0.48\linewidth]{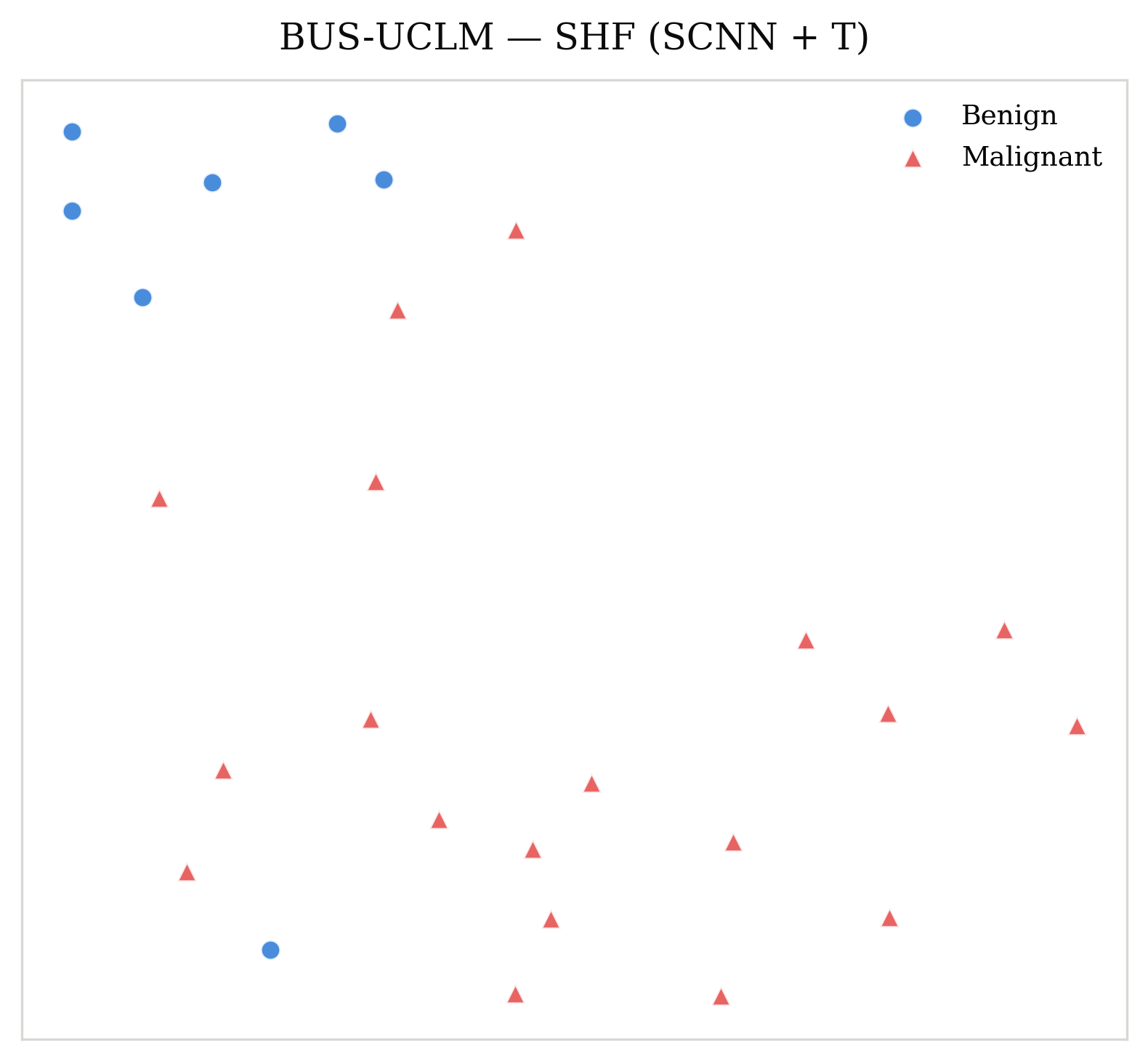}}

    \caption{UMAP projections of the fused embeddings on BUS-UCLM under the
    \textbf{SHF} strategy.}
    \label{fig:umap_busuclm_shf}
\end{figure}

\begin{figure}[htbp]
    \centering
    \subfloat[ResNet-18 + non-trainable]{%
        \includegraphics[width=0.48\linewidth]{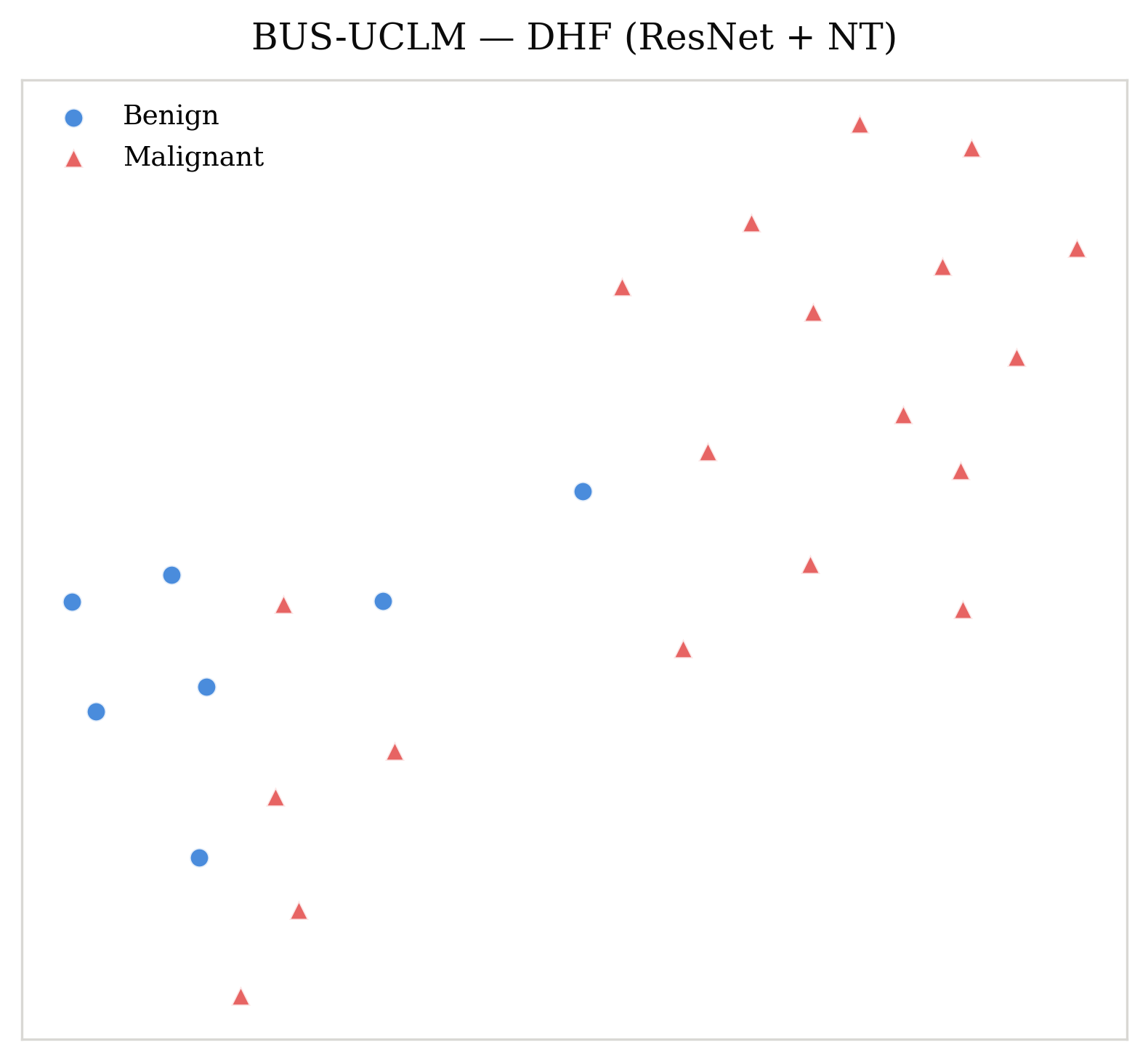}}
    \hfill
    \subfloat[ResNet-18 + trainable]{%
        \includegraphics[width=0.48\linewidth]{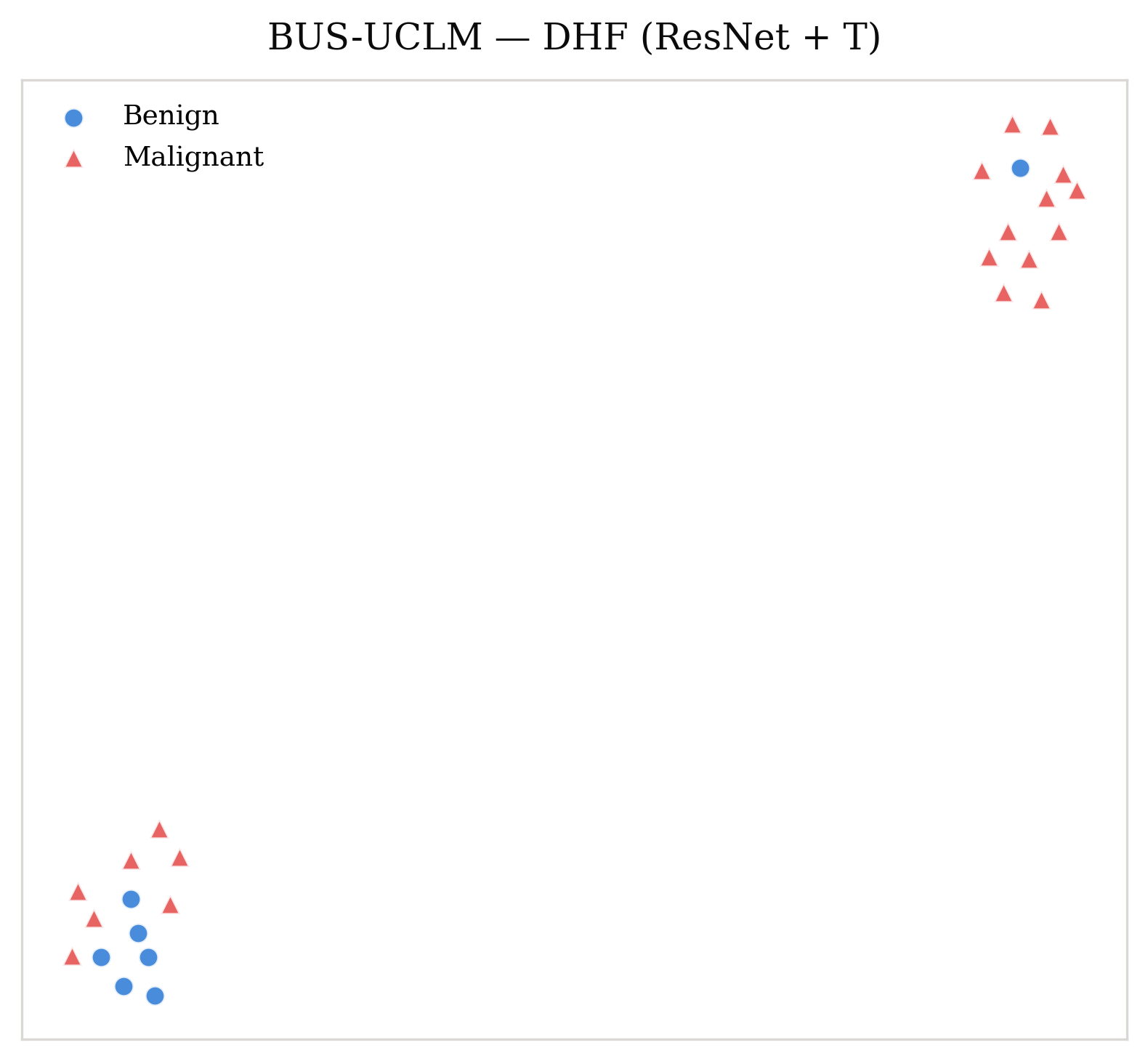}}

    \vspace{0.4em}

    \subfloat[SCNN + non-trainable]{%
        \includegraphics[width=0.48\linewidth]{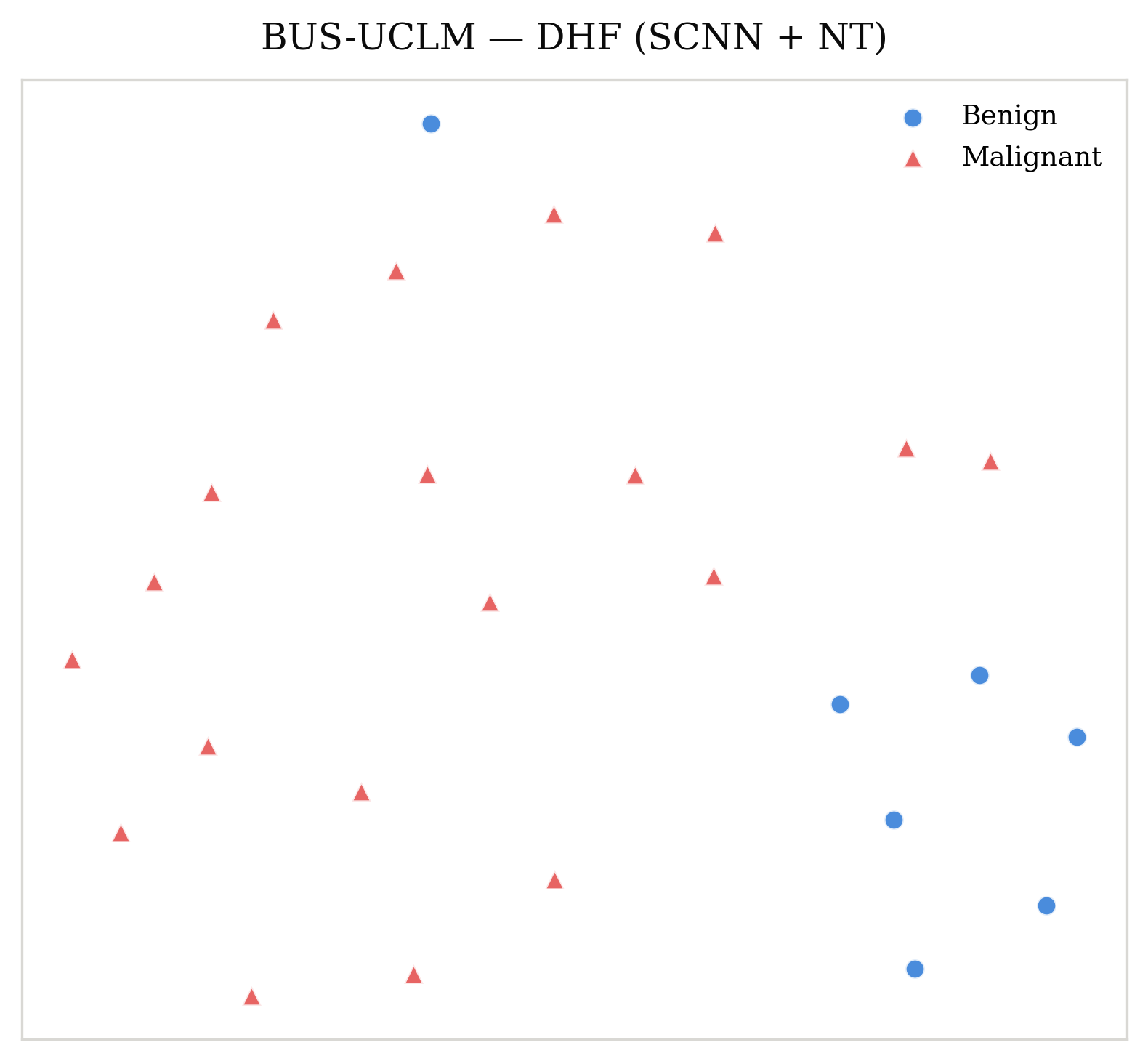}}
    \hfill
    \subfloat[SCNN + trainable]{%
        \includegraphics[width=0.48\linewidth]{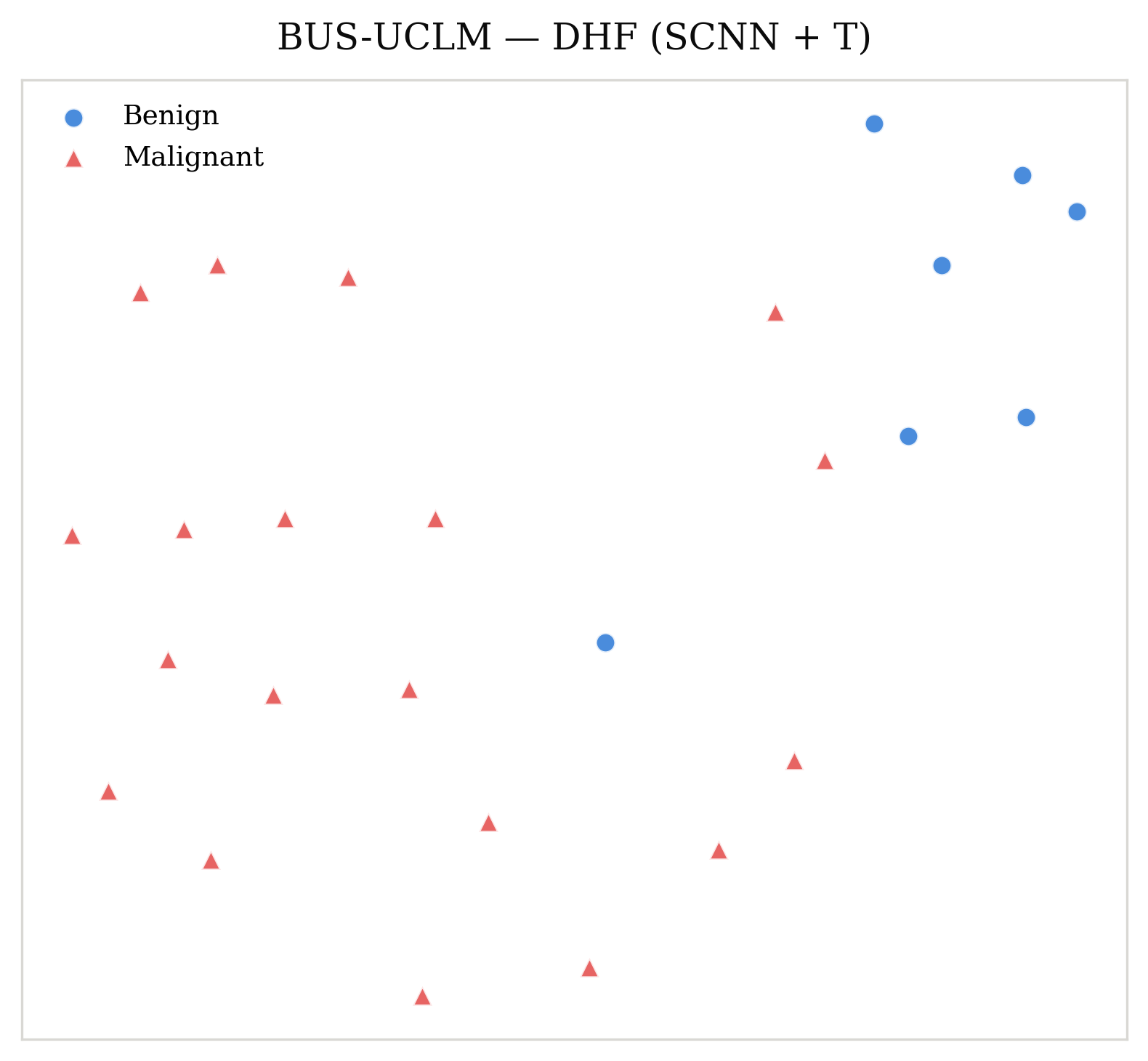}}

    \caption{UMAP projections of the fused embeddings on BUS-UCLM under the
    \textbf{DHF} strategy.}
    \label{fig:umap_busuclm_dhf}
\end{figure}

\begin{figure}[htbp]
    \centering
    \subfloat[ResNet-18 + non-trainable]{%
        \includegraphics[width=0.48\linewidth]{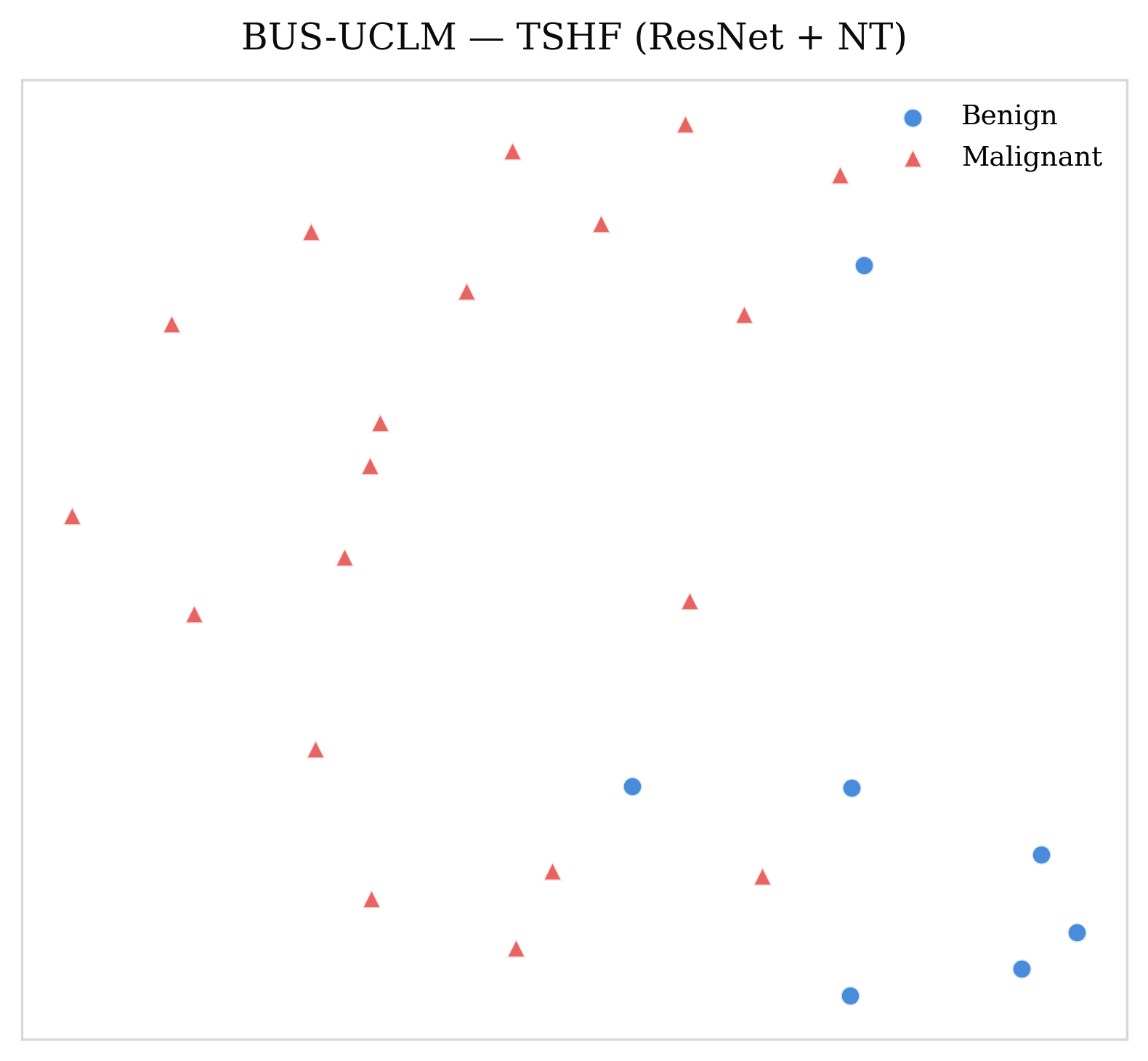}}
    \hfill
    \subfloat[ResNet-18 + trainable]{%
        \includegraphics[width=0.48\linewidth]{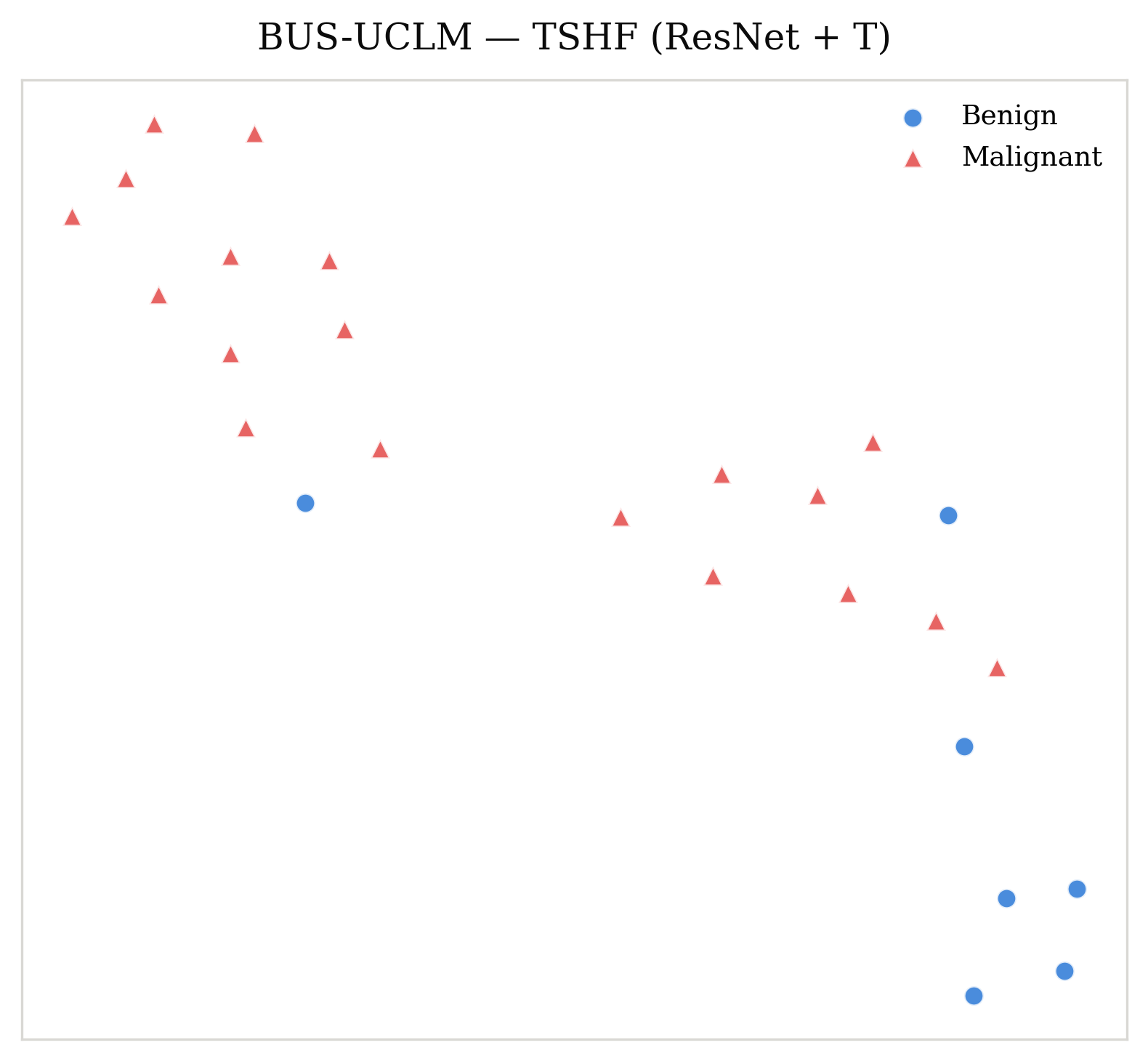}}

    \vspace{0.4em}

    \subfloat[SCNN + non-trainable]{%
        \includegraphics[width=0.48\linewidth]{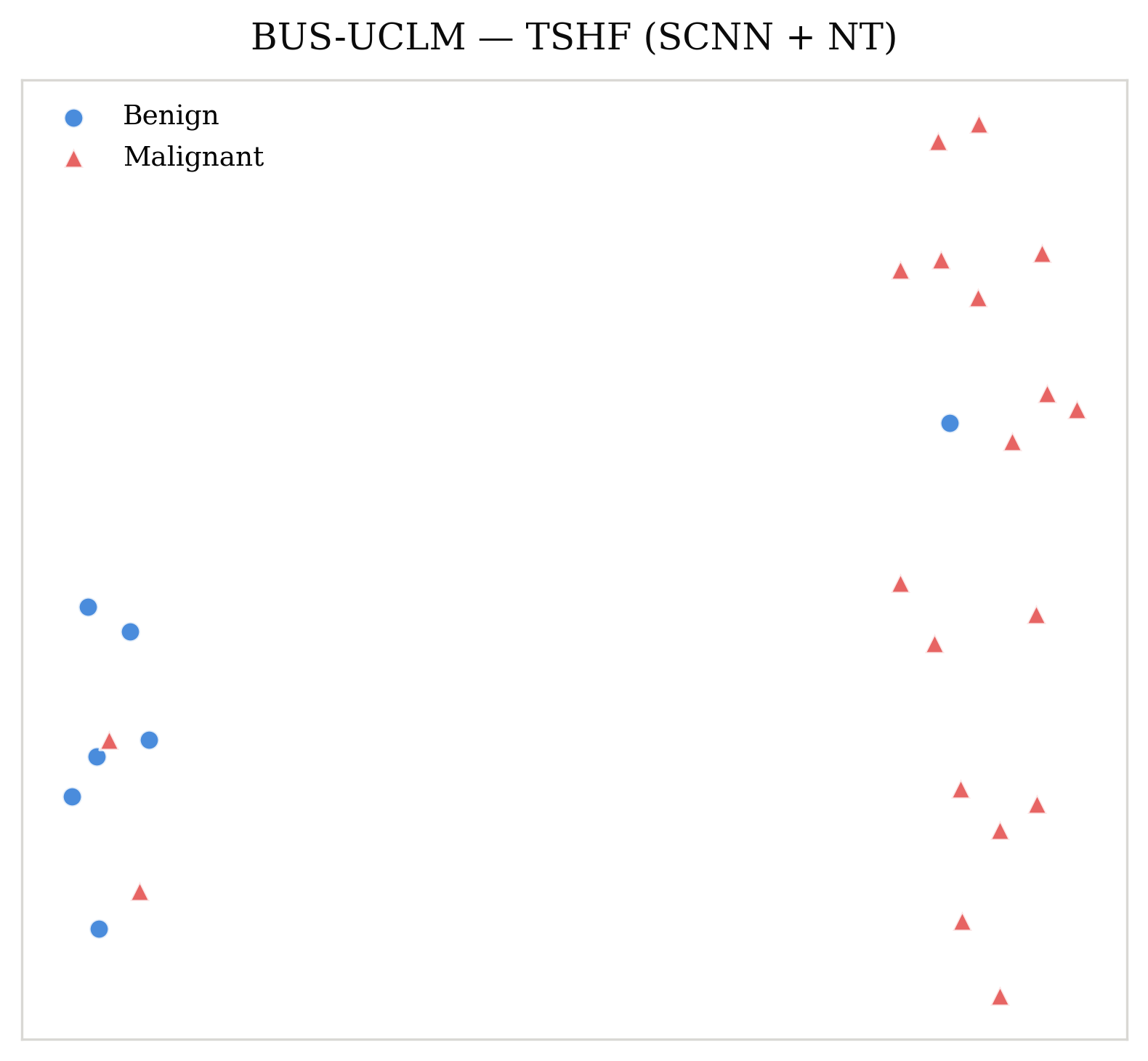}}
    \hfill
    \subfloat[SCNN + trainable]{%
        \includegraphics[width=0.48\linewidth]{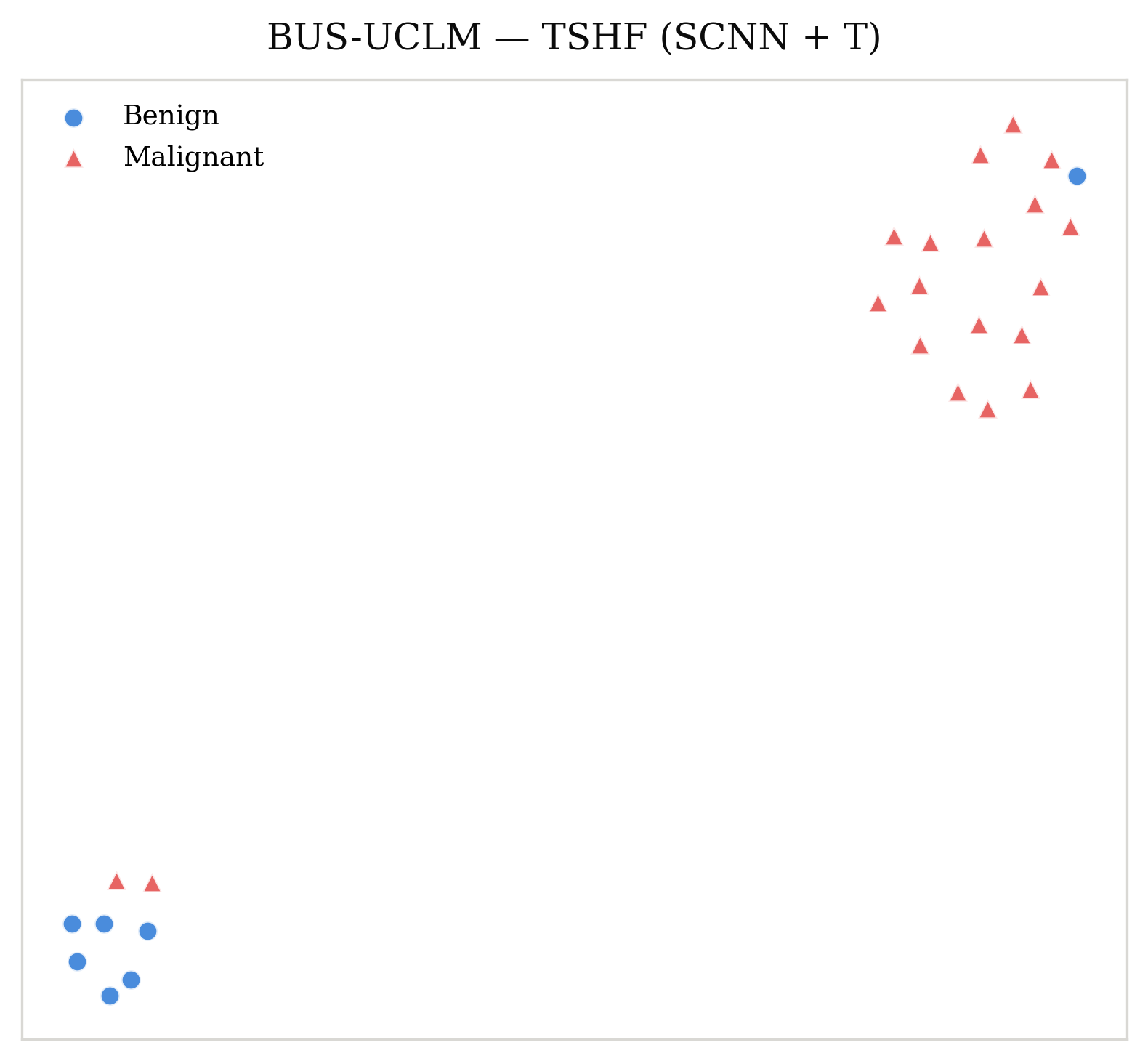}}

    \caption{UMAP projections of the fused embeddings on BUS-UCLM under the
    \textbf{TSHF} strategy.}
    \label{fig:umap_busuclm_tshf}
\end{figure}

Figures~\ref{fig:umap_inbreast_baselines}--\ref{fig:umap_inbreast_tshf} and
Figures~\ref{fig:umap_busuclm_baselines}--\ref{fig:umap_busuclm_tshf} show the
UMAP projections of the INbreast and BUS-UCLM test embeddings for the standalone
baselines and for each fusion strategy. Rows correspond to the classical backbone
and columns to the quantum circuit. As in the main text, these projections are
illustrative rather than quantitative, and are provided for completeness across
all configurations.

\clearpage

\section{Training Curves}
\label{app:f1_curves}

\begin{figure}[htbp]
    \centering
    \includegraphics[width=\linewidth]{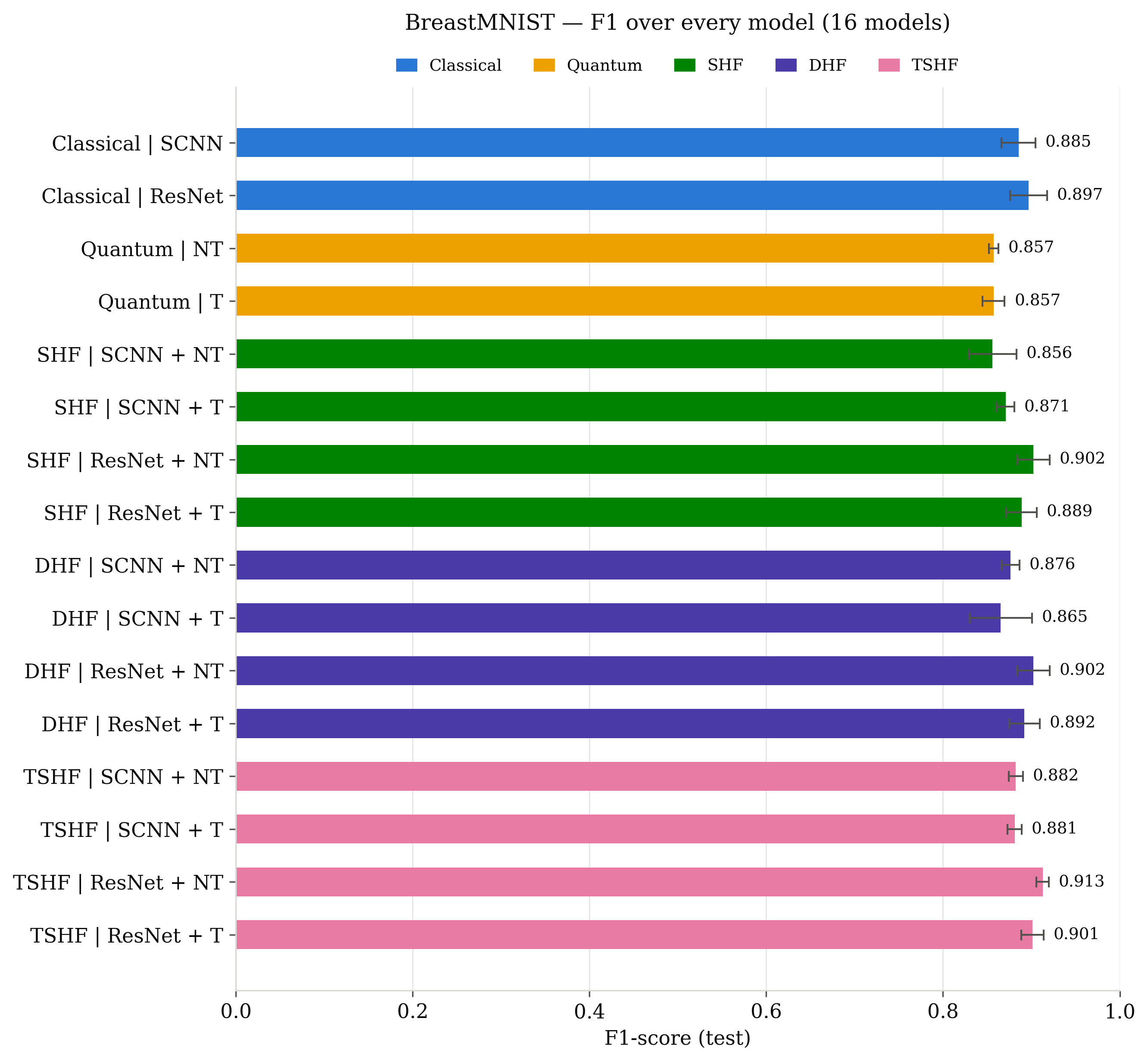}
    \caption{F1-score curves over training epochs for all models on
    BreastMNIST.}
    \label{fig:f1_combined_breastmnist}
\end{figure}

\begin{figure}[htbp]
    \centering
    \includegraphics[width=\linewidth]{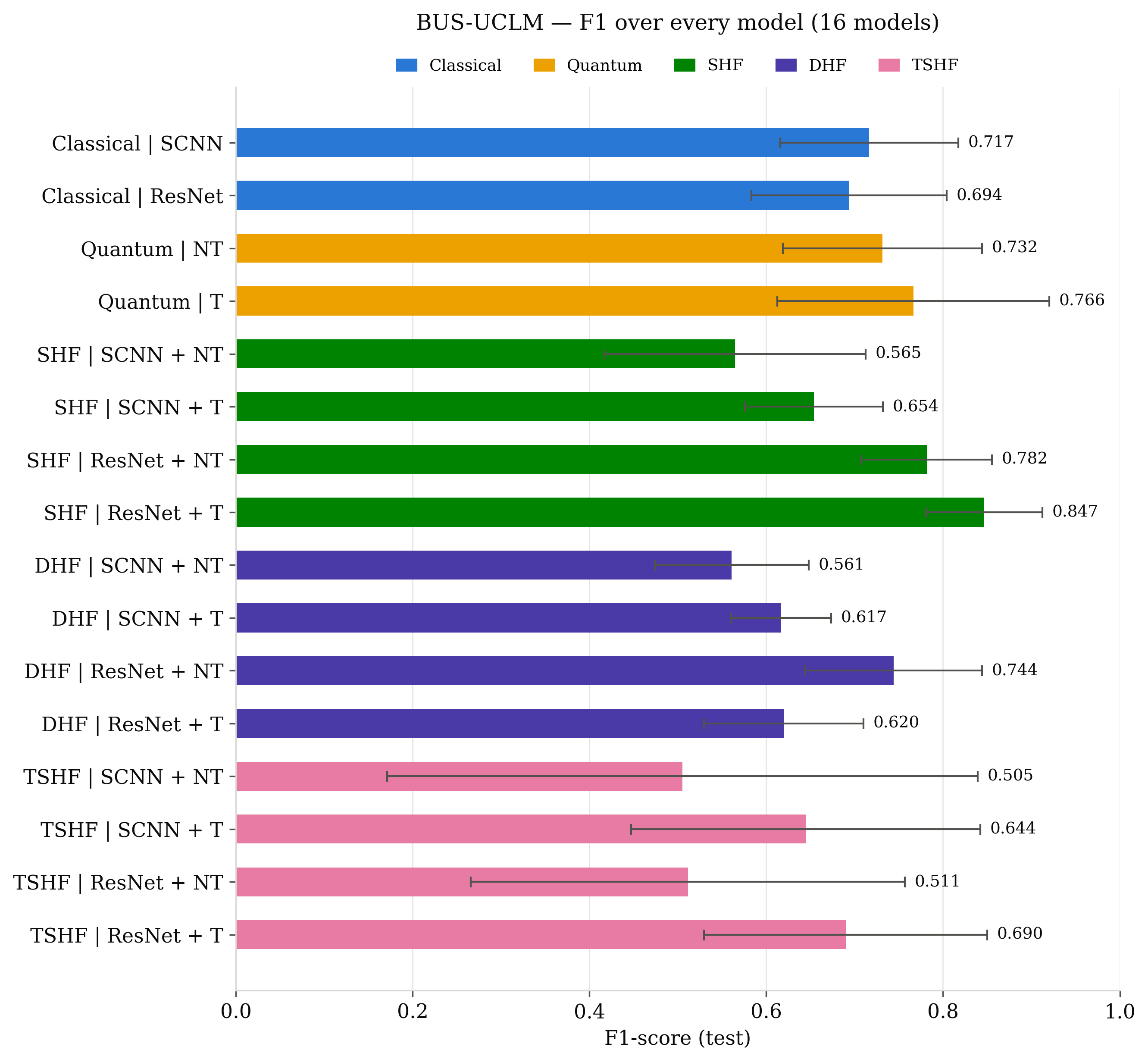}
    \caption{F1-score curves over training epochs for all models on BUS-UCLM.}
    \label{fig:f1_combined_busuclm}
\end{figure}

\begin{figure}[htbp]
    \centering
    \includegraphics[width=\linewidth]{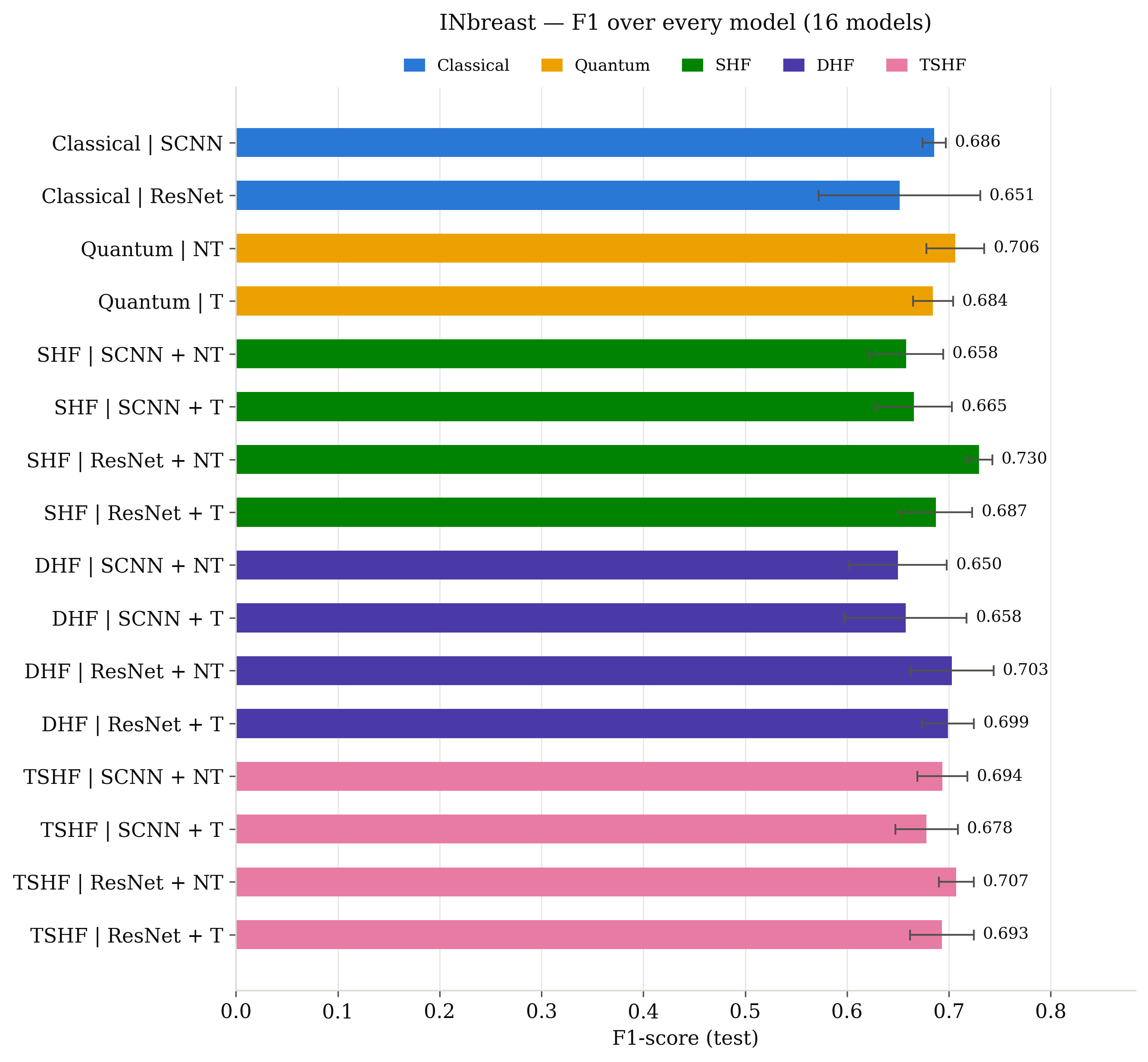}
    \caption{F1-score curves over training epochs for all models on INbreast.}
    \label{fig:f1_combined_inbreast}
\end{figure}

Figures~\ref{fig:f1_combined_breastmnist}--\ref{fig:f1_combined_inbreast} show
the F1-score over training epochs for all models on each dataset, providing a
view of convergence behavior that complements the final-epoch metrics reported
in the main text.

\clearpage

% =====================================================================
%  APPENDIX -- Confusion matrices, BreastMNIST (Figures D32--D35)
%  Paste at the END of the existing appendix.
%  Do NOT repeat \appendix; it is already declared earlier.
% =====================================================================
\section{Confusion Matrices}
\label{app:confusion}

This appendix reports the confusion matrices of every configuration evaluated in
this work, summed over the five independent runs of each configuration. The class order follows the label convention of each dataset, as indicated on the axes. Counts are therefore over $5 \times 156 = 780$ predictions
for BreastMNIST, of which 210 are malignant and 570 benign. The corresponding
metrics are reported in Tables~\ref{tab:standalone_baselines}
and~\ref{tab:metrics_breastmnist}.

\subsection{BreastMNIST: Standalone Baselines}
\label{app:cm_breastmnist_baselines}

\begin{figure}[htbp]
    \centering
    \subfloat[SCNN]{%
        \includegraphics[width=0.47\linewidth]{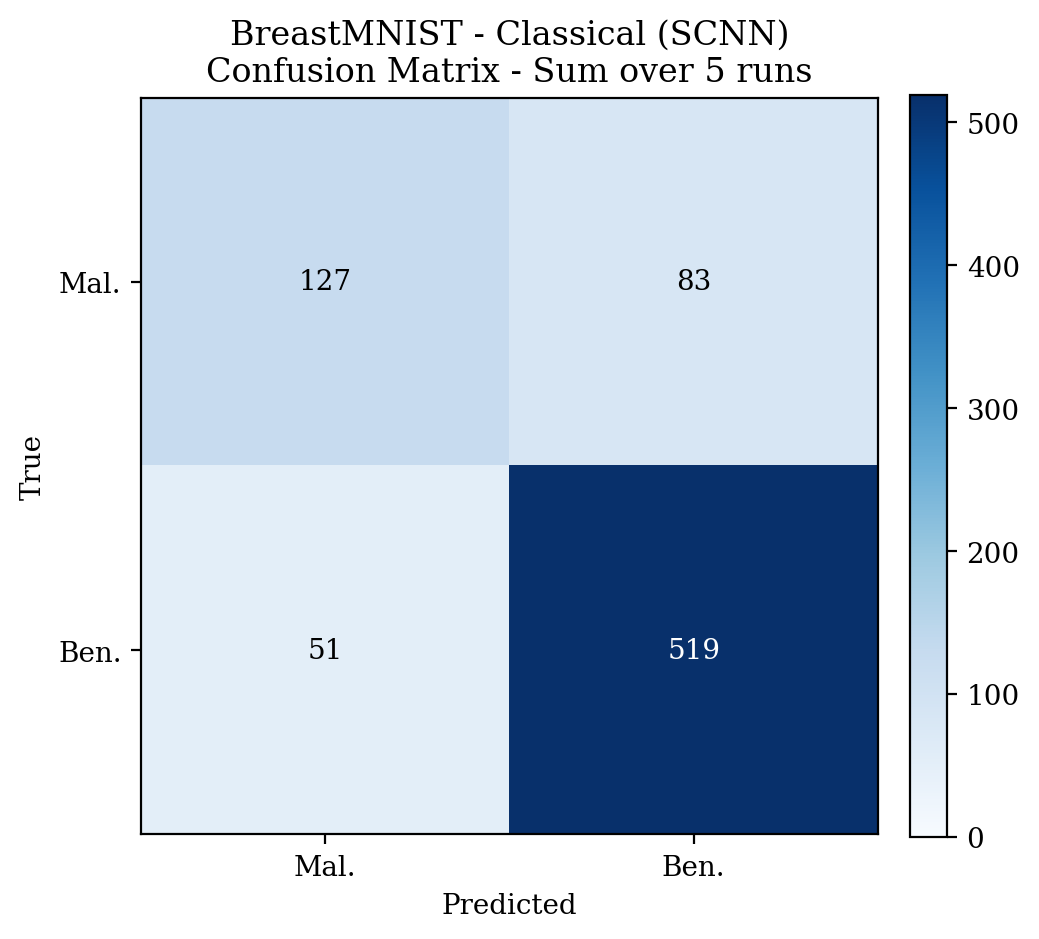}}
    \hfill
    \subfloat[ResNet-18]{%
        \includegraphics[width=0.47\linewidth]{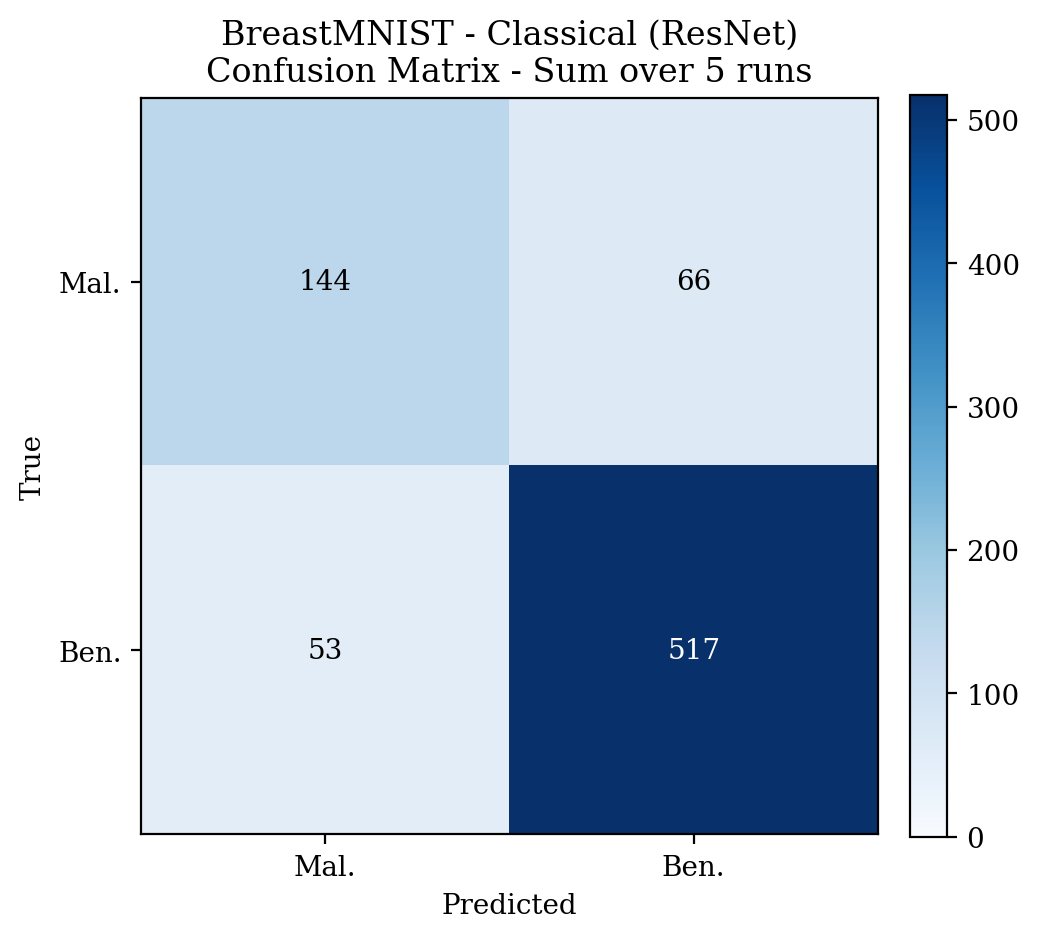}}

    \vspace{0.4em}

    \subfloat[Quantum, non-trainable]{%
        \includegraphics[width=0.47\linewidth]{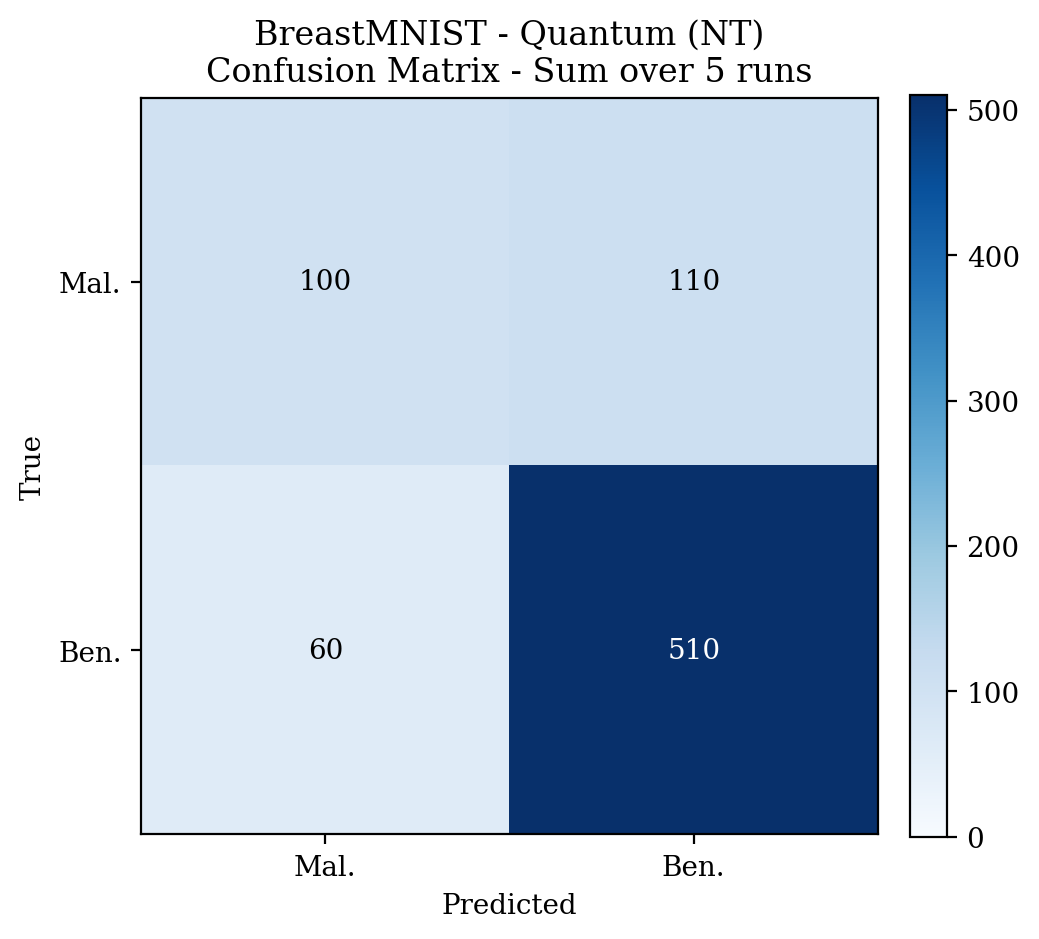}}
    \hfill
    \subfloat[Quantum, trainable]{%
        \includegraphics[width=0.47\linewidth]{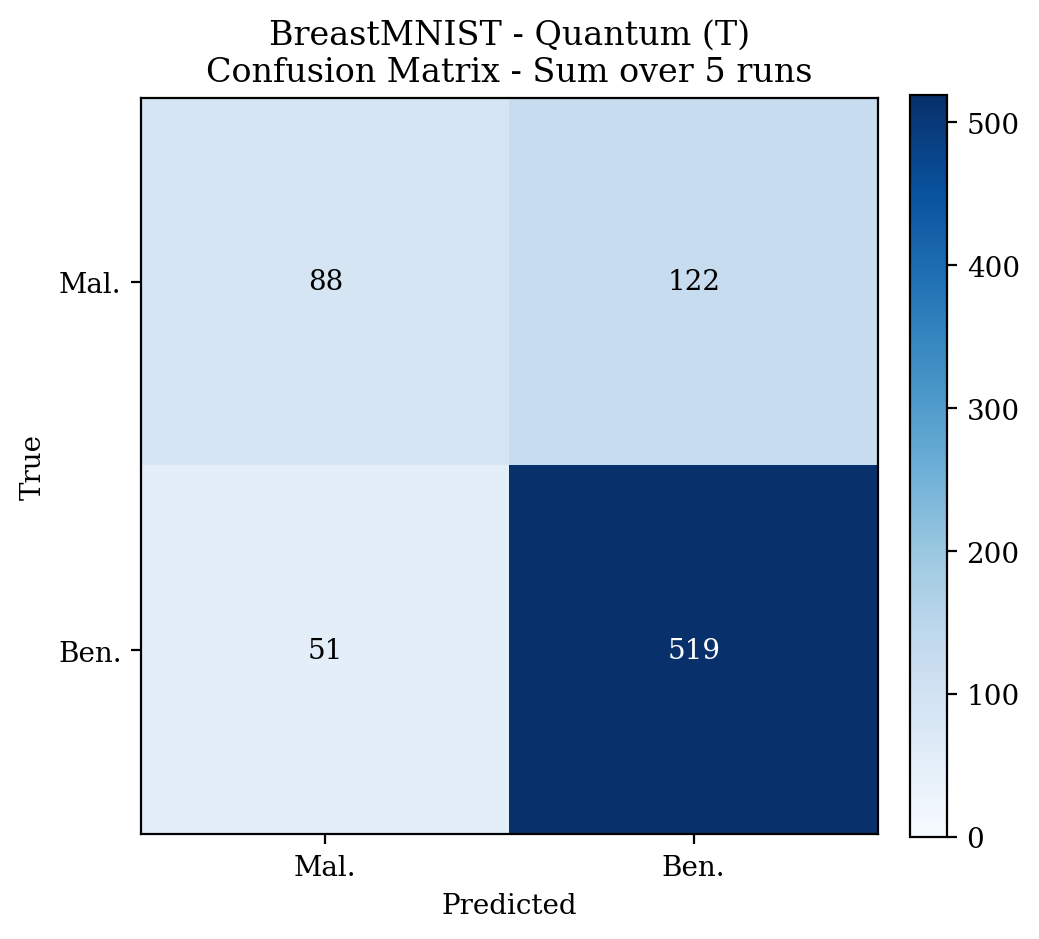}}

    \caption{Confusion matrices of the standalone baselines on the BreastMNIST
    test set, summed over five independent runs.}
    \label{fig:cm_breastmnist_baselines}
\end{figure}
\clearpage

\subsection{BreastMNIST: Static Hybrid Fusion}
\label{app:cm_breastmnist_shf}

\begin{figure}[htbp]
    \centering
    \subfloat[ResNet-18 + non-trainable]{%
        \includegraphics[width=0.47\linewidth]{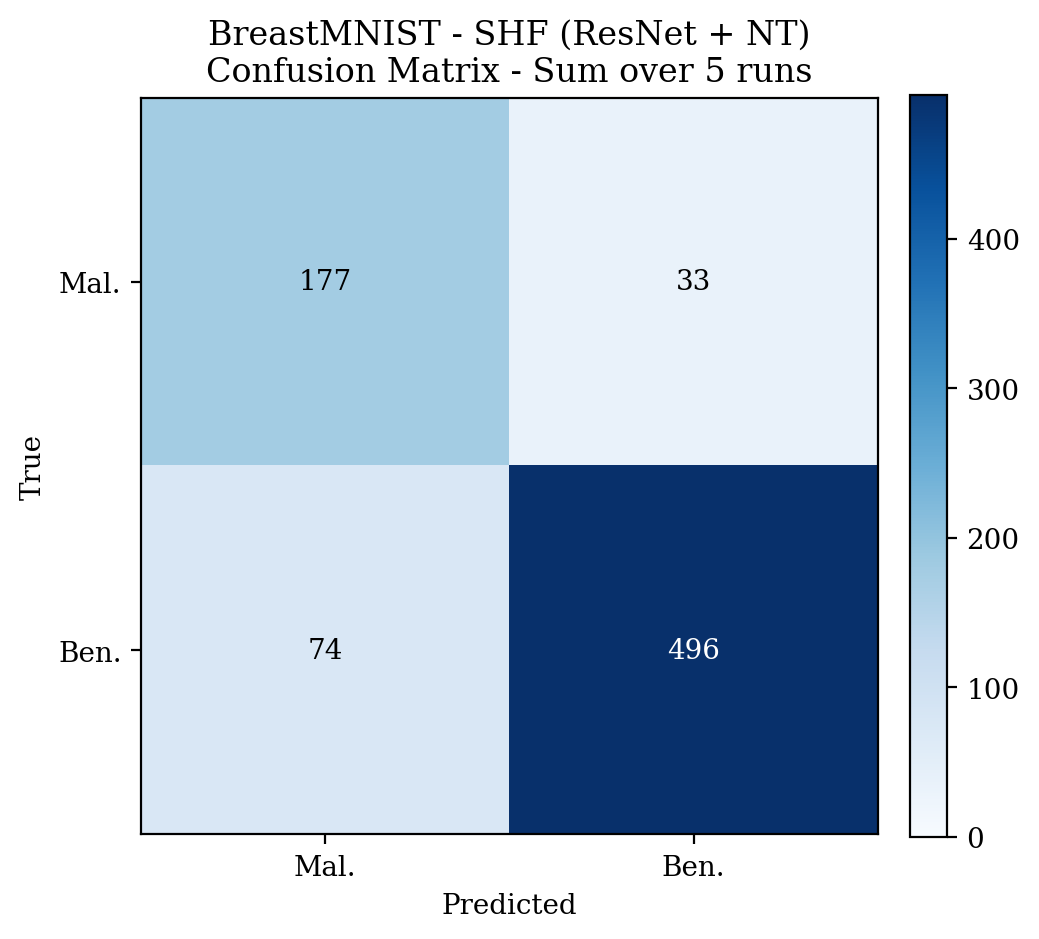}}
    \hfill
    \subfloat[ResNet-18 + trainable]{%
        \includegraphics[width=0.47\linewidth]{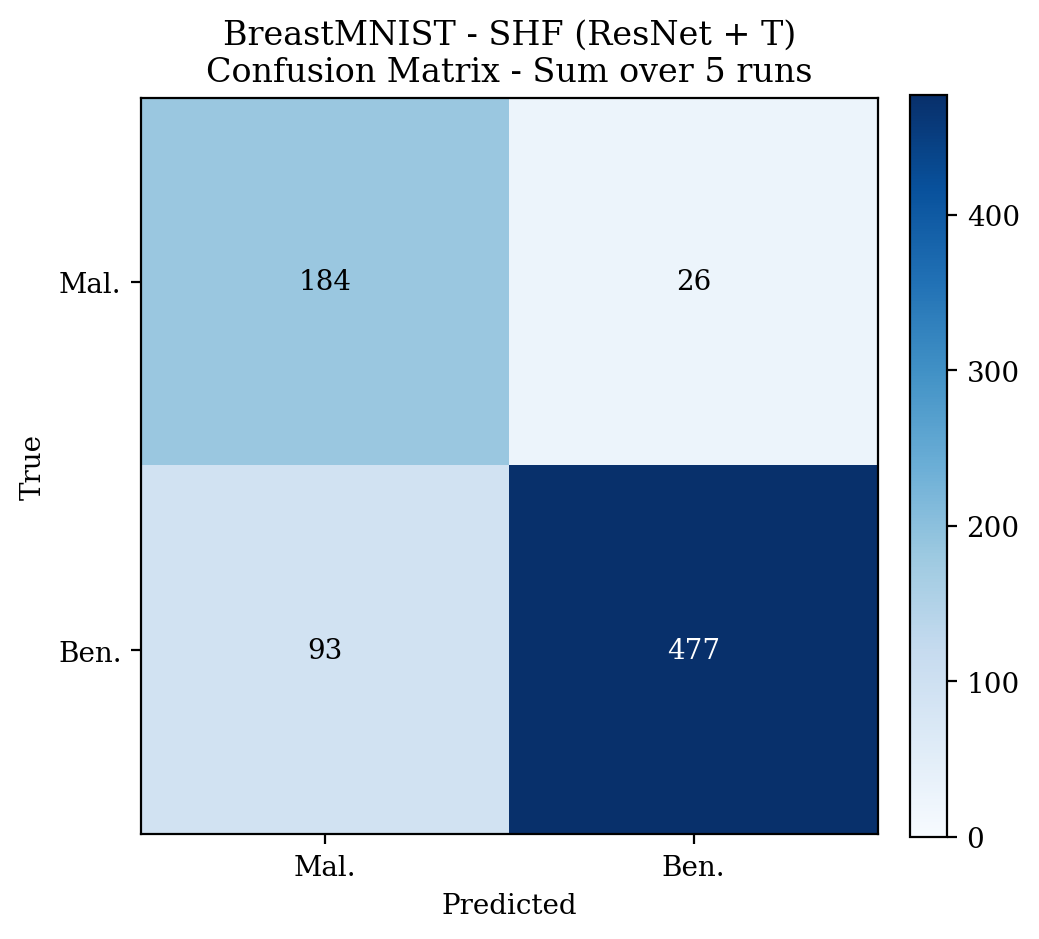}}

    \vspace{0.4em}

    \subfloat[SCNN + non-trainable]{%
        \includegraphics[width=0.47\linewidth]{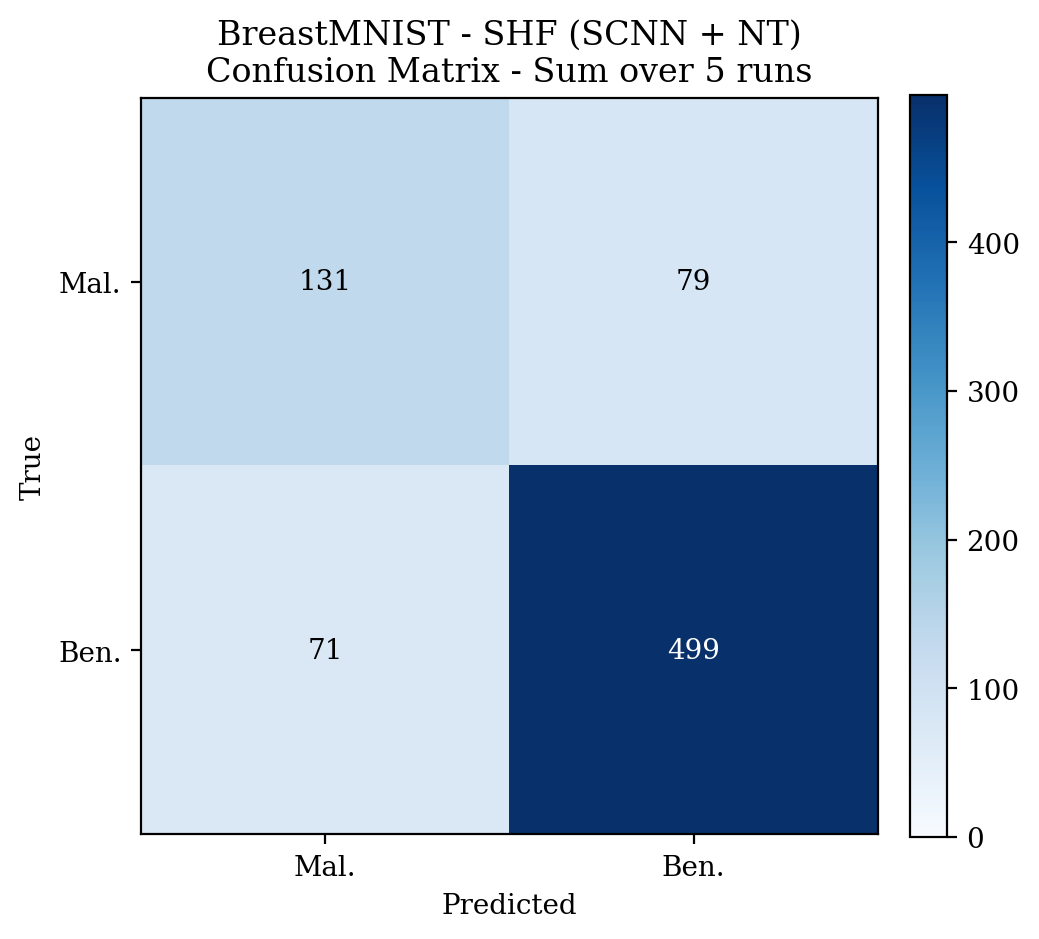}}
    \hfill
    \subfloat[SCNN + trainable]{%
        \includegraphics[width=0.47\linewidth]{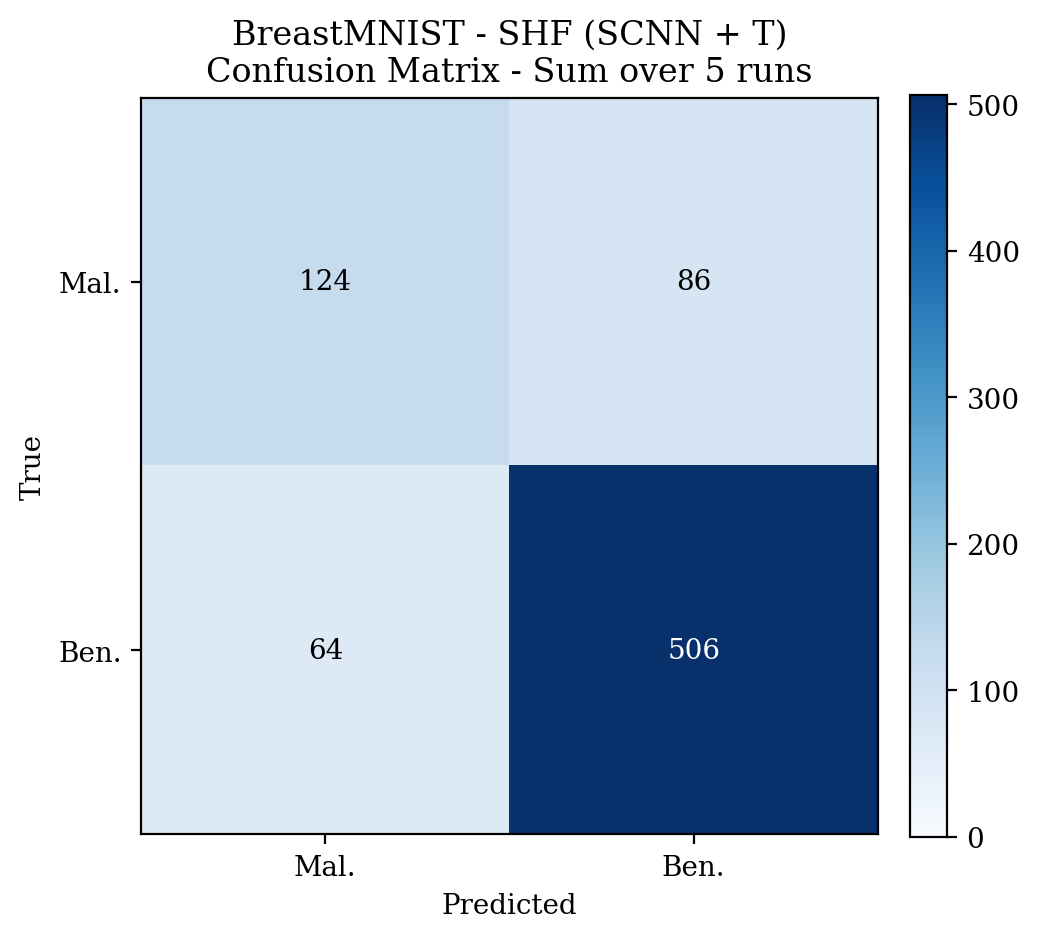}}

    \caption{Confusion matrices of the \textbf{SHF} configurations on the
    BreastMNIST test set, summed over five independent runs.}
    \label{fig:cm_breastmnist_shf}
\end{figure}
\clearpage

\subsection{BreastMNIST: Dynamic Hybrid Fusion}
\label{app:cm_breastmnist_dhf}

\begin{figure}[htbp]
    \centering
    \subfloat[ResNet-18 + non-trainable]{%
        \includegraphics[width=0.47\linewidth]{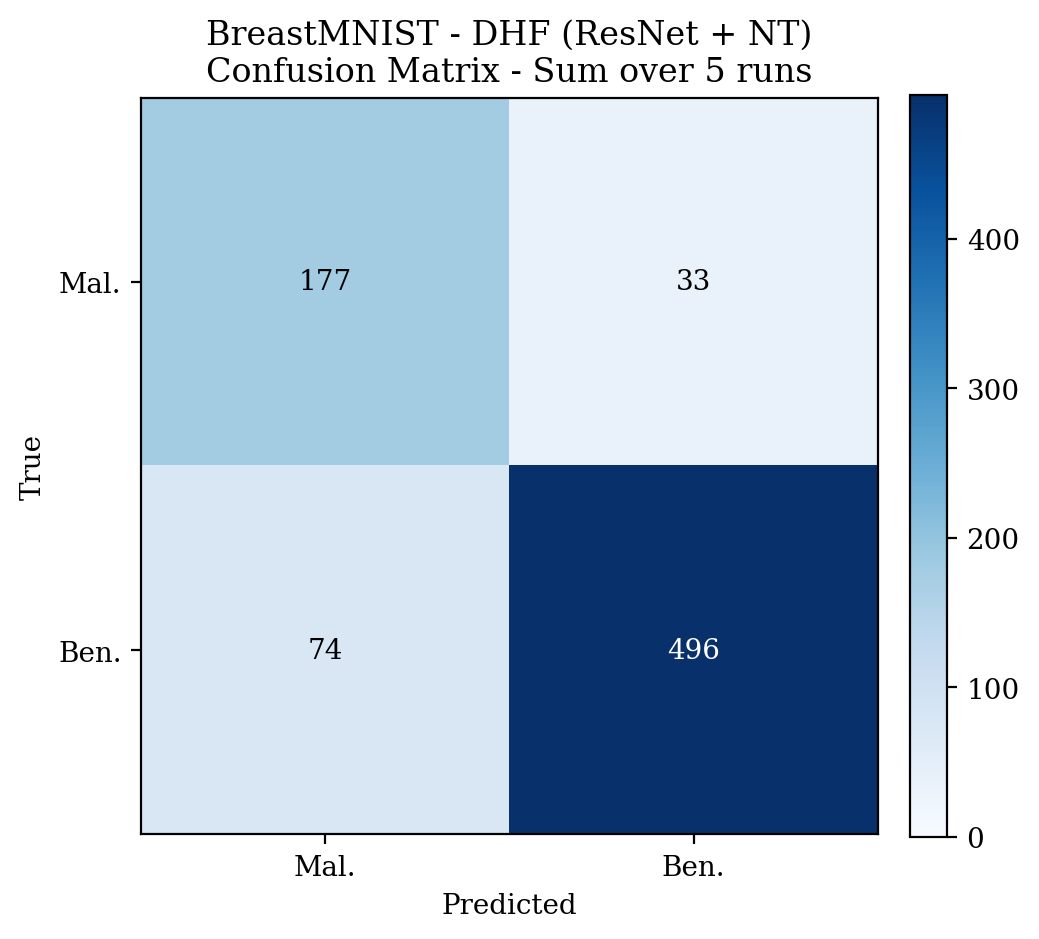}}
    \hfill
    \subfloat[ResNet-18 + trainable]{%
        \includegraphics[width=0.47\linewidth]{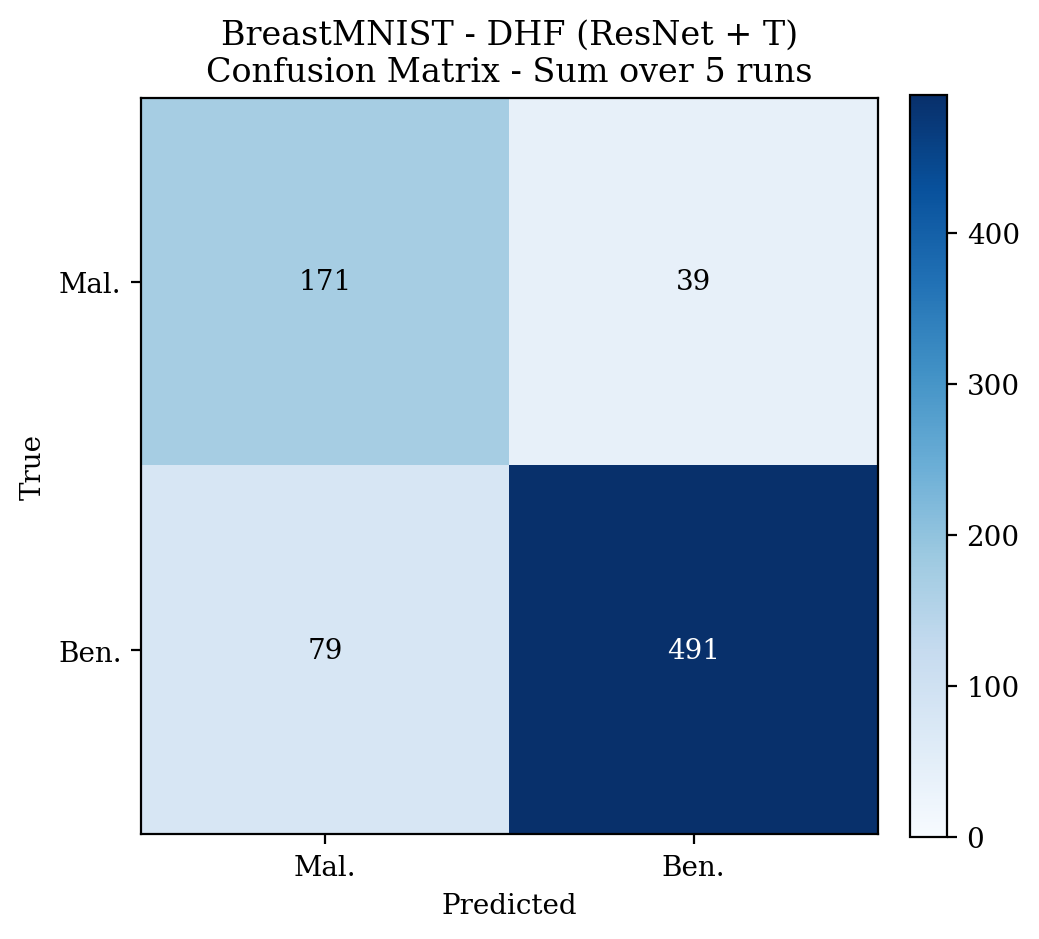}}

    \vspace{0.4em}

    \subfloat[SCNN + non-trainable]{%
        \includegraphics[width=0.47\linewidth]{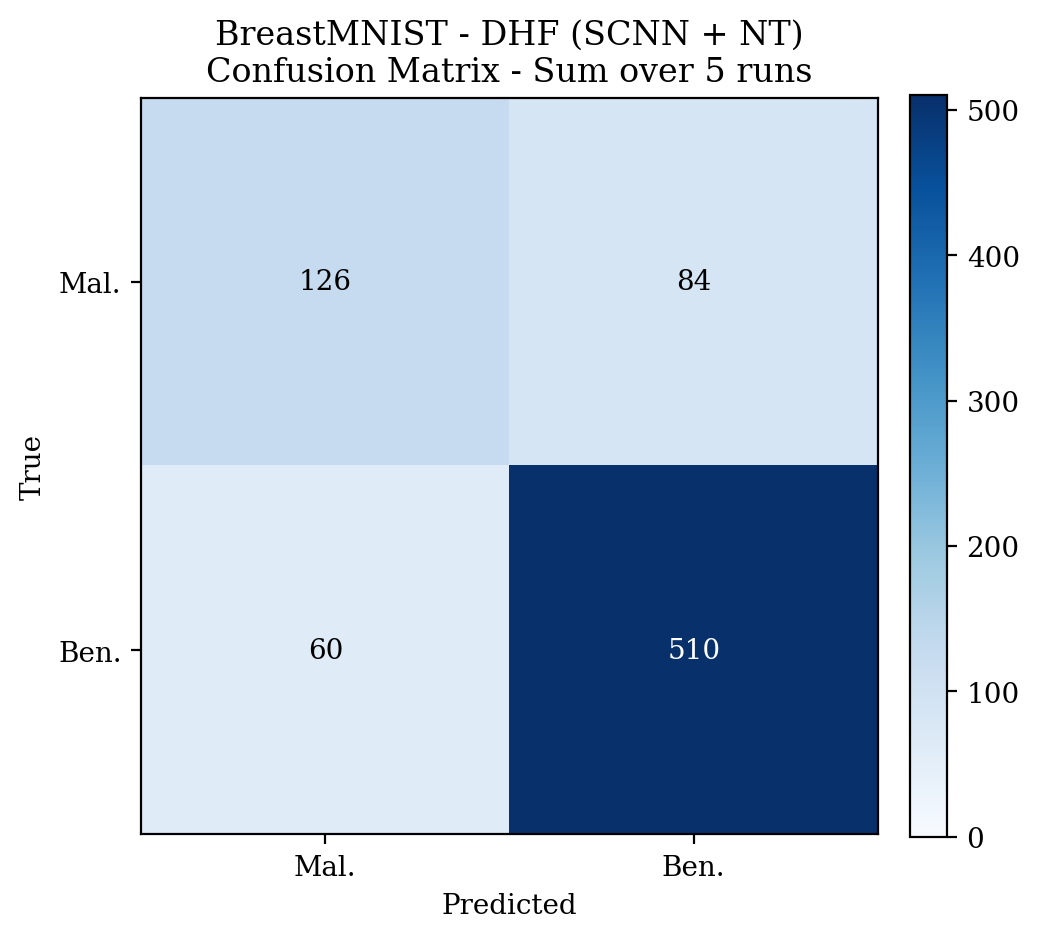}}
        \hfill
    \subfloat[SCNN + trainable]{%
        \includegraphics[width=0.47\linewidth]{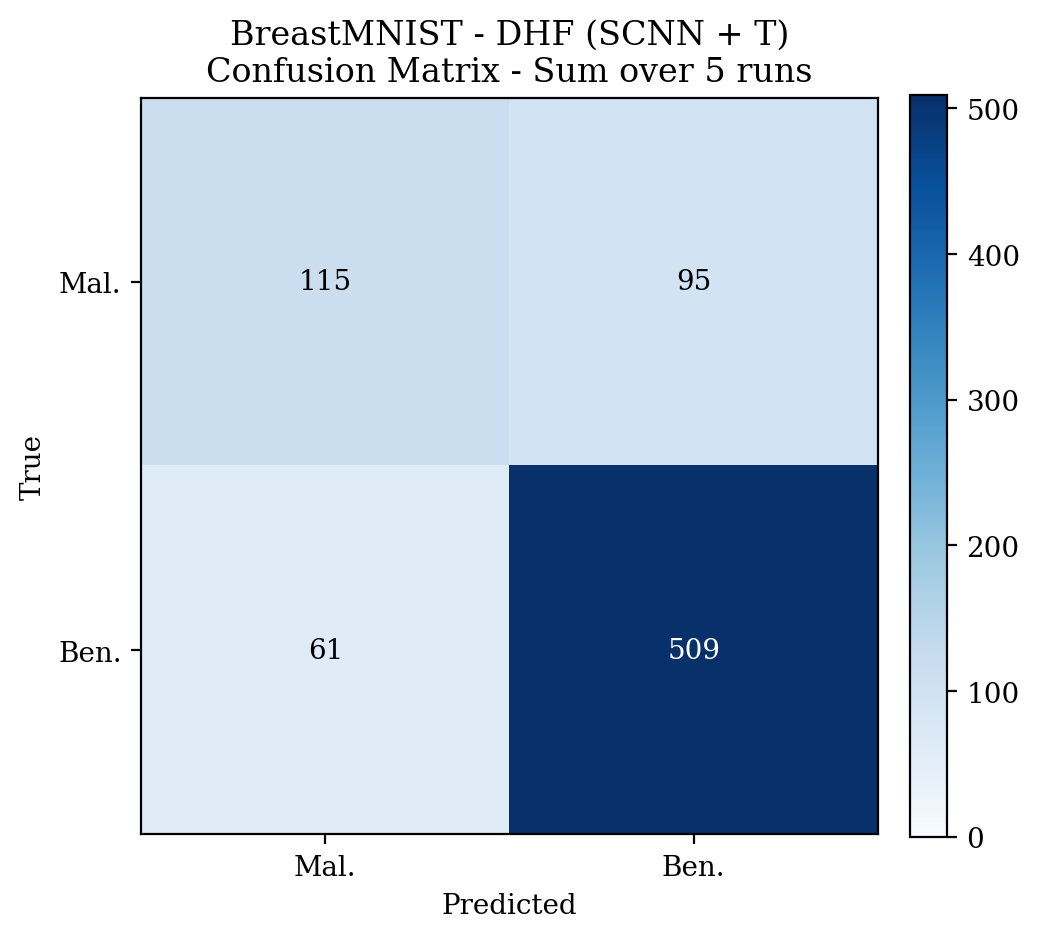}}
    \caption{Confusion matrices of the \textbf{DHF} configurations on the
    BreastMNIST test set, summed over five independent runs.}
\end{figure}
\clearpage

\subsection{BreastMNIST: Temperature-Scaled Hybrid Fusion}
\label{app:cm_breastmnist_tshf}

\begin{figure}[htbp]
    \centering
    \subfloat[ResNet-18 + non-trainable]{%
        \includegraphics[width=0.47\linewidth]{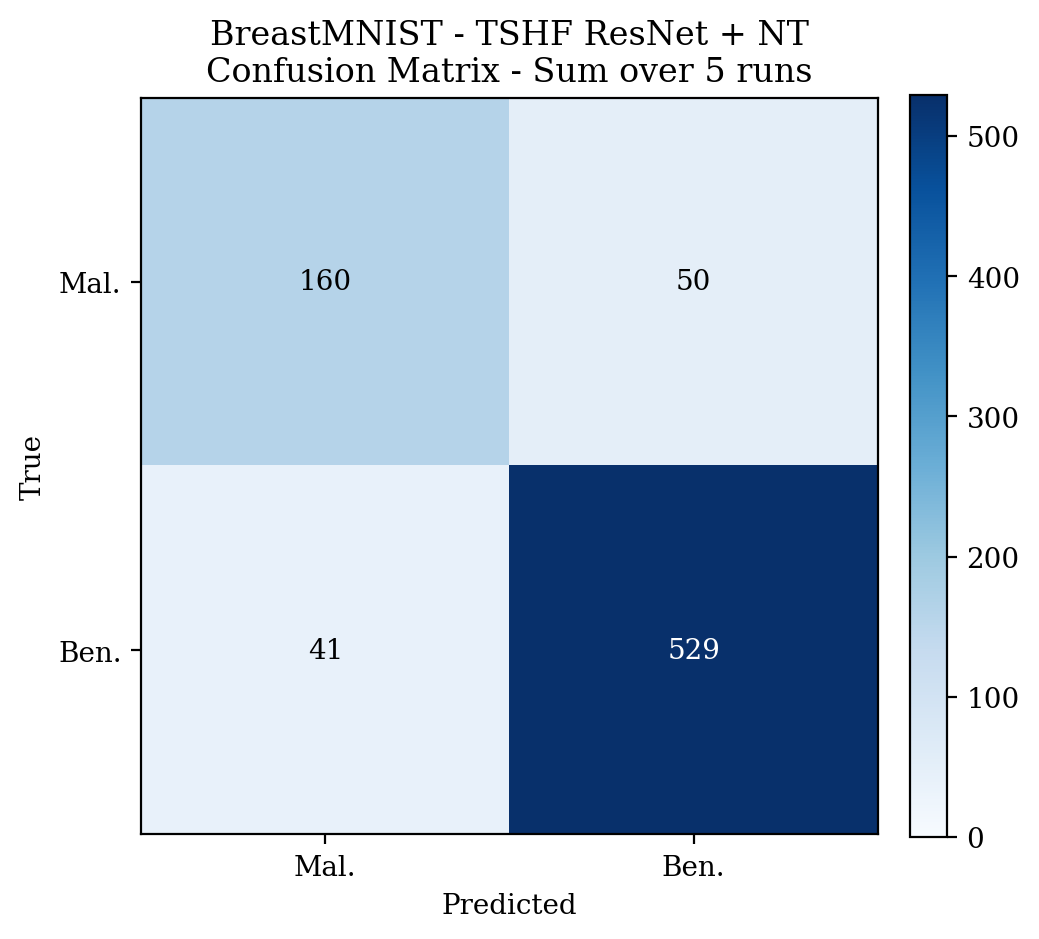}}
    \hfill
    \subfloat[ResNet-18 + trainable]{%
        \includegraphics[width=0.47\linewidth]{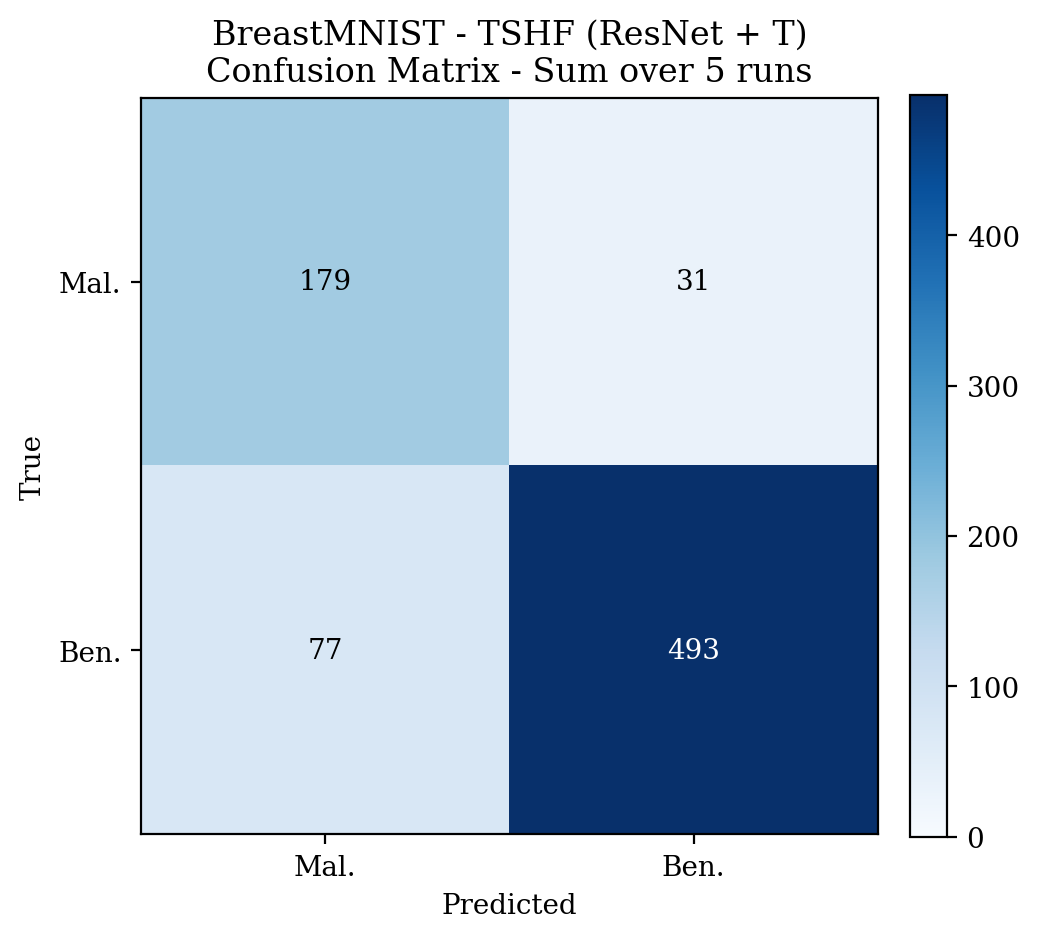}}

    \vspace{0.4em}

    \subfloat[SCNN + non-trainable]{%
        \includegraphics[width=0.47\linewidth]{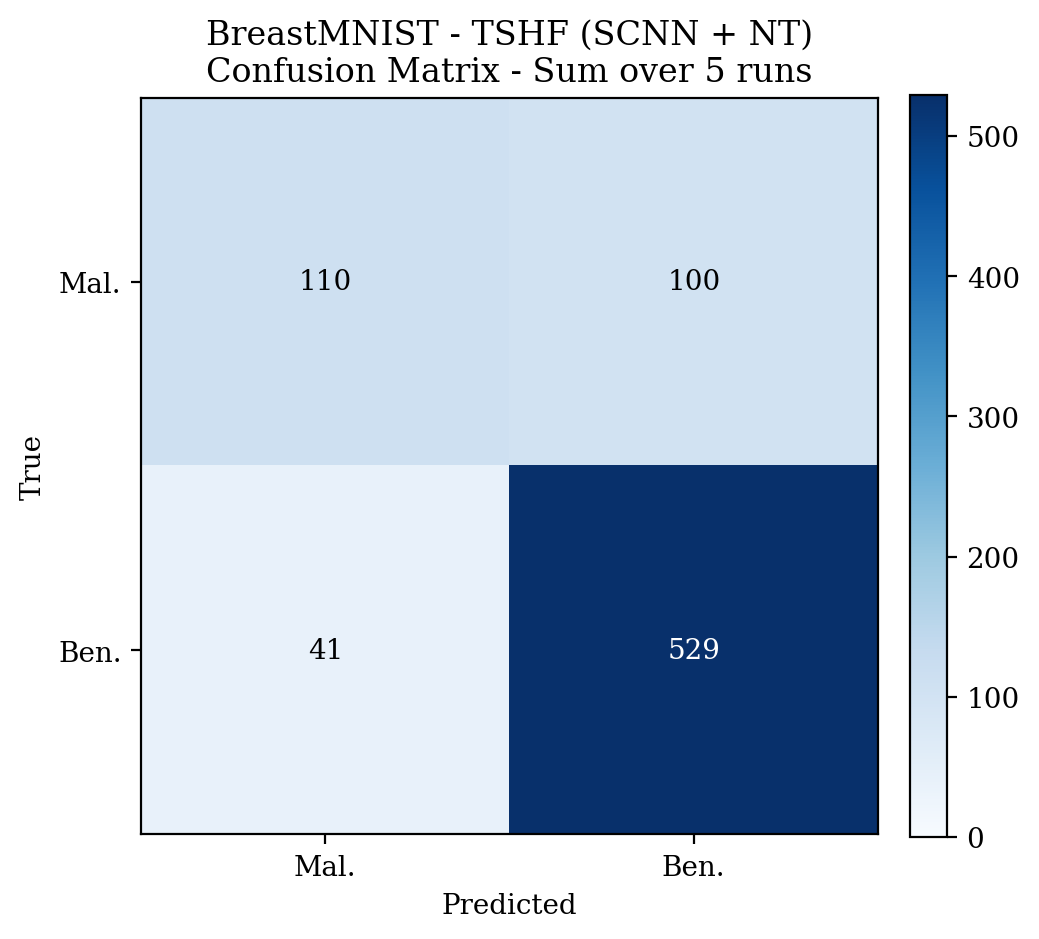}}
    \hfill
    \subfloat[SCNN + trainable]{%
        \includegraphics[width=0.47\linewidth]{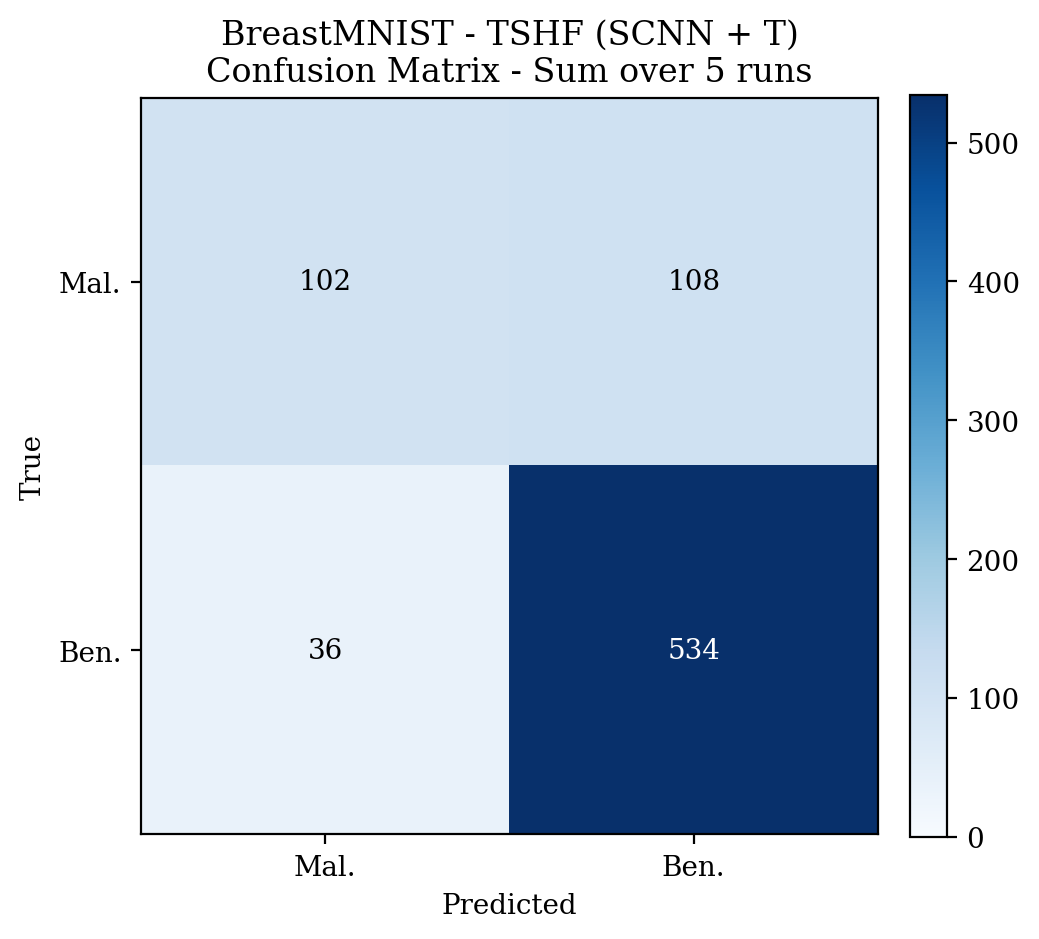}}

    \caption{Confusion matrices of the \textbf{TSHF} configurations on the
    BreastMNIST test set, summed over five independent runs.}
    \label{fig:cm_breastmnist_tshf}
\end{figure}
\clearpage

\subsection{INbreast: Standalone Baselines}
\label{app:cm_inbreast_baselines}

Counts are over $5 \times 96 = 480$ predictions, of which 250 are benign and 230
malignant. The corresponding metrics are reported in
Tables~\ref{tab:standalone_baselines} and~\ref{tab:metrics_inbreast}.

\begin{figure}[htbp]
    \centering
    \subfloat[SCNN]{%
        \includegraphics[width=0.47\linewidth]{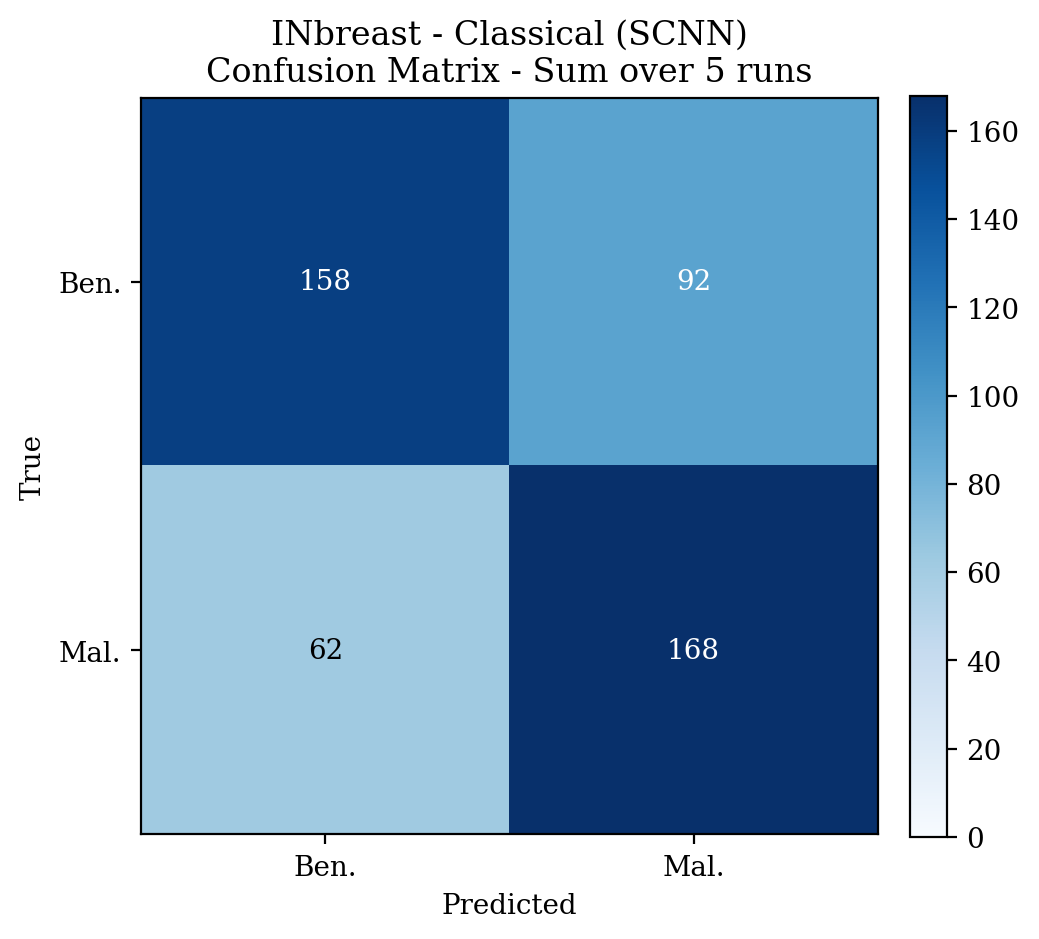}}
    \hfill
    \subfloat[ResNet-18]{%
        \includegraphics[width=0.47\linewidth]{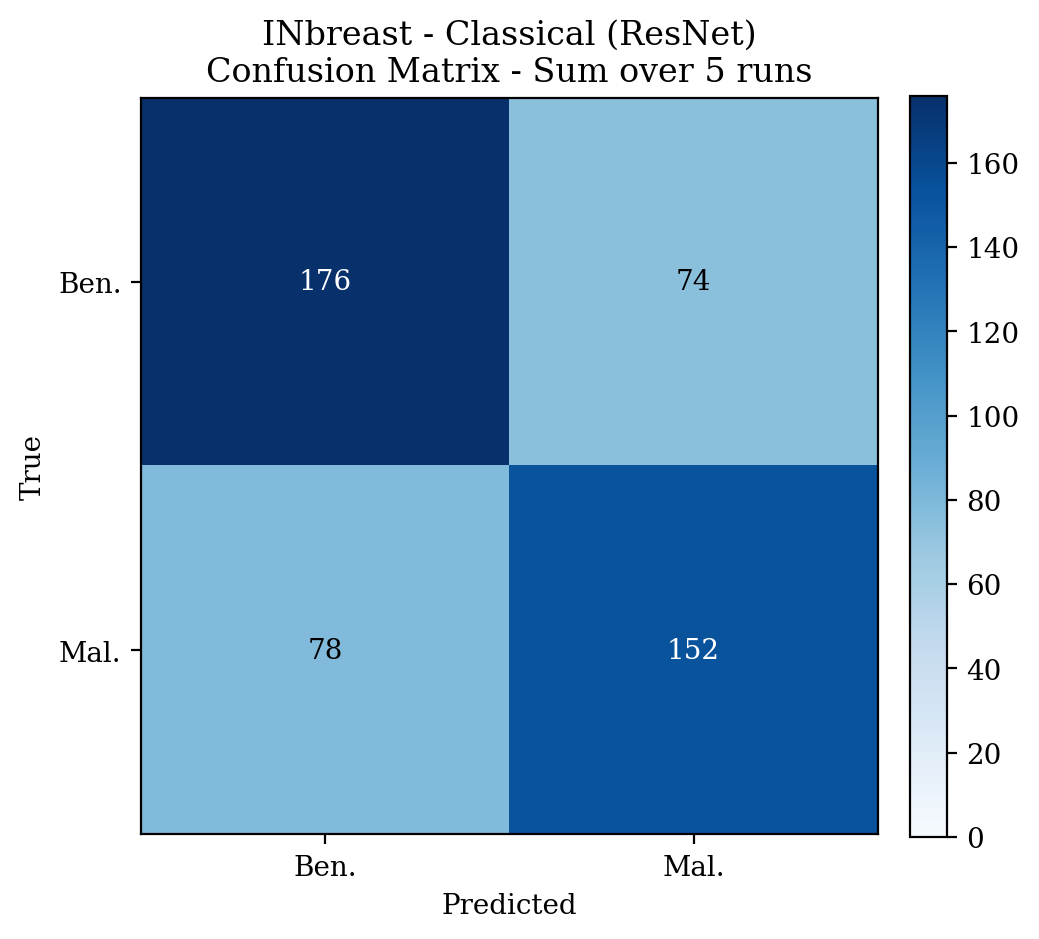}}

    \vspace{0.4em}

    \subfloat[Quantum, non-trainable]{%
        \includegraphics[width=0.47\linewidth]{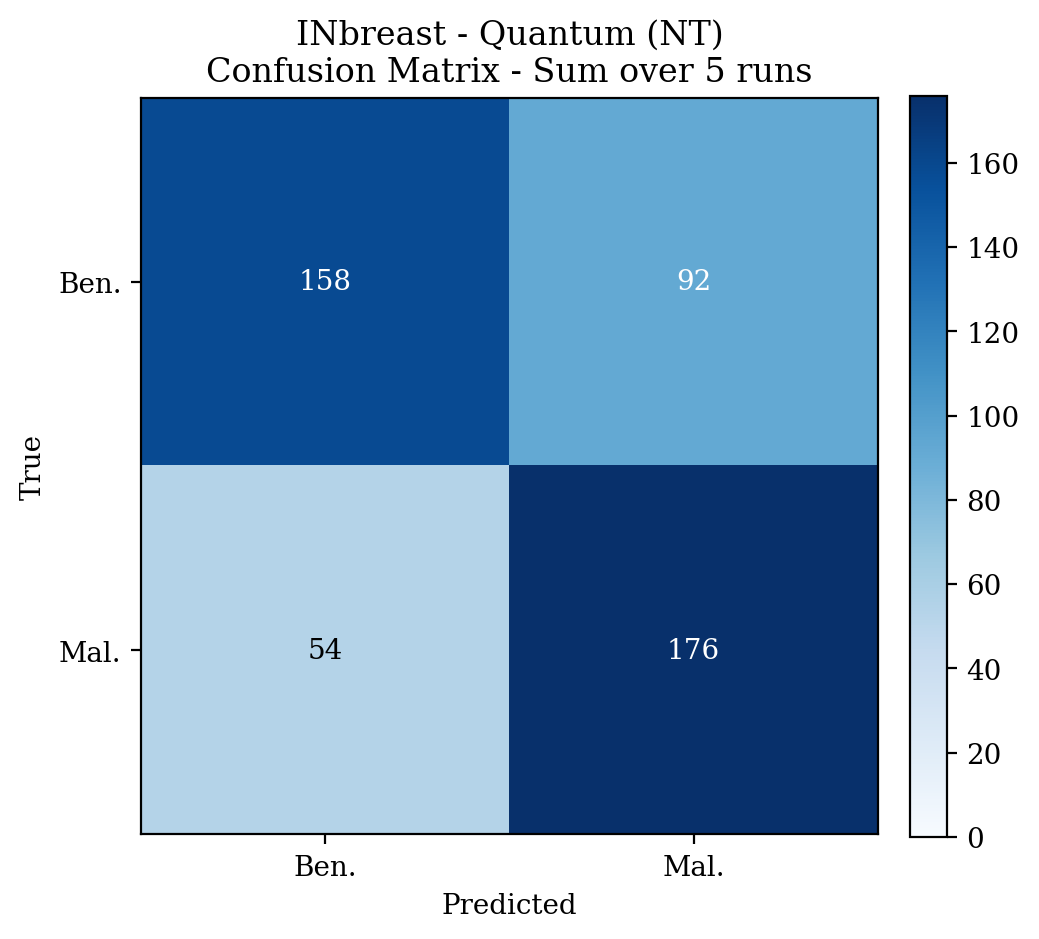}}
    \hfill
    \subfloat[Quantum, trainable]{%
        \includegraphics[width=0.47\linewidth]{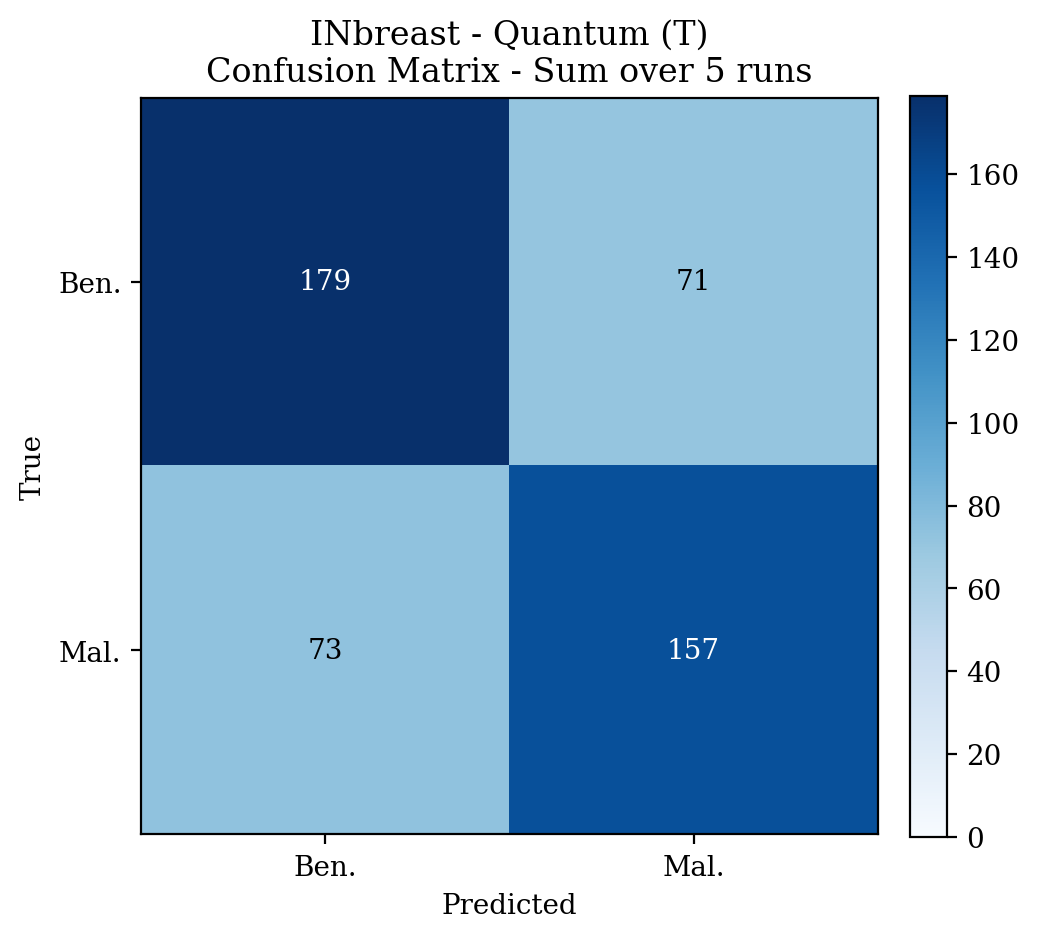}}

    \caption{Confusion matrices of the standalone baselines on the INbreast test
    set, summed over five independent runs.}
    \label{fig:cm_inbreast_baselines}
\end{figure}
\clearpage

\subsection{INbreast: Static Hybrid Fusion}
\label{app:cm_inbreast_shf}

\begin{figure}[htbp]
    \centering
    \subfloat[ResNet-18 + non-trainable]{%
        \includegraphics[width=0.47\linewidth]{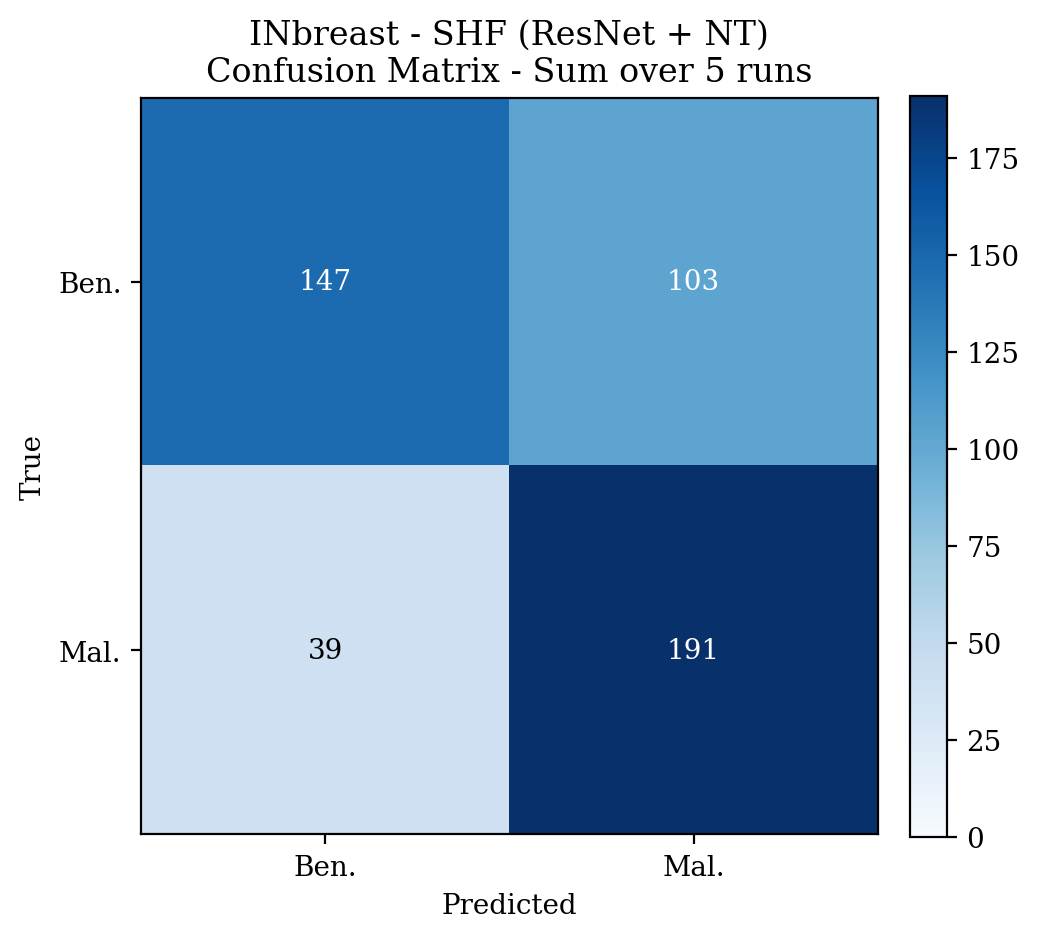}}
    \hfill
    \subfloat[ResNet-18 + trainable]{%
        \includegraphics[width=0.47\linewidth]{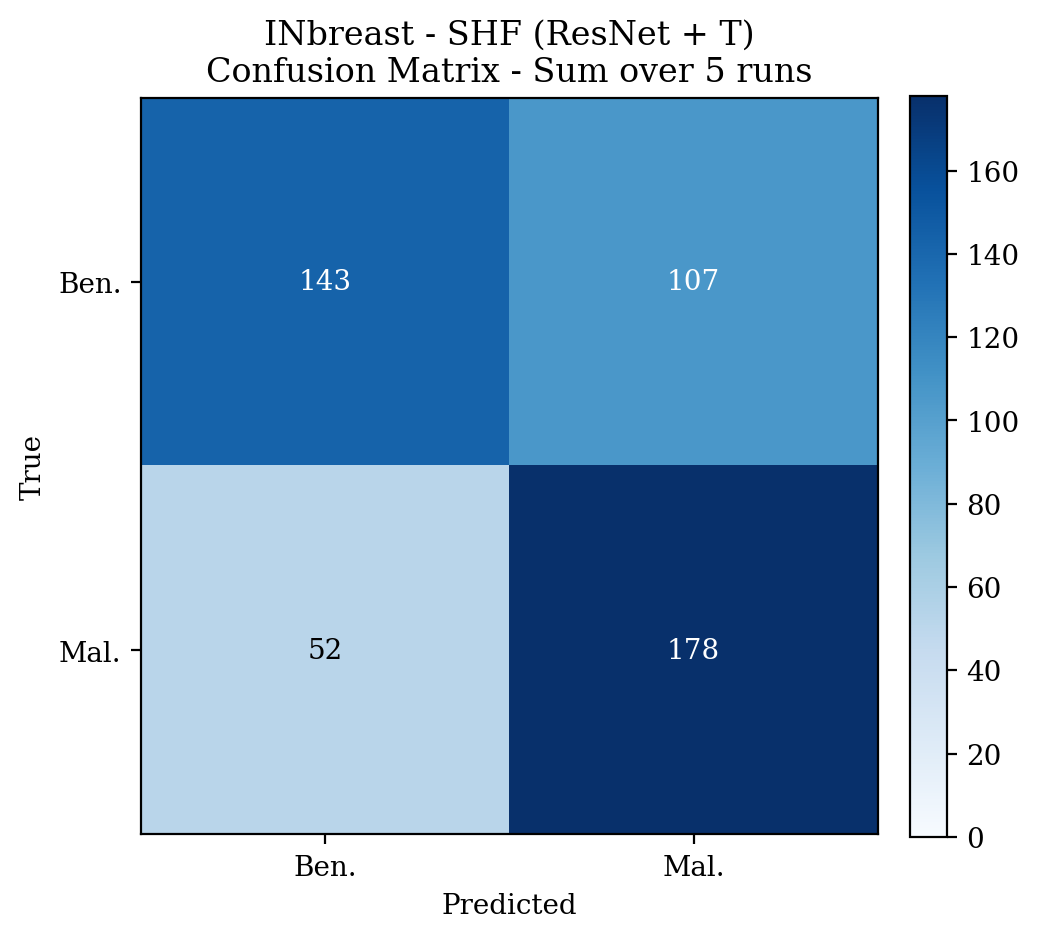}}

    \vspace{0.4em}

    \subfloat[SCNN + non-trainable]{%
        \includegraphics[width=0.47\linewidth]{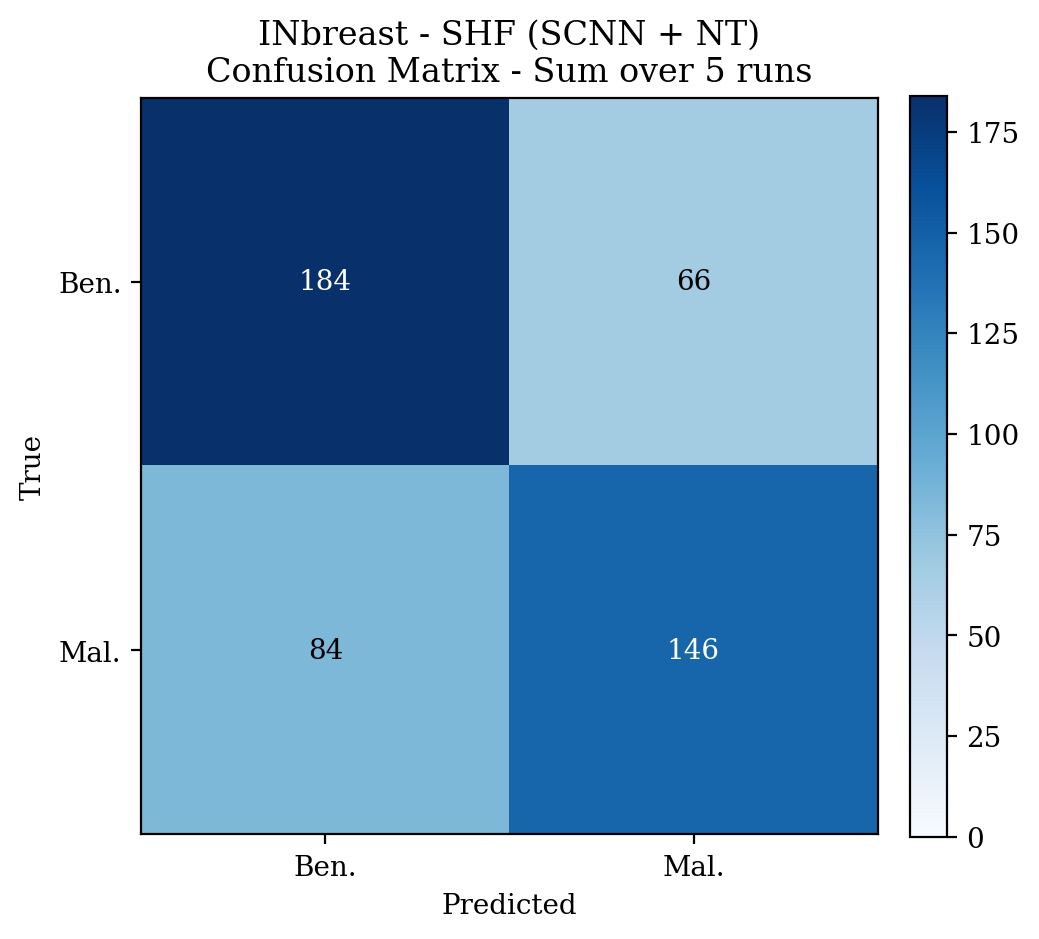}}
    \hfill
    \subfloat[SCNN + trainable]{%
        \includegraphics[width=0.47\linewidth]{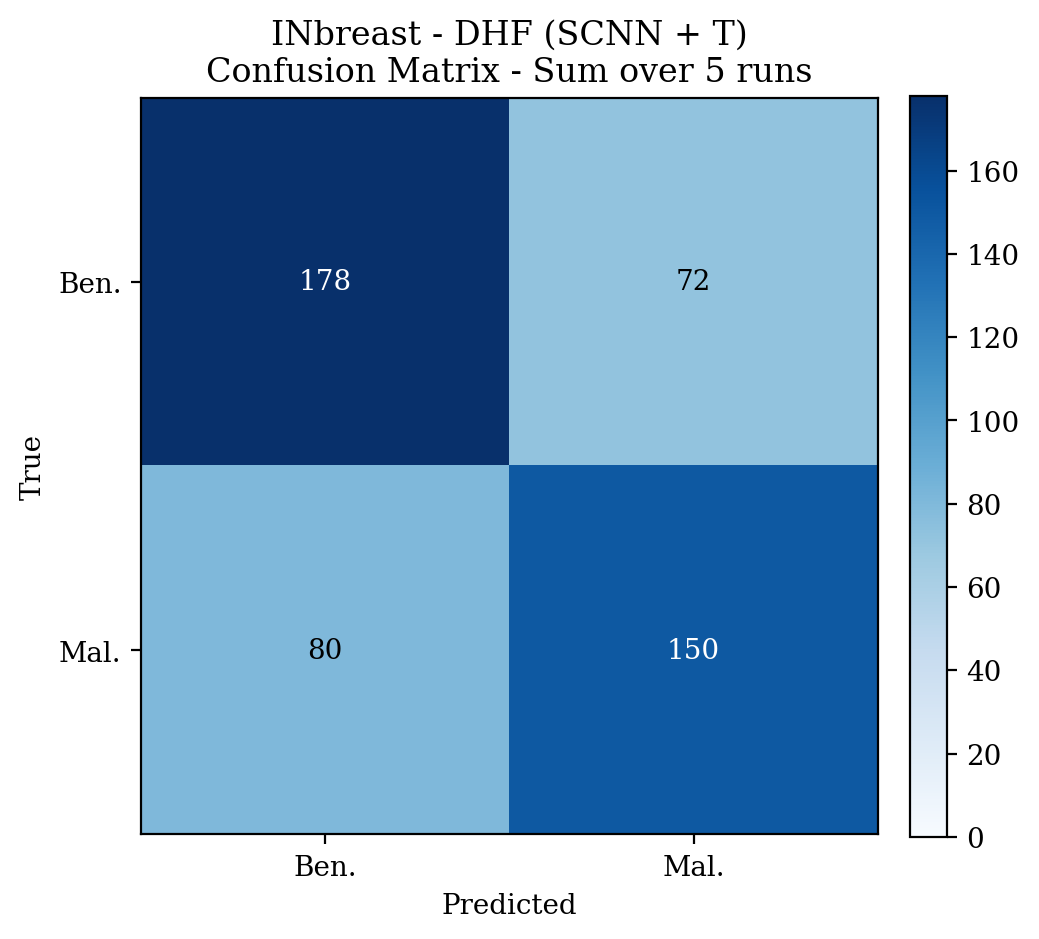}}

    \caption{Confusion matrices of the \textbf{SHF} configurations on the
    INbreast test set, summed over five independent runs.}
    \label{fig:cm_inbreast_shf}
\end{figure}
\clearpage

\subsection{INbreast: Dynamic Hybrid Fusion}
\label{app:cm_inbreast_dhf}

\begin{figure}[htbp]
    \centering
    \subfloat[ResNet-18 + non-trainable]{%
        \includegraphics[width=0.47\linewidth]{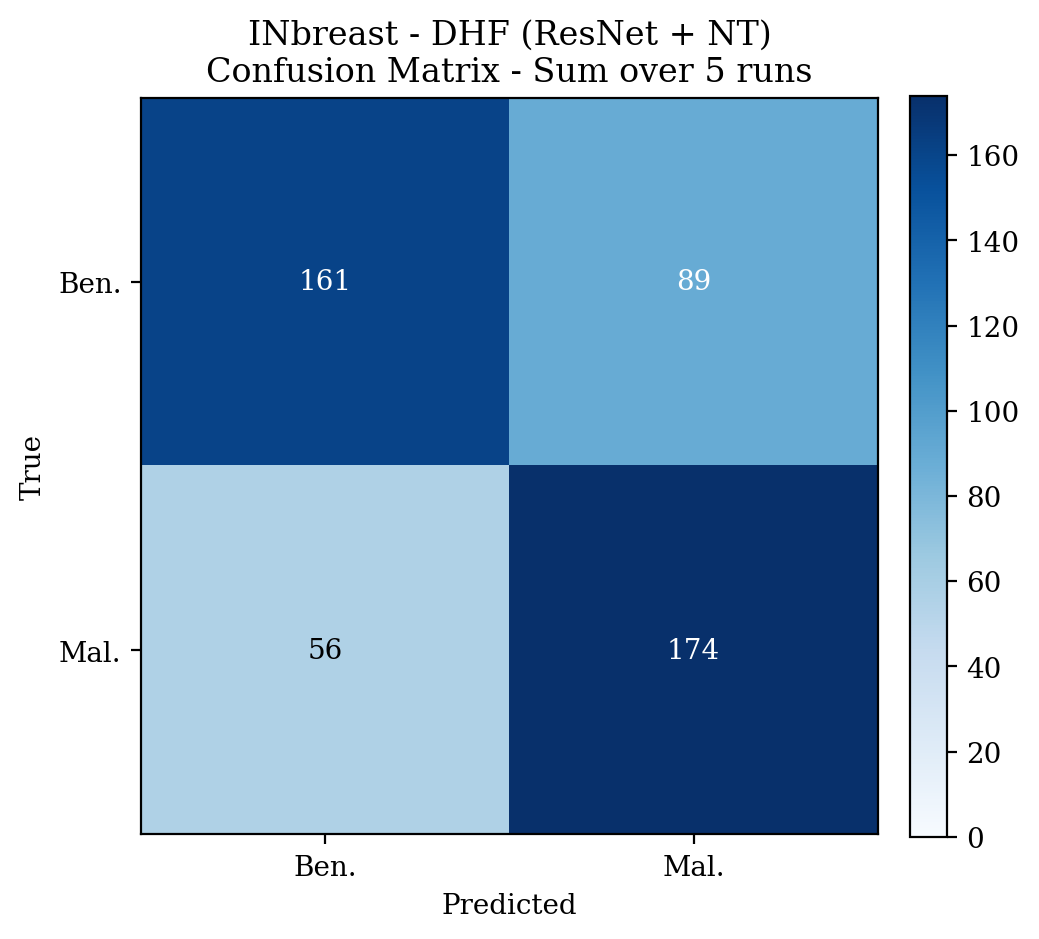}}
    \hfill
    \subfloat[ResNet-18 + trainable]{%
        \includegraphics[width=0.47\linewidth]{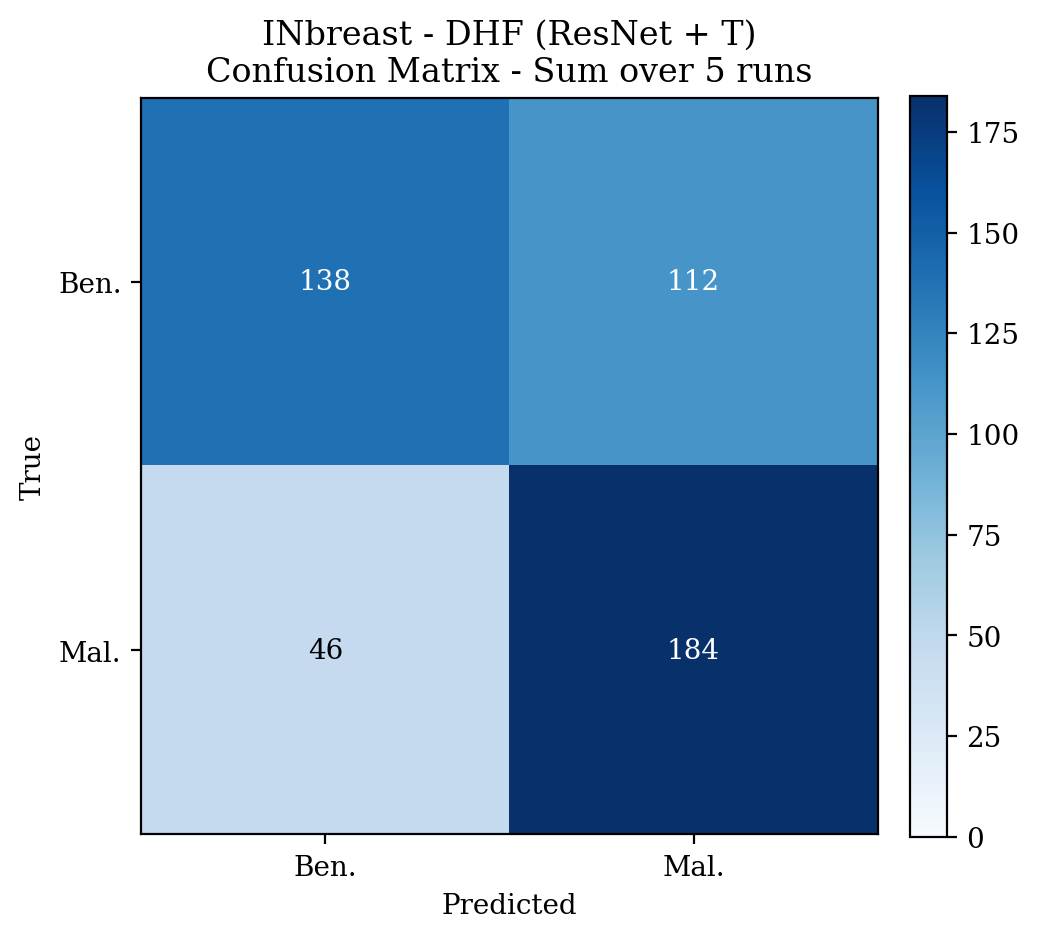}}

    \vspace{0.4em}

    \subfloat[SCNN + non-trainable]{%
        \includegraphics[width=0.47\linewidth]{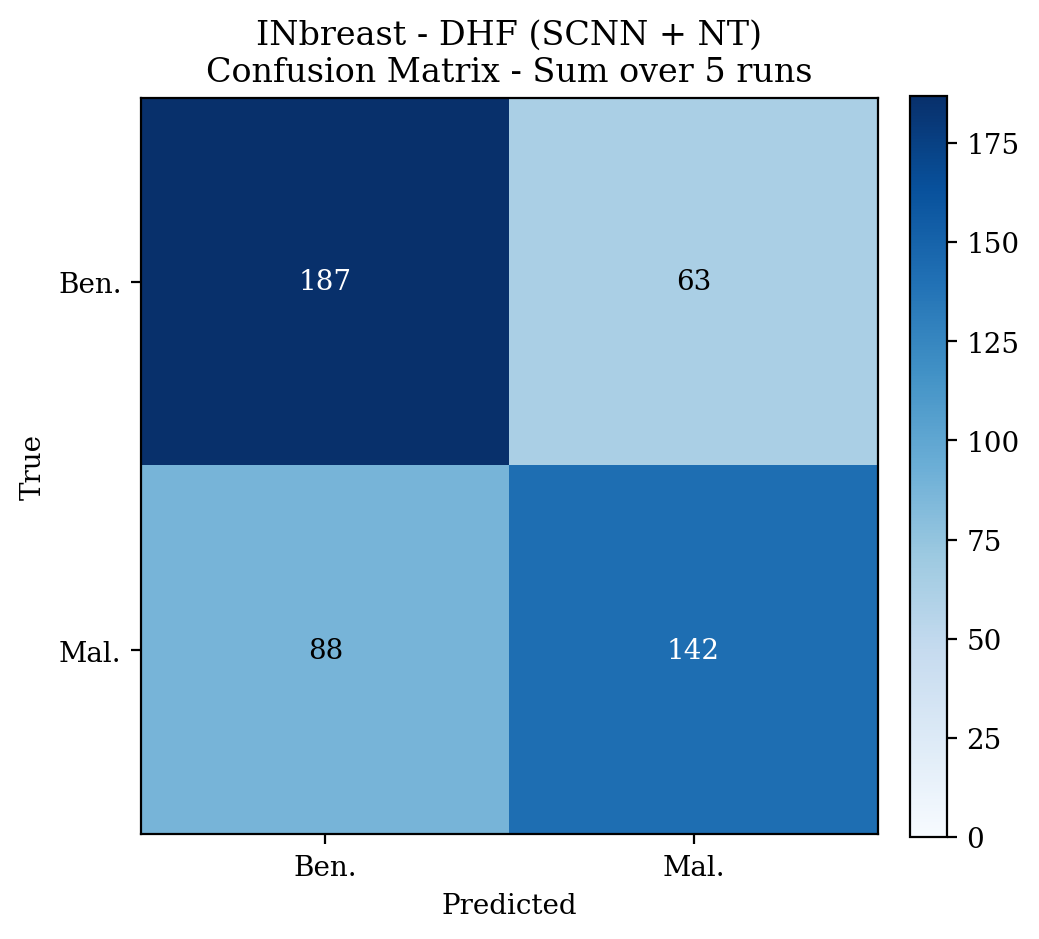}}
    \hfill
    \subfloat[SCNN + trainable]{%
        \includegraphics[width=0.47\linewidth]{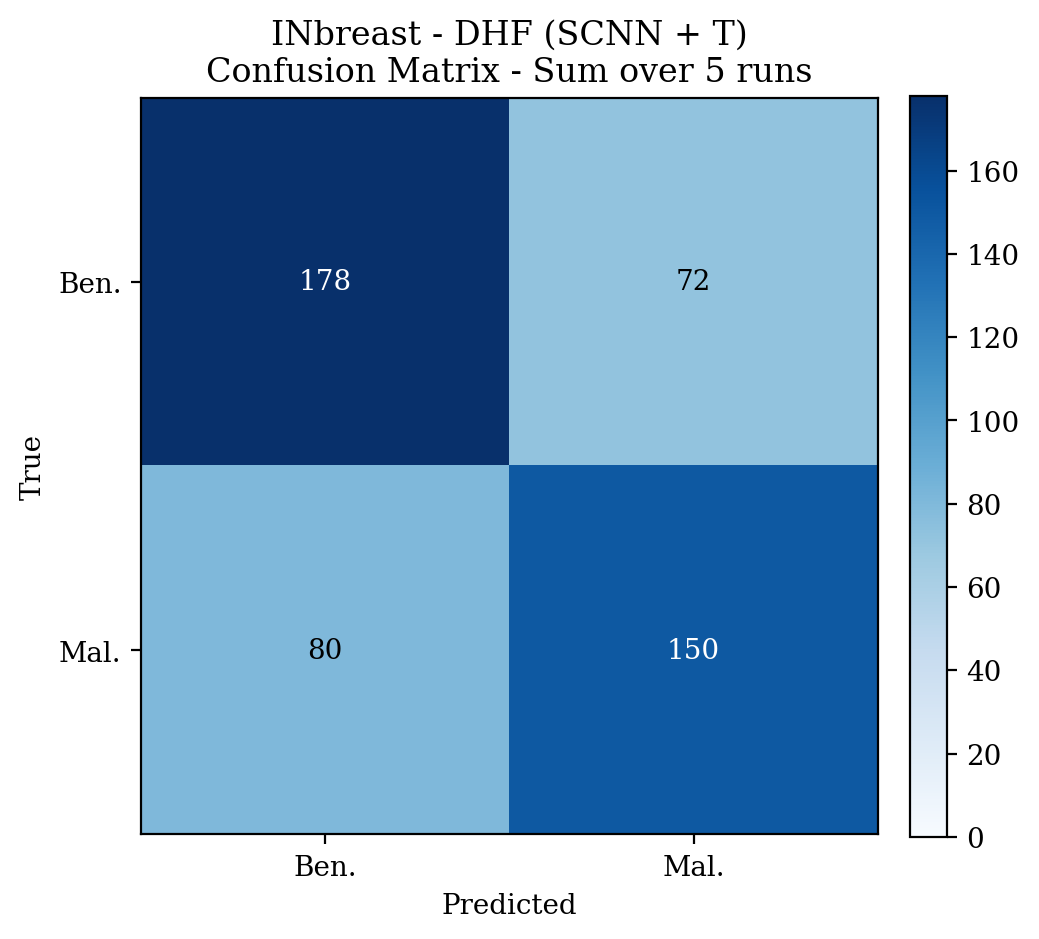}}

    \caption{Confusion matrices of the \textbf{DHF} configurations on the
    INbreast test set, summed over five independent runs.}
    \label{fig:cm_inbreast_dhf}
\end{figure}
\clearpage

\subsection{INbreast: Temperature-Scaled Hybrid Fusion}
\label{app:cm_inbreast_tshf}

\begin{figure}[htbp]
    \centering
    \subfloat[ResNet-18 + non-trainable]{%
        \includegraphics[width=0.47\linewidth]{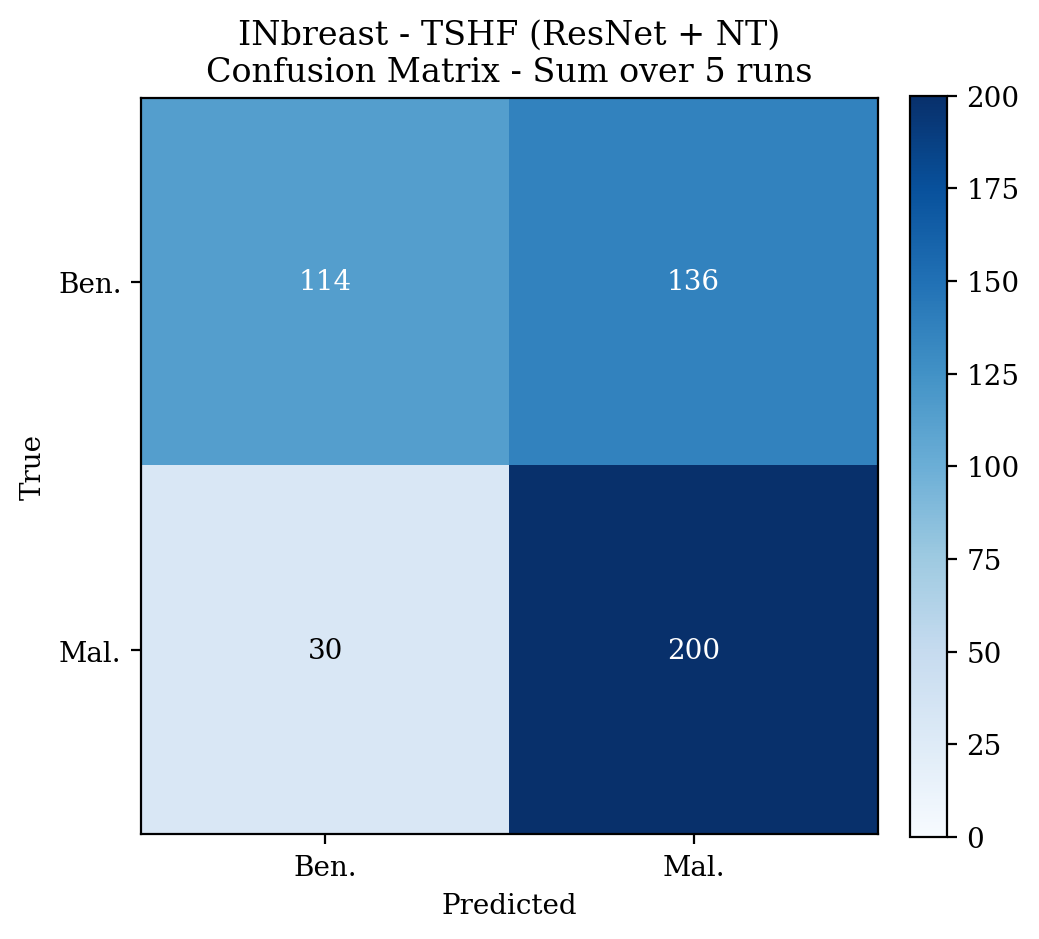}}
    \hfill
    \subfloat[ResNet-18 + trainable]{%
        \includegraphics[width=0.47\linewidth]{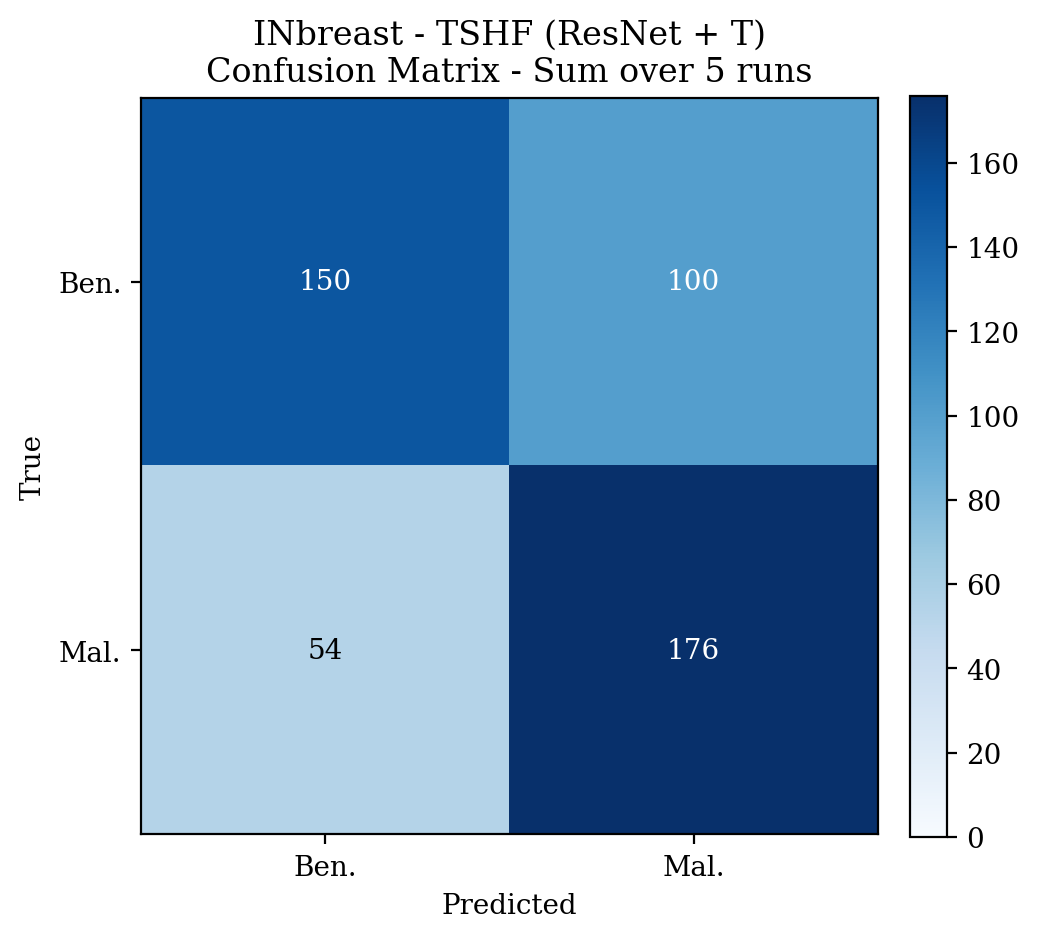}}

    \vspace{0.4em}

    \subfloat[SCNN + non-trainable]{%
        \includegraphics[width=0.47\linewidth]{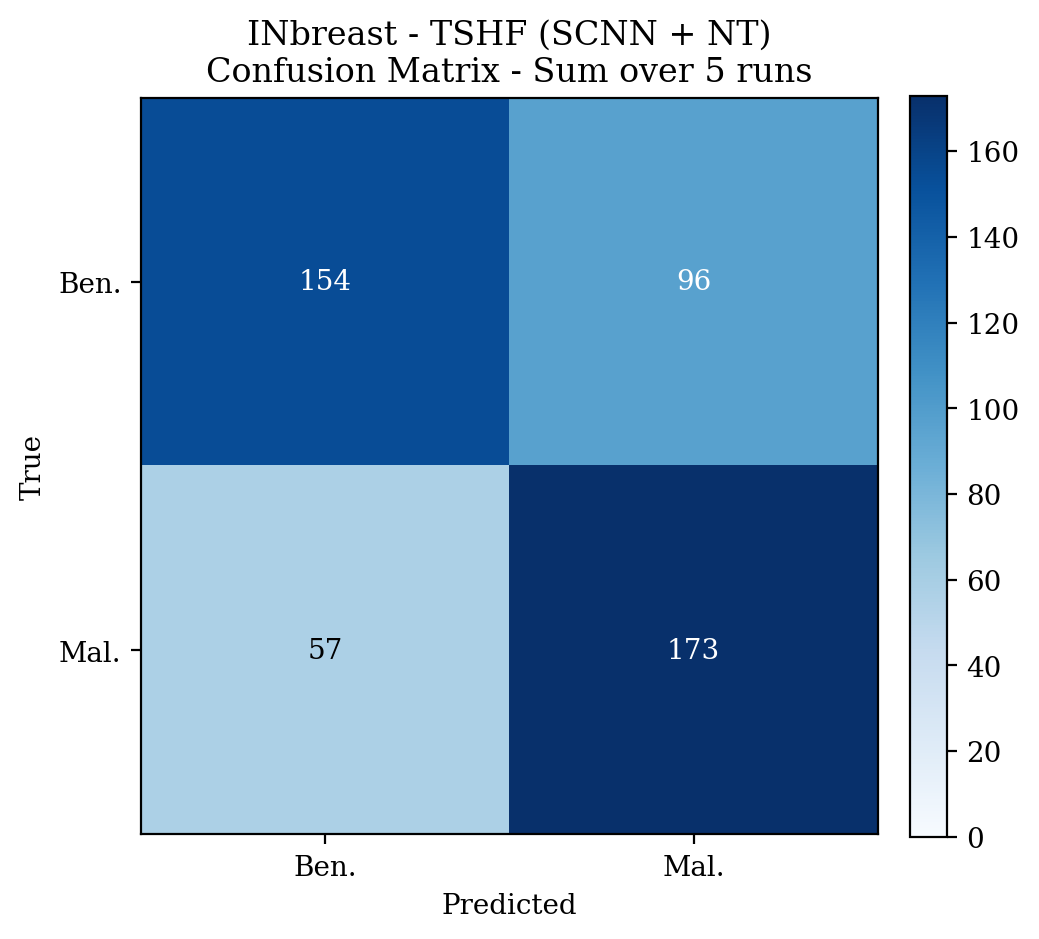}}
    \hfill
    \subfloat[SCNN + trainable]{%
        \includegraphics[width=0.47\linewidth]{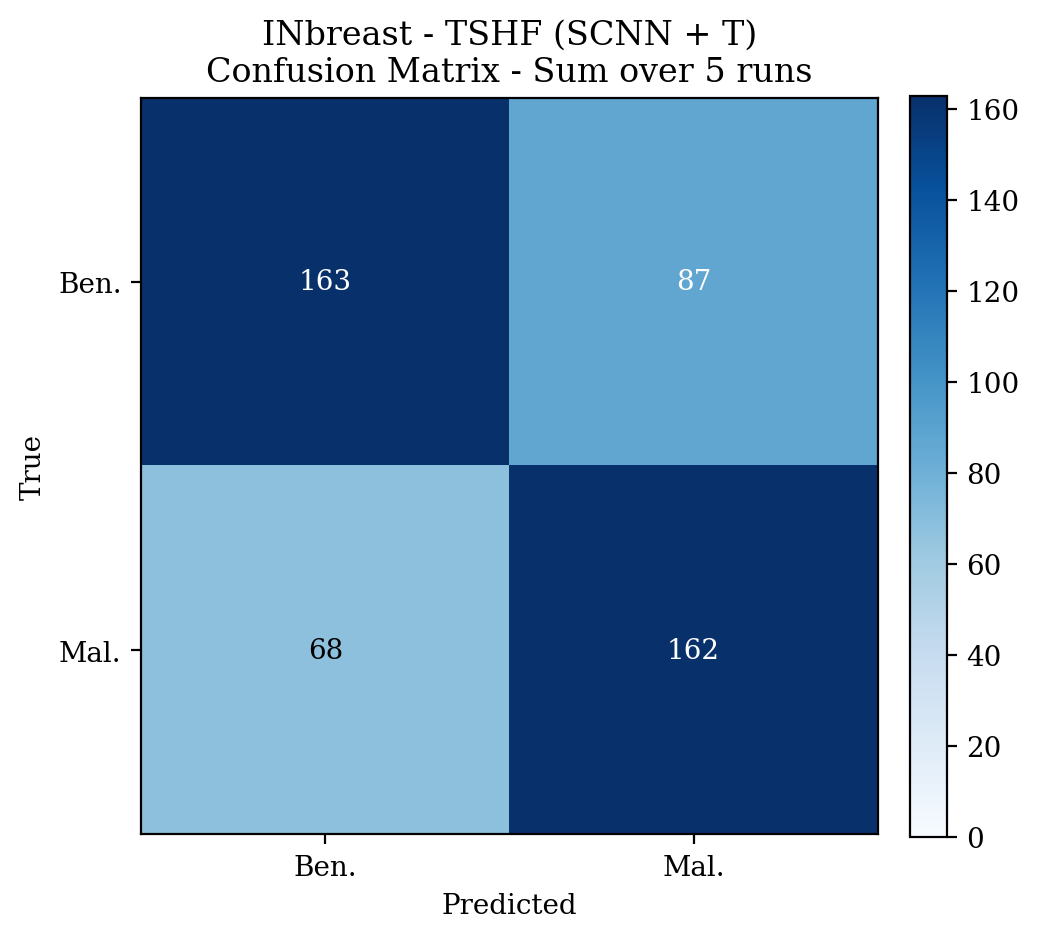}}

    \caption{Confusion matrices of the \textbf{TSHF} configurations on the
    INbreast test set, summed over five independent runs.}
    \label{fig:cm_inbreast_tshf}
\end{figure}
\clearpage

\subsection{BUS-UCLM: Standalone Baselines}
\label{app:cm_busuclm_baselines}

Counts are over $5 \times 26 = 130$ predictions, of which 35 are benign and 95
malignant. The corresponding metrics are reported in
Tables~\ref{tab:standalone_baselines} and~\ref{tab:metrics_busuclm}.

\begin{figure}[htbp]
    \centering
    \subfloat[SCNN]{%
        \includegraphics[width=0.47\linewidth]{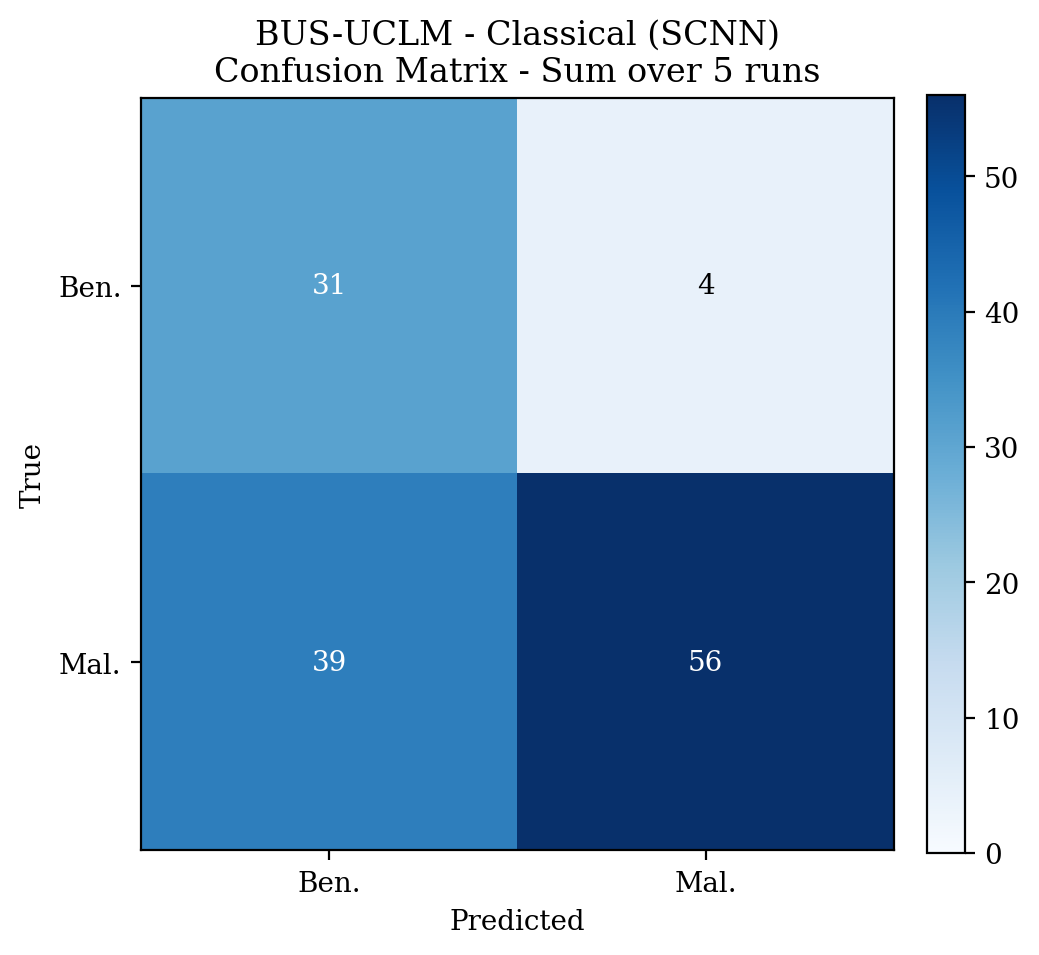}}
    \hfill
    \subfloat[ResNet-18]{%
        \includegraphics[width=0.47\linewidth]{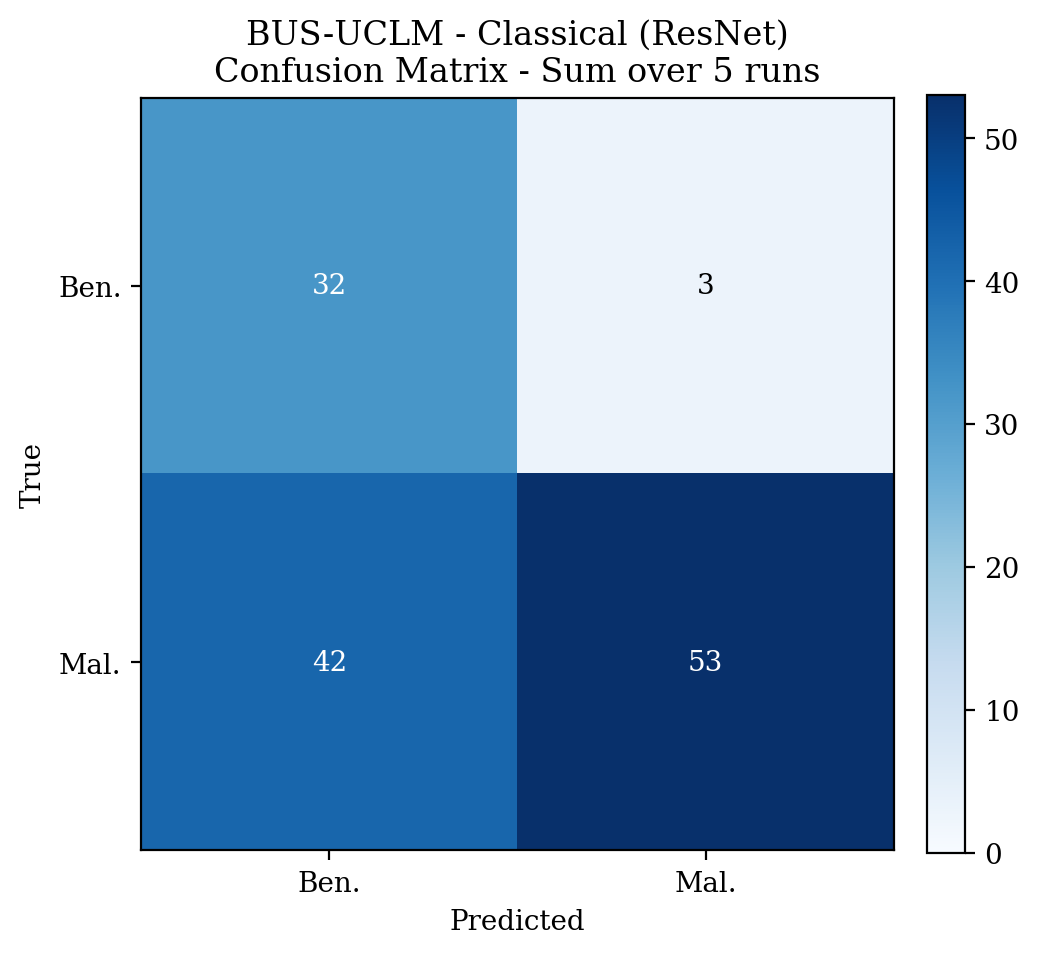}}

    \vspace{0.4em}

    \subfloat[Quantum, non-trainable]{%
        \includegraphics[width=0.47\linewidth]{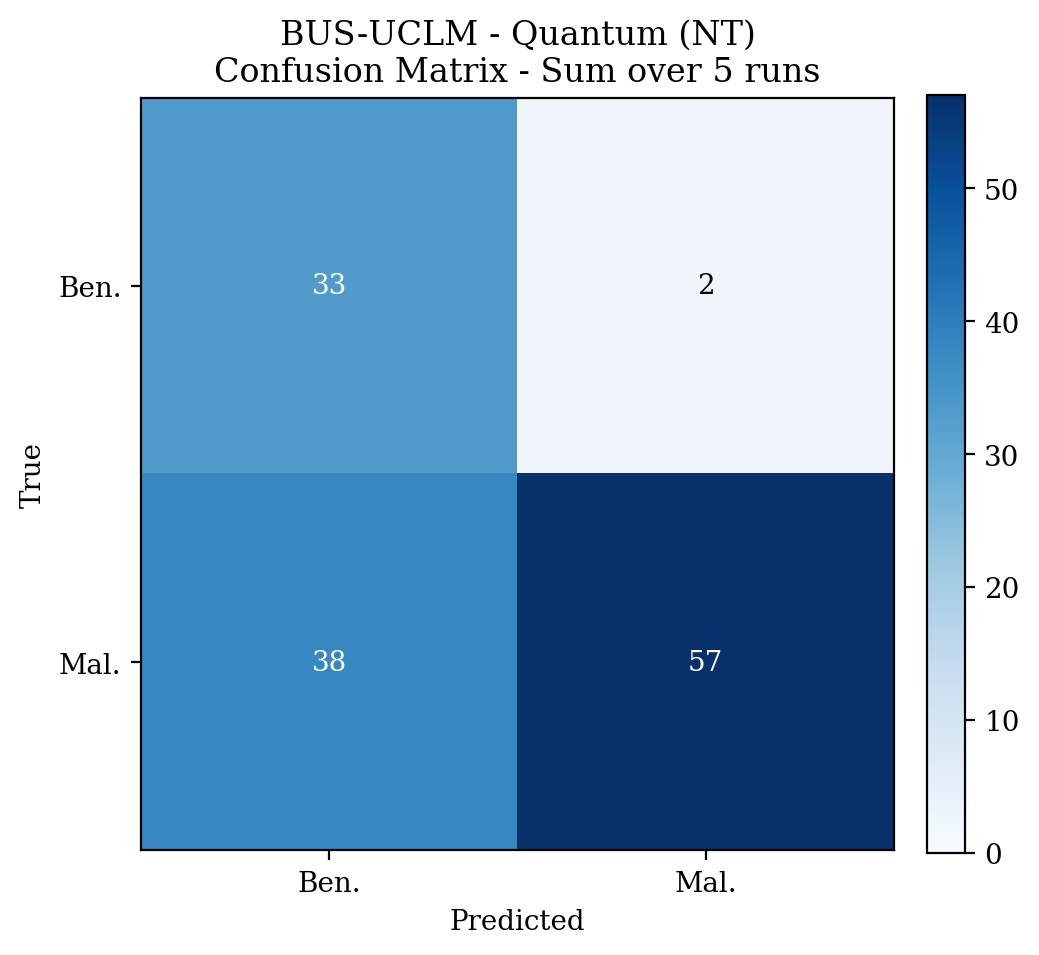}}
    \hfill
    \subfloat[Quantum, trainable]{%
        \includegraphics[width=0.47\linewidth]{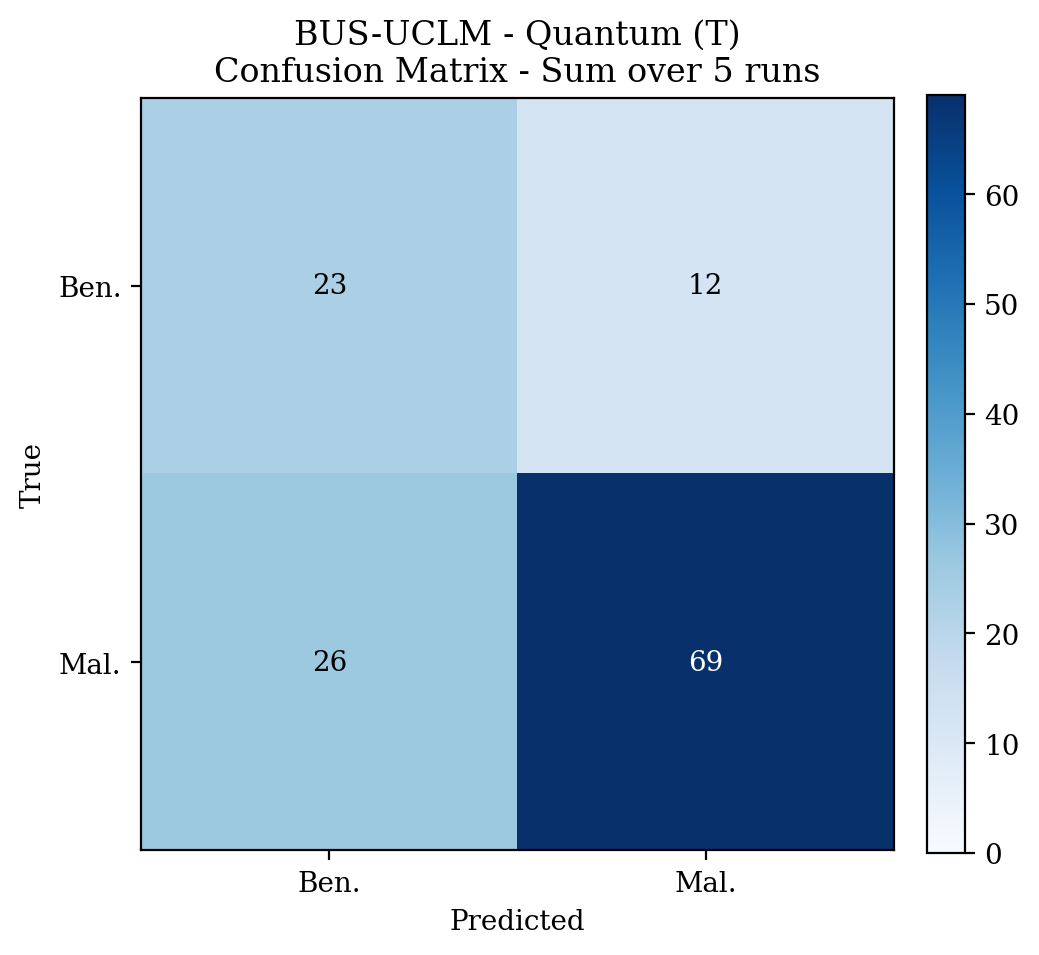}}

    \caption{Confusion matrices of the standalone baselines on the BUS-UCLM test
    set, summed over five independent runs.}
    \label{fig:cm_busuclm_baselines}
\end{figure}
\clearpage

\subsection{BUS-UCLM: Static Hybrid Fusion}
\label{app:cm_busuclm_shf}

\begin{figure}[htbp]
    \centering
    \subfloat[ResNet-18 + non-trainable]{%
        \includegraphics[width=0.47\linewidth]{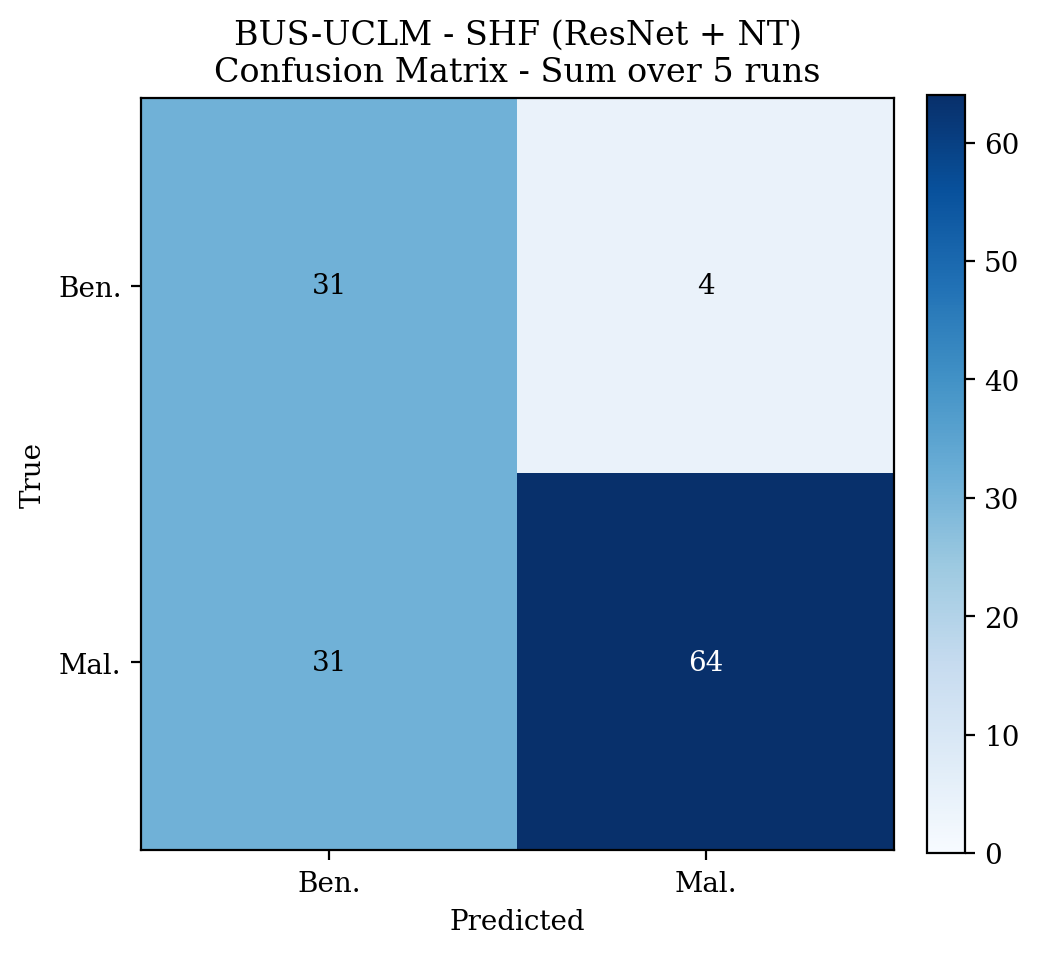}}
    \hfill
    \subfloat[ResNet-18 + trainable]{%
        \includegraphics[width=0.47\linewidth]{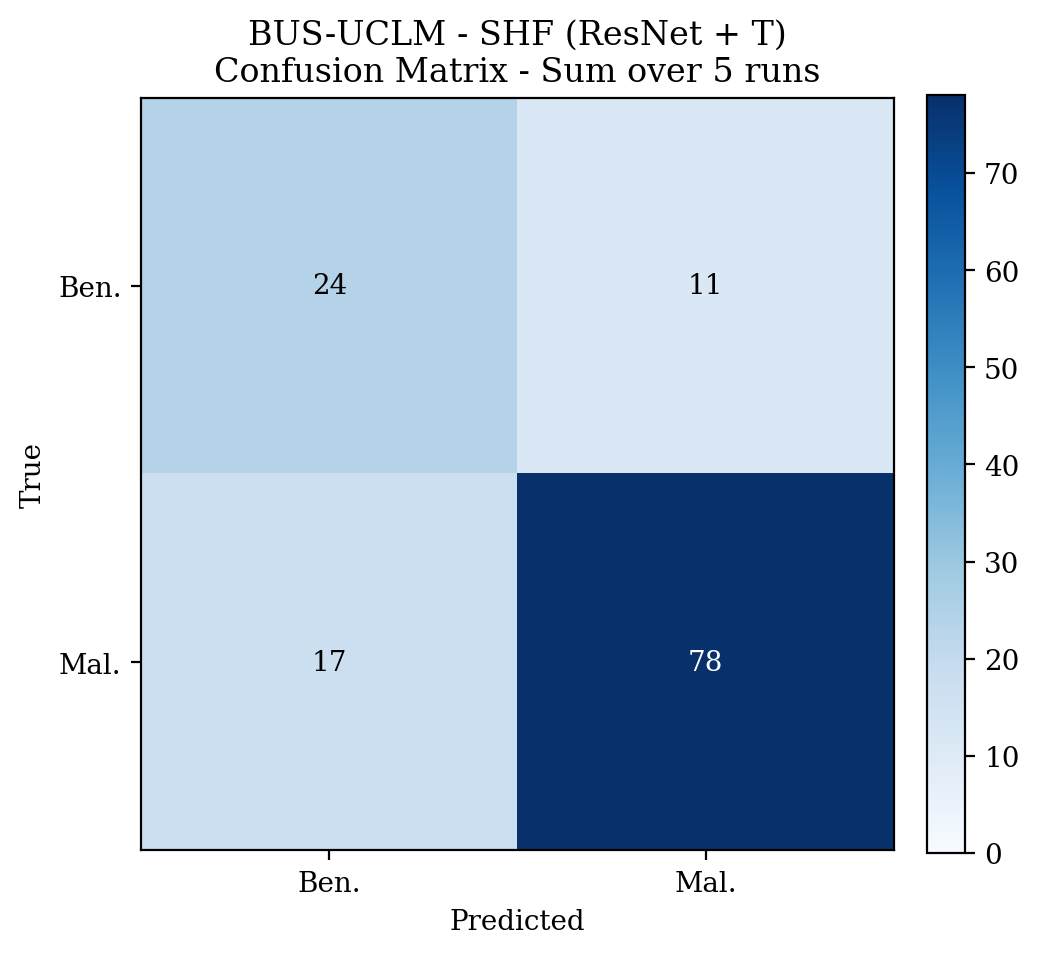}}

    \vspace{0.4em}

    \subfloat[SCNN + non-trainable]{%
        \includegraphics[width=0.47\linewidth]{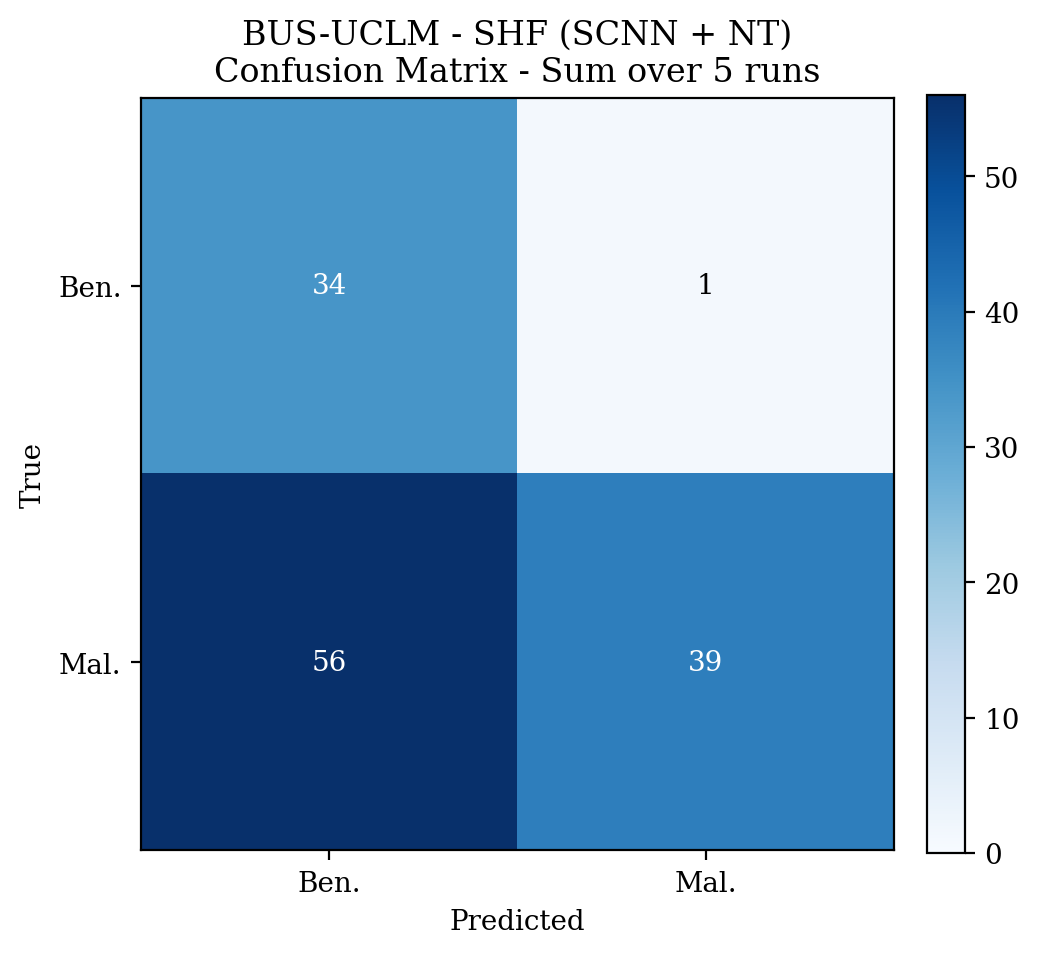}}
    \hfill
    \subfloat[SCNN + trainable]{%
        \includegraphics[width=0.47\linewidth]{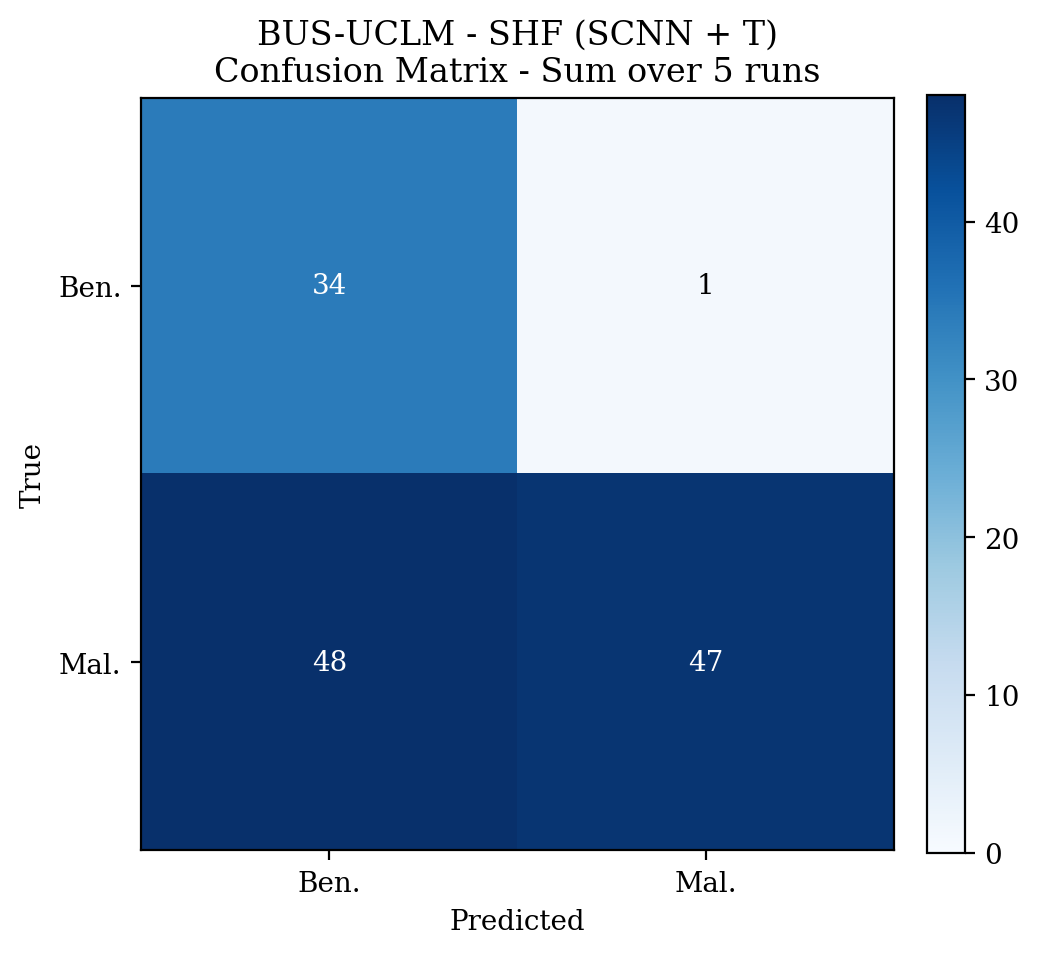}}

    \caption{Confusion matrices of the \textbf{SHF} configurations on the
    BUS-UCLM test set, summed over five independent runs.}
    \label{fig:cm_busuclm_shf}
\end{figure}
\clearpage

\subsection{BUS-UCLM: Dynamic Hybrid Fusion}
\label{app:cm_busuclm_dhf}

\begin{figure}[htbp]
    \centering
    \subfloat[ResNet-18 + non-trainable]{%
        \includegraphics[width=0.47\linewidth]{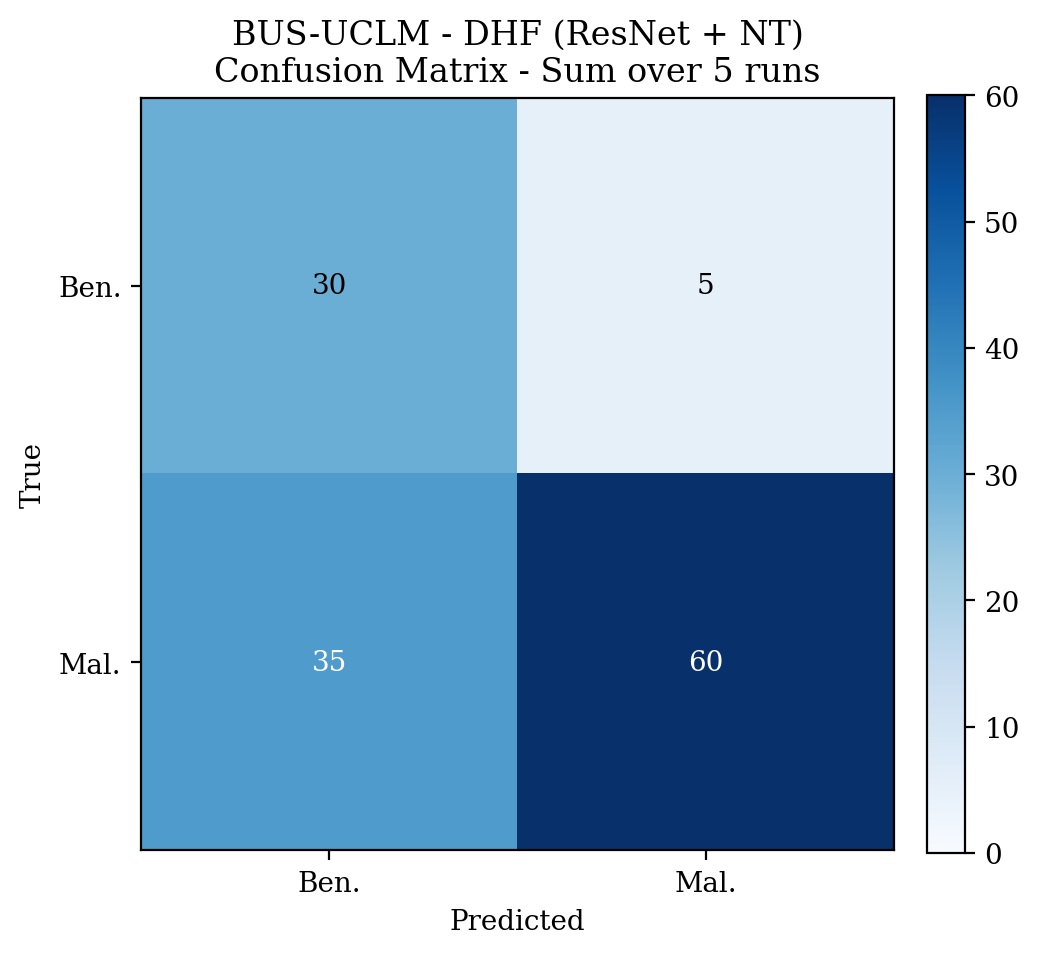}}
    \hfill
    \subfloat[ResNet-18 + trainable]{%
        \includegraphics[width=0.47\linewidth]{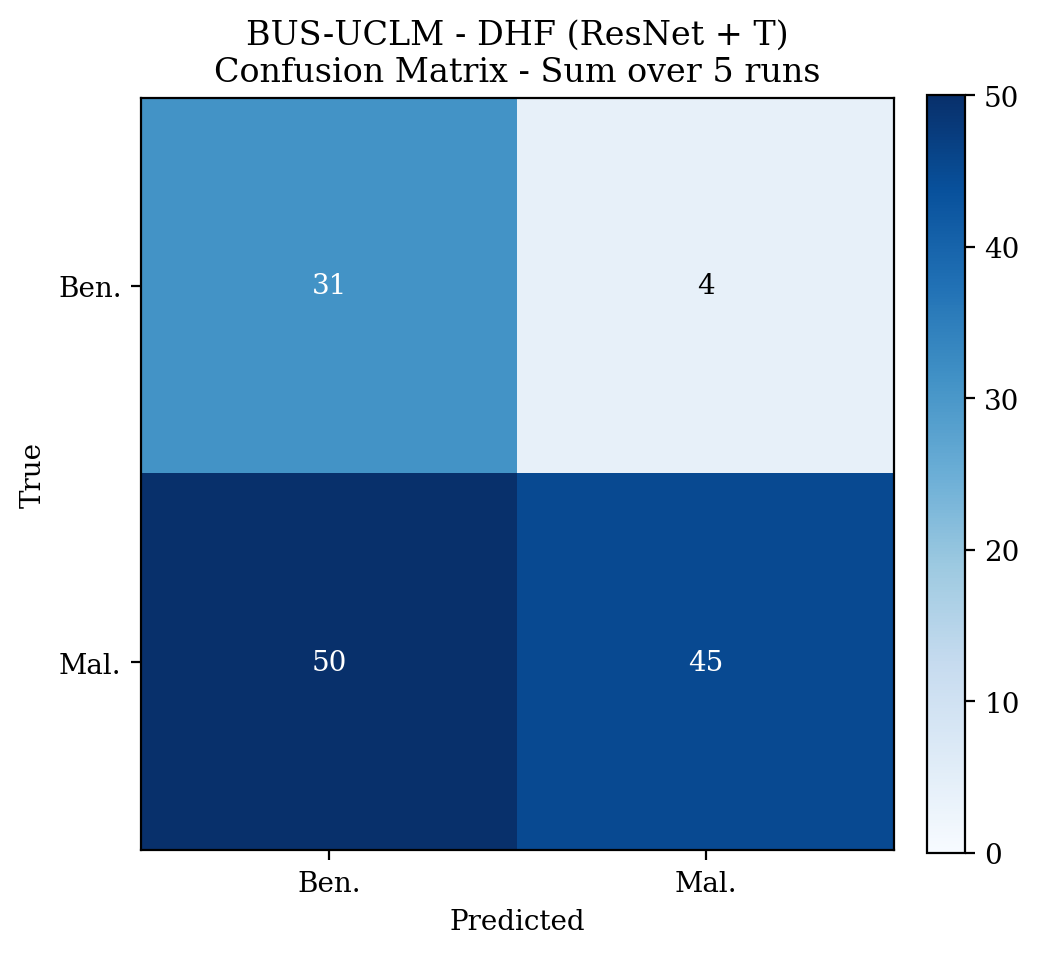}}

    \vspace{0.4em}

    \subfloat[SCNN + non-trainable]{%
        \includegraphics[width=0.47\linewidth]{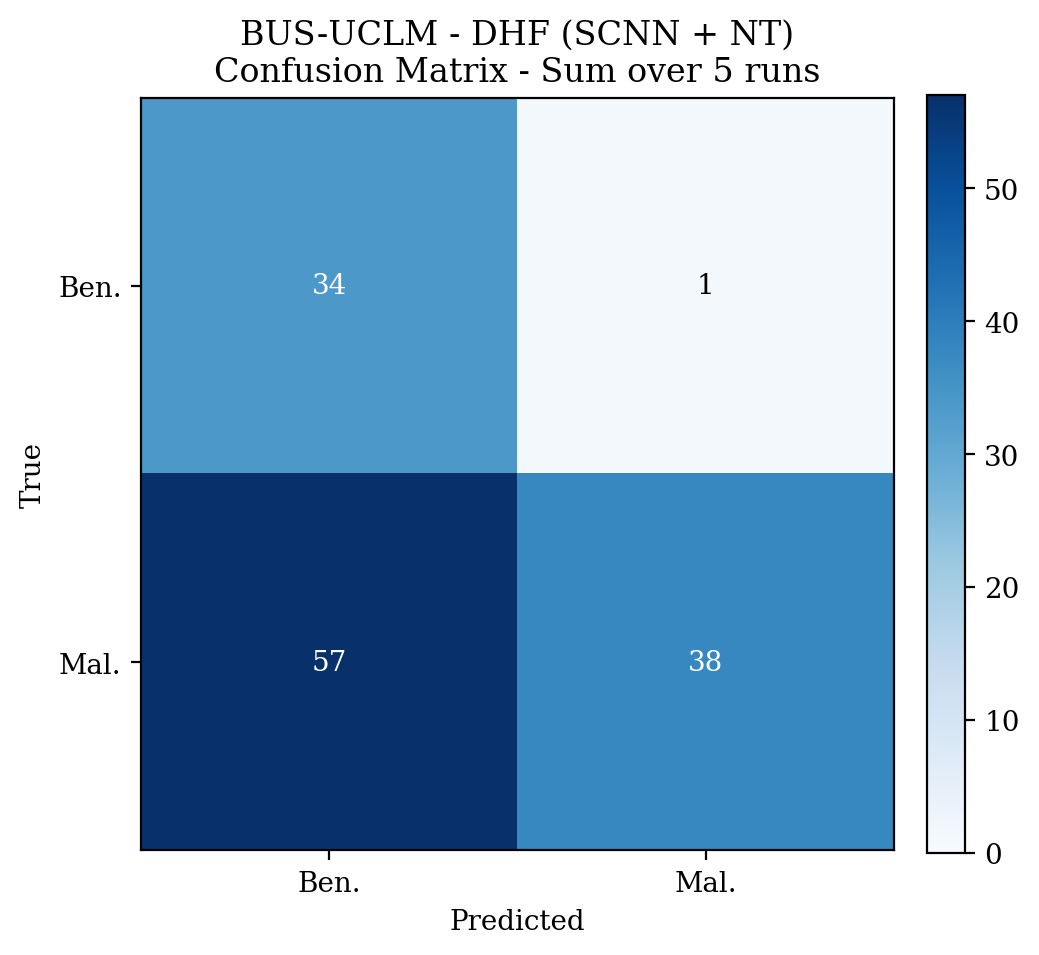}}
    \hfill
    \subfloat[SCNN + trainable]{%
        \includegraphics[width=0.47\linewidth]{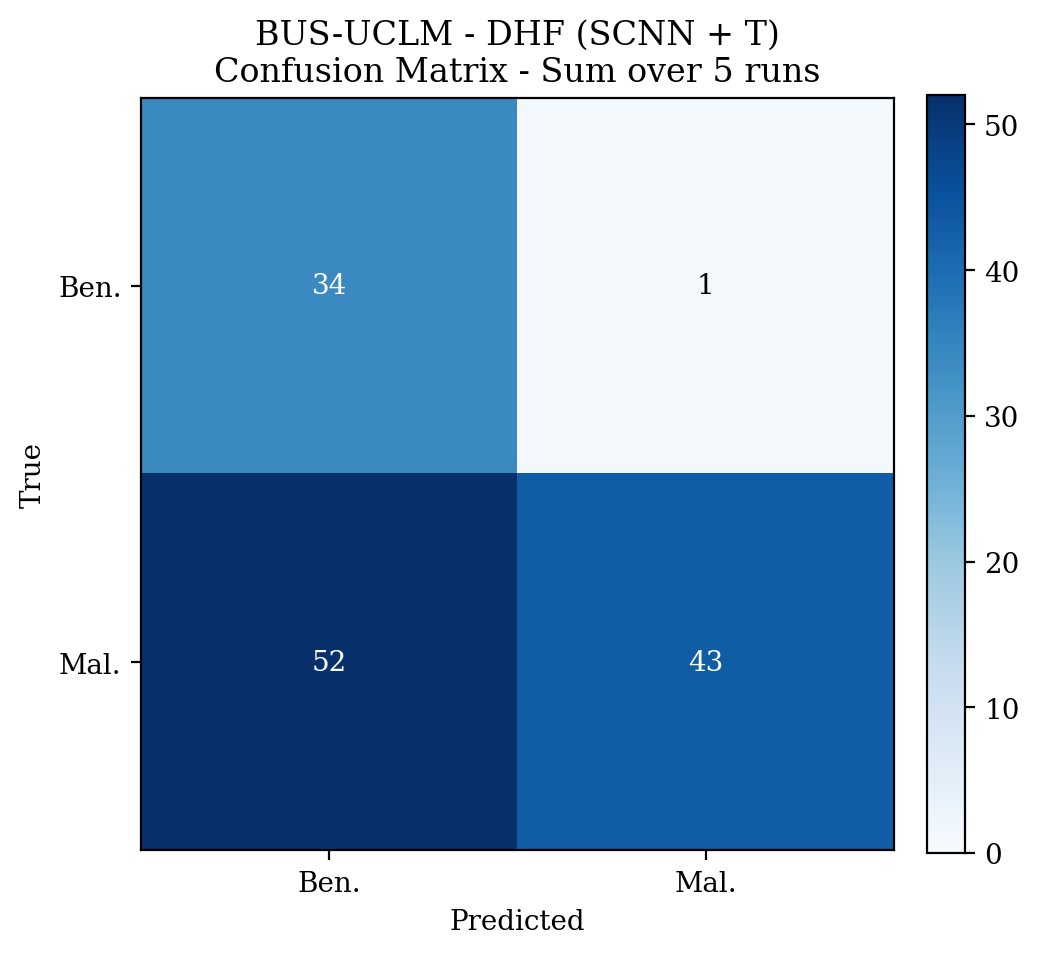}}

    \caption{Confusion matrices of the \textbf{DHF} configurations on the
    BUS-UCLM test set, summed over five independent runs.}
    \label{fig:cm_busuclm_dhf}
\end{figure}
\clearpage

\subsection{BUS-UCLM: Temperature-Scaled Hybrid Fusion}
\label{app:cm_busuclm_tshf}

\begin{figure}[htbp]
    \centering
    \subfloat[ResNet-18 + non-trainable]{%
        \includegraphics[width=0.47\linewidth]{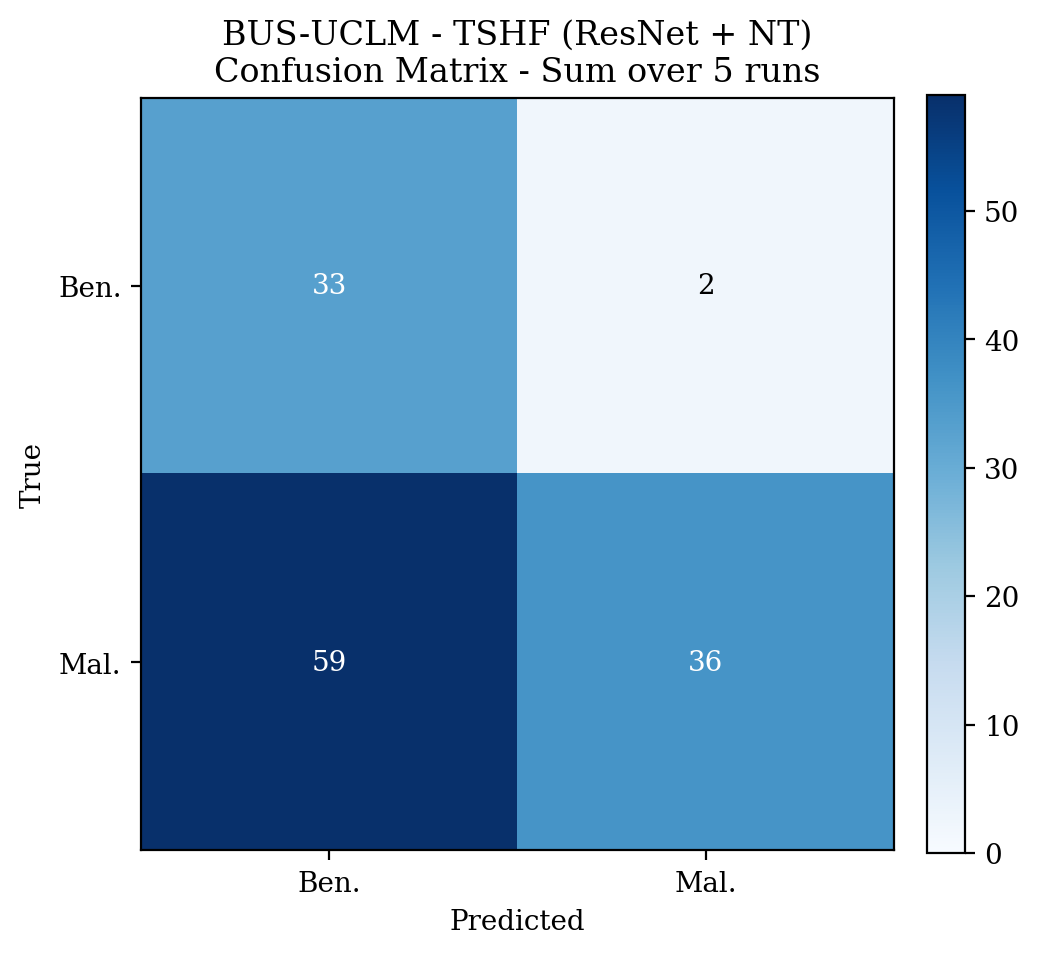}}
    \hfill
    \subfloat[ResNet-18 + trainable]{%
        \includegraphics[width=0.47\linewidth]{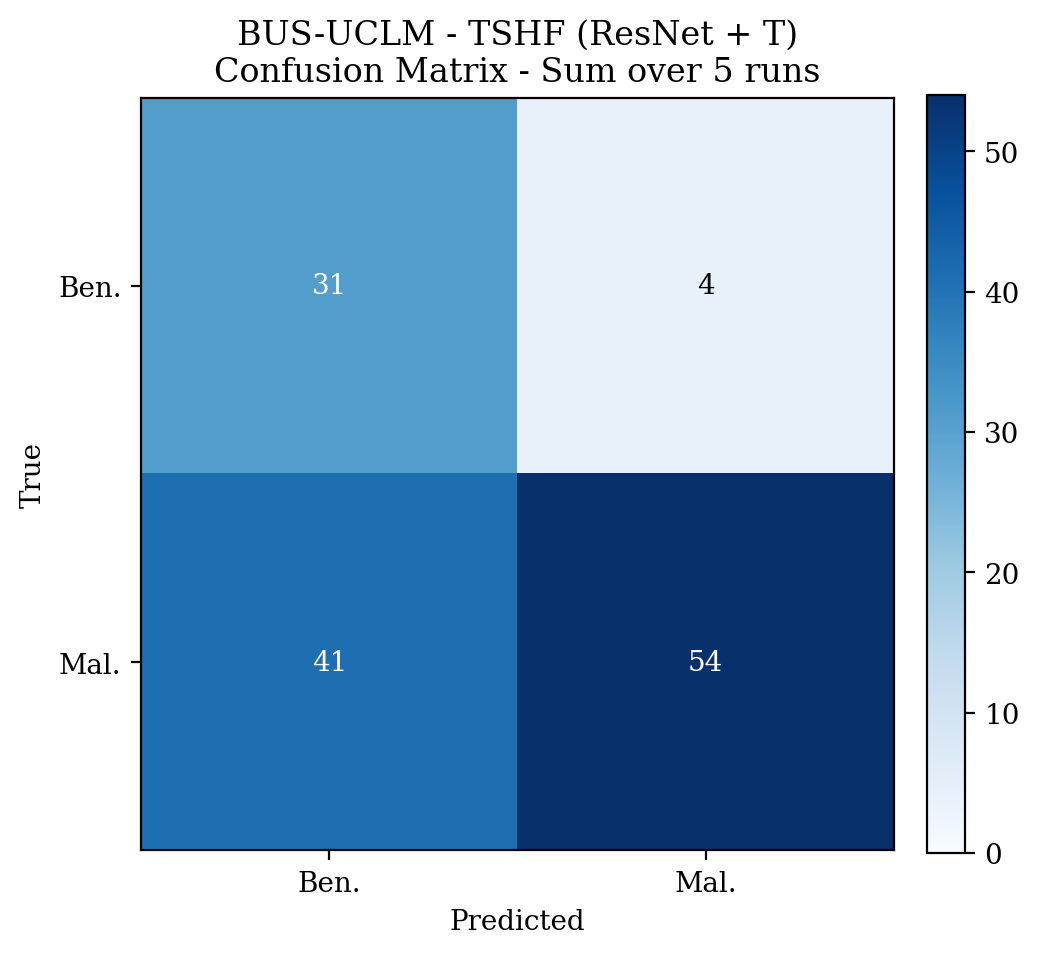}}

    \vspace{0.4em}

    \subfloat[SCNN + non-trainable]{%
        \includegraphics[width=0.47\linewidth]{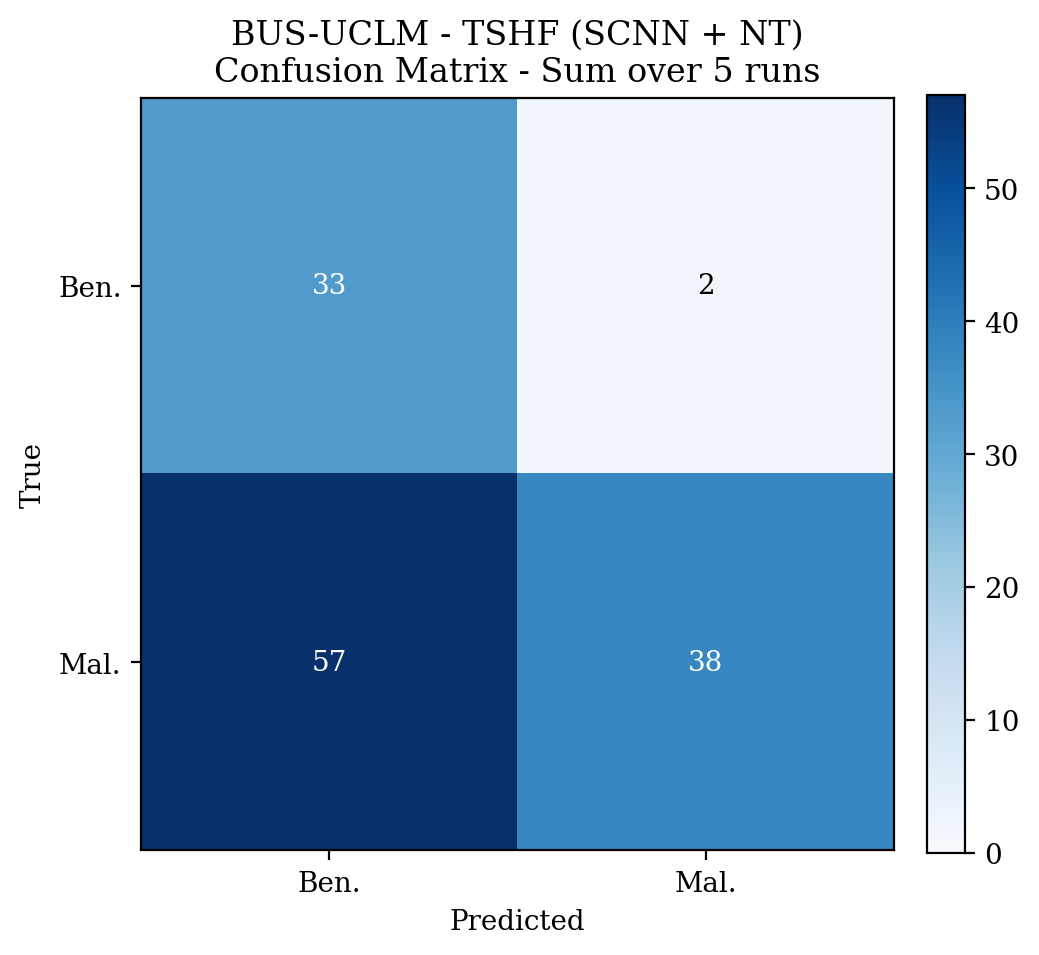}}
    \hfill
    \subfloat[SCNN + trainable]{%
        \includegraphics[width=0.47\linewidth]{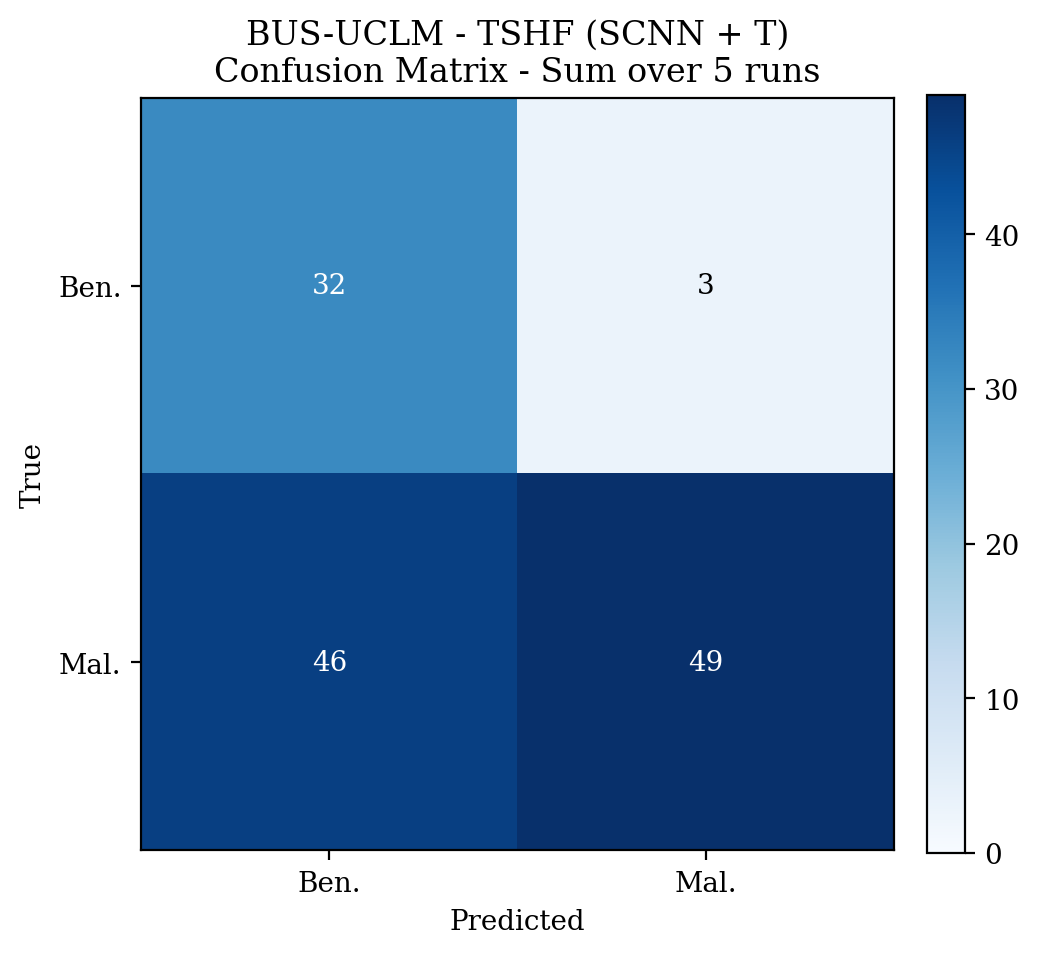}}

    \caption{Confusion matrices of the \textbf{TSHF} configurations on the
    BUS-UCLM test set, summed over five independent runs.}
    \label{fig:cm_busuclm_tshf}
\end{figure}

\clearpage

\section{Confidence intervals}
\label{app:ci}

\begin{table*}[htbp]
\centering
\caption{Test-set performance of the standalone classical and quantum
architectures, reported as mean with the 95\% confidence interval of the mean in brackets. All values in \%. The best
result per dataset and paradigm is highlighted in bold.}
\label{tab:standalone_baselinesb}

\footnotesize
\setlength{\tabcolsep}{4pt}
\renewcommand{\arraystretch}{1.2}
\begin{adjustbox}{max width=\linewidth}
\begin{tabular}{lll ccccc}
\toprule
\textbf{Dataset} & \textbf{Paradigm} & \textbf{Architecture} & \textbf{Acc.} & \textbf{Prec.} & \textbf{Rec.} & \textbf{F1} & \textbf{AUC} \\
\midrule
\multirow{4}{*}{BreastMNIST}
 & \multirow{2}{*}{Classical}
 & SCNN      & \makecell{82.82\\[-1pt]{\scriptsize [79.47, 86.17]}} & \makecell{86.27\\[-1pt]{\scriptsize [83.46, 89.08]}} & \makecell{\textbf{91.05}\\[-1pt]{\scriptsize [86.34, 95.76]}} & \makecell{88.55\\[-1pt]{\scriptsize [86.15, 90.95]}} & \makecell{84.49\\[-1pt]{\scriptsize [81.72, 87.26]}} \\
 & & ResNet-18 & \makecell{\textbf{84.74}\\[-1pt]{\scriptsize [81.03, 88.45]}} & \makecell{\textbf{88.67}\\[-1pt]{\scriptsize [86.57, 90.77]}} & \makecell{90.70\\[-1pt]{\scriptsize [87.20, 94.20]}} & \makecell{\textbf{89.67}\\[-1pt]{\scriptsize [87.06, 92.28]}} & \makecell{\textbf{87.08}\\[-1pt]{\scriptsize [83.27, 90.89]}} \\
\cmidrule{2-8}
 & \multirow{2}{*}{Quantum}
 & Non-trainable & \makecell{\textbf{78.21}\\[-1pt]{\scriptsize [75.75, 80.67]}} & \makecell{\textbf{82.69}\\[-1pt]{\scriptsize [76.73, 88.65]}} & \makecell{89.47\\[-1pt]{\scriptsize [83.36, 95.58]}} & \makecell{\textbf{85.73}\\[-1pt]{\scriptsize [85.06, 86.40]}} & \makecell{\textbf{79.42}\\[-1pt]{\scriptsize [75.53, 83.31]}} \\
 & & Trainable     & \makecell{77.82\\[-1pt]{\scriptsize [75.26, 80.38]}} & \makecell{81.18\\[-1pt]{\scriptsize [77.05, 85.31]}} & \makecell{\textbf{91.05}\\[-1pt]{\scriptsize [85.52, 96.58]}} & \makecell{85.71\\[-1pt]{\scriptsize [84.15, 87.27]}} & \makecell{78.32\\[-1pt]{\scriptsize [76.25, 80.39]}} \\
\midrule
\multirow{4}{*}{INbreast}
 & \multirow{2}{*}{Classical}
 & SCNN      & \makecell{67.92\\[-1pt]{\scriptsize [64.96, 70.88]}} & \makecell{65.05\\[-1pt]{\scriptsize [59.88, 70.22]}} & \makecell{\textbf{73.04}\\[-1pt]{\scriptsize [65.75, 80.33]}} & \makecell{\textbf{68.56}\\[-1pt]{\scriptsize [67.16, 69.96]}} & \makecell{\textbf{78.78}\\[-1pt]{\scriptsize [76.78, 80.78]}} \\
 & & ResNet-18 & \makecell{\textbf{68.33}\\[-1pt]{\scriptsize [64.95, 71.71]}} & \makecell{\textbf{72.91}\\[-1pt]{\scriptsize [55.69, 90.13]}} & \makecell{66.09\\[-1pt]{\scriptsize [36.24, 95.94]}} & \makecell{65.15\\[-1pt]{\scriptsize [55.30, 75.00]}} & \makecell{77.30\\[-1pt]{\scriptsize [71.66, 82.94]}} \\
\cmidrule{2-8}
 & \multirow{2}{*}{Quantum}
 & Non-trainable & \makecell{69.58\\[-1pt]{\scriptsize [66.49, 72.67]}} & \makecell{65.78\\[-1pt]{\scriptsize [62.42, 69.14]}} & \makecell{\textbf{76.52}\\[-1pt]{\scriptsize [69.28, 83.76]}} & \makecell{\textbf{70.63}\\[-1pt]{\scriptsize [67.09, 74.17]}} & \makecell{\textbf{78.63}\\[-1pt]{\scriptsize [76.99, 80.27]}} \\
 & & Trainable     & \makecell{\textbf{70.00}\\[-1pt]{\scriptsize [66.91, 73.09]}} & \makecell{\textbf{70.00}\\[-1pt]{\scriptsize [61.63, 78.37]}} & \makecell{68.26\\[-1pt]{\scriptsize [56.40, 80.12]}} & \makecell{68.44\\[-1pt]{\scriptsize [65.99, 70.89]}} & \makecell{77.15\\[-1pt]{\scriptsize [73.60, 80.70]}} \\
\midrule
\multirow{4}{*}{BUS-UCLM}
 & \multirow{2}{*}{Classical}
 & SCNN      & \makecell{\textbf{66.92}\\[-1pt]{\scriptsize [55.42, 78.42]}} & \makecell{93.20\\[-1pt]{\scriptsize [88.00, 98.40]}} & \makecell{\textbf{58.95}\\[-1pt]{\scriptsize [43.34, 74.56]}} & \makecell{\textbf{71.66}\\[-1pt]{\scriptsize [59.18, 84.14]}} & \makecell{\textbf{91.88}\\[-1pt]{\scriptsize [87.43, 96.33]}} \\
 & & ResNet-18 & \makecell{65.38\\[-1pt]{\scriptsize [54.18, 76.58]}} & \makecell{\textbf{95.14}\\[-1pt]{\scriptsize [89.45, 100.00]}} & \makecell{55.79\\[-1pt]{\scriptsize [38.75, 72.83]}} & \makecell{69.36\\[-1pt]{\scriptsize [55.65, 83.07]}} & \makecell{86.62\\[-1pt]{\scriptsize [80.08, 93.16]}} \\
\cmidrule{2-8}
 & \multirow{2}{*}{Quantum}
 & Non-trainable & \makecell{69.23\\[-1pt]{\scriptsize [58.03, 80.43]}} & \makecell{\textbf{97.24}\\[-1pt]{\scriptsize [92.53, 100.00]}} & \makecell{60.00\\[-1pt]{\scriptsize [42.34, 77.66]}} & \makecell{73.15\\[-1pt]{\scriptsize [59.17, 87.13]}} & \makecell{\textbf{86.47}\\[-1pt]{\scriptsize [81.58, 91.36]}} \\
 & & Trainable     & \makecell{\textbf{70.77}\\[-1pt]{\scriptsize [51.49, 90.05]}} & \makecell{86.67\\[-1pt]{\scriptsize [75.73, 97.61]}} & \makecell{\textbf{72.63}\\[-1pt]{\scriptsize [41.15, 100.00]}} & \makecell{\textbf{76.64}\\[-1pt]{\scriptsize [57.52, 95.76]}} & \makecell{81.80\\[-1pt]{\scriptsize [66.89, 96.71]}} \\
\bottomrule
\end{tabular}
\end{adjustbox}
\end{table*}

\begin{table*}[htbp]
\centering
\footnotesize
\setlength{\tabcolsep}{4pt}
\renewcommand{\arraystretch}{1.1}
\begin{adjustbox}{max width=\linewidth}
\begin{tabular}{lll ccccc c}
\toprule
\textbf{Fusion} & \textbf{Classical} & \textbf{Quantum} & \textbf{Acc.} & \textbf{Prec.} & \textbf{Rec.} & \textbf{F1} & \textbf{AUC} & \boldmath$\gamma$ \\
\midrule
\multirow{4}{*}{Baseline}
 & ResNet-18             & \multirow{2}{*}{N/A} & \makecell{84.74\\[-1pt]{\scriptsize [81.03, 88.45]}} & \makecell{88.67\\[-1pt]{\scriptsize [86.57, 90.77]}} & \makecell{90.70\\[-1pt]{\scriptsize [87.20, 94.20]}} & \makecell{89.67\\[-1pt]{\scriptsize [87.06, 92.28]}} & \makecell{87.08\\[-1pt]{\scriptsize [83.27, 90.89]}} & N/A \\
 & SCNN                  &                      & \makecell{82.82\\[-1pt]{\scriptsize [79.47, 86.17]}} & \makecell{86.27\\[-1pt]{\scriptsize [83.46, 89.08]}} & \makecell{91.05\\[-1pt]{\scriptsize [86.34, 95.76]}} & \makecell{88.55\\[-1pt]{\scriptsize [86.15, 90.95]}} & \makecell{84.49\\[-1pt]{\scriptsize [81.72, 87.26]}} & N/A \\
\cmidrule{2-9}
 & \multirow{2}{*}{N/A}  & Trainable            & \makecell{77.82\\[-1pt]{\scriptsize [75.26, 80.38]}} & \makecell{81.18\\[-1pt]{\scriptsize [77.05, 85.31]}} & \makecell{91.05\\[-1pt]{\scriptsize [85.52, 96.58]}} & \makecell{85.71\\[-1pt]{\scriptsize [84.15, 87.27]}} & \makecell{78.32\\[-1pt]{\scriptsize [76.25, 80.39]}} & N/A \\
 &                       & Non-trainable        & \makecell{78.21\\[-1pt]{\scriptsize [75.75, 80.67]}} & \makecell{82.69\\[-1pt]{\scriptsize [76.73, 88.65]}} & \makecell{89.47\\[-1pt]{\scriptsize [83.36, 95.58]}} & \makecell{85.73\\[-1pt]{\scriptsize [85.06, 86.40]}} & \makecell{79.42\\[-1pt]{\scriptsize [75.53, 83.31]}} & N/A \\
\midrule
\multirow{4}{*}{SHF}
 & \multirow{2}{*}{ResNet-18} & Trainable       & \makecell{84.74\\[-1pt]{\scriptsize [82.13, 87.35]}} & \makecell{\textbf{94.85}\\[-1pt]{\scriptsize [93.78, 95.92]}} & \makecell{83.68\\[-1pt]{\scriptsize [79.35, 88.01]}} & \makecell{88.88\\[-1pt]{\scriptsize [86.73, 91.03]}} & \makecell{92.12\\[-1pt]{\scriptsize [91.90, 92.34]}} & N/A \\
 &                            & Non-trainable   & \makecell{78.46\\[-1pt]{\scriptsize [74.87, 82.05]}} & \makecell{83.55\\[-1pt]{\scriptsize [78.19, 88.91]}} & \makecell{88.60\\[-1pt]{\scriptsize [77.49, 99.71]}} & \makecell{85.61\\[-1pt]{\scriptsize [82.26, 88.96]}} & \makecell{81.14\\[-1pt]{\scriptsize [79.60, 82.68]}} & N/A \\
\cmidrule{2-9}
 & \multirow{2}{*}{SCNN}      & Trainable       & \makecell{80.77\\[-1pt]{\scriptsize [79.18, 82.36]}} & \makecell{85.53\\[-1pt]{\scriptsize [83.51, 87.55]}} & \makecell{88.77\\[-1pt]{\scriptsize [85.12, 92.42]}} & \makecell{87.08\\[-1pt]{\scriptsize [85.83, 88.33]}} & \makecell{83.56\\[-1pt]{\scriptsize [82.08, 85.04]}} & N/A \\
 &                            & Non-trainable   & \makecell{80.77\\[-1pt]{\scriptsize [78.91, 82.63]}} & \makecell{86.37\\[-1pt]{\scriptsize [84.43, 88.31]}} & \makecell{87.54\\[-1pt]{\scriptsize [84.15, 90.93]}} & \makecell{86.92\\[-1pt]{\scriptsize [85.50, 88.34]}} & \makecell{83.07\\[-1pt]{\scriptsize [82.25, 83.89]}} & N/A \\
\midrule
\multirow{4}{*}{DHF}
 & \multirow{2}{*}{ResNet-18} & Trainable       & \makecell{84.87\\[-1pt]{\scriptsize [82.65, 87.09]}} & \makecell{92.96\\[-1pt]{\scriptsize [88.54, 97.38]}} & \makecell{86.14\\[-1pt]{\scriptsize [78.31, 93.97]}} & \makecell{89.21\\[-1pt]{\scriptsize [87.06, 91.36]}} & \makecell{\textbf{92.48}\\[-1pt]{\scriptsize [91.83, 93.13]}} & N/A \\
 &                            & Non-trainable   & \makecell{86.28\\[-1pt]{\scriptsize [83.55, 89.01]}} & \makecell{93.83\\[-1pt]{\scriptsize [92.09, 95.57]}} & \makecell{87.02\\[-1pt]{\scriptsize [81.61, 92.43]}} & \makecell{90.22\\[-1pt]{\scriptsize [87.94, 92.50]}} & \makecell{92.39\\[-1pt]{\scriptsize [92.09, 92.69]}} & N/A \\
\cmidrule{2-9}
 & \multirow{2}{*}{SCNN}      & Trainable       & \makecell{80.00\\[-1pt]{\scriptsize [75.63, 84.37]}} & \makecell{84.55\\[-1pt]{\scriptsize [81.17, 87.93]}} & \makecell{89.30\\[-1pt]{\scriptsize [77.79, 100.00]}} & \makecell{86.52\\[-1pt]{\scriptsize [82.12, 90.92]}} & \makecell{82.45\\[-1pt]{\scriptsize [81.77, 83.13]}} & N/A \\
 &                            & Non-trainable   & \makecell{81.54\\[-1pt]{\scriptsize [79.81, 83.27]}} & \makecell{85.88\\[-1pt]{\scriptsize [84.34, 87.42]}} & \makecell{89.47\\[-1pt]{\scriptsize [86.59, 92.35]}} & \makecell{87.62\\[-1pt]{\scriptsize [86.34, 88.90]}} & \makecell{83.46\\[-1pt]{\scriptsize [82.39, 84.53]}} & N/A \\
\midrule
\multirow{4}{*}{TSHF}
 & \multirow{2}{*}{ResNet-18} & Trainable       & \makecell{86.15\\[-1pt]{\scriptsize [84.31, 87.99]}} & \makecell{94.12\\[-1pt]{\scriptsize [92.89, 95.35]}} & \makecell{86.49\\[-1pt]{\scriptsize [82.75, 90.23]}} & \makecell{90.11\\[-1pt]{\scriptsize [88.55, 91.67]}} & \makecell{92.18\\[-1pt]{\scriptsize [91.91, 92.45]}} & \makecell{1.280\\[-1pt]{\scriptsize [0.869, 1.691]}} \\
 &                            & Non-trainable   & \makecell{\textbf{88.33}\\[-1pt]{\scriptsize [85.85, 90.81]}} & \makecell{91.56\\[-1pt]{\scriptsize [86.98, 96.14]}} & \makecell{92.81\\[-1pt]{\scriptsize [88.85, 96.77]}} & \makecell{\textbf{92.09}\\[-1pt]{\scriptsize [90.54, 93.64]}} & \makecell{92.09\\[-1pt]{\scriptsize [91.54, 92.64]}} & \makecell{1.025\\[-1pt]{\scriptsize [0.977, 1.073]}} \\
\cmidrule{2-9}
 & \multirow{2}{*}{SCNN}      & Trainable       & \makecell{81.54\\[-1pt]{\scriptsize [80.50, 82.58]}} & \makecell{83.30\\[-1pt]{\scriptsize [80.46, 86.14]}} & \makecell{\textbf{93.68}\\[-1pt]{\scriptsize [88.66, 98.70]}} & \makecell{88.10\\[-1pt]{\scriptsize [87.11, 89.09]}} & \makecell{83.71\\[-1pt]{\scriptsize [82.92, 84.50]}} & \makecell{0.930\\[-1pt]{\scriptsize [0.864, 0.996]}} \\
 &                            & Non-trainable   & \makecell{81.92\\[-1pt]{\scriptsize [80.74, 83.10]}} & \makecell{84.15\\[-1pt]{\scriptsize [82.52, 85.78]}} & \makecell{92.81\\[-1pt]{\scriptsize [89.16, 96.46]}} & \makecell{88.23\\[-1pt]{\scriptsize [87.22, 89.24]}} & \makecell{83.29\\[-1pt]{\scriptsize [82.25, 84.33]}} & \makecell{1.019\\[-1pt]{\scriptsize [0.948, 1.090]}} \\
\bottomrule
\end{tabular}
\end{adjustbox}
\caption{Test-set performance of the fusion strategies on \textbf{BreastMNIST},
reported as mean with the 95\% confidence interval of the mean in brackets, clipped to the $[0,100]$ range of the metrics. Metric values are given in \%, whereas
$\gamma$ is the learned fusion scaling factor of TSHF (dimensionless). The best
result per metric is highlighted in bold.}
\label{tab:metrics_breastmnistb}
\end{table*}

\begin{table*}[htbp]
\centering
\caption{Test-set performance of the fusion strategies on \textbf{INbreast},
reported as mean with the 95\% confidence interval of the mean in brackets, clipped to the $[0,100]$ range of the metrics. Metric values are given in \%, whereas
$\gamma$ is the learned fusion scaling factor of TSHF (dimensionless). The best
result per metric is highlighted in bold.}
\label{tab:metrics_inbreastb}
\footnotesize
\setlength{\tabcolsep}{4pt}
\renewcommand{\arraystretch}{1.1}
\begin{adjustbox}{max width=\linewidth}
\begin{tabular}{lll ccccc c}
\toprule
\textbf{Fusion} & \textbf{Classical} & \textbf{Quantum} & \textbf{Acc.} & \textbf{Prec.} & \textbf{Rec.} & \textbf{F1} & \textbf{AUC} & \boldmath$\gamma$ \\
\midrule
\multirow{4}{*}{Baseline}
 & ResNet-18             & \multirow{2}{*}{N/A} & \makecell{68.33\\[-1pt]{\scriptsize [64.95, 71.71]}} & \makecell{\textbf{72.91}\\[-1pt]{\scriptsize [55.69, 90.13]}} & \makecell{66.09\\[-1pt]{\scriptsize [36.24, 95.94]}} & \makecell{65.15\\[-1pt]{\scriptsize [55.30, 75.00]}} & \makecell{77.30\\[-1pt]{\scriptsize [71.66, 82.94]}} & N/A \\
 & SCNN                  &                      & \makecell{67.92\\[-1pt]{\scriptsize [64.96, 70.88]}} & \makecell{65.05\\[-1pt]{\scriptsize [59.88, 70.22]}} & \makecell{73.04\\[-1pt]{\scriptsize [65.75, 80.33]}} & \makecell{68.56\\[-1pt]{\scriptsize [67.16, 69.96]}} & \makecell{78.78\\[-1pt]{\scriptsize [76.78, 80.78]}} & N/A \\
\cmidrule{2-9}
 & \multirow{2}{*}{N/A}  & Trainable            & \makecell{70.00\\[-1pt]{\scriptsize [66.91, 73.09]}} & \makecell{70.00\\[-1pt]{\scriptsize [61.63, 78.37]}} & \makecell{68.26\\[-1pt]{\scriptsize [56.40, 80.12]}} & \makecell{68.44\\[-1pt]{\scriptsize [65.99, 70.89]}} & \makecell{77.15\\[-1pt]{\scriptsize [73.60, 80.70]}} & N/A \\
 &                       & Non-trainable        & \makecell{69.58\\[-1pt]{\scriptsize [66.49, 72.67]}} & \makecell{65.78\\[-1pt]{\scriptsize [62.42, 69.14]}} & \makecell{76.52\\[-1pt]{\scriptsize [69.28, 83.76]}} & \makecell{70.63\\[-1pt]{\scriptsize [67.09, 74.17]}} & \makecell{78.63\\[-1pt]{\scriptsize [76.99, 80.27]}} & N/A \\
\midrule
\multirow{4}{*}{SHF}
 & \multirow{2}{*}{ResNet-18} & Trainable       & \makecell{66.87\\[-1pt]{\scriptsize [61.51, 72.23]}} & \makecell{64.83\\[-1pt]{\scriptsize [53.82, 75.84]}} & \makecell{77.39\\[-1pt]{\scriptsize [55.83, 98.95]}} & \makecell{68.73\\[-1pt]{\scriptsize [64.30, 73.16]}} & \makecell{79.62\\[-1pt]{\scriptsize [79.51, 79.73]}} & N/A \\
 &                            & Non-trainable   & \makecell{\textbf{70.42}\\[-1pt]{\scriptsize [65.79, 75.05]}} & \makecell{65.74\\[-1pt]{\scriptsize [58.39, 73.09]}} & \makecell{83.04\\[-1pt]{\scriptsize [74.21, 91.87]}} & \makecell{\textbf{72.96}\\[-1pt]{\scriptsize [71.36, 74.56]}} & \makecell{\textbf{80.28}\\[-1pt]{\scriptsize [79.71, 80.85]}} & N/A \\
\cmidrule{2-9}
 & \multirow{2}{*}{SCNN}      & Trainable       & \makecell{68.33\\[-1pt]{\scriptsize [67.18, 69.48]}} & \makecell{68.16\\[-1pt]{\scriptsize [61.18, 75.14]}} & \makecell{66.96\\[-1pt]{\scriptsize [50.72, 83.20]}} & \makecell{66.55\\[-1pt]{\scriptsize [61.88, 71.22]}} & \makecell{76.83\\[-1pt]{\scriptsize [74.10, 79.56]}} & N/A \\
 &                            & Non-trainable   & \makecell{68.75\\[-1pt]{\scriptsize [66.92, 70.58]}} & \makecell{69.57\\[-1pt]{\scriptsize [63.59, 75.55]}} & \makecell{63.48\\[-1pt]{\scriptsize [51.81, 75.15]}} & \makecell{65.81\\[-1pt]{\scriptsize [61.30, 70.32]}} & \makecell{77.70\\[-1pt]{\scriptsize [76.48, 78.92]}} & N/A \\
\midrule
\multirow{4}{*}{DHF}
 & \multirow{2}{*}{ResNet-18} & Trainable       & \makecell{67.08\\[-1pt]{\scriptsize [64.40, 69.76]}} & \makecell{62.18\\[-1pt]{\scriptsize [60.13, 64.23]}} & \makecell{80.00\\[-1pt]{\scriptsize [73.28, 86.72]}} & \makecell{69.91\\[-1pt]{\scriptsize [66.73, 73.09]}} & \makecell{74.67\\[-1pt]{\scriptsize [73.11, 76.23]}} & N/A \\
 &                            & Non-trainable   & \makecell{69.79\\[-1pt]{\scriptsize [67.37, 72.21]}} & \makecell{66.51\\[-1pt]{\scriptsize [62.85, 70.17]}} & \makecell{75.65\\[-1pt]{\scriptsize [61.06, 90.24]}} & \makecell{70.30\\[-1pt]{\scriptsize [65.22, 75.38]}} & \makecell{76.25\\[-1pt]{\scriptsize [73.72, 78.78]}} & N/A \\
\cmidrule{2-9}
 & \multirow{2}{*}{SCNN}      & Trainable       & \makecell{68.33\\[-1pt]{\scriptsize [65.21, 71.45]}} & \makecell{68.47\\[-1pt]{\scriptsize [62.44, 74.50]}} & \makecell{65.22\\[-1pt]{\scriptsize [46.72, 83.72]}} & \makecell{65.76\\[-1pt]{\scriptsize [58.32, 73.20]}} & \makecell{76.55\\[-1pt]{\scriptsize [73.47, 79.63]}} & N/A \\
 &                            & Non-trainable   & \makecell{68.54\\[-1pt]{\scriptsize [65.06, 72.02]}} & \makecell{69.98\\[-1pt]{\scriptsize [62.89, 77.07]}} & \makecell{61.74\\[-1pt]{\scriptsize [49.44, 74.04]}} & \makecell{64.99\\[-1pt]{\scriptsize [59.04, 70.94]}} & \makecell{77.12\\[-1pt]{\scriptsize [75.26, 78.98]}} & N/A \\
\midrule
\multirow{4}{*}{TSHF}
 & \multirow{2}{*}{ResNet-18} & Trainable       & \makecell{67.92\\[-1pt]{\scriptsize [66.00, 69.84]}} & \makecell{64.35\\[-1pt]{\scriptsize [59.86, 68.84]}} & \makecell{76.52\\[-1pt]{\scriptsize [62.32, 90.72]}} & \makecell{69.32\\[-1pt]{\scriptsize [65.45, 73.19]}} & \makecell{76.53\\[-1pt]{\scriptsize [74.98, 78.08]}} & \makecell{1.080\\[-1pt]{\scriptsize [0.870, 1.290]}} \\
 &                            & Non-trainable   & \makecell{65.42\\[-1pt]{\scriptsize [60.81, 70.03]}} & \makecell{59.79\\[-1pt]{\scriptsize [55.63, 63.95]}} & \makecell{\textbf{86.96}\\[-1pt]{\scriptsize [81.24, 92.68]}} & \makecell{70.72\\[-1pt]{\scriptsize [68.60, 72.84]}} & \makecell{78.12\\[-1pt]{\scriptsize [76.61, 79.63]}} & \makecell{1.095\\[-1pt]{\scriptsize [0.909, 1.281]}} \\
\cmidrule{2-9}
 & \multirow{2}{*}{SCNN}      & Trainable       & \makecell{67.71\\[-1pt]{\scriptsize [59.69, 75.73]}} & \makecell{66.53\\[-1pt]{\scriptsize [56.47, 76.59]}} & \makecell{70.43\\[-1pt]{\scriptsize [61.76, 79.10]}} & \makecell{67.80\\[-1pt]{\scriptsize [63.99, 71.61]}} & \makecell{75.26\\[-1pt]{\scriptsize [72.11, 78.41]}} & \makecell{0.977\\[-1pt]{\scriptsize [0.911, 1.043]}} \\
 &                            & Non-trainable   & \makecell{68.13\\[-1pt]{\scriptsize [63.78, 72.48]}} & \makecell{64.65\\[-1pt]{\scriptsize [59.46, 69.84]}} & \makecell{75.22\\[-1pt]{\scriptsize [68.71, 81.73]}} & \makecell{69.36\\[-1pt]{\scriptsize [66.31, 72.41]}} & \makecell{75.70\\[-1pt]{\scriptsize [72.27, 79.13]}} & \makecell{1.045\\[-1pt]{\scriptsize [0.941, 1.149]}} \\
\bottomrule
\end{tabular}
\end{adjustbox}

\end{table*}

\begin{table*}[htbp]
\centering
\caption{Test-set performance of the fusion strategies on \textbf{BUS-UCLM},
reported as mean with the 95\% confidence interval of the mean in brackets, clipped to the $[0,100]$ range of the metrics. Metric values are given in \%, whereas
$\gamma$ is the learned fusion scaling factor of TSHF (dimensionless). The best
result per metric is highlighted in bold.}
\label{tab:metrics_busuclmb}
\footnotesize
\setlength{\tabcolsep}{4pt}
\renewcommand{\arraystretch}{1.1}
\begin{adjustbox}{max width=\linewidth}
\begin{tabular}{lll ccccc c}
\toprule
\textbf{Fusion} & \textbf{Classical} & \textbf{Quantum} & \textbf{Acc.} & \textbf{Prec.} & \textbf{Rec.} & \textbf{F1} & \textbf{AUC} & \boldmath$\gamma$ \\
\midrule
\multirow{4}{*}{Baseline}
 & ResNet-18             & \multirow{2}{*}{N/A} & \makecell{65.38\\[-1pt]{\scriptsize [54.18, 76.58]}} & \makecell{95.14\\[-1pt]{\scriptsize [89.45, 100.00]}} & \makecell{55.79\\[-1pt]{\scriptsize [38.75, 72.83]}} & \makecell{69.36\\[-1pt]{\scriptsize [55.65, 83.07]}} & \makecell{86.62\\[-1pt]{\scriptsize [80.08, 93.16]}} & N/A \\
 & SCNN                  &                      & \makecell{66.92\\[-1pt]{\scriptsize [55.42, 78.42]}} & \makecell{93.20\\[-1pt]{\scriptsize [88.00, 98.40]}} & \makecell{58.95\\[-1pt]{\scriptsize [43.34, 74.56]}} & \makecell{71.66\\[-1pt]{\scriptsize [59.18, 84.14]}} & \makecell{\textbf{91.88}\\[-1pt]{\scriptsize [87.43, 96.33]}} & N/A \\
\cmidrule{2-9}
 & \multirow{2}{*}{N/A}  & Trainable            & \makecell{70.77\\[-1pt]{\scriptsize [51.49, 90.05]}} & \makecell{86.67\\[-1pt]{\scriptsize [75.73, 97.61]}} & \makecell{72.63\\[-1pt]{\scriptsize [41.15, 100.00]}} & \makecell{76.64\\[-1pt]{\scriptsize [57.52, 95.76]}} & \makecell{81.80\\[-1pt]{\scriptsize [66.89, 96.71]}} & N/A \\
 &                       & Non-trainable        & \makecell{69.23\\[-1pt]{\scriptsize [58.03, 80.43]}} & \makecell{97.24\\[-1pt]{\scriptsize [92.53, 100.00]}} & \makecell{60.00\\[-1pt]{\scriptsize [42.34, 77.66]}} & \makecell{73.15\\[-1pt]{\scriptsize [59.17, 87.13]}} & \makecell{86.47\\[-1pt]{\scriptsize [81.58, 91.36]}} & N/A \\
\midrule
\multirow{4}{*}{SHF}
 & \multirow{2}{*}{ResNet-18} & Trainable       & \makecell{\textbf{78.46}\\[-1pt]{\scriptsize [67.47, 89.45]}} & \makecell{89.32\\[-1pt]{\scriptsize [77.25, 100.00]}} & \makecell{\textbf{82.11}\\[-1pt]{\scriptsize [67.06, 97.16]}} & \makecell{\textbf{84.66}\\[-1pt]{\scriptsize [76.50, 92.82]}} & \makecell{89.47\\[-1pt]{\scriptsize [85.40, 93.54]}} & N/A \\
 &                            & Non-trainable   & \makecell{73.08\\[-1pt]{\scriptsize [63.53, 82.63]}} & \makecell{94.17\\[-1pt]{\scriptsize [89.92, 98.42]}} & \makecell{67.37\\[-1pt]{\scriptsize [53.97, 80.77]}} & \makecell{78.16\\[-1pt]{\scriptsize [69.00, 87.32]}} & \makecell{90.38\\[-1pt]{\scriptsize [87.78, 92.98]}} & N/A \\
\cmidrule{2-9}
 & \multirow{2}{*}{SCNN}      & Trainable       & \makecell{62.31\\[-1pt]{\scriptsize [53.77, 70.85]}} & \makecell{98.00\\[-1pt]{\scriptsize [92.45, 100.00]}} & \makecell{49.47\\[-1pt]{\scriptsize [38.53, 60.41]}} & \makecell{65.39\\[-1pt]{\scriptsize [55.69, 75.09]}} & \makecell{87.52\\[-1pt]{\scriptsize [84.24, 90.80]}} & N/A \\
 &                            & Non-trainable   & \makecell{56.15\\[-1pt]{\scriptsize [44.17, 68.13]}} & \makecell{\textbf{98.33}\\[-1pt]{\scriptsize [93.70, 100.00]}} & \makecell{41.05\\[-1pt]{\scriptsize [22.92, 59.18]}} & \makecell{56.46\\[-1pt]{\scriptsize [38.10, 74.82]}} & \makecell{90.23\\[-1pt]{\scriptsize [85.08, 95.38]}} & N/A \\
\midrule
\multirow{4}{*}{DHF}
 & \multirow{2}{*}{ResNet-18} & Trainable       & \makecell{58.46\\[-1pt]{\scriptsize [49.27, 67.65]}} & \makecell{91.92\\[-1pt]{\scriptsize [85.86, 97.98]}} & \makecell{47.37\\[-1pt]{\scriptsize [34.30, 60.44]}} & \makecell{61.98\\[-1pt]{\scriptsize [50.79, 73.17]}} & \makecell{82.56\\[-1pt]{\scriptsize [74.58, 90.54]}} & N/A \\
 &                            & Non-trainable   & \makecell{69.23\\[-1pt]{\scriptsize [58.03, 80.43]}} & \makecell{92.05\\[-1pt]{\scriptsize [89.82, 94.28]}} & \makecell{63.16\\[-1pt]{\scriptsize [47.84, 78.48]}} & \makecell{74.41\\[-1pt]{\scriptsize [61.99, 86.83]}} & \makecell{89.17\\[-1pt]{\scriptsize [84.96, 93.38]}} & N/A \\
\cmidrule{2-9}
 & \multirow{2}{*}{SCNN}      & Trainable       & \makecell{59.23\\[-1pt]{\scriptsize [53.78, 64.68]}} & \makecell{98.00\\[-1pt]{\scriptsize [92.45, 100.00]}} & \makecell{45.26\\[-1pt]{\scriptsize [37.81, 52.71]}} & \makecell{61.69\\[-1pt]{\scriptsize [54.69, 68.69]}} & \makecell{89.77\\[-1pt]{\scriptsize [86.78, 92.76]}} & N/A \\
 &                            & Non-trainable   & \makecell{55.38\\[-1pt]{\scriptsize [48.14, 62.62]}} & \makecell{\textbf{98.33}\\[-1pt]{\scriptsize [93.70, 100.00]}} & \makecell{40.00\\[-1pt]{\scriptsize [27.26, 52.74]}} & \makecell{56.10\\[-1pt]{\scriptsize [45.31, 66.89]}} & \makecell{88.57\\[-1pt]{\scriptsize [84.03, 93.11]}} & N/A \\
\midrule
\multirow{4}{*}{TSHF}
 & \multirow{2}{*}{ResNet-18} & Trainable       & \makecell{65.38\\[-1pt]{\scriptsize [48.49, 82.27]}} & \makecell{92.93\\[-1pt]{\scriptsize [86.62, 99.24]}} & \makecell{56.84\\[-1pt]{\scriptsize [32.65, 81.03]}} & \makecell{68.98\\[-1pt]{\scriptsize [49.08, 88.88]}} & \makecell{83.16\\[-1pt]{\scriptsize [66.47, 99.85]}} & \makecell{0.989\\[-1pt]{\scriptsize [0.933, 1.045]}} \\
 &                            & Non-trainable   & \makecell{53.08\\[-1pt]{\scriptsize [35.40, 70.76]}} & \makecell{95.83\\[-1pt]{\scriptsize [88.52, 100.00]}} & \makecell{37.89\\[-1pt]{\scriptsize [12.41, 63.37]}} & \makecell{51.13\\[-1pt]{\scriptsize [20.65, 81.61]}} & \makecell{81.95\\[-1pt]{\scriptsize [68.18, 95.72]}} & \makecell{1.005\\[-1pt]{\scriptsize [0.975, 1.035]}} \\
\cmidrule{2-9}
 & \multirow{2}{*}{SCNN}      & Trainable       & \makecell{62.31\\[-1pt]{\scriptsize [43.39, 81.23]}} & \makecell{94.53\\[-1pt]{\scriptsize [88.09, 100.00]}} & \makecell{51.58\\[-1pt]{\scriptsize [25.28, 77.88]}} & \makecell{64.45\\[-1pt]{\scriptsize [39.93, 88.97]}} & \makecell{88.57\\[-1pt]{\scriptsize [83.44, 93.70]}} & \makecell{0.959\\[-1pt]{\scriptsize [0.882, 1.036]}} \\
 &                            & Non-trainable   & \makecell{54.62\\[-1pt]{\scriptsize [29.49, 79.75]}} & \makecell{76.67\\[-1pt]{\scriptsize [23.18, 100.00]}} & \makecell{40.00\\[-1pt]{\scriptsize [3.43, 76.57]}} & \makecell{50.51\\[-1pt]{\scriptsize [9.01, 92.01]}} & \makecell{87.67\\[-1pt]{\scriptsize [81.86, 93.48]}} & \makecell{1.025\\[-1pt]{\scriptsize [0.969, 1.081]}} \\
\bottomrule
\end{tabular}
\end{adjustbox}

\end{table*}

\begin{table*}[htbp]
\centering
\caption{Ablation on the temperature scalar $\gamma$, evaluated on the
\textbf{BreastMNIST} test set with the ResNet-18 backbone. All configurations
share the same training protocol and differ only in how the quantum feature
vector is scaled prior to fusion: no scaling (DHF), a fixed scalar, $\ell_1$
normalization, or a scalar learned end-to-end (TSHF). Results are reported as mean with the 95\% confidence interval of the mean in
brackets (Student's $t$, $n=5$); metric values in \%, while
$\gamma_{\text{eff}}$ is dimensionless. The best result per metric within each
quantum configuration is highlighted in bold.}
\label{tab:ablation_gammab}
\footnotesize
\setlength{\tabcolsep}{4pt}
\renewcommand{\arraystretch}{1.1}
\begin{adjustbox}{max width=\linewidth}
\begin{tabular}{ll ccccc c}
\toprule
\textbf{Quantum} & \textbf{Scaling} & \textbf{Acc.} & \textbf{Prec.} & \textbf{Rec.} & \textbf{F1} & \textbf{AUC} & \boldmath$\gamma_{\text{eff}}$ \\
\midrule
\multirow{8}{*}{Non-trainable}
 & none (DHF)       & \makecell{86.28\\[-1pt]{\scriptsize [83.55, 89.01]}} & \makecell{\textbf{93.83}\\[-1pt]{\scriptsize [92.09, 95.57]}} & \makecell{87.02\\[-1pt]{\scriptsize [81.61, 92.43]}} & \makecell{90.22\\[-1pt]{\scriptsize [87.94, 92.50]}} & \makecell{\textbf{92.39}\\[-1pt]{\scriptsize [92.09, 92.69]}} & N/A \\
 & $\gamma = 0.001$ & \makecell{85.00\\[-1pt]{\scriptsize [82.78, 87.22]}} & \makecell{92.83\\[-1pt]{\scriptsize [88.36, 97.30]}} & \makecell{86.49\\[-1pt]{\scriptsize [78.47, 94.51]}} & \makecell{89.32\\[-1pt]{\scriptsize [87.15, 91.49]}} & \makecell{92.14\\[-1pt]{\scriptsize [92.02, 92.26]}} & 0.001 \\
 & $\gamma = 0.01$  & \makecell{85.64\\[-1pt]{\scriptsize [82.64, 88.64]}} & \makecell{91.25\\[-1pt]{\scriptsize [84.04, 98.46]}} & \makecell{89.65\\[-1pt]{\scriptsize [81.44, 97.86]}} & \makecell{90.11\\[-1pt]{\scriptsize [88.10, 92.12]}} & \makecell{91.93\\[-1pt]{\scriptsize [91.62, 92.24]}} & 0.010 \\
 & $\gamma = 0.1$   & \makecell{84.74\\[-1pt]{\scriptsize [82.19, 87.29]}} & \makecell{92.65\\[-1pt]{\scriptsize [89.38, 95.92]}} & \makecell{86.14\\[-1pt]{\scriptsize [79.32, 92.96]}} & \makecell{89.13\\[-1pt]{\scriptsize [86.83, 91.43]}} & \makecell{91.92\\[-1pt]{\scriptsize [91.57, 92.27]}} & 0.100 \\
 & $\gamma = 0.5$   & \makecell{85.26\\[-1pt]{\scriptsize [82.03, 88.49]}} & \makecell{92.77\\[-1pt]{\scriptsize [89.06, 96.48]}} & \makecell{86.84\\[-1pt]{\scriptsize [78.69, 94.99]}} & \makecell{89.51\\[-1pt]{\scriptsize [86.67, 92.35]}} & \makecell{91.83\\[-1pt]{\scriptsize [90.24, 93.42]}} & 0.500 \\
 & $\gamma = 2$     & \makecell{86.54\\[-1pt]{\scriptsize [84.95, 88.13]}} & \makecell{92.30\\[-1pt]{\scriptsize [89.37, 95.23]}} & \makecell{89.12\\[-1pt]{\scriptsize [84.94, 93.30]}} & \makecell{90.62\\[-1pt]{\scriptsize [89.32, 91.92]}} & \makecell{91.53\\[-1pt]{\scriptsize [90.38, 92.68]}} & 2.000 \\
 & $\ell_1$ norm.   & \makecell{87.82\\[-1pt]{\scriptsize [85.50, 90.14]}} & \makecell{92.99\\[-1pt]{\scriptsize [90.59, 95.39]}} & \makecell{90.18\\[-1pt]{\scriptsize [87.26, 93.10]}} & \makecell{91.54\\[-1pt]{\scriptsize [89.88, 93.20]}} & \makecell{90.99\\[-1pt]{\scriptsize [89.60, 92.38]}} & N/A \\
 & learnable (TSHF) & \makecell{\textbf{88.33}\\[-1pt]{\scriptsize [85.85, 90.81]}} & \makecell{91.56\\[-1pt]{\scriptsize [86.98, 96.14]}} & \makecell{\textbf{92.81}\\[-1pt]{\scriptsize [88.85, 96.77]}} & \makecell{\textbf{92.09}\\[-1pt]{\scriptsize [90.54, 93.64]}} & \makecell{92.09\\[-1pt]{\scriptsize [91.54, 92.64]}} & \makecell{1.025\\[-1pt]{\scriptsize [0.977, 1.073]}} \\
\midrule
\multirow{8}{*}{Trainable}
 & none (DHF)       & \makecell{84.87\\[-1pt]{\scriptsize [82.65, 87.09]}} & \makecell{92.96\\[-1pt]{\scriptsize [88.54, 97.38]}} & \makecell{86.14\\[-1pt]{\scriptsize [78.31, 93.97]}} & \makecell{89.21\\[-1pt]{\scriptsize [87.06, 91.36]}} & \makecell{\textbf{92.48}\\[-1pt]{\scriptsize [91.83, 93.13]}} & N/A \\
 & $\gamma = 0.001$ & \makecell{86.54\\[-1pt]{\scriptsize [85.29, 87.79]}} & \makecell{89.42\\[-1pt]{\scriptsize [84.09, 94.75]}} & \makecell{\textbf{92.98}\\[-1pt]{\scriptsize [87.02, 98.94]}} & \makecell{90.99\\[-1pt]{\scriptsize [90.36, 91.62]}} & \makecell{92.13\\[-1pt]{\scriptsize [91.65, 92.61]}} & 0.001 \\
 & $\gamma = 0.01$  & \makecell{84.23\\[-1pt]{\scriptsize [82.39, 86.07]}} & \makecell{94.32\\[-1pt]{\scriptsize [92.43, 96.21]}} & \makecell{83.51\\[-1pt]{\scriptsize [79.26, 87.76]}} & \makecell{88.53\\[-1pt]{\scriptsize [86.90, 90.16]}} & \makecell{92.29\\[-1pt]{\scriptsize [92.15, 92.43]}} & 0.010 \\
 & $\gamma = 0.1$   & \makecell{82.82\\[-1pt]{\scriptsize [79.03, 86.61]}} & \makecell{91.36\\[-1pt]{\scriptsize [82.04, 100.00]}} & \makecell{85.79\\[-1pt]{\scriptsize [75.77, 95.81]}} & \makecell{87.95\\[-1pt]{\scriptsize [85.73, 90.17]}} & \makecell{92.08\\[-1pt]{\scriptsize [91.89, 92.27]}} & 0.100 \\
 & $\gamma = 0.5$   & \makecell{86.15\\[-1pt]{\scriptsize [82.83, 89.47]}} & \makecell{90.01\\[-1pt]{\scriptsize [84.91, 95.11]}} & \makecell{91.58\\[-1pt]{\scriptsize [83.86, 99.30]}} & \makecell{90.58\\[-1pt]{\scriptsize [87.97, 93.19]}} & \makecell{91.92\\[-1pt]{\scriptsize [90.95, 92.89]}} & 0.500 \\
 & $\gamma = 2$     & \makecell{86.28\\[-1pt]{\scriptsize [83.72, 88.84]}} & \makecell{94.30\\[-1pt]{\scriptsize [93.24, 95.36]}} & \makecell{86.49\\[-1pt]{\scriptsize [81.77, 91.21]}} & \makecell{90.18\\[-1pt]{\scriptsize [88.08, 92.28]}} & \makecell{92.46\\[-1pt]{\scriptsize [92.12, 92.80]}} & 2.000 \\
 & $\ell_1$ norm.   & \makecell{\textbf{87.69}\\[-1pt]{\scriptsize [85.88, 89.50]}} & \makecell{\textbf{94.93}\\[-1pt]{\scriptsize [93.48, 96.38]}} & \makecell{87.89\\[-1pt]{\scriptsize [84.09, 91.69]}} & \makecell{\textbf{91.24}\\[-1pt]{\scriptsize [89.73, 92.75]}} & \makecell{90.96\\[-1pt]{\scriptsize [89.32, 92.60]}} & N/A \\
 & learnable (TSHF) & \makecell{86.15\\[-1pt]{\scriptsize [84.31, 87.99]}} & \makecell{94.12\\[-1pt]{\scriptsize [92.89, 95.35]}} & \makecell{86.49\\[-1pt]{\scriptsize [82.75, 90.23]}} & \makecell{90.11\\[-1pt]{\scriptsize [88.55, 91.67]}} & \makecell{92.18\\[-1pt]{\scriptsize [91.91, 92.45]}} & \makecell{1.280\\[-1pt]{\scriptsize [0.869, 1.691]}} \\
\bottomrule
\end{tabular}
\end{adjustbox}

\end{table*}

\begin{table*}[htbp]
\centering
\caption{Comparison of the proposed fusion strategies against prior hybrid
quantum--classical models on \textbf{BreastMNIST}. All works use the official
MedMNIST test split, in which the positive class is the benign category. Our
results are reported as mean with the 95\% confidence interval of the mean in
brackets (Student's $t$, $n=5$) and correspond to the best configuration of each fusion strategy; prior works
report single runs. Values in \%.}
\label{tab:across_literatureb}
\footnotesize
\setlength{\tabcolsep}{5pt}
\renewcommand{\arraystretch}{1.15}
\begin{adjustbox}{max width=\linewidth}
\begin{tabular}{lccccc}
\toprule
\textbf{Reference} & \textbf{Acc.} & \textbf{Prec.} & \textbf{Recall} & \textbf{F1-score} & \textbf{AUC} \\
\midrule
Matondo \textit{et al.}~\cite{matondo2024breast}   & 76.66  & 71.00  & 67.00 & 68.00 & -- \\
Yurtseven \textit{et al.}~\cite{yurtseven2025parallel} & 86.54 & 86.70 & \textbf{96.32} & 91.31 & -- \\
Sobrinho \textit{et al.}~\cite{sobrinho2025hybrid} & 82.10  & 81.70  & 82.00 & 80.30 & 82.60 \\
\midrule
ResNet-18 + Train.\ $|$ SHF (Ours)      & \makecell{84.74\\[-1pt]{\scriptsize [82.13, 87.35]}} & \makecell{\textbf{94.85}\\[-1pt]{\scriptsize [93.78, 95.92]}} & \makecell{83.68\\[-1pt]{\scriptsize [79.35, 88.01]}} & \makecell{88.88\\[-1pt]{\scriptsize [86.73, 91.03]}} & \makecell{92.12\\[-1pt]{\scriptsize [91.90, 92.34]}} \\
ResNet-18 + Train.\ $|$ DHF (Ours)      & \makecell{84.87\\[-1pt]{\scriptsize [82.65, 87.09]}} & \makecell{92.96\\[-1pt]{\scriptsize [88.54, 97.38]}} & \makecell{86.14\\[-1pt]{\scriptsize [78.31, 93.97]}} & \makecell{89.21\\[-1pt]{\scriptsize [87.06, 91.36]}} & \makecell{\textbf{92.48}\\[-1pt]{\scriptsize [91.83, 93.13]}} \\
ResNet-18 + Non-Train.\ $|$ DHF (Ours)  & \makecell{86.28\\[-1pt]{\scriptsize [83.55, 89.01]}} & \makecell{93.83\\[-1pt]{\scriptsize [92.09, 95.57]}} & \makecell{87.02\\[-1pt]{\scriptsize [81.61, 92.43]}} & \makecell{90.22\\[-1pt]{\scriptsize [87.94, 92.50]}} & \makecell{92.39\\[-1pt]{\scriptsize [92.09, 92.69]}} \\
ResNet-18 + Non-Train.\ $|$ TSHF (Ours) & \makecell{\textbf{88.33}\\[-1pt]{\scriptsize [85.85, 90.81]}} & \makecell{91.56\\[-1pt]{\scriptsize [86.98, 96.14]}} & \makecell{92.81\\[-1pt]{\scriptsize [88.85, 96.77]}} & \makecell{\textbf{92.09}\\[-1pt]{\scriptsize [90.54, 93.64]}} & \makecell{92.09\\[-1pt]{\scriptsize [91.54, 92.64]}} \\
\bottomrule
\end{tabular}
\end{adjustbox}

\end{table*}




% %% For numbered reference style

%   \textit{\LaTeX: a document preparation system},
%   Addison Wesley, Massachusetts,

% \end{thebibliography}
\end{document}